\documentclass[dvipsnames,format=sigconf,anonymous=false,review=false]{acmart}
\AtBeginDocument{%
  }

\usepackage{overpic}

\usepackage{svg}

\usepackage{xcolor}

\definecolor{orange_histograms}{RGB}{255,127,14}
\definecolor{blue_histograms}{RGB}{87,153,199}
\copyrightyear{2026}
\acmYear{2026}
\setcopyright{cc}
\setcctype{by}
\acmConference[GECCO '26]{Genetic and Evolutionary Computation Conference}{July 13--17, 2026}{San Jose, Costa Rica}
\acmBooktitle{Genetic and Evolutionary Computation Conference (GECCO '26), July 13--17, 2026, San Jose, Costa Rica}
\acmDOI{10.1145/3795095.3805190}
\acmISBN{979-8-4007-2487-9/2026/07}

\usepackage{xr-hyper}
\usepackage{hyperref}
\usepackage{cleveref}

\usepackage{graphicx}
\usepackage{subcaption}

\usepackage{enumitem}

\crefname{figure}{Fig.}{Figs.} 
\crefformat{supp}{Supplementary #2#1#3}

\begin{document}
\sloppy

\title{Does Dimensionality Reduction via Random Projections Preserve Landscape Features?}

\author{Iv{\'a}n Olarte Rodr{\'i}guez}
\email{i.olarte.rodriguez@liacs.leidenuniv.nl}
\orcid{0009-0005-0748-9069}
\affiliation{%
  \institution{Leiden Institute of Advanced Computer Science, Leiden University}
  \city{Leiden}
  \country{The Netherlands}
}

\author{Anja Jankovic}
\email{jankovic@aim.rwth-aachen.de}
\orcid{0000-0001-9267-4595}
\affiliation{%
  \institution{RTWH Aachen University}
  \city{Aachen}
  \country{Germany}
}

\author{Thomas B{\"a}ck}
\email{t.h.w.baeck@liacs.leidenuniv.nl}
\orcid{0000-0001-6768-1478}
\affiliation{%
  \institution{Leiden Institute of Advanced Computer Science, Leiden University}
  \city{Leiden}
  \country{The Netherlands}
}

\author{Elena Raponi}
\email{e.raponi@liacs.leidenuniv.nl}
\orcid{0000-0001-6841-7409}

\affiliation{%
  \institution{Leiden Institute of Advanced Computer Science, Leiden University}
  \city{Leiden}
  \country{The Netherlands}
}

\renewcommand{\shortauthors}{Olarte Rodr{\'i}guez et al.}

\begin{abstract}
%
%
%
Exploratory Landscape Analysis (ELA) provides numerical features for characterizing black-box optimization problems. In high-dimensional settings, however, ELA suffers from sparsity effects, high estimator variance, and the prohibitive cost of computing several feature classes. Dimensionality reduction has therefore been proposed as a way to make ELA applicable in such settings, but it remains unclear whether features computed in reduced spaces still reflect intrinsic properties of the original landscape.

In this work, we investigate the robustness of ELA features under dimensionality reduction via Random Gaussian Embeddings (RGEs). Starting from the same sampled points and objective values, we compute ELA features in projected spaces and compare them to those obtained in the original search space across multiple sample budgets and embedding dimensions.

Our results show that linear random projections often alter the geometric and topological structure relevant to ELA, yielding feature values that are no longer representative of the original problem. While a small subset of features remains comparatively stable, most are highly sensitive to the embedding. Moreover, robustness under projection does not necessarily imply informativeness, as apparently robust features may still reflect projection-induced artifacts rather than intrinsic landscape characteristics.
\end{abstract}






\maketitle

\section{Introduction}
\label{sec:introduction}
Identifying informative and reliable problem characteristics is a key requirement for a wide range of downstream tasks in numerical optimization, including landscape similarity analysis, problem classification, and automated algorithm configuration and selection \cite{kerschke_automated_2018, kerschke_automated_2019}. In recent years, Exploratory Landscape Analysis (ELA) \cite{mersmann_exploratory_2011} has emerged as a prominent data-driven framework for this purpose. By mapping a set of sampled points and their corresponding objective values to a numerical feature representation, ELA enables the characterization of black-box optimization problems without requiring explicit access to their analytical structure.

Despite their demonstrated effectiveness, ELA-based pipelines have limited applicability in expensive, large-scale optimization settings. A key limitation is that many ELA features require relatively large sample sizes to yield stable and informative estimates \cite{munoz_analyzing_2022}, which is often infeasible when function evaluations are costly. In addition, feature estimates can exhibit considerable stochastic variability, with values differing substantially across samples of identical size \cite{renau_exploratory_2020}. Furthermore, several widely used ELA features have been shown to be highly correlated with one another and to display limited variation across problem classes, thereby diminishing their discriminative capability \cite{renau_expressiveness_2019}.

An additional and often overlooked limitation of ELA concerns its scalability with respect to problem dimensionality. As pointed out by \citet{tanabe_towards_2021}, most existing ELA studies focus on low- to moderate-dimensional problems (particularly for $D \leq 20 $). However, many real-world optimization problems are inherently high-dimensional, and the computational cost of feature computation can grow rapidly with dimension. In particular, feature sets such as meta-model and level set become prohibitively expensive to compute in large dimensions due to repeated classifier training or regression model fitting. Similarly, features based on cell mappings \cite{kerschke_cell_2014} suffer from exponential growth in complexity, rendering them impractical beyond very small dimensions (typically $D \leq 5$). As a result, several feature classes that have proven important in low-dimensional settings cannot be applied to large-scale problems in practice.

Previous studies have shown that the limited scalability of ELA-based approaches substantially restricts its applicability to large-scale black-box optimization problems \cite{tanabe_towards_2021, renau_expressiveness_2019}. To mitigate this limitation, dimensionality reduction has been proposed as a viable strategy for extending ELA to high-dimensional settings \cite{tanabe_towards_2021}. Rather than computing landscape features directly in the original search space, this approach projects sampled points into a lower-dimensional representation and performs ELA feature extraction in the reduced space.
\citet{tanabe_towards_2021} addressed the scalability of computationally expensive ELA features via dimensionality reduction, focusing on cell mapping, level set, and meta-model features. Building on a PCA-inspired approach proposed for high-dimensional Bayesian optimization \cite{raponi_high_2020, antonovHighDimensionalBayesian2022a}, they showed that combining features computed in a reduced space with inexpensive features can improve classification performance when the original features are intractable in high dimensions.

Despite these results, two questions remain open. First, the effect of dimensionality reduction on other ELA feature sets is not well understood. Second, it is unclear whether features computed in reduced spaces reflect intrinsic properties of the objective function or are influenced by distortions introduced by the transformation.


In this work, we position ourselves in the setting where we are already given a sampled set of points of small size relative to the problem dimension, and we need to characterize the objective landscape based on this limited information. We aim to study how projection-based dimensionality reduction affects the characterization of the objective landscapes through ELA. 
Specifically, we use Random Gaussian Embeddings (RGEs) \cite{dasgupta_learning_1999} as dimensionality reduction technique, which embeds points into lower-dimensional manifolds independently of the sampled data. Unlike PCA-based approaches \cite{tanabe_towards_2021}, RGEs define random linear projections that do not depend on the function-instance pair, enabling us to analyze how multiple projections affect sampled points and the resulting ELA features.

%
\paragraph{Our contributions}
\label{par:contributions}
We systematically investigate the impact of RGEs on ELA features, using the noiseless Black-Box Optimization Benchmarking (BBOB) problem collection from the COmparing Continuous Optimizers (COCO) platform \cite{hansen_coco_2021} as testbed. ELA features are computed in both the original search space and in the corresponding reduced spaces across multiple problem instances. 
Specifically, our contributions are:
\begin{itemize}[leftmargin=*]
    \item We apply \textbf{RGEs} as a dimensionality reduction mechanism \textbf{in the ELA setting}, enabling the extraction of multiple low-dimensional representations from the same original problem instance and sample set. 
    \item We systematically study \textbf{ELA feature distortion under dimensionality reduction} considering both limited ($10 \cdot D$) and abundant ($100 \cdot D$) evaluation budgets, to reflect realistic data regimes in black-box optimization.
    \item We analyze the \textbf{distributional variability induced by the embedding procedure}, highlighting which ELA features are stable versus sensitive, and providing insights into the preservation of landscape characterization when using random projections. 
\end{itemize}


%
Our analysis identifies a subset of ELA features that remains robust under dimensionality reduction for a fixed sample size, providing guidance for constructing feature sets resilient to perturbations and outliers. However, robustness alone does not ensure informativeness, as some features preserve or display small changes 
due to limited sensitivity to underlying landscape variations. Furthermore, dimensionality reduction may alter how landscape properties are reflected in ELA features, raising questions about whether observed preservation corresponds to intrinsic problem characteristics. These findings suggest that relying on such features could adversely affect downstream applications, even though these effects are not directly evaluated in this study.
\paragraph{Code Availability} \url{https://github.com/olarterodriguezivan/Dimensionality-Reduction-and-Projections} 
\paragraph{Supplementary Material} Supplementary material is available as a separate file.

\section{Preliminaries}
In Section~\ref{subsec:ELA}, we provide preliminaries on ELA and motivate our selection of landscape features, while Section~\ref{subsec:random_gaussian_embeddings} introduces RGEs as a dimensionality reduction technique in the ELA context.

\subsection{Exploratory Landscape Analysis}
\label{subsec:ELA}


ELA ~\cite{mersmann_exploratory_2011} provides a methodological framework to characterize black-box optimization problems by extracting numerical features from a finite set of points and their objective values.
It facilitates a quantitative understanding of the landscape of a given problem, and serves as a key component in downstream tasks such as automated algorithm selection and configuration~\cite{kerschke_automated_2018,kerschke_automated_2019,jankovic2020landscape,kostovska2022per}.
Different feature sets have been proposed over time to capture various aspects of the problem landscape.
They can be broadly categorized into \textit{cheap} features \cite{saleem_direct_2019}, which are computed using only the initial fixed set of evaluations, and \textit{expensive} features, which require additional sampling (\textit{e.g.,} local search iterations) to be effectively computed.
In this work, we focus only on the former, \textit{i.e.,} we consider only those features whose computation does not incur additional cost in terms of function evaluations.

An overview of the selected feature sets can be found in Table~\ref{tab:feature_classes}. 
Among those, we choose three out of six feature sets originally proposed in~\cite{mersmann_exploratory_2011}: \texttt{ela\_distr}, \texttt{ela\_level} and \texttt{ela\_meta}.
Distribution features (\texttt{ela\_distr}) provide a global baseline characterization of problem difficulty by analyzing the statistical distribution of the objective values (\textit{i.e.,} how they are spread, skewed, or concentrated), independent of the location of the points in the search space.
Level set features ({\texttt{ela\_level}) analyze the geometry and connectivity of different sublevel sets, \textit{i.e.,} regions of the search space where the objective value is below a certain predefined threshold; intuitively, they capture how feasible regions change when the level of the objective value changes.
The meta-model feature set (\texttt{ela\_meta}) captures smoothness, non-linearity and multimodality of a problem by fitting linear and quadratic regression models to the sampled data and measuring their predictive error and model coefficients.

We also include five other prominent feature sets that do not require additional evaluations.
The near-better clustering features~\cite{kerschke_detecting_2015,preuss_improved_2012} (\texttt{nbc}) analyze the spatial relationships between each observation, its nearest neighbors, and its nearest better neighbors (\textit{i.e.,} neighbors with superior objective values) to capture the local clustering and funnel structures.
Dispersion features~\cite{lunacek_dispersion_2006} (\texttt{disp}) measure how points with similar objective values are distributed in the search space (\textit{i.e.,} whether they are clustered or scattered), giving insight into multimodality and basin structure.
Information content features~\cite{munoz_exploratory_2015} (\texttt{ic}) quantify ruggedness and neutrality of the landscape by looking at differences between the objective values along random walk trajectories.
Fitness distance correlation features~\cite{jones_fdc_1995} (\texttt{fitness-distance}) measure the statistical relationship between objective values and distance to the best-known solution, capturing deception and global structure.
Finally, principal component analysis features~\cite{kerschke_comprehensive_2017} (\texttt{pca}) assess the effective dimensionality and anisotropy of the problem by analyzing the relative amount of principal components required to explain a predefined amount of variance. 
%
\paragraph{Observation.}
The evaluation of ELA features in projected spaces substantially affects feature classes such as \texttt{nbc} and \texttt{ic}, which are explicitly designed to characterize local neighborhood structure and inverse pairwise distance information. In reduced spaces, random projections may induce artificial clustering or point densification, thereby distorting neighborhood relations and distance distributions. 
As we will discuss in Section~\ref{par:induced_distortion}, projection can compress and distort the geometry of the search space, leading to artificial indications of multimodality and high ruggedness. 
These effects inherently bias neighborhood-based feature classes: for \texttt{nbc}, measures such as the correlation of nearest-better points or the proportion of partially sensitive variables are affected, while for \texttt{ic}, projection-induced distortions alter the symbolic sequences obtained from random walks in the search space, leading to biased information content and partial information content values.

\paragraph{Methodological decision.}
To investigate whether these projection-induced biases could be mitigated, we additionally considered a subsampling strategy in which ELA features were computed on subsets of the projected data rather than on the full sample set. However, the observed biases persisted at comparable levels to those obtained when computing \texttt{nbc} and \texttt{ic} features on the complete projected dataset. Although subsampling increased the dispersion of the resulting feature values, the dominant source of distortion remained the embedding itself rather than the sample size. Consequently, and to avoid confusing the reader, we report in this paper only results obtained by computing ELA features using the full set of projected samples and corresponding objective values.

\begin{table}[t]
\centering
\caption{Feature sets considered in this study.}
\label{tab:feature_classes}
\renewcommand{\arraystretch}{1.05}
\resizebox{\columnwidth}{!}{%
\begin{tabular}{llc}
\toprule
\textbf{Feature set} & \textbf{Name} & \textbf{Num. features}\\
\midrule
\texttt{ela\_distr}  & distribution-based features \cite{mersmann_exploratory_2011} &  3\\
\texttt{ela\_level}  & level-set features \cite{mersmann_exploratory_2011}          & 9\\
\texttt{ela\_meta}   & meta-model features \cite{mersmann_exploratory_2011}         & 9\\
\texttt{nbc}         & nearest-better clustering \cite{kerschke_detecting_2015,preuss_improved_2012} &  5\\
\texttt{disp}        & dispersion features \cite{lunacek_dispersion_2006}           & 16\\
\texttt{ic}          & information content features \cite{munoz_exploratory_2015}    &  5\\
\texttt{fitness\allowbreak-distance}
                    & fitness distance correlation \cite{jones_fdc_1995}            & 6\\
\texttt{pca}         & principal component analysis \cite{kerschke_comprehensive_2017}                                  & 8\\
\bottomrule
\end{tabular}%
}
\end{table}

\subsection{Random Gaussian Embeddings for Dimensionality Reduction}
\label{subsec:random_gaussian_embeddings}

\paragraph{Generalities}
\label{par:random_gaussian_embeddings_generalities}

RGEs~\cite{dasgupta_learning_1999} are a simple and widely adopted class of dimensionality reduction techniques based on random linear projections. Given an original search space of dimension $D$, RGEs map points $\mathbf{x} \in \mathbb{R}^D$ to a \allowbreak lower\allowbreak-\allowbreak dimensional space $\mathbb{R}^d$, with $d \ll D$, through multiplication by a random projection matrix $\mathbf{A} \in \mathbb{R}^{d \times D}$. The entries of $\mathbf{A}$ are assumed to be independent, zero-mean, unit-variance random variables, and are commonly drawn from a standard normal distribution. With appropriate normalization, the resulting projection is given by:
\begin{equation}
\mathbf{z} = \frac{1}{\sqrt{d}} \mathbf{A}\mathbf{x}.
\label{eq:gaussian_random_embedding}
\end{equation}

The theoretical justification for random projection methods is provided by the Johnson–Lindenstrauss (JL) lemma \cite{jl_extensions_1984}, which guarantees that, for any finite set of points, a random linear mapping into a lower-dimensional space preserves pairwise Euclidean distances up to a bounded distortion with high probability. Crucially, this probability depends on the reduced dimensionality: higher target dimensions yield stronger concentration guarantees and lower distortion, while overly aggressive dimensionality reduction degrades distance preservation.

As a result, RGEs preserve the global geometric structure of a finite set of points in terms of their pairwise Euclidean distances while reducing the dimensionality of the representation. In contrast to commonly used linear dimensionality reduction techniques such as PCA, RGEs are data-agnostic, as the embedding is constructed independently of the observed data. 
More broadly, dimensionality reduction often entails a trade-off between preserving global geometry (\textit{e.g.,} distances) and preserving local or topological structure (\textit{e.g.,} neighborhood relations and connectivity). Since ELA features capture both geometric and neighborhood-based properties, distortions in either aspect can affect the resulting landscape characterization.

In black-box optimization, random embedding methods have been explored as a strategy to alleviate the curse of dimensionality, most notably in the \texttt{REMBO} algorithm~\cite{wang_bayesian_2016}. These approaches rely on the assumption that the objective function depends on a low-dimensional active subspace, which a random linear embedding intersects with high probability. 
However, while random projections approximately preserve pairwise distances, this property alone does not guarantee the preservation of high-level landscape characteristics.
In particular, projecting the search space can alter the qualitative structure of level sets and introduce spurious landscape features in the reduced space.

In contrast, other dimensionality reduction methods, such as Uniform Manifold Approximation and Projection (UMAP)~\cite{mcinnes_umap_2020}, prioritize the preservation of local neighborhood structure, but typically at the expense of global geometric properties. Most dimensionality reduction techniques commonly used for tabular data involve an explicit trade-off between geometric fidelity and neighborhood preservation.
This highlights that dimensionality reduction inevitably introduces a ``price to pay'' and motivates studying how embedding-induced geometric and topological distortions propagate to ELA feature values.


Moreover, some ELA feature classes, notably \texttt{ela\_level} and \texttt{ela\_meta}, rely on classification or regression models fitted to sampled objective values. When combined with dimensionality reduction, the loss function governing the classification or regression task is typically not aligned with the projection step. This mismatch can distort landscape properties such as modality, basin structure, or ruggedness, thereby affecting the resulting feature values and reducing their reliability.

\paragraph{Visual Proof of Induced Landscape Distortion by Projections}
\label{par:induced_distortion}

Regarding the aforementioned distortion in the topology of the landscape, a simple visual proof is shown in Figure \ref{fig:example_aliasing}, where a total of 20 points were initially sampled and the 11th instance of the noiseless two-dimensional Rosenbrock Function (f8) of the BBOB problem collection \cite{hansen_coco_2021} was evaluated at each point.

\begin{figure}[ht]
  \centering



  \includegraphics[trim=2.2cm 0.3cm 1cm 0.15cm, clip,width=1\linewidth]{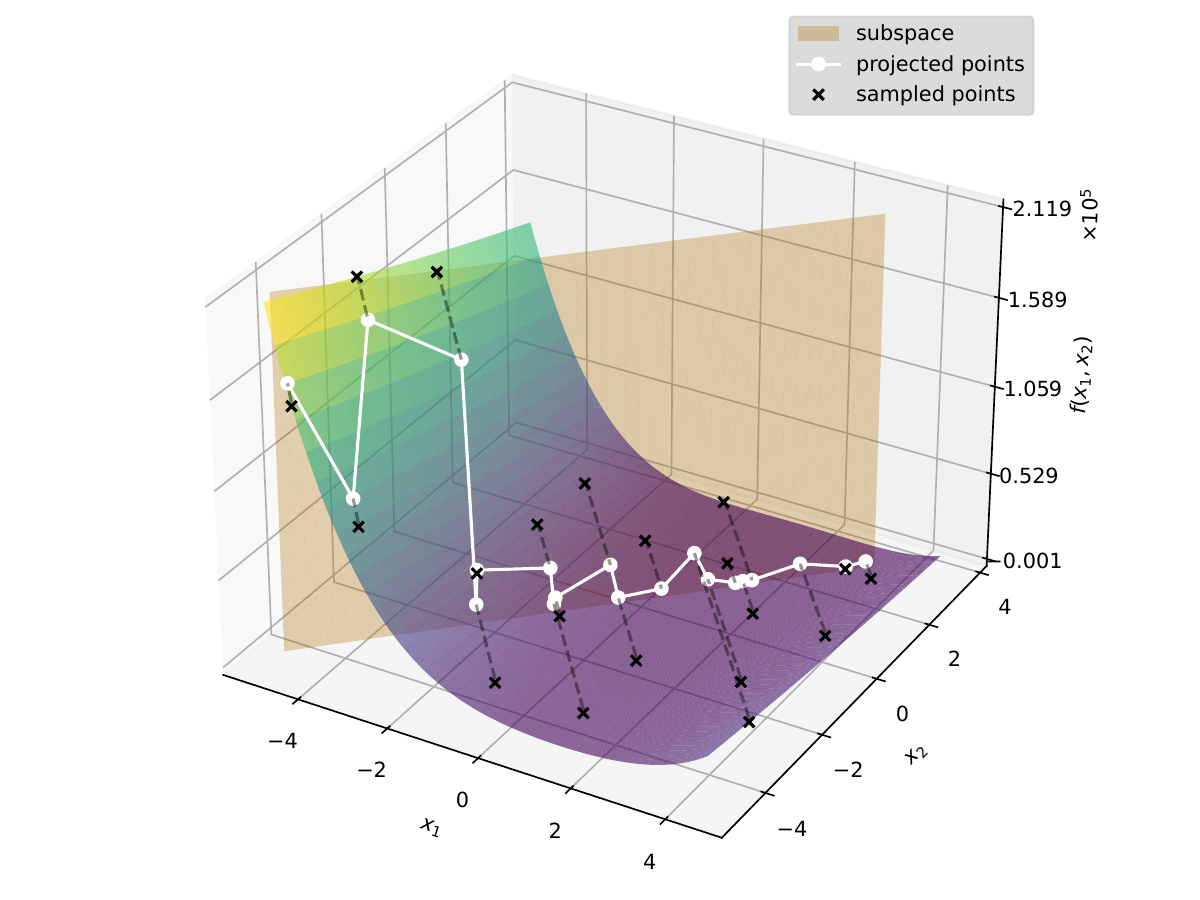}

  \caption{Surface plot and pointwise projection onto a lower-dimensional space of the 11th instance of the two-dimensional Rosenbrock function (f8).
  }
  \label{fig:example_aliasing}
\end{figure}

In this case, Figure~\ref{fig:example_aliasing} shows that the projected landscape exhibits spurious multimodality, while the strong conditioning present in the original space is no longer observable. Although the initial experimental design was generated using a highly space-filling Latin hypercube sampling (LHS) scheme, this property is not preserved under projection. Instead, the embedding induces sampling artifacts, manifested as a non-uniform and anisotropic distribution of points in the reduced space. Because the projection operator is non-injective, multiple distinct points in the original design space collapse onto identical projected coordinates, while the objective function varies along the corresponding fibers. As a result, structural patterns observed in the projected space are projection-induced distortions of the sampling design rather than intrinsic characteristics of the objective function.

This effect violates a core assumption underlying many ELA features, namely that the sample set provides approximately uniform and isotropic coverage of the search space~\cite{mersmann_exploratory_2011, renau_exploratory_2020}. Consequently, projection-induced clustering is systematically misinterpreted as landscape properties such as multimodality, ruggedness, or ill-conditioning, leading to biased and misleading feature values.

When ELA features are computed in the reduced space, several fundamental questions arise: (i) how to obtain reliable and stable ELA estimates when the projected design is no longer space-filling; (ii) how to determine whether essential landscape characteristics are preserved under projection; and (iii) to what extent ELA features remain invariant under changes in the projection, such as different random embeddings or reduced dimensionalities. The following experimental setup is designed to empirically analyze these questions.

\section{Experimental Setup}
\label{sec:experimental_setup}
The experiments were conducted on the BBOB problem collection provided by the COCO framework. The BBOB suite comprises 24 noiseless single-objective benchmark functions defined on bounded continuous search spaces. For each function, multiple problem instances are generated through transformations such as translations, rotations and scaling. 

In this study, we work in dimension $D=20$, and consider 15 instances each of the 24 BBOB functions, which adds up to a total of 360 problem instances.
All benchmark problems were accessed via the \texttt{IOHexperimenter} framework \cite{IOHexperimenter}.

These functions span the five function groups defined in the BBOB taxonomy~\citep{bbob2019}, namely \emph{separable}, \emph{low or moderate conditioning}, \emph{high conditioning and unimodal}, \emph{multi-modal with adequate global structure}, and \emph{multi-modal with weak global structure}}. 

To compute the ELA features, we generated 80 LHS designs
\(\mathbf{X_l} = \{\mathbf{x}_i\}_{i=1}^{S}, \text{for } l=1,\dots,80\), by sampling from the domain
\([-5,5]^{20}\). 
40 of these designs were created with a sample size of \(S = 10D\) and the
remaining 40 with \(S = 100D\), where \(D = 20\). This setup reflects scenarios
with limited and abundant data, respectively. For each sampling design, all 360 problem instances were evaluated, after which the ELA features were computed.

The feature sets considered in this study are introduced in Section~\ref{subsec:ELA} and presented in Table~\ref{tab:feature_classes}.
These particular feature sets were selected to avoid any form of resampling, thereby ensuring that all feature sets are computed from datasets that contain the same amount of information. Feature evaluation was carried out using the \texttt{Python} implementation of the \texttt{flacco} library \cite{kerschke_comprehensive_2017}. The computational runtime of individual feature sets is deliberately omitted, as it is not descriptive of the underlying landscape characteristics.

We considered three reduced dimensionalities, namely $d = \{2,5,10 \}$, corresponding to compression ratios $r = d / D$ of $\{0.1,0.25,0.5\}$, respectively, and constructed 40 independent realizations of random RGEs for each of them.
All embeddings were generated using the \texttt{scikit\allowbreak-\allowbreak learn} framework~\cite{scikit-learn}. Employing multiple embedding realizations constitutes a replicated experimental design that enables a systematic assessment of the robustness and stability of ELA features with respect to projection-induced variability. Rather than relying on a single embedding instance, this setup treats the projection as a controlled stochastic factor and allows variability in the extracted features to be attributed explicitly to the embedding mechanism. Since the computational overhead associated with generating embeddings and projecting samples is negligible, this replicated design permits repeated perturbations of the original sample set without introducing additional sources of noise. Importantly, generating multiple embeddings requires no additional objective evaluations, making it particularly suited for expensive-to-evaluate problems.

\subsection{Evaluation Metric}
\label{subsec:evaluation_metric}

We introduce a distributional formulation of ELA features that explicitly characterizes estimator bias induced by random embeddings.
Let \((\mathbf{X}_l, f_{m,n}(\mathbf{X}_l))\) denote a dataset generated according to
sampling design \(l\) for objective function \(f_{m,n}\), where $m$ corresponds to the function identifier according to the notation used in BBOB and $n$ the corresponding instance identifier, where $n=\{0,1,2, \dots, 14 \}$.
The computation of ELA features is formalized through the mapping
\begin{equation}
\mathbf{t}_{l,m,n,\star}
:= \phi\!\left(\mathbf{X}_l, f_{m,n}(\mathbf{X}_l)\right) \in \mathbb{R}^p,
\label{eq:general_ELA_mapping}
\end{equation}
where \(\mathbf{t}_{l,m,n,\star}\) denotes the vector of \(p\) ELA feature values.
The symbol \(\star\) refers to a reference feature vector computed using the full
dataset associated with sampling design \(l\), consistent with standard practice
in ELA.
In the present work, this reference serves as a baseline for assessing the
sensitivity of feature estimates to embedding and sampling effects.

\medskip
To explicitly model such perturbations, we define an extended feature mapping
\begin{equation}
\tilde{\mathbf{t}}_{l,m,n,k}
:=
\tilde{\phi}\!\left(
\mathbf{A}_k,\,
\mathbf{X}_l,\,
f_{m,n}(\mathbf{X}_l)
\right)
\in \mathbb{R}^p,
\label{eq:sampling_projections_ELA_mapping}
\end{equation}
where \(\mathbf{A}_k \in \mathbb{R}^{d \times D}\) denotes a random linear embedding
operator. Each realization \(\tilde{\mathbf{t}}_{l,m,n,k}\) constitutes a perturbed estimate
of the reference feature vector \(\mathbf{t}_{l,m,n,\star}\).
Varying the embedding index \(k\) induces a probability distribution over ELA feature values, with randomness arising from the embedding operator.
For a fixed problem instance \((l,m,n)\) and feature index \(q\), we define the relative feature shift generated by an embedding as:
\begin{equation}
\delta^{(q)}_{l,m,n,k}
=
\frac{
\tilde{t}^{(q)}_{l,m,n,k}
-
t^{(q)}_{l,m,n,\star}
}{
\left| t^{(q)}_{l,m,n,\star} \right| + \epsilon
},
\label{eq:bias_embedding}
\end{equation}
where $\epsilon$ is a small numerical value set to $10^{-9}$.

\begin{figure*}[t]
    \centering
    \includegraphics[width=0.8\linewidth]{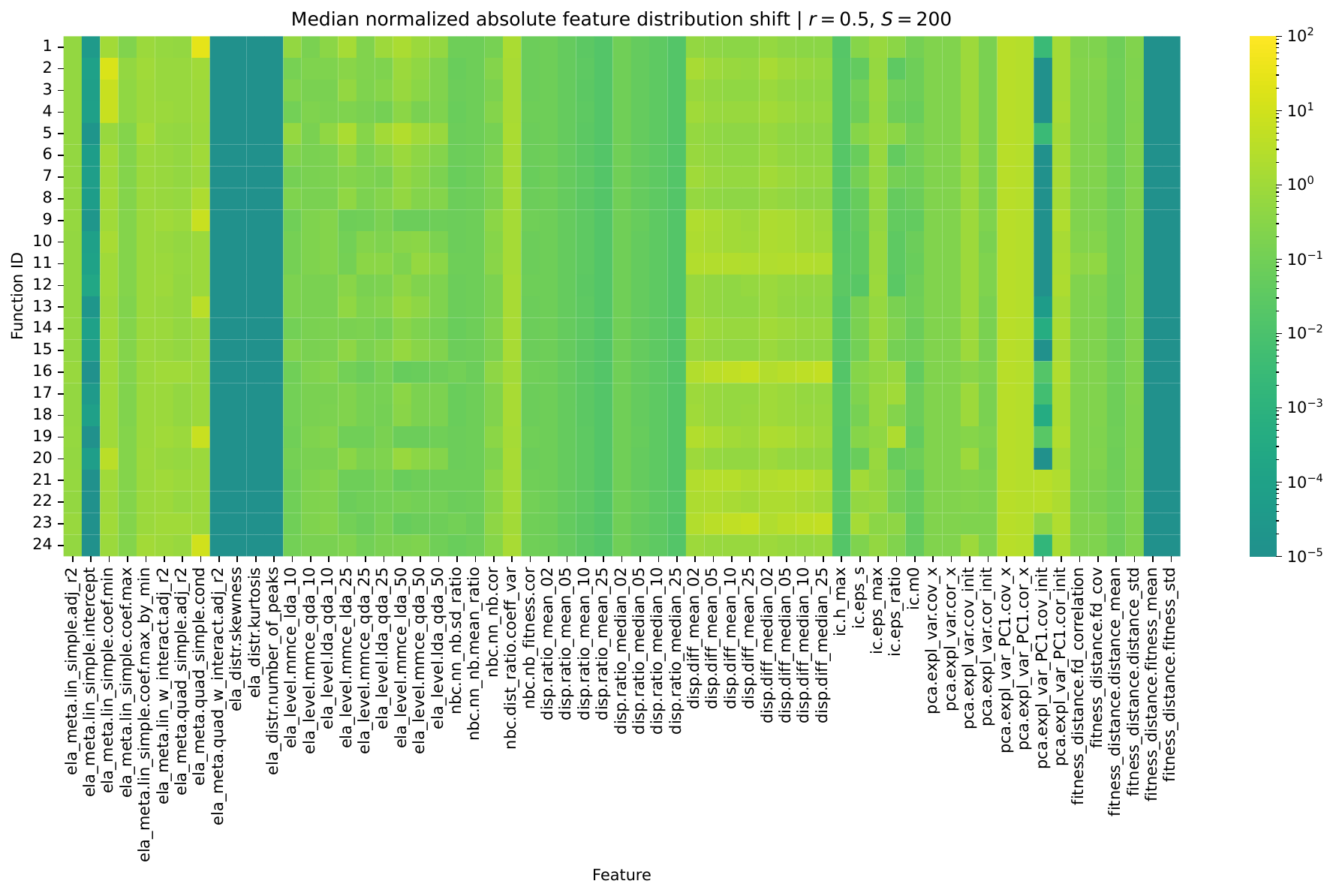}
    \caption{Heatmap of the median normalized absolute feature distribution shift for datasets of size 200 with a compression ratio of $0.5$. Results shown for all BBOB functions.}
    \label{fig:heatmap_oneshot_200_0.5}
\end{figure*}

Equation~\eqref{eq:bias_embedding} defines a normalized measure of the shift induced by an embedding, enabling meaningful comparison and aggregation across different instances of the same function as well as across datasets. If a feature $\phi^{(q)}$ exhibits limited sensitivity to distance-preserving transformations, then random linear embeddings introduce only a controlled distortion in the aggregated feature estimates. In particular, as the embedding dimension $d$ increases relative to the original dimension $D$, the induced shift $\delta^{(q)}_{l,m,n,k}$ remains small and does not grow with problem size. 

In the presence of random RGEs, the induced shift is generally nonzero and tends to increase as the compression ratio $r$ decreases, reflecting the stronger geometric distortion associated with more aggressive dimensionality reduction. We therefore define a feature to be \textit{robust} if the induced shift remains limited across embedding dimensions, and \textit{invariant} if the shift is identically zero for all embedding dimensions.

Invariance is not observed for all considered ELA features except those belonging to the \texttt{ela-distr} feature class, as well as the
\texttt{fitness\allowbreak-distance\allowbreak.\allowbreak fitness\allowbreak\_\allowbreak mean} and
\texttt{fitness\allowbreak-distance\allowbreak.fitness\allowbreak\_std} features.
These features depend exclusively on the set of objective function evaluations $f_{m,n}(\mathbf{X}_l)$ and are therefore invariant under linear embeddings of the input space. An analogous pseudo-invariance property is observed for the
\texttt{ela\allowbreak-\allowbreak meta\allowbreak.\allowbreak lin\allowbreak\_\allowbreak simple\allowbreak.\allowbreak intercept} feature, which is only marginally affected, as LHS designs yield sample mean vectors $\mathbf{x}$ close to $\mathbf{0}$ (see supplementary material for the theoretical proof).


\section{Results}
\label{sec:Results}

\paragraph{Changes in shift by feature and function}

The first part of the results section analyzes how the induced feature distribution shift varies across functions. To this end, we consider an \textit{absolute shift}, defined as the absolute value of the quantity in Equation~\eqref{eq:bias_embedding}. For this matter, we aggregate the values per each $(l,n,k)$ triplet in order to have a distribution per feature over a distinct function $m$. The corresponding results are shown in Figure~\ref{fig:heatmap_oneshot_200_0.5}, obtained for a compression ratio of $0.5$ and a sample size of $200$. Similar qualitative conclusions are observed for the remaining sample sizes and compression ratios considered (plots available as supplementary material).

The variation in shift across functions is, in most cases, predominantly driven by the feature under consideration rather than by the specific function. This is evidenced by the vertical stripe patterns, indicating that the projection induces similar relative shifts across functions from different BBOB categories, with no systematic dependence on the underlying function landscapes. 
Since the shift is measured relative to feature values computed in the original space, comparable colors across functions reflect comparable \textit{relative} distortions induced by the projection, even if the absolute feature values differ.
Moreover, because the comparison is performed on a relative scale, the remaining variability is mainly determined by the compression ratio and, to a lesser extent, by the small sample size considered in this case. 

Notable exceptions are the \texttt{ela\_level} and \texttt{ic} feature sets, for which a more pronounced variability across functions is observed. 
This suggests a stronger dependence on the underlying landscape and points to a non-negligible feature–function interaction, where projection-induced distortions affect these feature sets differently.
The higher across-function variability of the \texttt{ela\_level} and \texttt{ic} features can be attributed to the fact that both feature sets depend on level set geometry and neighborhood-induced local transitions, respectively. These properties are strongly landscape-dependent and particularly sensitive to projection-induced changes in neighborhood relations, conditioning, and basin structure.


\begin{figure*}[tbh]
    \centering
    \includegraphics[width=0.85\textwidth,trim=0cm 7.5cm 0cm 0cm,clip]{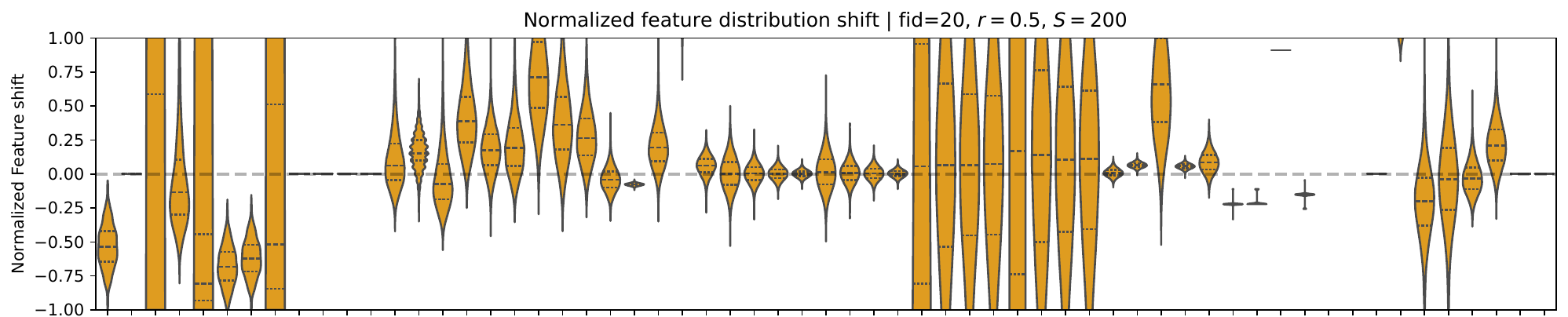}

    \includegraphics[width=0.85\linewidth]{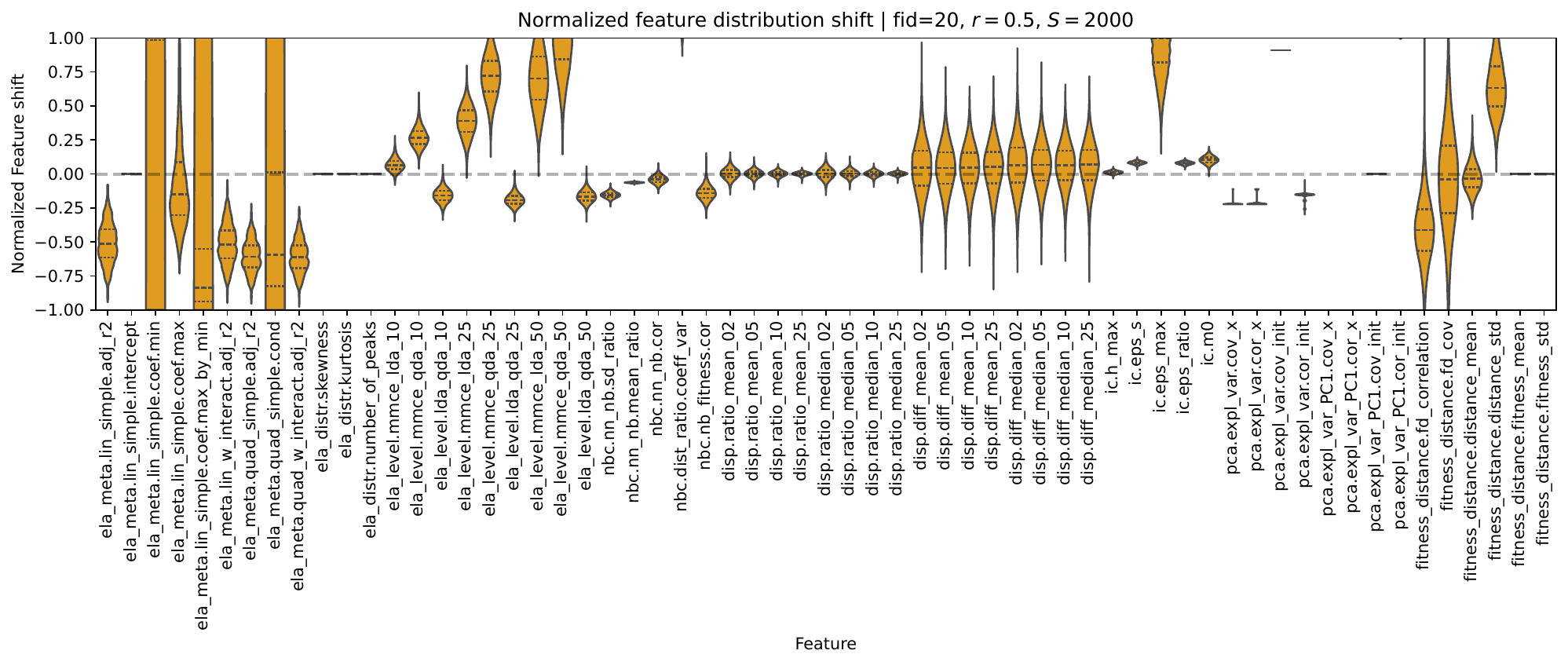}
    \caption{Normalized aggregated feature distribution shift of Schwefel (f20) function with $S=200$ and $S=2000$ for fixed compression ratio $r=0.5$. The horizontal dashed line denotes a normalized reference corresponding to the median of each feature distribution in the original search space. To enhance visualization, the limits of the Normalized Feature shift has been set to $[-1,1 ]$.}
    \label{fig:violin_plot_comp}
\end{figure*}

The smallest absolute median shift values are observed for the \texttt{ela\allowbreak\_distr} feature set, as well as for the features \texttt{fitness\allowbreak\_mean} and \texttt{fitness\allowbreak-distance\allowbreak.fitness\allowbreak\_std}. As discussed in Section~\ref{subsec:evaluation_metric}, these features depend only on the fitness values and are therefore unaffected by changes in the search space representation. Moreover, features such as \texttt{ela\allowbreak\_meta\allowbreak.lin\allowbreak\_simple\allowbreak.intercept} and \texttt{pca\allowbreak.expl\allowbreak\_var\allowbreak\_PC1\allowbreak.cov\allowbreak\_init} also exhibit small shifts, despite not being purely fitness-based. The stability of the former can be attributed to the approximately centered sampling designs, which render the intercept largely insensitive to changes in the coordinate representation. The latter is computed from the global covariance structure of the sample points and captures coarse-grained geometric properties that are preserved under random projection, resulting in limited sensitivity to dimensionality reduction.

\paragraph{An analysis towards stability and robustness}
Building on the previous analysis, which revealed similar feature shift patterns across different functions, we focus here on a single representative case.
Figure~\ref{fig:violin_plot_comp} depicts the normalized distribution shift, as defined in Equation~\eqref{eq:bias_embedding}, of ELA feature estimators under dimensionality reduction of the Schwefel function (f20). The features are computed using the two considered sample sizes, while the comparison focuses on a compression ratio of 0.5. Each violin plot summarizes the empirical distribution of the normalized shift across multiple function instances, providing insight into both the relative median and its dispersion induced by projection. Similar plots for compression ratios $r=\{0.1,0.25\}$ and the other functions considered in this study are provided as supplementary material.

Overall, both figures reveal qualitatively similar patterns across ELA feature sets, indicating that the effect of dimensionality reduction on feature estimators is largely consistent across the configurations considered. In both cases, a consistent portion of feature values remains concentrated near zero, suggesting limited systematic bias introduced by projection. This behavior highlights subsets of ELA features that are, on average, robust to projection-induced distortions.

Nevertheless, marked differences in the dispersion of the resulting distributions provide important insights into feature robustness and estimator variance. In contrast to the dimensionality reduction strength, which is held fixed in the presented comparisons, the initial sample size plays a key role: as the number of sampled points decreases, increased estimator variance becomes apparent and propagates through the dimensionality reduction process.

In particular, the \texttt{disp\allowbreak.ratio} features exhibit a quite stable behavior, remaining largely unbiased even when the sample size is reduced. In contrast, several other feature estimators tend to converge to non-zero values, with changes in sample size primarily affecting the dispersion of their distributions rather than their median. This pattern is especially pronounced for features belonging to the \texttt{nbc} and \texttt{ic} sets, and to a lesser extent for those in the \texttt{fitness\allowbreak\_distance} set. 

Compared to the findings of \citet{renau_towards_2021}, the features identified in our study as exhibiting low variance and limited bias largely coincide with the reduced feature subset proposed therein. While we initially attributed this robustness to the capture of coarse-grained, global landscape properties, our results indicate that this explanation is insufficient.

In particular, although information content features depend explicitly on the metric, they appear comparatively robust under projection, likely due to their thresholded and aggregated construction. In contrast, most of the meta-model features are strongly affected by projection. Level-set features exhibit pronounced and highly non-uniform variability, suggesting a strong dependence on the geometric structure of sublevel regions, while nearest-better clustering features show moderate sensitivity due to their reliance on local neighborhood relations.

Overall, these findings indicate that projection robustness is governed less by the nominal spatial scale of a feature than by its dependence on the geometric organization of the sample. Features based on thresholding and aggregation tend to be more stable, whereas those relying on connectivity, neighborhood structure, or surrogate-model geometry are more susceptible to projection-induced bias and variance.


Taken together, these observations indicate that dimensionality reduction affects ELA feature estimators in a heterogeneous manner, with sample size acting as a critical moderating factor. While some features exhibit stable and well-bounded behavior across different dataset sizes, others (most notably \texttt{ela\_meta} and \texttt{ela\_level} feature sets) show substantial sensitivity. Such behavior can complicate the interpretation of ELA features and may introduce systematic bias in downstream tasks such as algorithm selection and configuration, particularly when feature estimates are computed from low-dimensional projections or limited sample sizes.




\paragraph{Features measured in reduced projected spaces are not representative of intrinsic function properties}

Although the previous sections identified subsets of ELA features that remain robust and stable under dimensionality reduction for a fixed sample size, it remains unclear whether these features can still capture the intrinsic properties of the objective function that are detectable in the original high-dimensional space given sufficient sampling, but are eventually lost with small sample sizes. Therefore, in this section we examine whether ELA features computed in projected spaces are able to reveal such intrinsic properties.
To this end, we compare empirical feature distributions obtained from different sample sizes, both in the original space and after projection, for specific, representative functions. We consider the examples below, for specific function--instance pairs, to illustrate that projection can yield feature estimators that present different stability properties with respect to sampling and compression ratios.


%
\begin{figure}[t]
  \centering
  \begin{tiny}
    \includesvg[width=\linewidth]{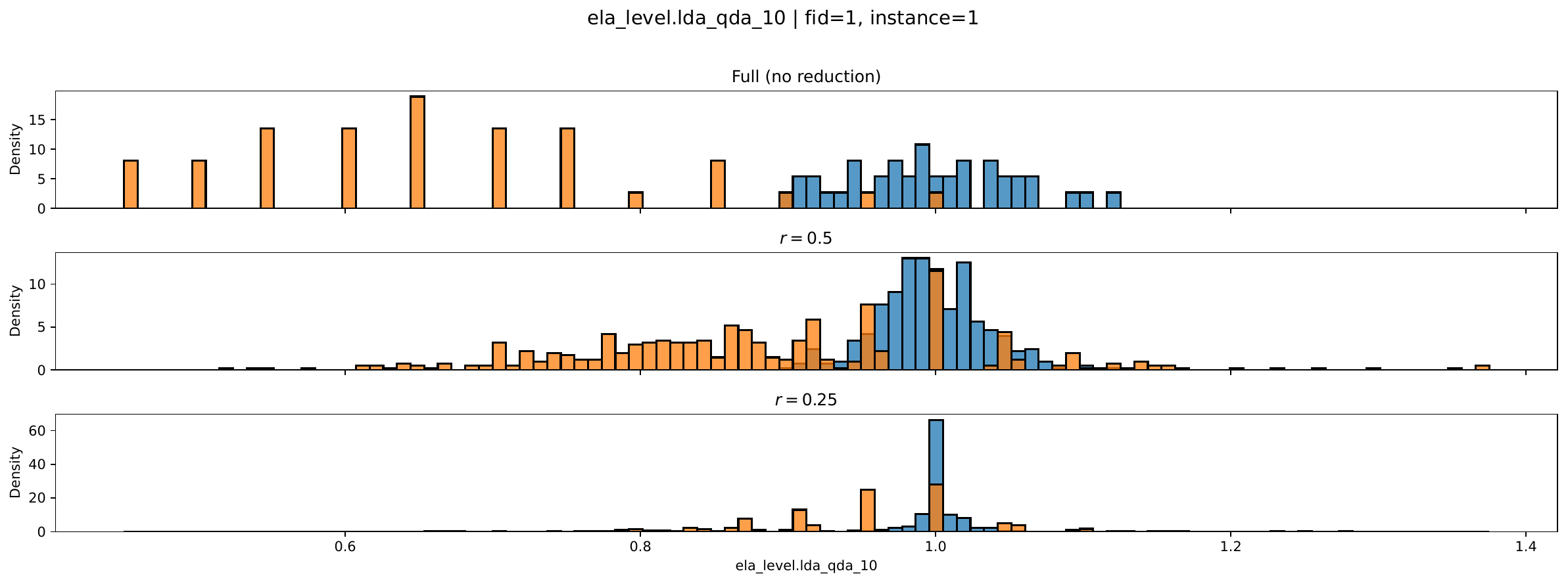}
    \end{tiny}
  \caption{Distribution of the~\texttt{ela\_level.lda\_qda\_10}~feature of instance 1 of the Sphere (f1) function. The \textcolor{blue_histograms}{blue} distribution corresponds to the measured feature by using $S=2000$ and the \textcolor{orange_histograms}{orange} corresponds to the measurement of the feature by using $S=200$.}
  \label{fig:histogram_lda_qda_sphere}
\end{figure}
%

An illustrative case is shown in Figure~\ref{fig:histogram_lda_qda_sphere}, which reports the empirical distribution of \texttt{ela\allowbreak\_level\allowbreak.lda\allowbreak\_qda\_10} for instance $1$ of the Sphere function (f1). The feature \texttt{ela\allowbreak-level\allowbreak.lda\allowbreak\_qda\_10} measures the relative separability of the best $10\%$ of sampled points by comparing the misclassification error of a linear (LDA) versus a quadratic (QDA) discriminant classifier. This means that it captures whether the set of best points is linearly separable from the rest in the search space, or whether a more complex boundary is needed (often linked to multimodality).
In the original space, the feature exhibits substantial variability and does not indicate an aligned trend for the two sample sizes. After projection, however, the empirical distributions of the feature estimates obtained using sample sizes $S=2000$ and $S=200$ become increasingly aligned in terms of their mean values. This apparent agreement suggests that the feature estimator has stabilized, even though it converges to a projection-specific value rather than to the corresponding estimate in the original space with the largest sample set.
We refer to this behavior as \textit{projection-consistent convergence}. It is particularly pronounced for \texttt{ela\_level} and \texttt{ela\_meta} features, and is consistent with projection-induced distortions that can alter the qualitative structure of the sampled landscape (e.g., inducing apparent multimodality, which increases the difficulty for both LDA and QDA classifiers to identify a reliable separating boundary).

\begin{figure}[t]
  \centering
  \begin{tiny}
    \includesvg[width=\linewidth]{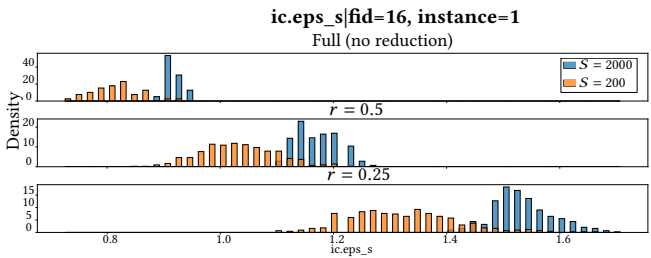}
    \end{tiny}
  \caption{Distribution of the~\texttt{ic\allowbreak.eps\allowbreak\_s}~feature of instance 1 of the Weierstrass (f16) function. The \textcolor{blue_histograms}{blue} distribution corresponds to the measured feature by using $S=2000$ and the \textcolor{orange_histograms}{orange} corresponds to the measurement of the feature by using $S=200$}
  \label{fig:histogram_ic_eps_s_weierstrass}
\end{figure}

A second pattern is illustrated in Figure~\ref{fig:histogram_ic_eps_s_weierstrass}, which shows the empirical distribution of the \texttt{ic\allowbreak.eps\allowbreak.s} feature for instance 1 of the Weierstrass function (f16). This function is known to exhibit multimodal landscape characteristics; consequently, for sufficiently dense sampling, elevated feature values are expected. 
As shown in the figure, increasing the level of dimensionality reduction leads to a systematic rightward shift of the empirical feature distributions for both considered dataset sizes. This shift indicates an increase in the estimated feature that is induced by the projection rather than by changes in the underlying objective function. In particular, the reduction of the embedding dimension increases the effective point density in the projected space, which in turn inflates the \texttt{ic\allowbreak.eps\allowbreak.s} estimators. 
The fact that this effect is observed consistently across both dataset sizes suggests that it is primarily driven by projection-induced changes in point density rather than by sampling variability. We refer to this behavior as \textit{projection-induced shift}. It is predominantly observed for feature estimators belonging to the \texttt{nbc} and \texttt{ic} feature sets, and can be explained by the increased effective point density in the reduced space, which inflates neighborhood- and transition-based statistics.

    

In summary, the two observed behaviors indicate that dimensionality reduction can introduce systematic distortions in the estimation of certain ELA features. These effects cannot always be avoided and may lead to feature estimates that reflect properties induced by the projection rather than intrinsic characteristics of the underlying objective function. As a consequence, such projection-induced effects can be misinterpreted as genuine landscape properties and can systematically bias downstream tasks, in particular algorithm configuration and selection, which rely on an accurate characterization of problem features.
\section{Conclusions}
\label{sec:Conclusions}

This work investigated the effects of random Gaussian embeddings on Exploratory Landscape Analysis (ELA) features and identified a subset of features that exhibit comparatively stable behavior under dimensionality reduction. 
In particular, several 
fitness-only statistics (e.g., \texttt{fitness\_distance}), dispersion-related measures (e.g., \texttt{disp.ratio}), selected meta-model coefficients (including \texttt{ela\_meta.lin\_simple.intercept}), and some PCA-derived descriptors (such as \texttt{pca.expl\_var\_PC1.cov\_init}) show lower sensitivity to both the choice of embedding and the target dimension in our experiments. 
These features appear to capture coarse-grained properties of the objective landscape that are, to some extent, preserved under random projection. 

At the same time, our results indicate that dimensionality reduction via random projections can significantly alter landscape representations. 
In particular, we observe that projection may introduce distortions in the induced sample distribution, which can manifest as artificial multimodality or increased ruggedness in certain ELA descriptors. 
Such effects can bias the interpretation of features and, consequently, impact downstream tasks such as algorithm selection and configuration. 
Therefore, ELA features computed in projected spaces should not, in general, be considered directly interchangeable with their high-dimensional counterparts.


Finally, our results suggest that combining limited sampling with random projections does not fully mitigate the challenges associated with estimating ELA features in high-dimensional settings. 
In particular, projection does not consistently compensate for the loss of information due to small sample sizes. 
Future work will therefore investigate the use of structured or \textit{a priori} embeddings, with the goal of enabling more effective allocation of sampling budgets and improving the reliability of landscape characterization under tight evaluation constraints.

\bibliographystyle{ACM-Reference-Format}
\bibliography{sample-base}

\appendix










\end{document}


\maketitle

\appendix

\section{Proof that \texttt{ela-meta.lin\_simple.intercept} is near invariant under linear projections}
\label{sec:proof_invariance}

Let $\{(\mathbf{x}_i, y_i)\}_{i=1}^n$ be a dataset with predictors
$\mathbf{x}_i \in \mathbb{R}^d$ and losses $y_i \in \mathbb{R}$.
Consider ordinary least squares (OLS) with an intercept:
\begin{equation}
\min_{\beta_0,\boldsymbol{\beta}} \;
\sum_{i=1}^n \big(y_i - \beta_0 - \boldsymbol{\beta}^\top \mathbf{x}_i\big)^2 .
\end{equation}
We define the sample means:
\[
\bar{\mathbf{x}} := \frac{1}{n}\sum_{i=1}^n \mathbf{x}_i,
\qquad
\bar{y} := \frac{1}{n}\sum_{i=1}^n y_i .
\]

\paragraph{Lemma 1 (Intercept identity).}
For OLS with an intercept, any optimal solution
$(\hat\beta_0,\hat{\boldsymbol{\beta}})$ satisfies:
\begin{equation}
\hat\beta_0 = \bar{y} - \hat{\boldsymbol{\beta}}^\top \bar{\mathbf{x}} .
\end{equation}

\paragraph{Proof.}
Differentiating the objective with respect to $\beta_0$ and setting the
derivative to zero yields:
\[
0 = \sum_{i=1}^n (y_i - \beta_0 - \boldsymbol{\beta}^\top \mathbf{x}_i),
\]
which implies:
\[
n\beta_0 = \sum_{i=1}^n y_i -
\boldsymbol{\beta}^\top \sum_{i=1}^n \mathbf{x}_i .
\]
Dividing by $n$ gives the stated result. 

\paragraph{Independent OLS Estimators.}
Consider two independent OLS problems. The first is in the original space $\mathbb{R}^d$:
\begin{equation}
(\hat{\beta}_0, \hat{\boldsymbol{\beta}}) = \arg\min_{\beta_0, \boldsymbol{\beta}} \sum_{i=1}^n (y_i - \beta_0 - \boldsymbol{\beta}^\top \mathbf{x}_i)^2
\end{equation}
The second is in the reduced space $\mathbb{R}^k$ using $\mathbf{z}_i = \mathbf{A}\mathbf{x}_i$:
\begin{equation}
(\hat{\alpha}_0, \hat{\boldsymbol{\alpha}}) = \arg\min_{\alpha_0, \boldsymbol{\alpha}} \sum_{i=1}^n (y_i - \alpha_0 - \boldsymbol{\alpha}^\top \mathbf{z}_i)^2
\end{equation}

\paragraph{Analysis of the Gradient.}
While the reduced estimator $\hat{\boldsymbol{\alpha}}$ effectively optimizes over the transformation $\mathbf{A}$, the original estimator $\hat{\boldsymbol{\beta}}$ is free to utilize all $d$ dimensions. We define the discrepancy vector $\boldsymbol{\delta}$ as:
\[
\boldsymbol{\delta} = \hat{\boldsymbol{\beta}} - \mathbf{A}^\top \hat{\boldsymbol{\alpha}}
\]
This vector $\boldsymbol{\delta}$ represents the information in $\mathbf{y}$ that is correlated with the components of $\mathbf{x}$ discarded by the dimensionality reduction $\mathbf{A}$. Specifically, if $\mathbf{P}_A = \mathbf{A}^\top(\mathbf{A}\mathbf{A}^\top)^{-1}\mathbf{A}$ is the projection onto the row space of $\mathbf{A}$, then $\mathbf{A}^\top \hat{\boldsymbol{\alpha}}$ is constrained to this subspace, whereas $\hat{\boldsymbol{\beta}}$ is not.

\paragraph{Theorem: Intercept Non-Equivalence.}
The intercepts of the two independent models satisfy the relationship:
\begin{equation}
\hat{\beta}_0 = \hat{\alpha}_0 - \boldsymbol{\delta}^\top \bar{\mathbf{x}}
\end{equation}
Equivalently, the difference is $\hat{\alpha}_0 - \hat{\beta}_0 = (\hat{\boldsymbol{\beta}} - \mathbf{A}^\top \hat{\boldsymbol{\alpha}})^\top \bar{\mathbf{x}}$.

\paragraph{Proof.}
By the first-order conditions of OLS, the intercept always centers the model at the sample means:
\begin{align}
\hat{\beta}_0 &= \bar{y} - \hat{\boldsymbol{\beta}}^\top \bar{\mathbf{x}} \\
\hat{\alpha}_0 &= \bar{y} - \hat{\boldsymbol{\alpha}}^\top \bar{\mathbf{z}} = \bar{y} - \hat{\boldsymbol{\alpha}}^\top \mathbf{A}\bar{\mathbf{x}}
\end{align}
Subtracting the two yields:
\begin{align}
\hat{\alpha}_0 - \hat{\beta}_0 &= (\bar{y} - \hat{\boldsymbol{\alpha}}^\top \mathbf{A}\bar{\mathbf{x}}) - (\bar{y} - \hat{\boldsymbol{\beta}}^\top \bar{\mathbf{x}}) \\
&= \hat{\boldsymbol{\beta}}^\top \bar{\mathbf{x}} - \hat{\boldsymbol{\alpha}}^\top \mathbf{A}\bar{\mathbf{x}} \\
&= (\hat{\boldsymbol{\beta}} - \mathbf{A}^\top \hat{\boldsymbol{\alpha}})^\top \bar{\mathbf{x}}
\end{align}
The proof demonstrates that $\hat{\alpha}_0 = \hat{\beta}_0$ if and only if the discrepancy $\boldsymbol{\delta}$ is orthogonal to the mean vector $\bar{\mathbf{x}}$, or if $\bar{\mathbf{x}} = \mathbf{0}$. In all other cases, the "reduction" introduces a bias in the intercept relative to the full-dimensional model.

\section{Subsampling Strategy}

For the purpose of subsampling such that the concentration phenomena caused by the projection is reduced, we define an extended feature mapping as:
\begin{equation}
\hat{\mathbf{t}}_{l,m,n,k,o}
:=
\hat{\phi}\!\left(
\mathbf{A}_k,\,
\mathbf{X}_l,\,
f_{m,n}(\mathbf{X}_l),\,
\Pi_{s,o}
\right)
\in \mathbb{R}^p,
\label{eq:meta_sampling_ELA_mapping}
\end{equation}
where \(\mathbf{A}_k \in \mathbb{R}^{d \times D}\) denotes a random linear embedding
operator, and \(\Pi_{s,o}(\cdot)\) a subsampling or permutation operator.
Each realization $\hat{\mathbf{t}}_{l,m,n,k,o}$ represents a perturbed estimate of the reference feature vector $\mathbf{t}_{l,m,n,\star}$. Varying the embedding index $k$ induces a probability distribution over the ELA feature values, with randomness arising from the embedding operator $\mathbf{A}_k$ and, when applicable, the permutation operator $\Pi_{s,o}$.

Subsampling was performed by drawing samples from the reduced space $\mathbf{z}_l$, with the sample size chosen to preserve the same proportion as in the ambient space. Specifically, if the global sample size was $100D$, the corresponding subsampled dataset consisted of $100d$ points. This subsampling procedure was repeated for 30 rounds, and the ELA features were estimated independently for each subsampled dataset.

This experimental design aims to assess both the distortion and stability of the extracted features, thereby quantifying the impact of the squashing effect induced by the projection operator by computing the feature with even less data than in ambient space.

\section{Additional Heatmaps}
\label{sec:add_heatmaps}

\subsection{One-shot Distributions}
\label{subsec:one_shot_distr}
\begin{figure}[hbtp]
    \centering
    \includegraphics[width=1\linewidth]{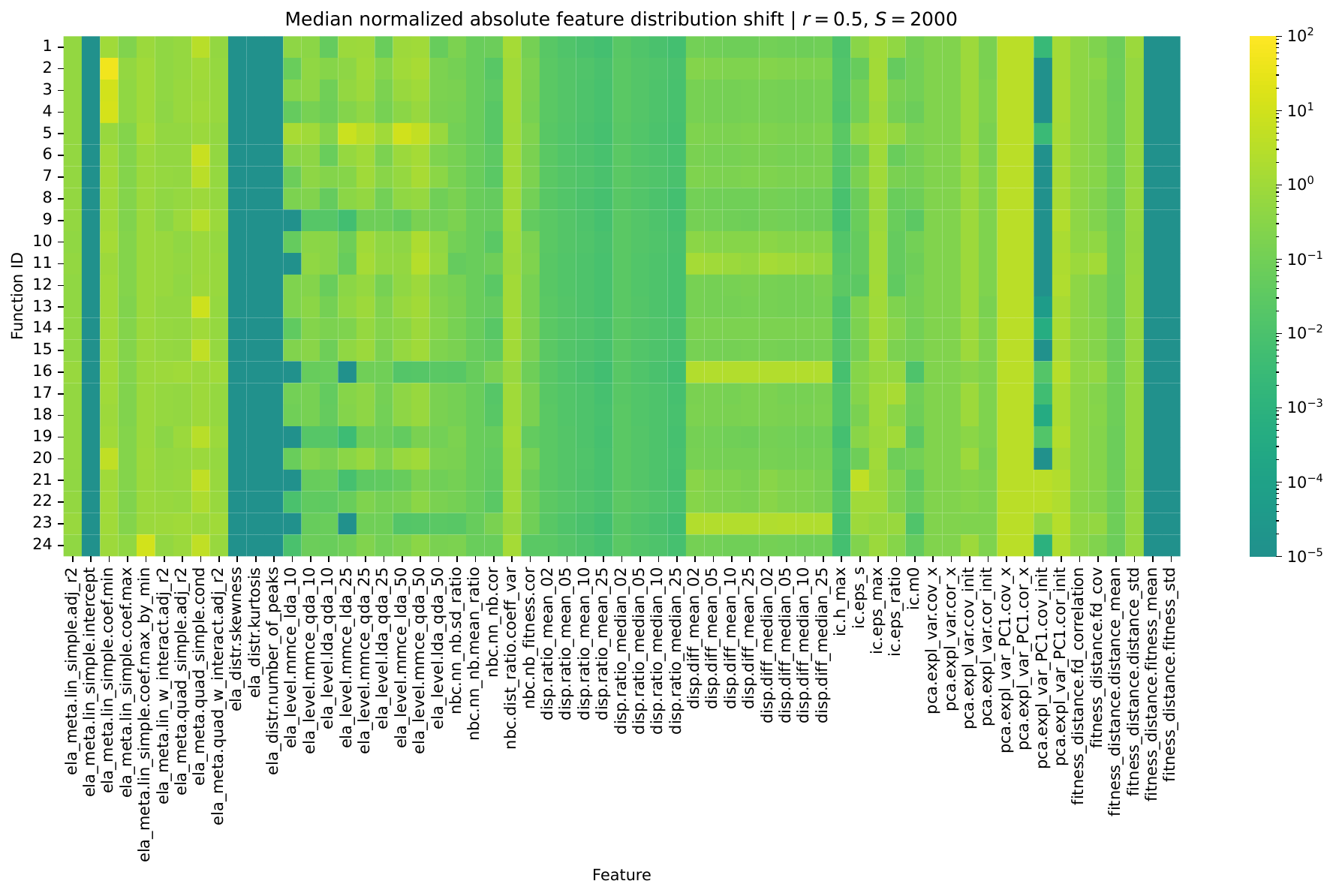}
    \caption{Heatmap of the median normalized absolute feature distribution shift for \textbf{datasets of size 2000} with a \textbf{compression ratio of 0.5}. Results shown for all BBOB functions.}
    \label{fig:heatmap_median_oneshot_r0.5_n2000}
\end{figure}
\begin{figure}[hbtp]\ContinuedFloat
    \centering
    \includegraphics[width=1\linewidth]{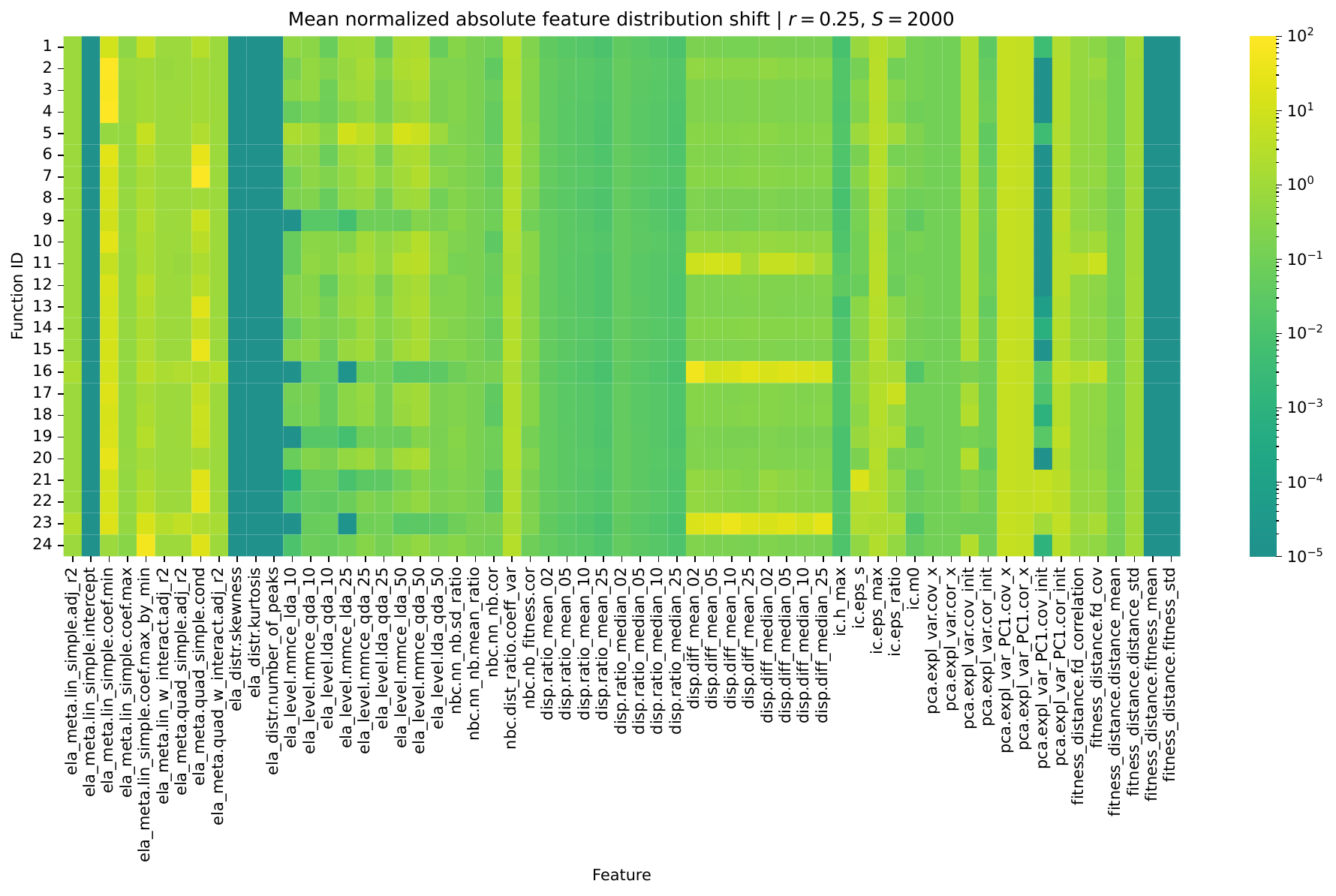}
    \caption{Heatmap of the median normalized absolute feature distribution shift for \textbf{datasets of size 2000} with a \textbf{compression ratio of 0.25}. Results shown for all BBOB functions.}
    \label{fig:heatmap_mean_oneshot_r0.25_n2000}
\end{figure}
\begin{figure}[hbtp]\ContinuedFloat
    \centering
    \includegraphics[width=1\linewidth]{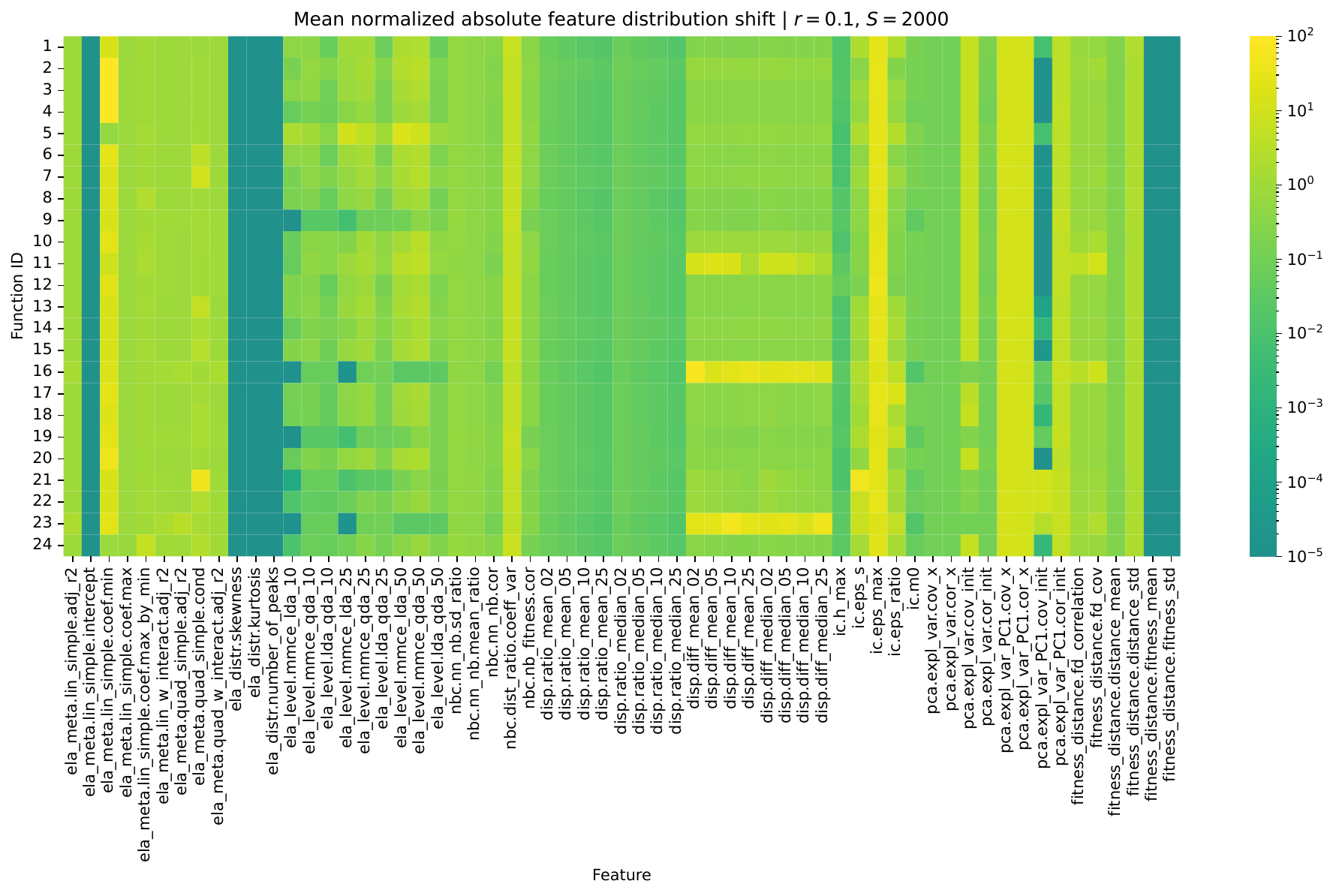}
    \caption{Heatmap of the median normalized absolute feature distribution shift for \textbf{datasets of size 2000} with a \textbf{compression ratio of 0.1}. Results shown for all BBOB functions.}
    \label{fig:heatmap_mean_oneshot_r0.25_n2000}
\end{figure}
\begin{figure}[hbtp]\ContinuedFloat
    \centering
    \includegraphics[width=1\linewidth]{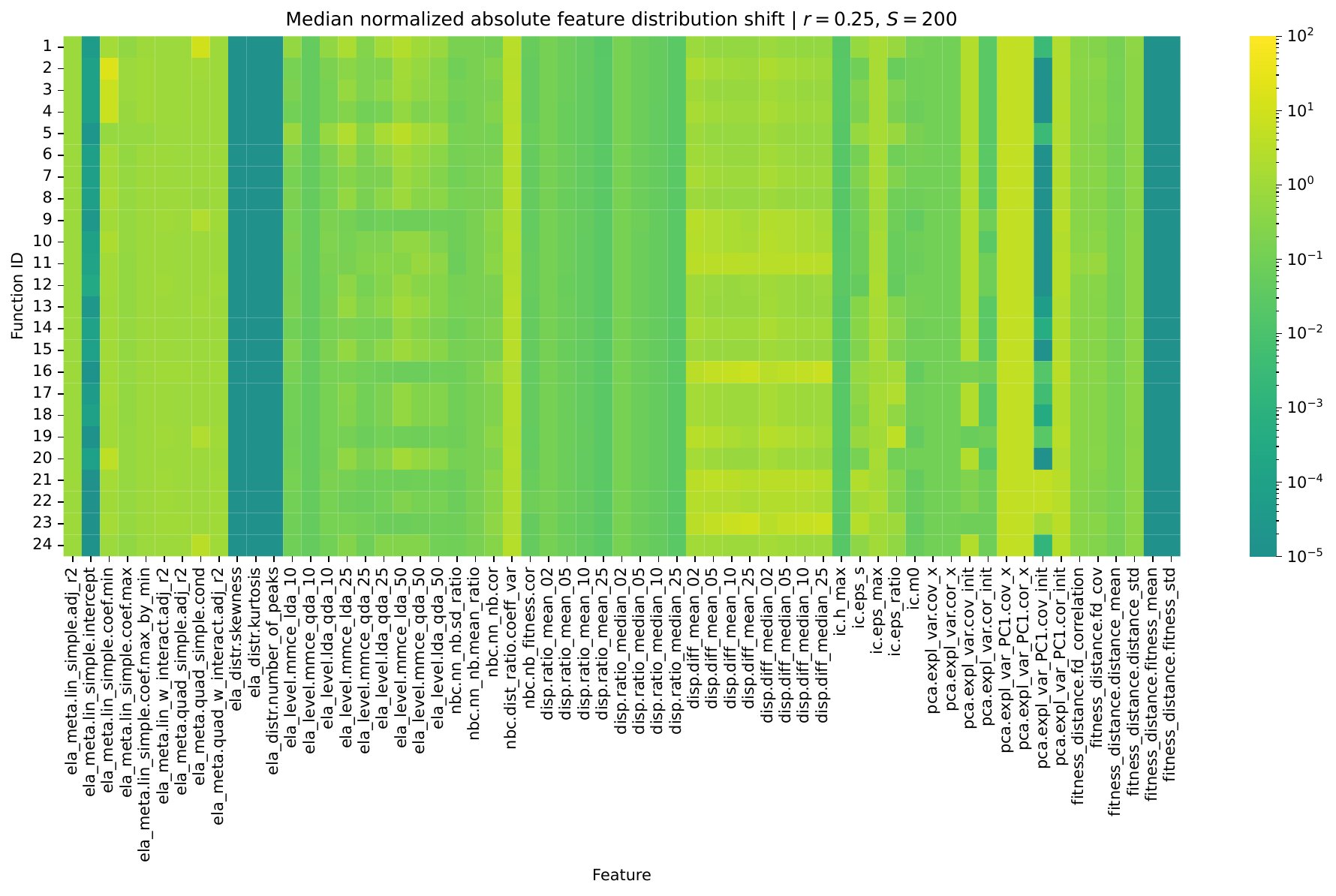}
    \caption{Heatmap of the median normalized absolute feature distribution shift for \textbf{datasets of size 200} with a \textbf{compression ratio of 0.25}. Results shown for all BBOB functions.}
    \label{fig:heatmap_median_oneshot_r0.25_n200}
\end{figure}

\begin{figure}[hbtp]\ContinuedFloat
    \centering
    \includegraphics[width=1\linewidth]{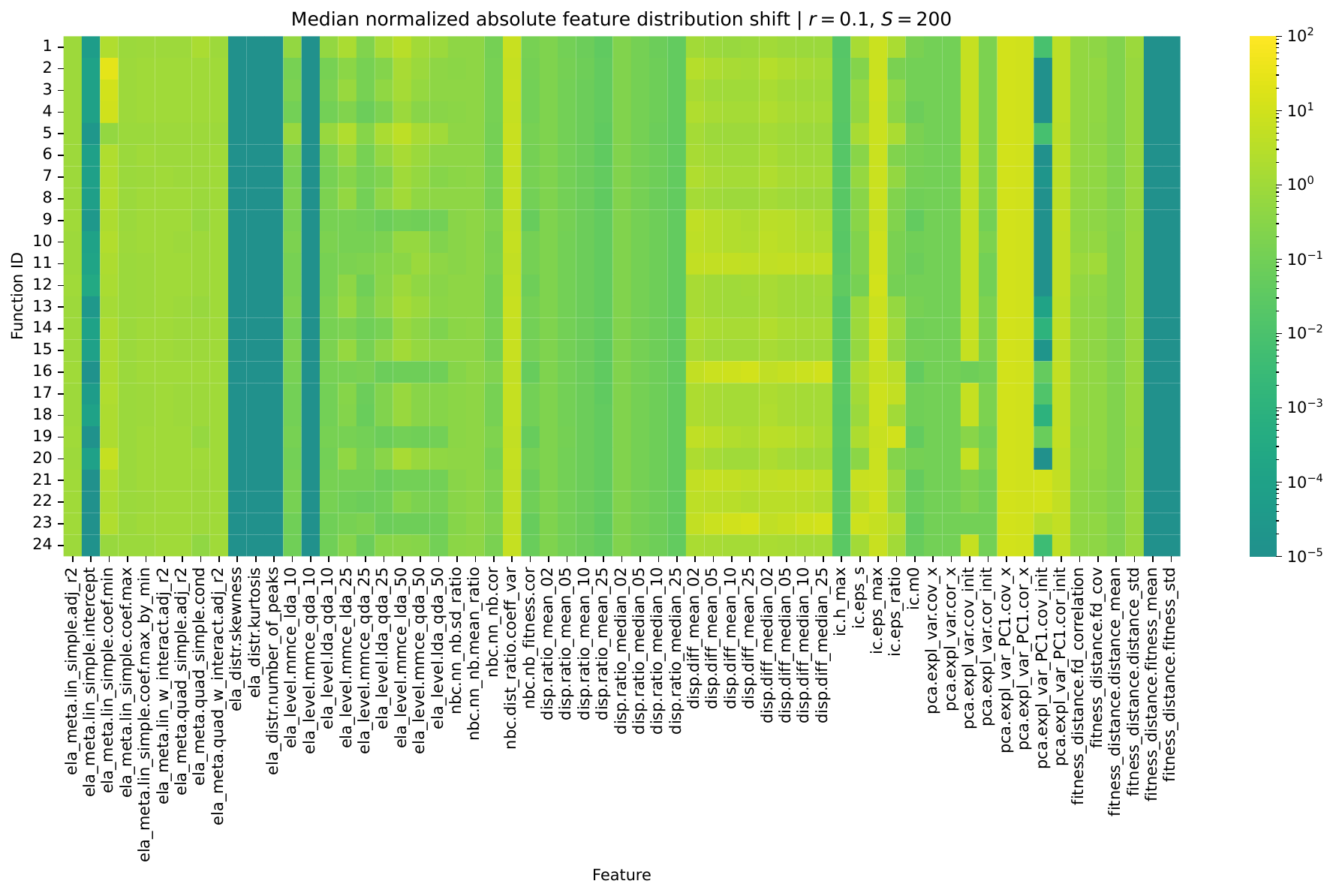}
    \caption{Heatmap of the median normalized absolute feature distribution shift for \textbf{datasets of size 200} with a \textbf{compression ratio of 0.1}. Results shown for all BBOB functions.}
    \label{fig:heatmap_median_oneshot_r0.25_n200}
\end{figure}

\clearpage
\subsection{Distributions with Resampling}
\label{subsec:resampling_distr}

\begin{figure}[hbtp]
    \centering
    \includegraphics[width=1\linewidth]{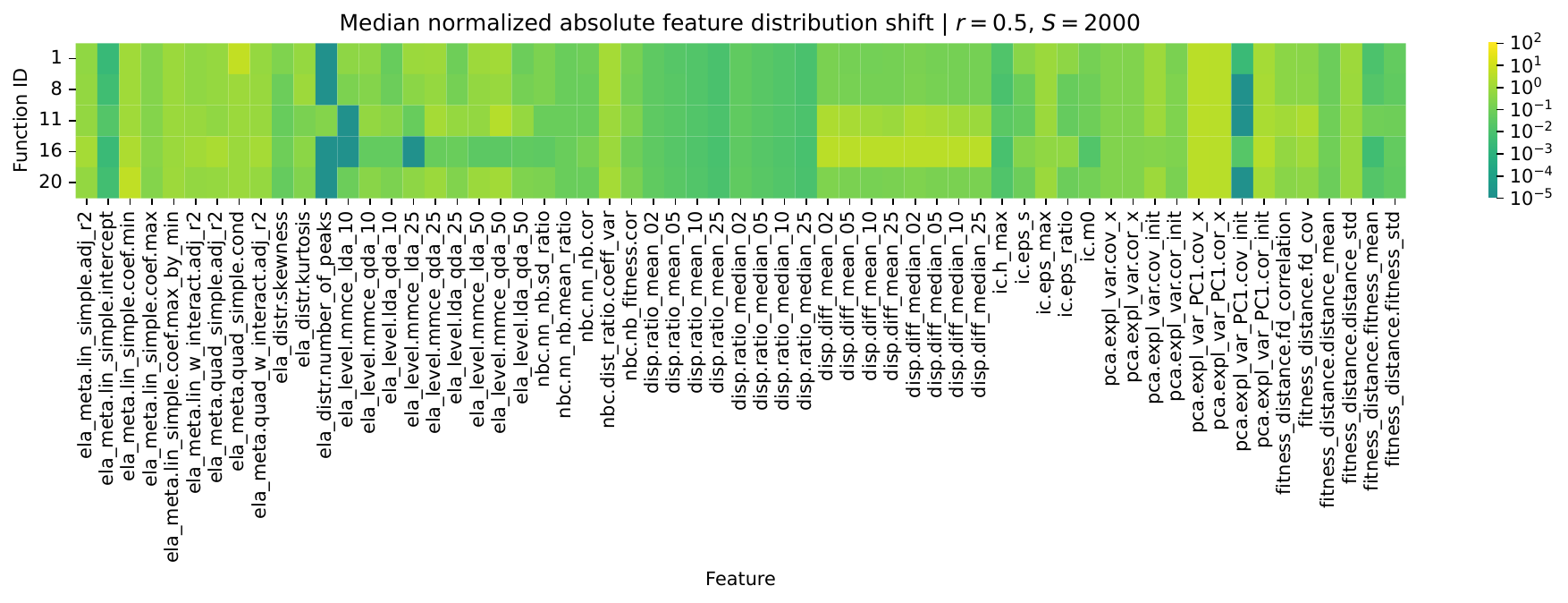}
    \caption{Heatmap of the median normalized absolute feature distribution shift for \textbf{datasets of size 2000} with a \textbf{compression ratio of 0.5} with \textbf{resampling}. Results shown for all BBOB function category representatives: f1, f8, f11, f16, and f20.}
    \label{fig:heatmap_median_standard_r0.5_n2000}
\end{figure}

\begin{figure}[hbtp]\ContinuedFloat
    \centering
    \includegraphics[width=1\linewidth]{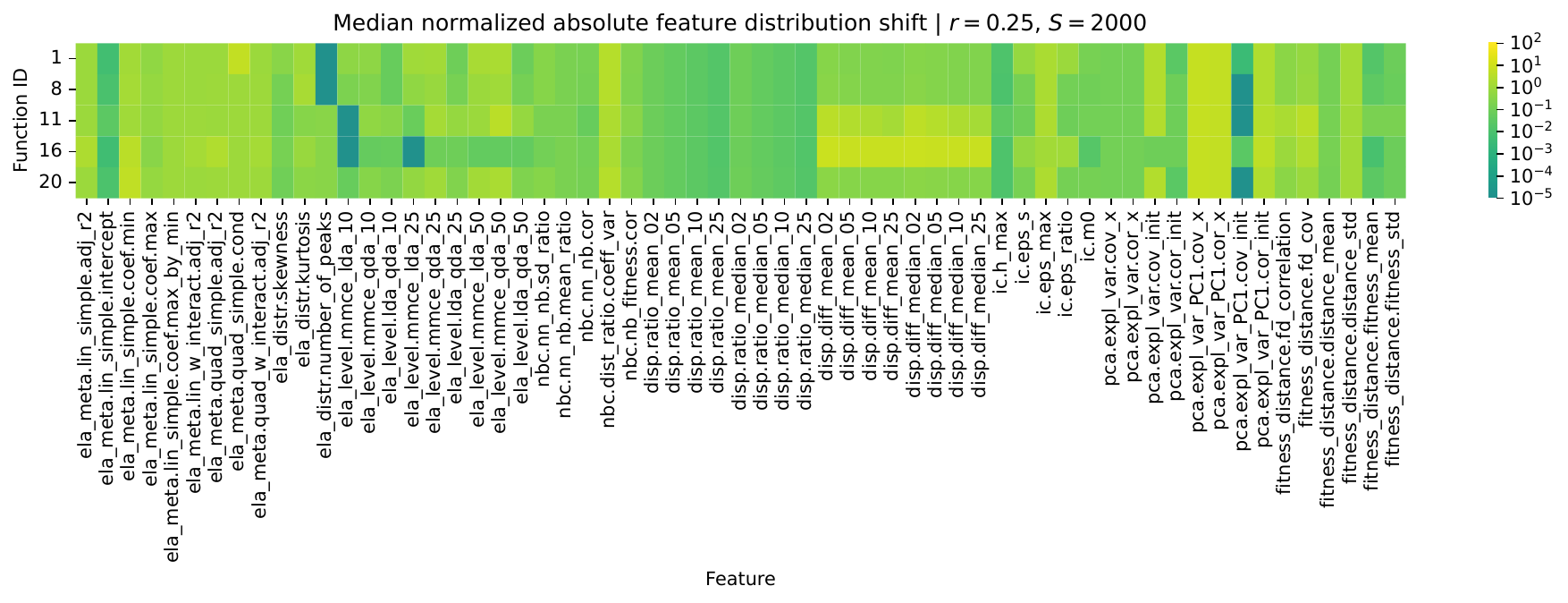}
     \caption{Heatmap of the median normalized absolute feature distribution shift for \textbf{datasets of size 2000} with a \textbf{compression ratio of 0.25} with \textbf{resampling}. Results shown for all BBOB function category representatives: f1, f8, f11, f16, and f20.}
    \label{fig:heatmap_median_standard_r0.25_n2000}
\end{figure}

\begin{figure}[hbtp]
    \centering
    \includegraphics[width=1\linewidth]{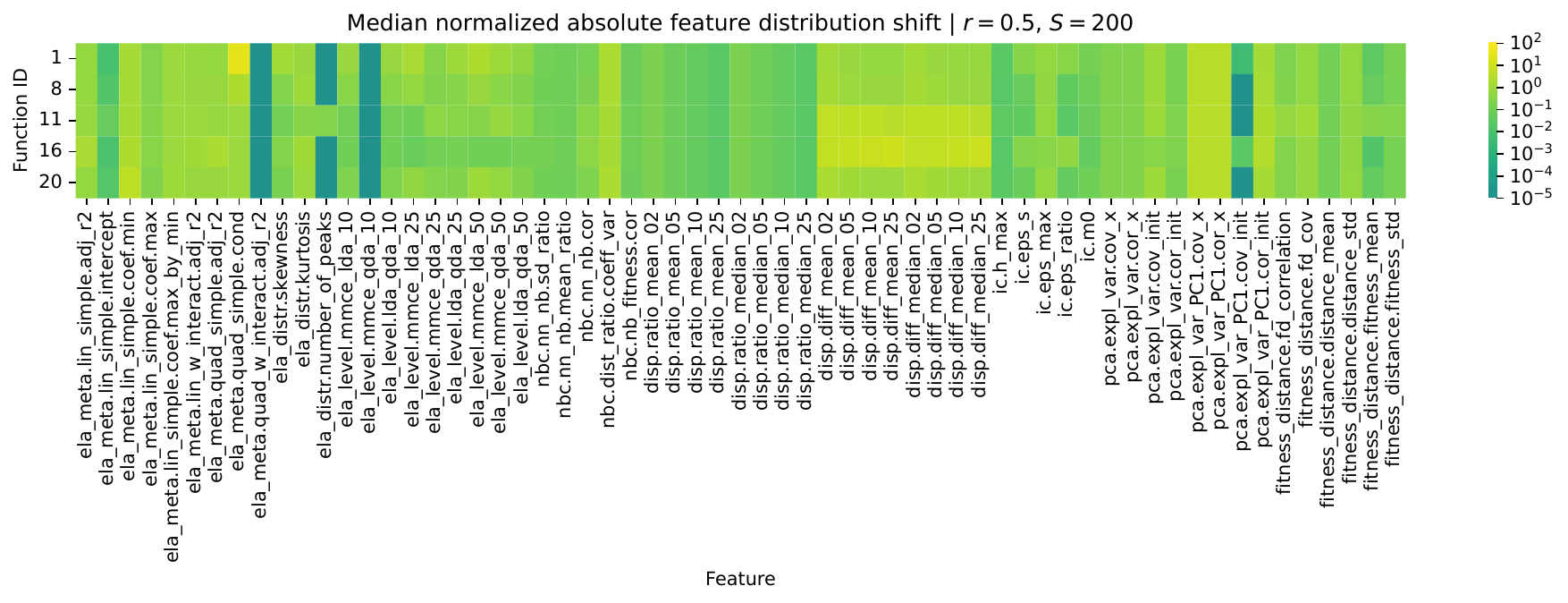}
    \caption{Heatmap of the median normalized absolute feature distribution shift for \textbf{datasets of size 200} with a \textbf{compression ratio of 0.5} with \textbf{resampling}. Results shown for all BBOB function category representatives: f1, f8, f11, f16, and f20.}
    \label{fig:heatmap_median_standard_r0.5_n200}
\end{figure}

\clearpage
\section{Violin Plots of Normalized feature distribution shift per function}

\subsection{One-shot Distributions}

\begin{figure}[hbtp]
    \centering
    \includegraphics[width=.8\linewidth,trim=0cm 7.5cm 0cm 0cm,clip]{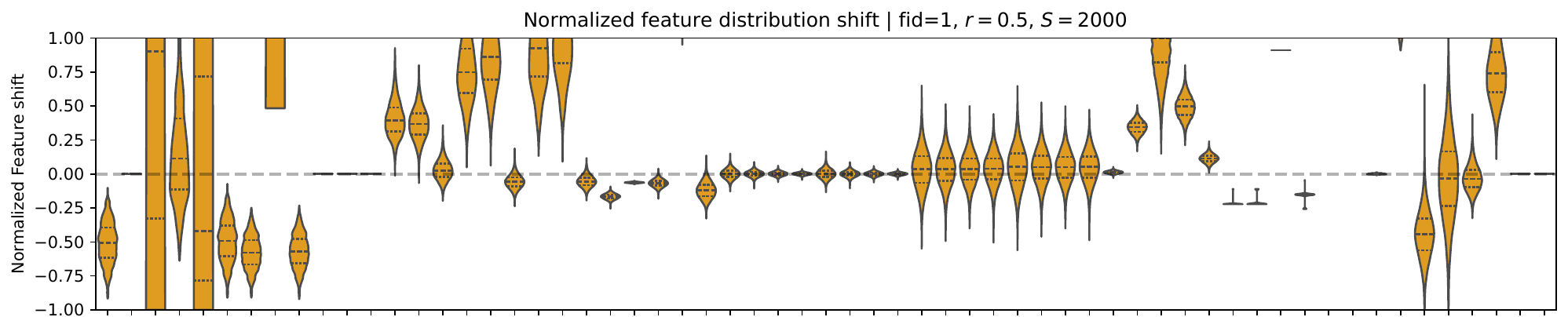}
    \includegraphics[width=.8\linewidth,trim=0cm 7.5cm 0cm 0cm,clip]{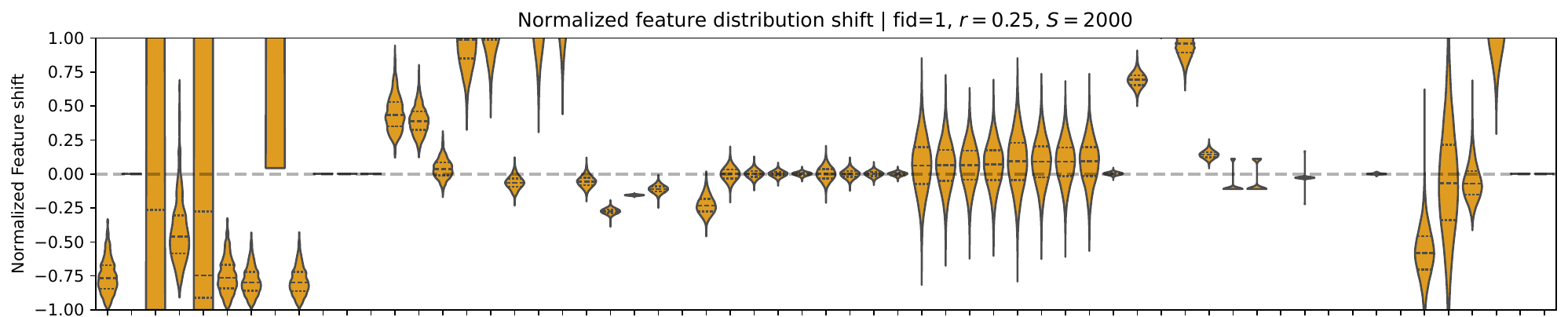}
    \includegraphics[width=.8\linewidth,trim=0cm .7cm 0cm 0cm,clip]{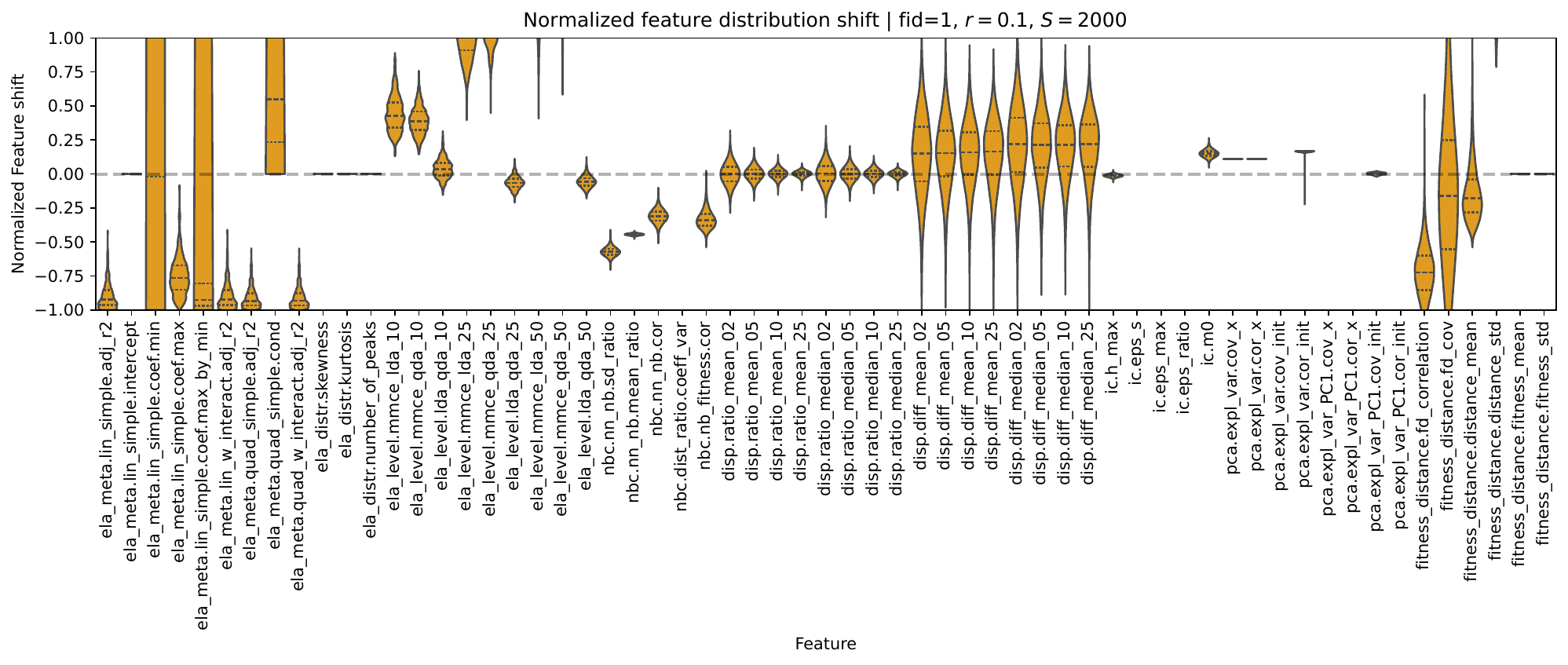}
    \caption{
    Normalized aggregated feature distribution shift of \textbf{Sphere (f1) function} with $\boldsymbol{S=2000}$ for compression ratios $r=\{0.5,0.25,0.1\}$. The dashed line denotes a normalized reference corresponding to the median of each feature distribution in the original search space. To enhance visualization, the limits of the Normalized Feature shift has been set to $[-1,1 ]$.
    }
    \label{fig:violin_f1_n2000}
\end{figure}

\begin{figure}[hbtp]\ContinuedFloat
    \centering
    \includegraphics[width=.8\linewidth,trim=0cm 7.5cm 0cm 0cm,clip]{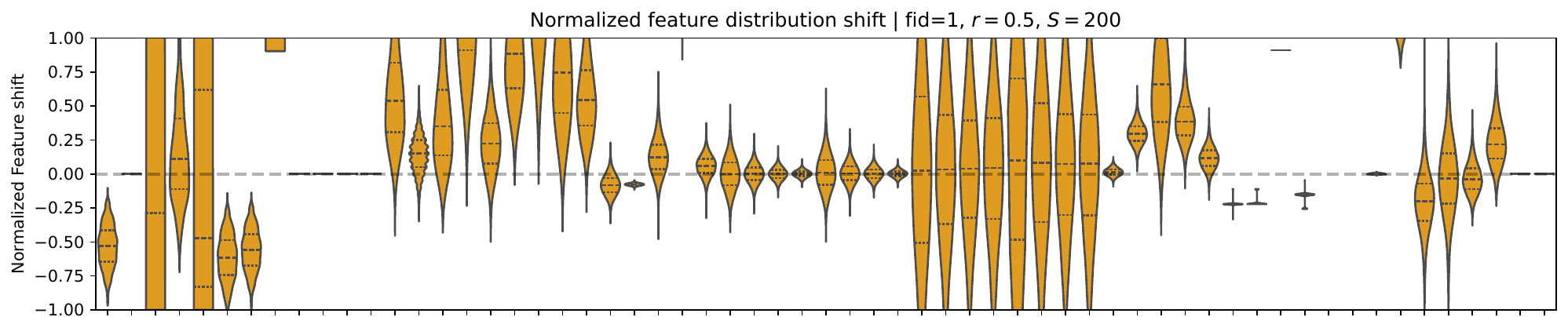}
    \includegraphics[width=.8\linewidth,trim=0cm 7.5cm 0cm 0cm,clip]{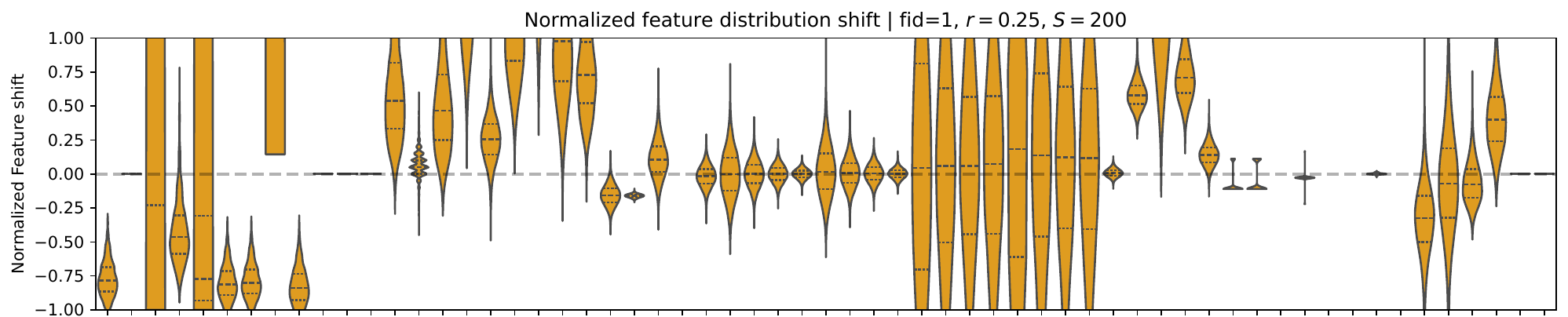}
    \includegraphics[width=.8\linewidth,trim=0cm .7cm 0cm 0cm,clip]{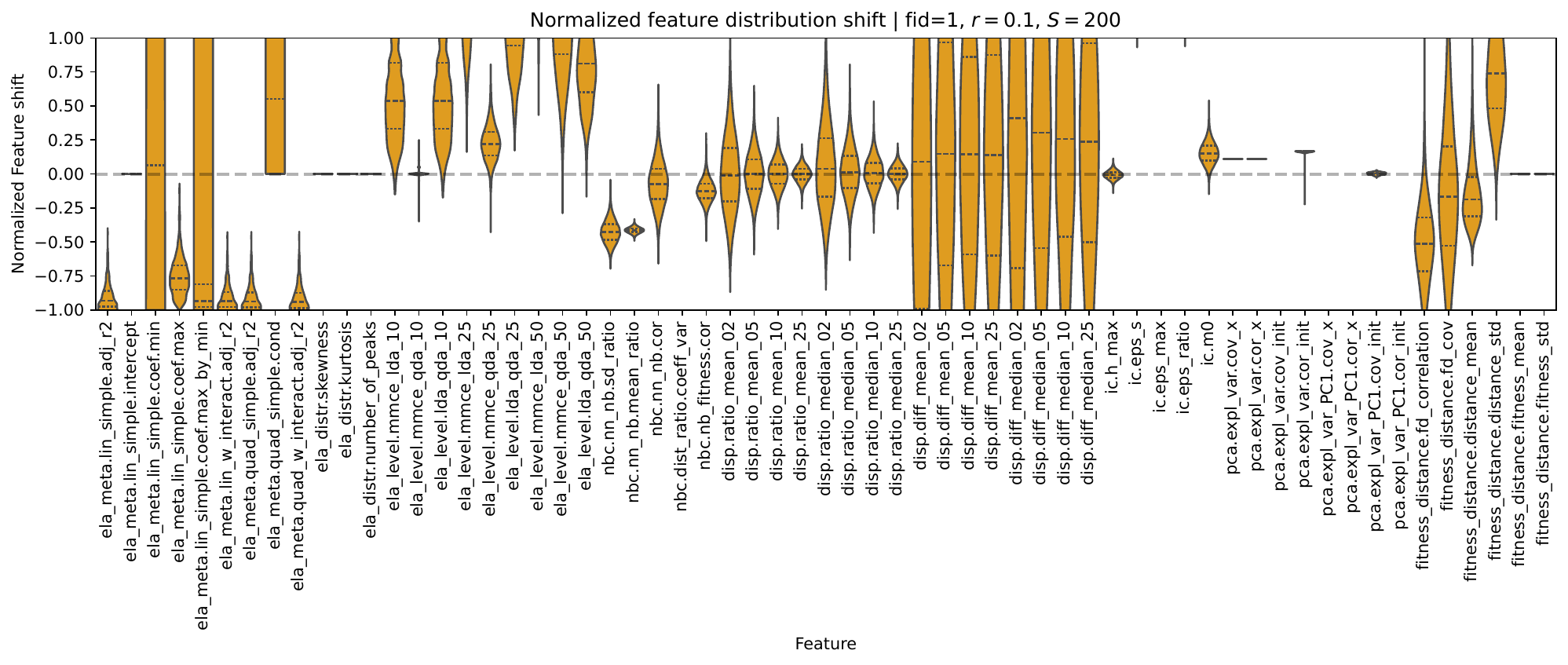}
  \caption{
  Same as above, for $\boldsymbol{S=200}$.
  }
    \label{fig:violin_f1_n200}
\end{figure}


\begin{figure}[hbtp]
    \centering
    \includegraphics[width=.8\linewidth,trim=0cm 7.5cm 0cm 0cm,clip]{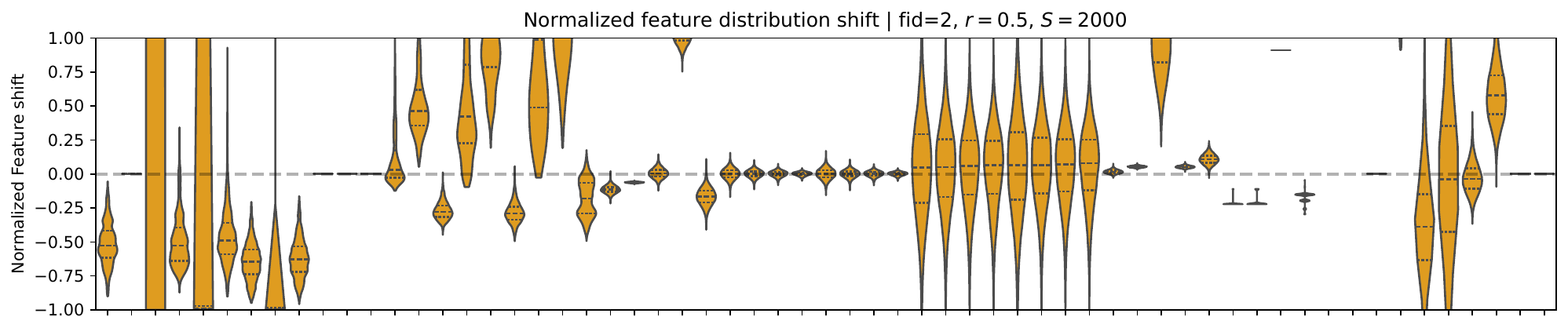}
    \includegraphics[width=.8\linewidth,trim=0cm 7.5cm 0cm 0cm,clip]{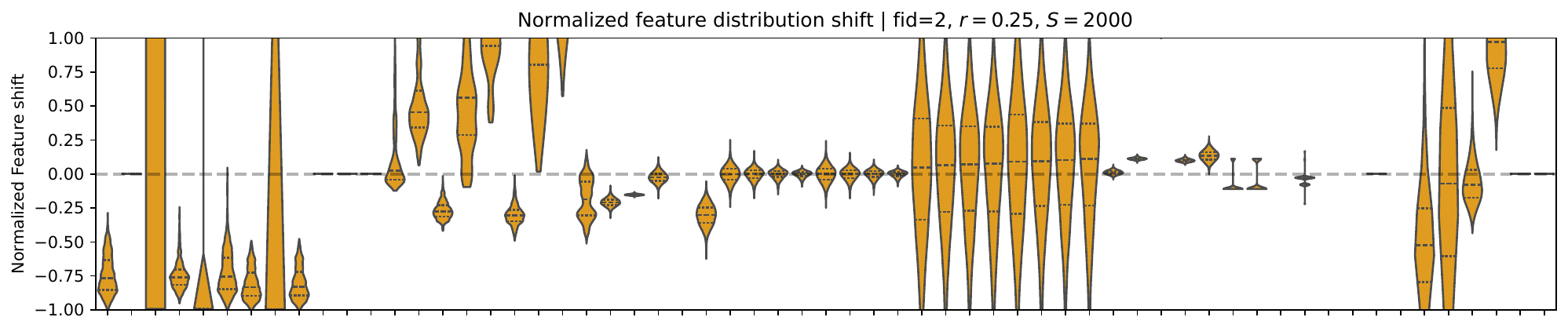}
    \includegraphics[width=.8\linewidth,trim=0cm .7cm 0cm 0cm,clip]{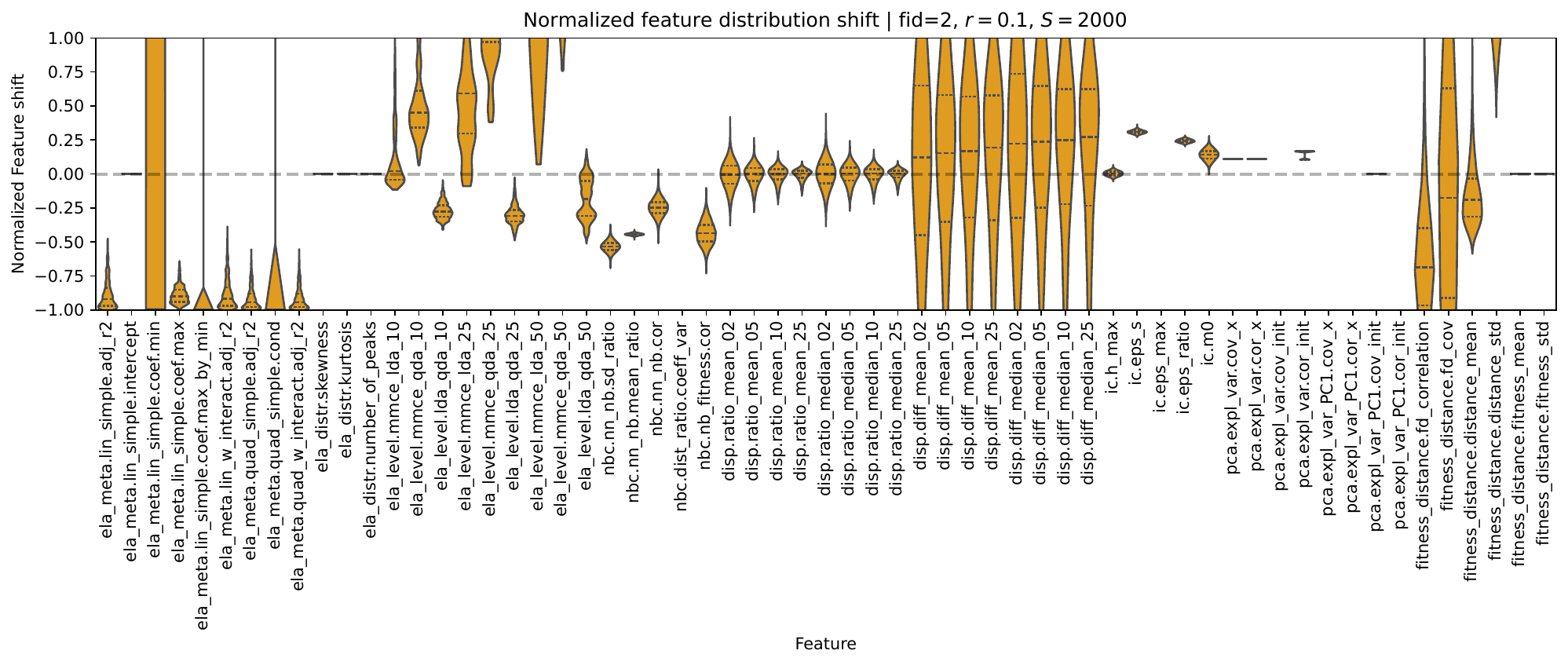}
    \caption{
    Normalized aggregated feature distribution shift of \textbf{Ellipsoid separable (f2) function} with $\boldsymbol{S=2000}$ for compression ratios $r=\{0.5,0.25,0.1\}$. The dashed line denotes a normalized reference corresponding to the median of each feature distribution in the original search space. To enhance visualization, the limits of the Normalized Feature shift has been set to $[-1,1 ]$.
    }
    \label{fig:violin_f2_n2000}
\end{figure}

\begin{figure}[hbtp]\ContinuedFloat
    \centering
    \includegraphics[width=.8\linewidth,trim=0cm 7.5cm 0cm 0cm,clip]{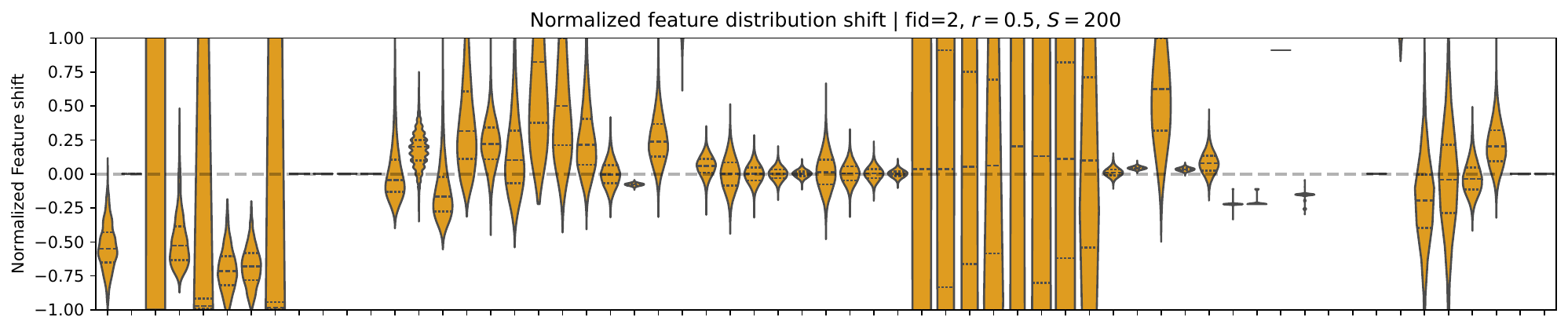}
    \includegraphics[width=.8\linewidth,trim=0cm 7.5cm 0cm 0cm,clip]{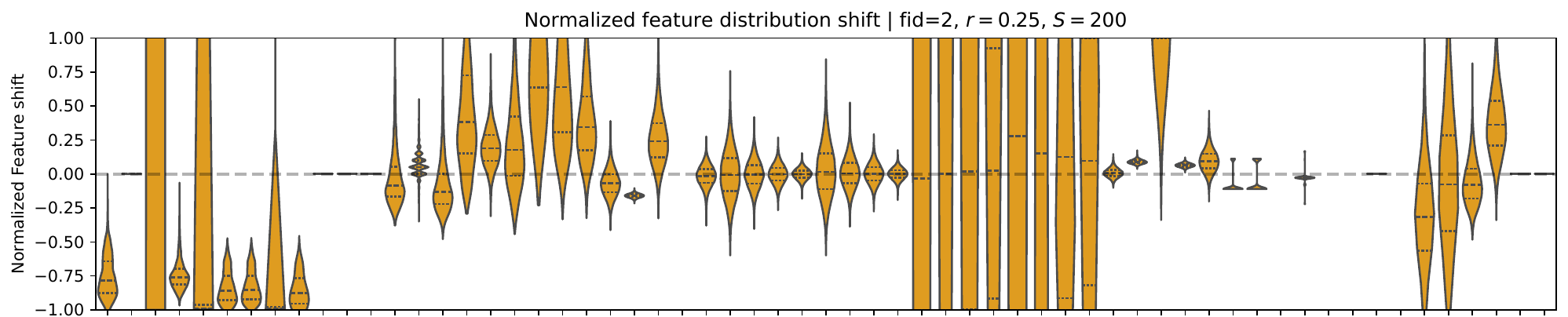}
    \includegraphics[width=.8\linewidth,trim=0cm .7cm 0cm 0cm,clip]{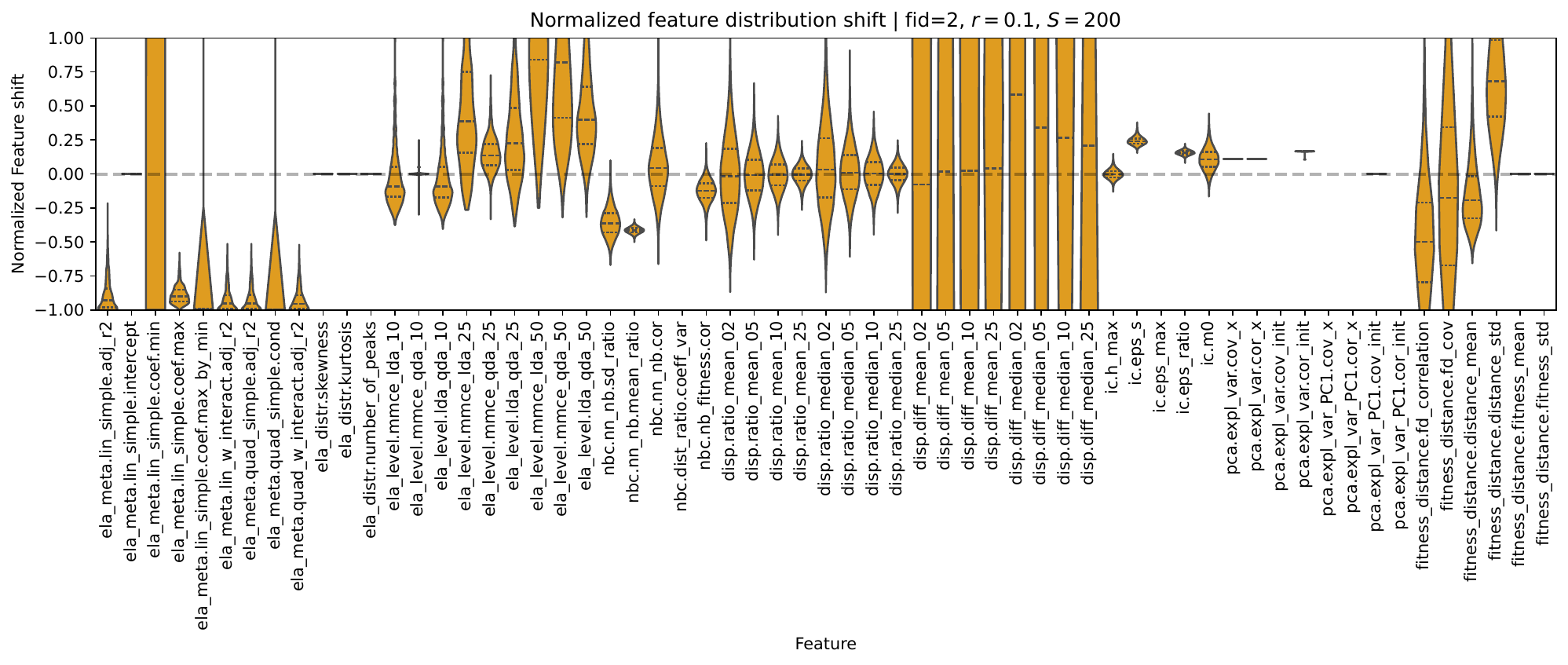}
  \caption{
  Same as above, for $\boldsymbol{S=200}$.
  }
    \label{fig:violin_f2_n200}
\end{figure}


\begin{figure}[hbtp]
    \centering
    \includegraphics[width=.8\linewidth,trim=0cm 7.5cm 0cm 0cm,clip]{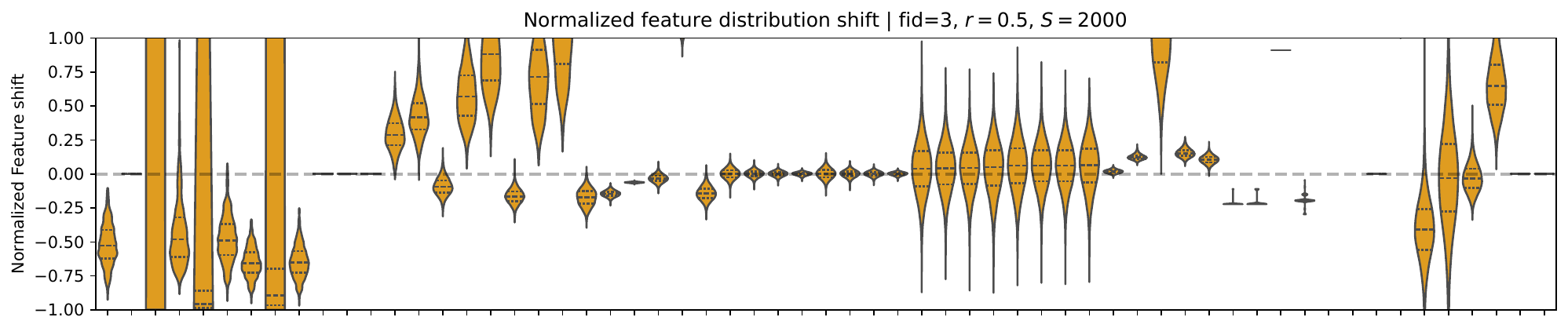}
    \includegraphics[width=.8\linewidth,trim=0cm 7.5cm 0cm 0cm,clip]{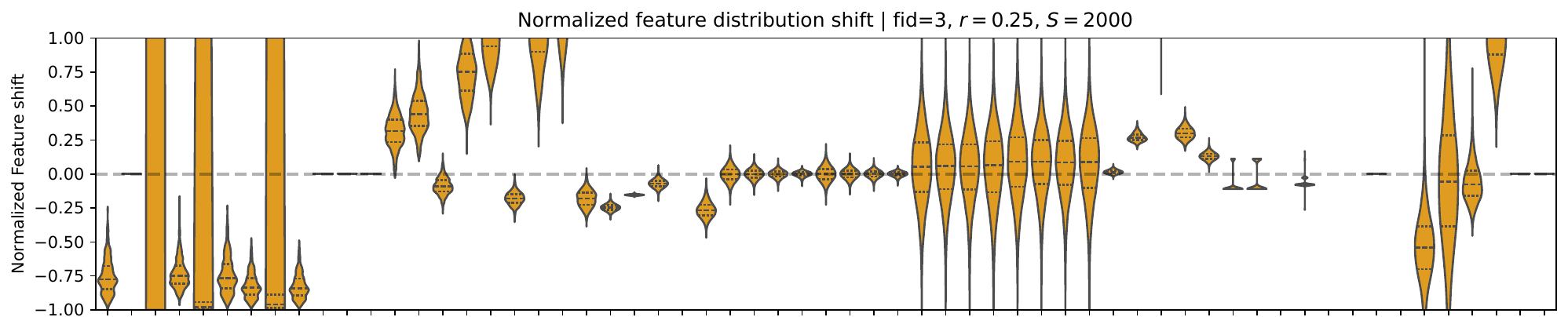}
    \includegraphics[width=.8\linewidth,trim=0cm .7cm 0cm 0cm,clip]{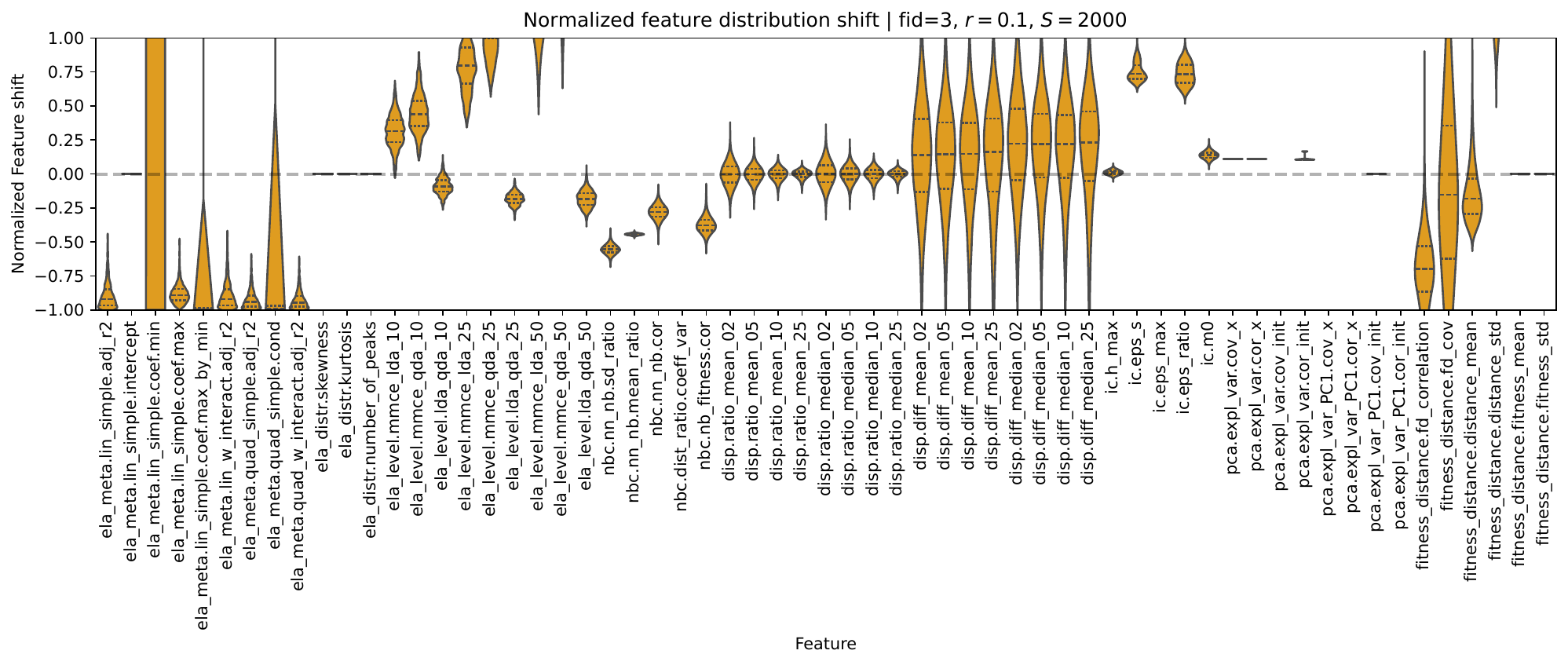}
    \caption{
    Normalized aggregated feature distribution shift of \textbf{Rastrigin separable (f3) function} with $\boldsymbol{S=2000}$ for compression ratios $r=\{0.5,0.25,0.1\}$. The dashed line denotes a normalized reference corresponding to the median of each feature distribution in the original search space. To enhance visualization, the limits of the Normalized Feature shift has been set to $[-1,1 ]$.
    }
    \label{fig:violin_f3_n2000}
\end{figure}

\begin{figure}[hbtp]\ContinuedFloat
    \centering
    \includegraphics[width=.8\linewidth,trim=0cm 7.5cm 0cm 0cm,clip]{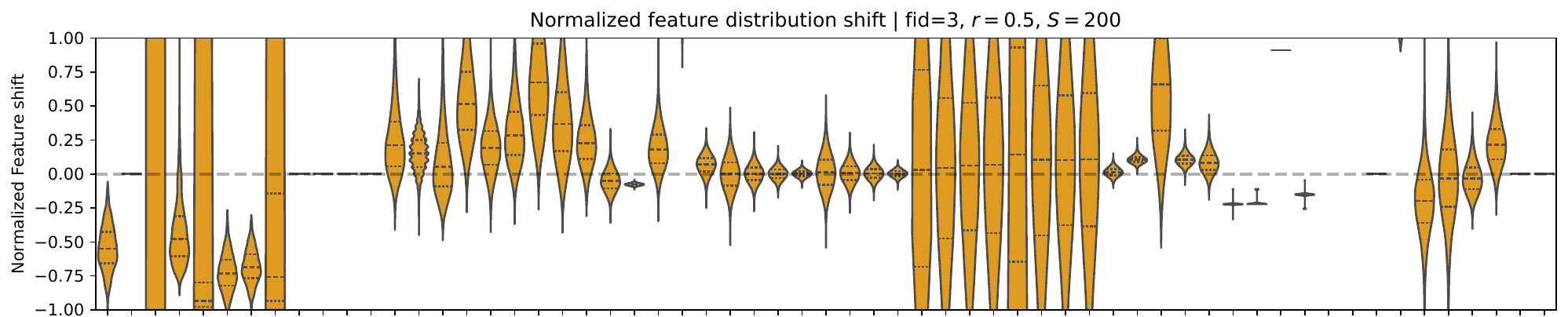}
    \includegraphics[width=.8\linewidth,trim=0cm 7.5cm 0cm 0cm,clip]{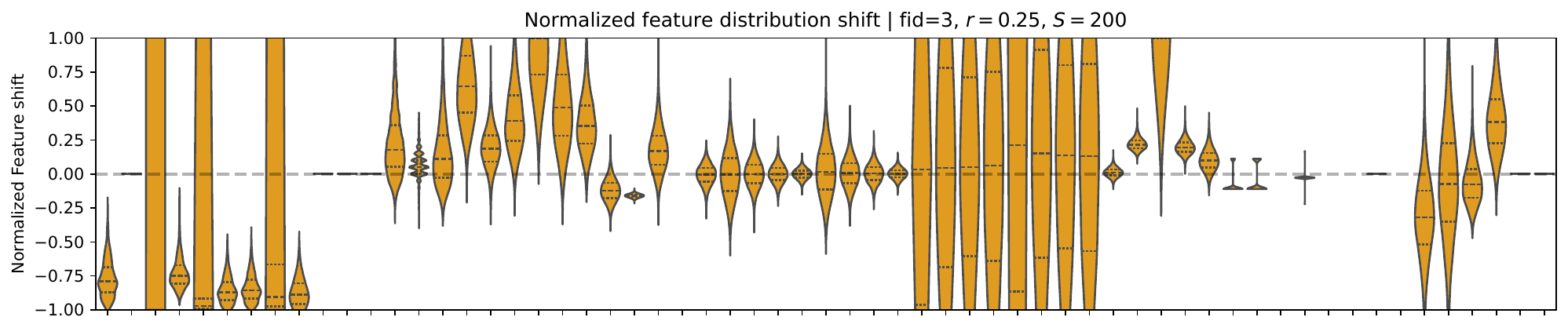}
    \includegraphics[width=.8\linewidth,trim=0cm .7cm 0cm 0cm,clip]{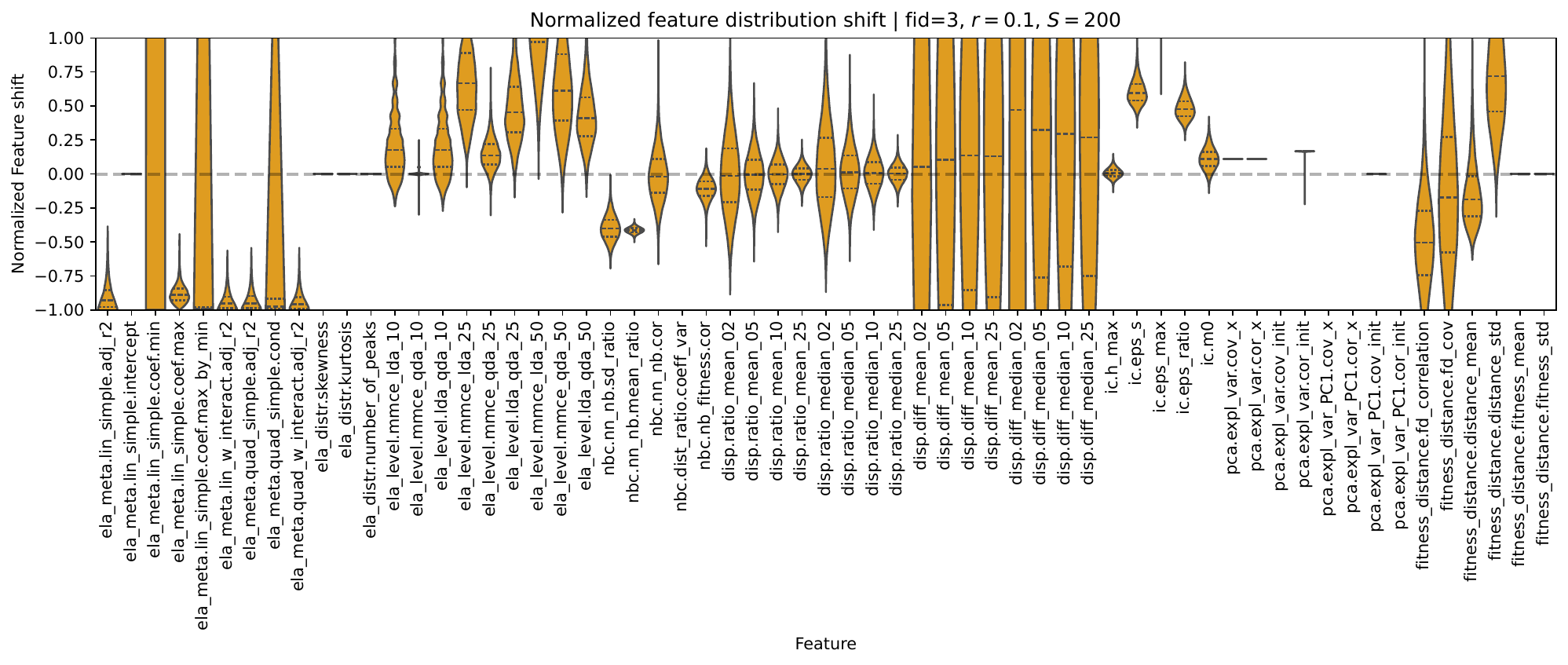}
  \caption{
  Same as above, for $\boldsymbol{S=200}$.
  }
    \label{fig:violin_f3_n200}
\end{figure}

\begin{figure}[hbtp]
    \centering
    \includegraphics[width=.8\linewidth,trim=0cm 7.5cm 0cm 0cm,clip]{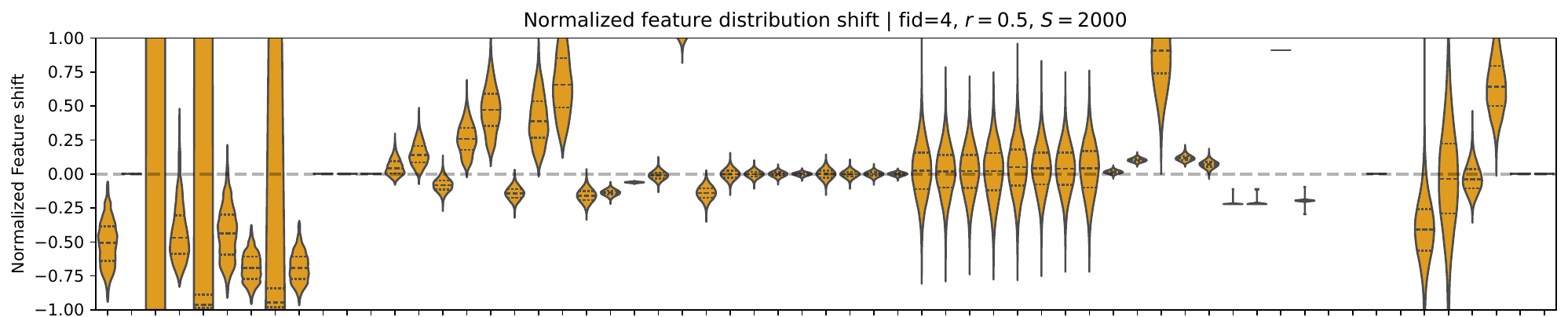}
    \includegraphics[width=.8\linewidth,trim=0cm 7.5cm 0cm 0cm,clip]{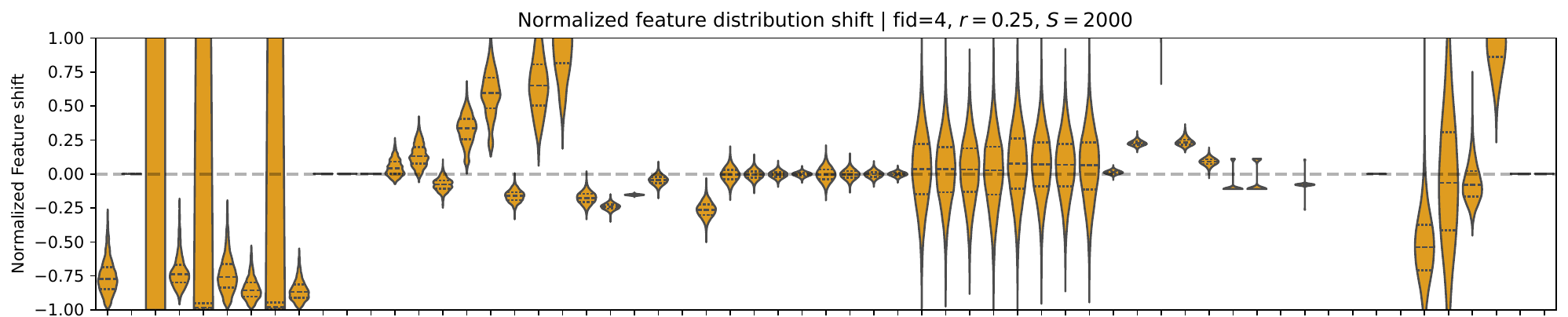}
    \includegraphics[width=.8\linewidth,trim=0cm .7cm 0cm 0cm,clip]{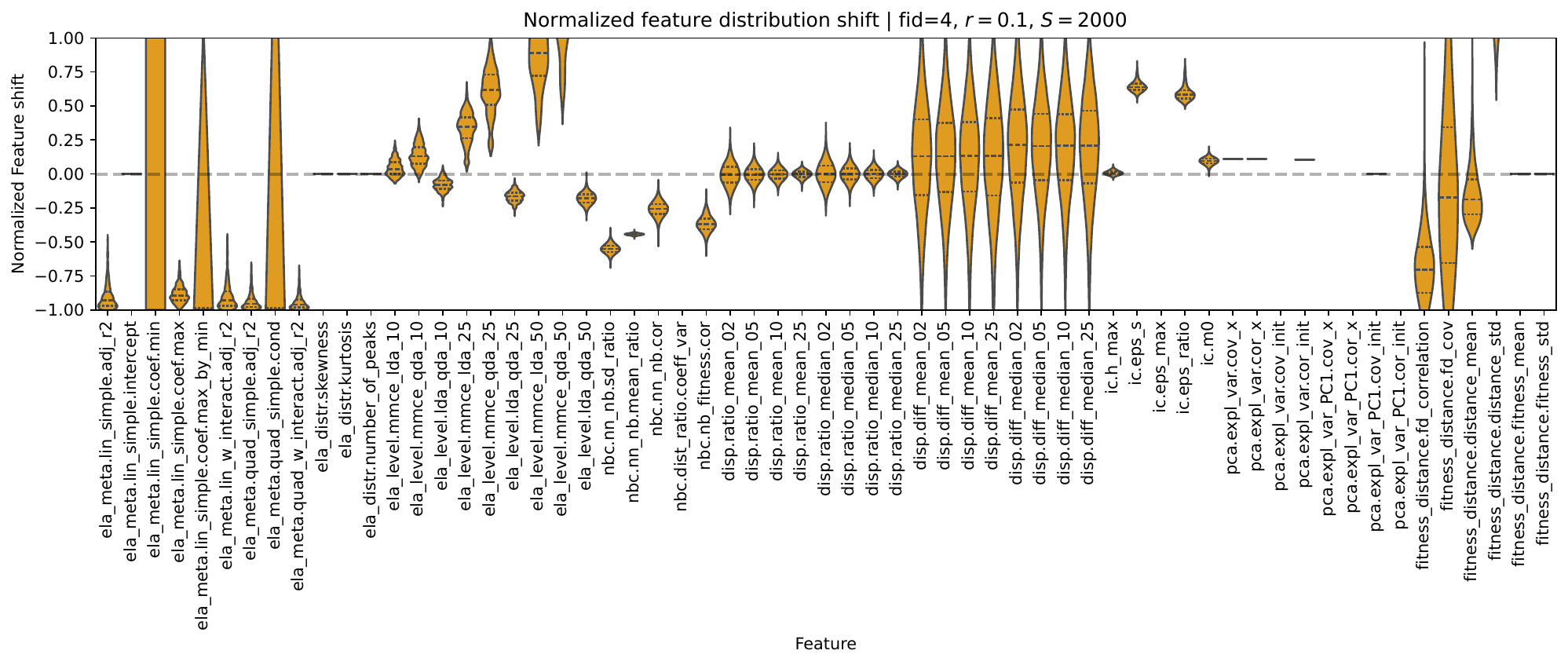}
    \caption{
    Normalized aggregated feature distribution shift of \textbf{Skew Rastrigin-Bueche (f4) function} with $\boldsymbol{S=2000}$ for compression ratios $r=\{0.5,0.25,0.1\}$. The dashed line denotes a normalized reference corresponding to the median of each feature distribution in the original search space. To enhance visualization, the limits of the Normalized Feature shift has been set to $[-1,1 ]$.
    }
    \label{fig:violin_f4_n2000}
\end{figure}

\begin{figure}[hbtp]\ContinuedFloat
    \centering
    \includegraphics[width=.8\linewidth,trim=0cm 7.5cm 0cm 0cm,clip]{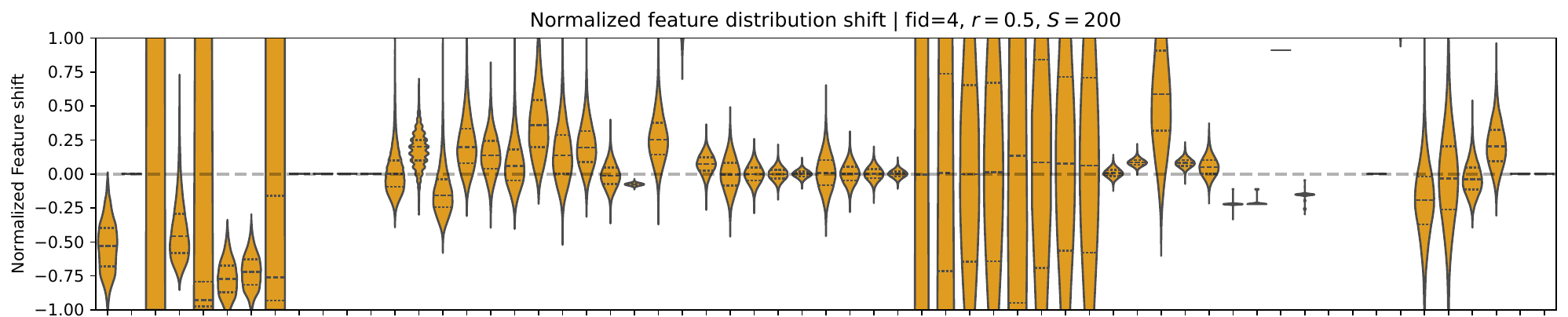}
    \includegraphics[width=.8\linewidth,trim=0cm 7.5cm 0cm 0cm,clip]{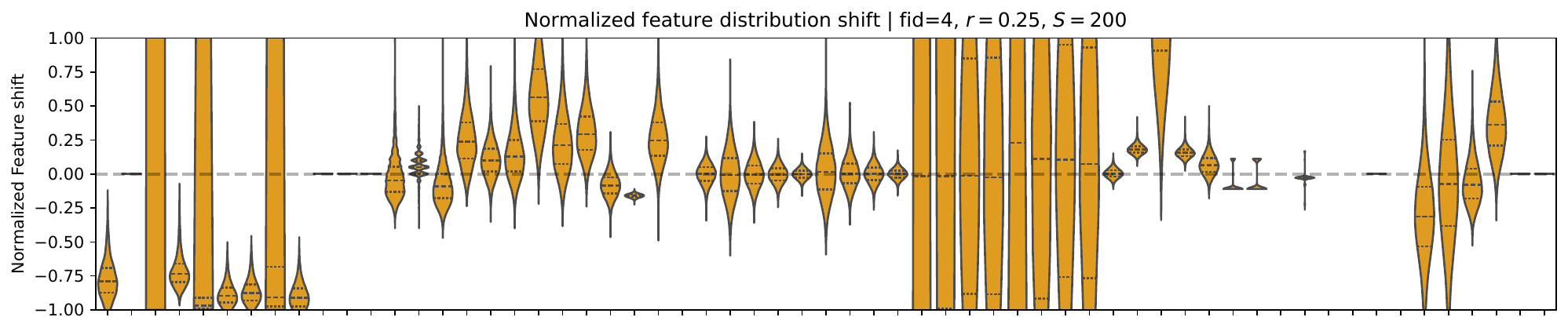}
    \includegraphics[width=.8\linewidth,trim=0cm .7cm 0cm 0cm,clip]{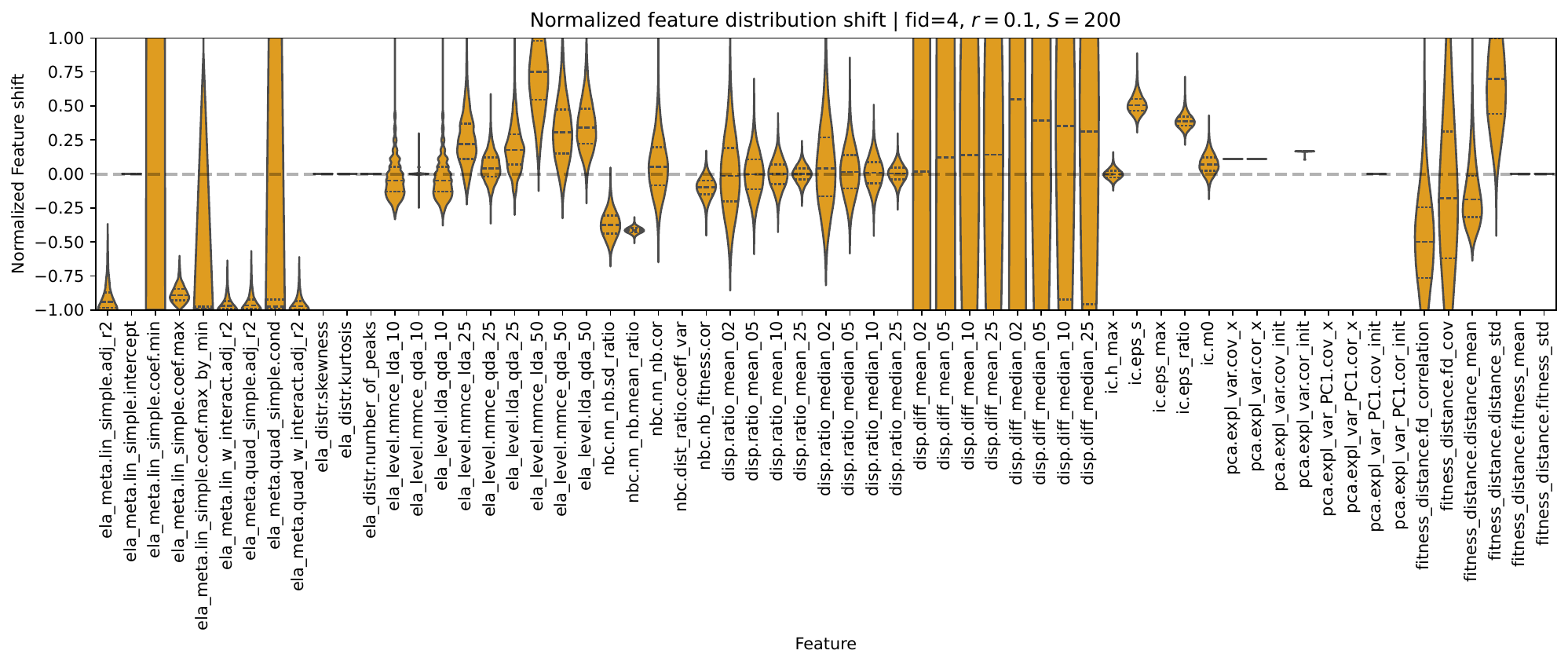}
  \caption{
  Same as above, for $\boldsymbol{S=200}$.
  }
    \label{fig:violin_f4_n200}
\end{figure}

\begin{figure}[hbtp]
    \centering
    \includegraphics[width=.8\linewidth,trim=0cm 7.5cm 0cm 0cm,clip]{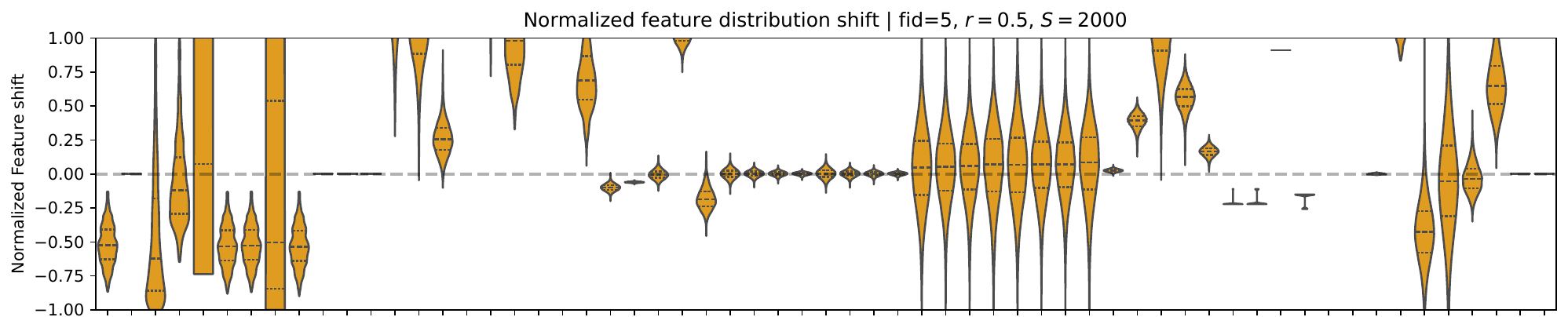}
    \includegraphics[width=.8\linewidth,trim=0cm 7.5cm 0cm 0cm,clip]{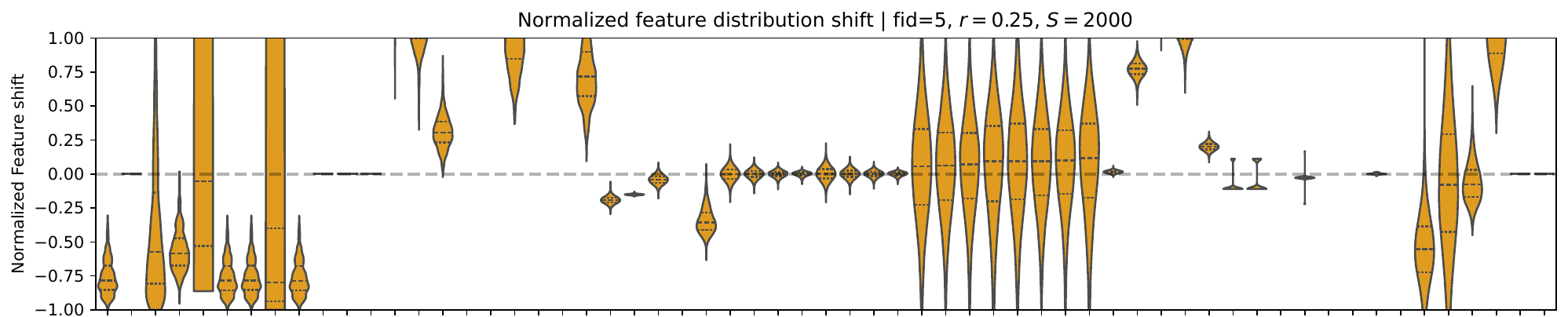}
    \includegraphics[width=.8\linewidth,trim=0cm .7cm 0cm 0cm,clip]{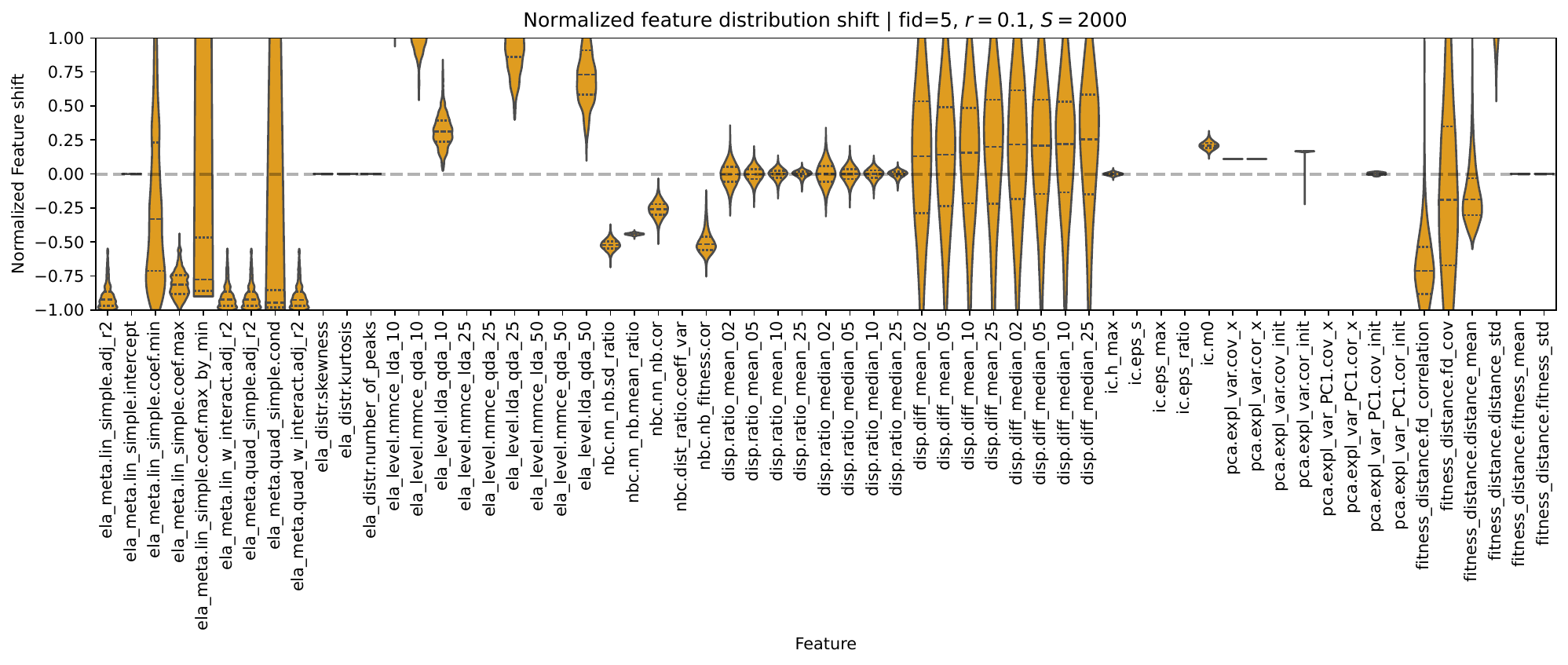}
    \caption{
    Normalized aggregated feature distribution shift of \textbf{Linear slope (f5) function} with $\boldsymbol{S=2000}$ for compression ratios $r=\{0.5,0.25,0.1\}$. The dashed line denotes a normalized reference corresponding to the median of each feature distribution in the original search space. To enhance visualization, the limits of the Normalized Feature shift has been set to $[-1,1 ]$.
    }
    \label{fig:violin_f5_n2000}
\end{figure}

\begin{figure}[hbtp]\ContinuedFloat
    \centering
    \includegraphics[width=.8\linewidth,trim=0cm 7.5cm 0cm 0cm,clip]{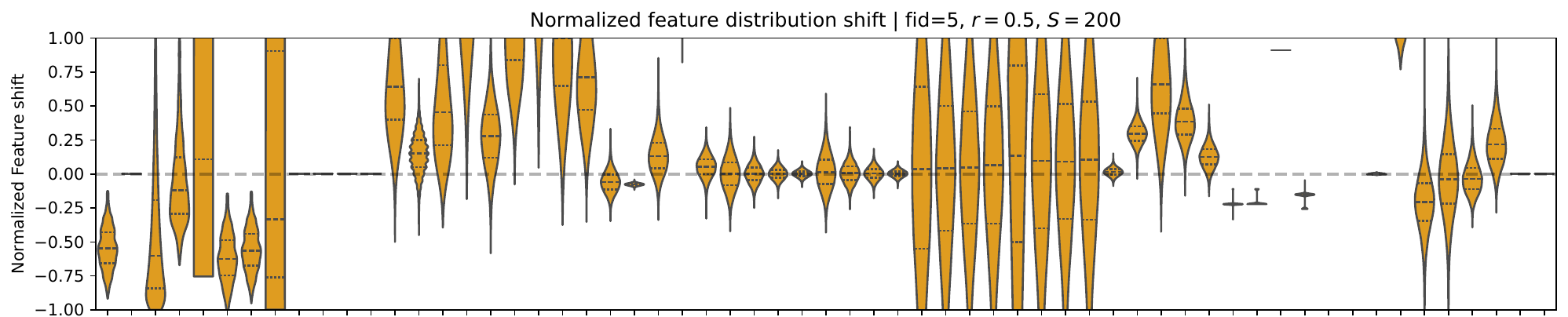}
    \includegraphics[width=.8\linewidth,trim=0cm 7.5cm 0cm 0cm,clip]{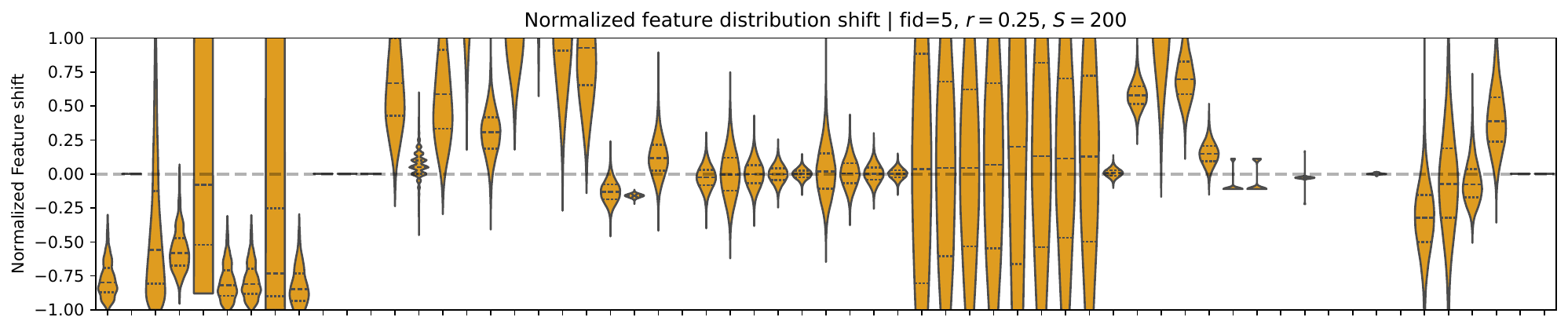}
    \includegraphics[width=.8\linewidth,trim=0cm .7cm 0cm 0cm,clip]{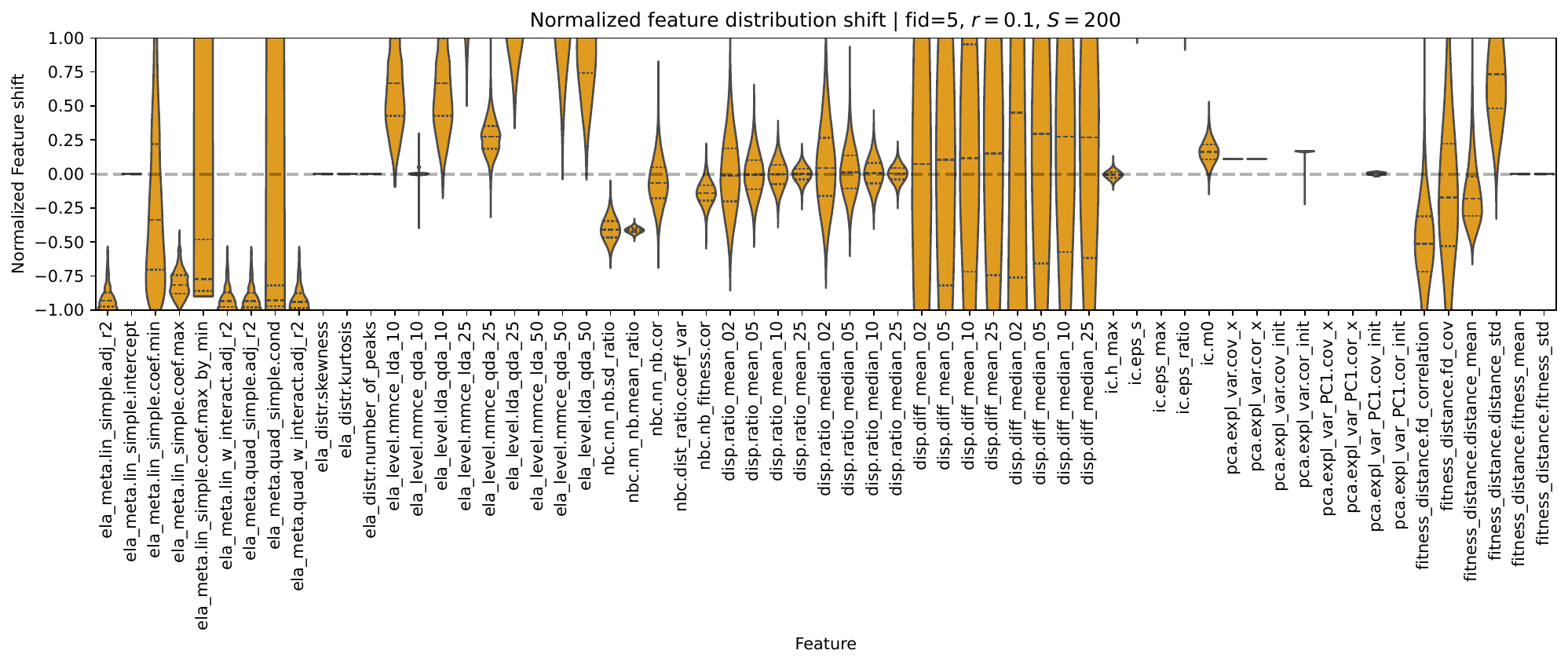}
  \caption{
  Same as above, for $\boldsymbol{S=200}$.
  }
    \label{fig:violin_f5_n200}
\end{figure}

\begin{figure}[hbtp]
    \centering
    \includegraphics[width=.8\linewidth,trim=0cm 7.5cm 0cm 0cm,clip]{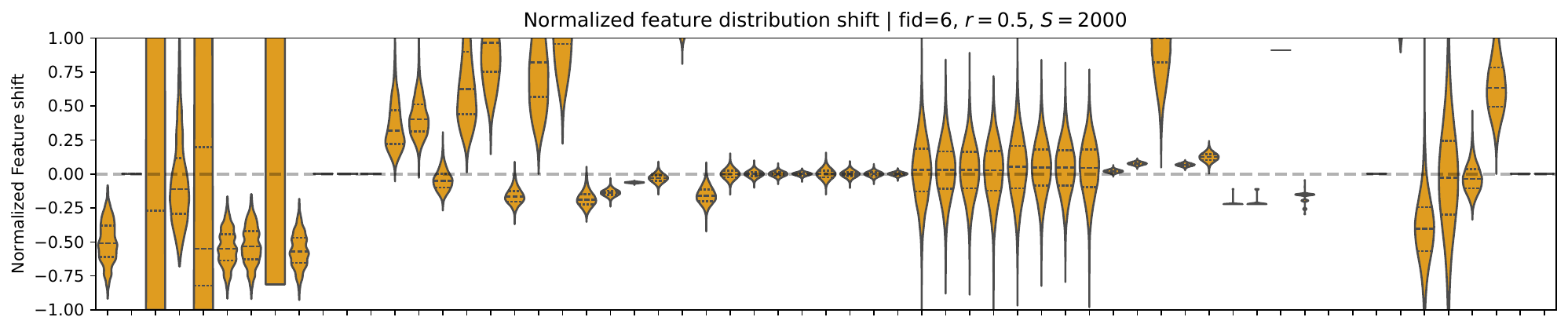}
    \includegraphics[width=.8\linewidth,trim=0cm 7.5cm 0cm 0cm,clip]{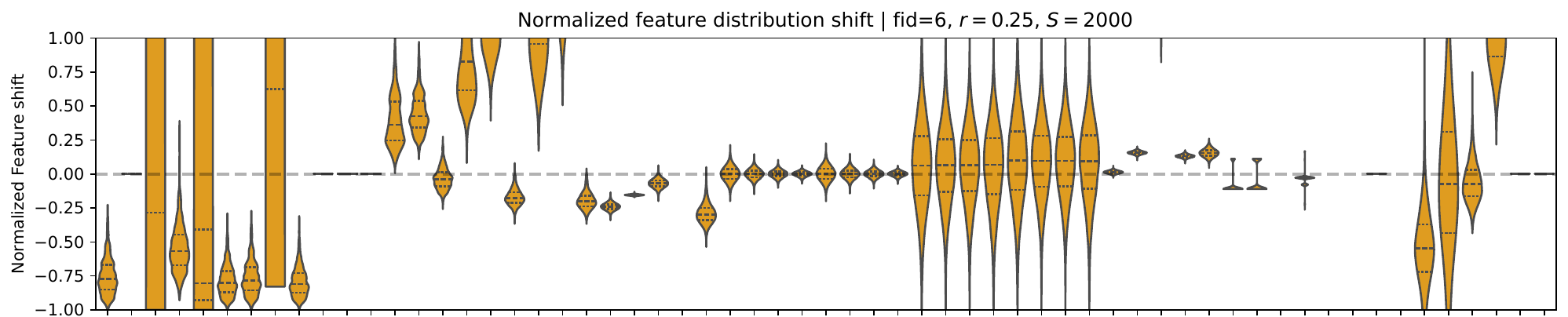}
    \includegraphics[width=.8\linewidth,trim=0cm .7cm 0cm 0cm,clip]{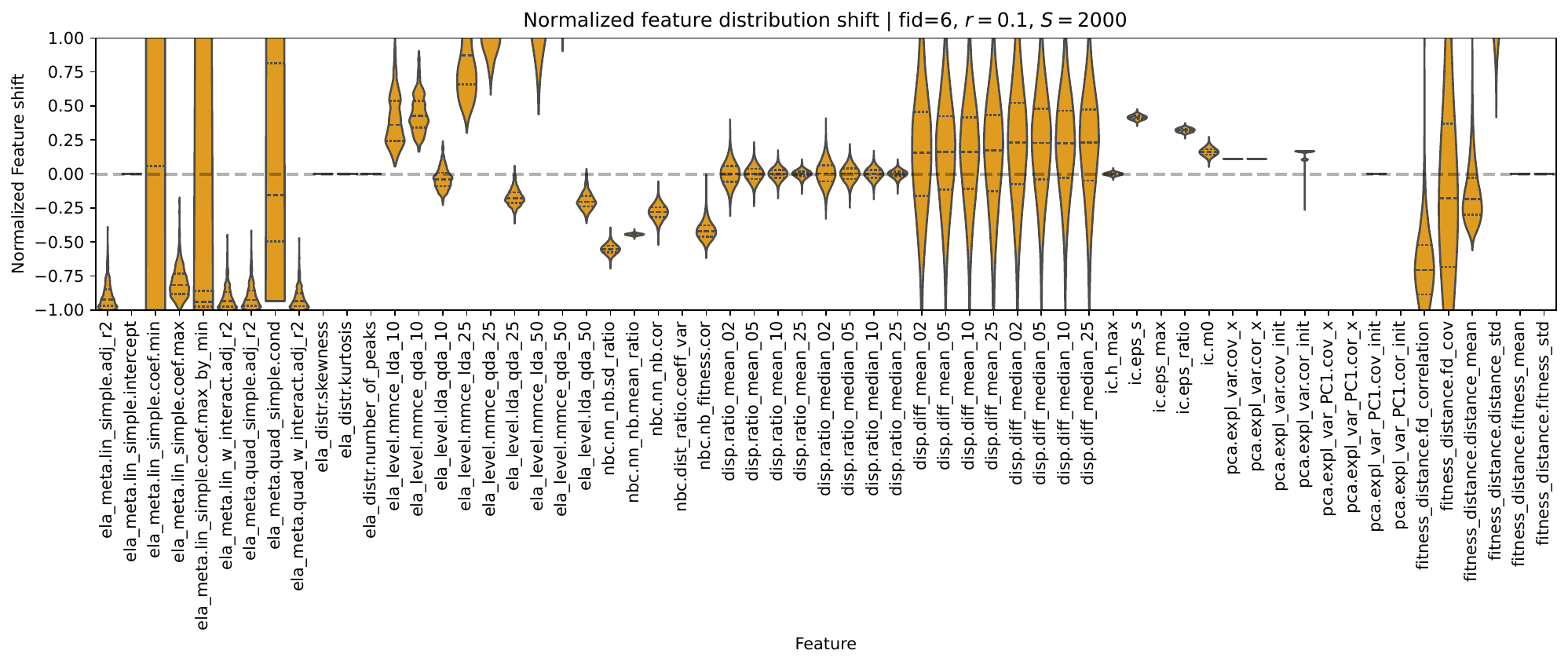}
    \caption{
    Normalized aggregated feature distribution shift of \textbf{Attractive sector (f6) function} with $\boldsymbol{S=2000}$ for compression ratios $r=\{0.5,0.25,0.1\}$. The dashed line denotes a normalized reference corresponding to the median of each feature distribution in the original search space. To enhance visualization, the limits of the Normalized Feature shift has been set to $[-1,1 ]$.
    }
    \label{fig:violin_f6_n2000}
\end{figure}

\begin{figure}[hbtp]\ContinuedFloat
    \centering
    \includegraphics[width=.8\linewidth,trim=0cm 7.5cm 0cm 0cm,clip]{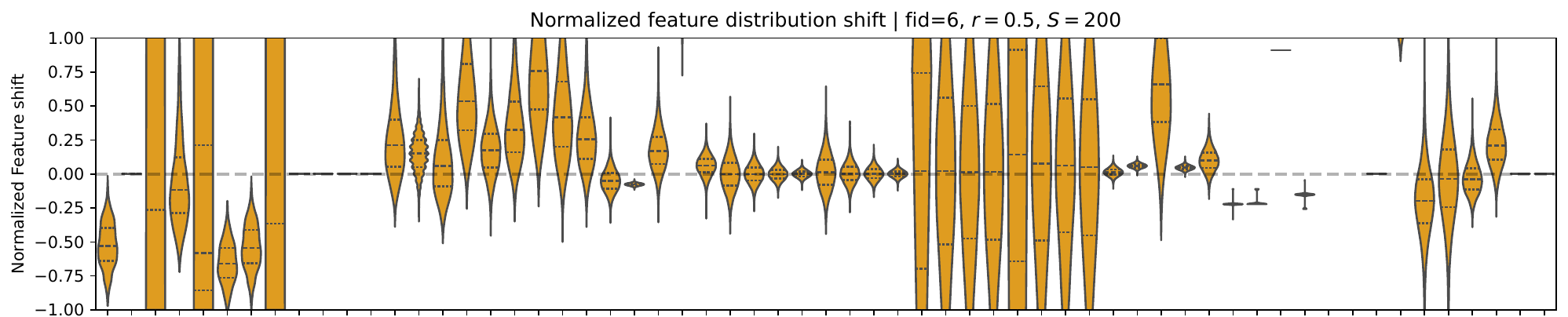}
    \includegraphics[width=.8\linewidth,trim=0cm 7.5cm 0cm 0cm,clip]{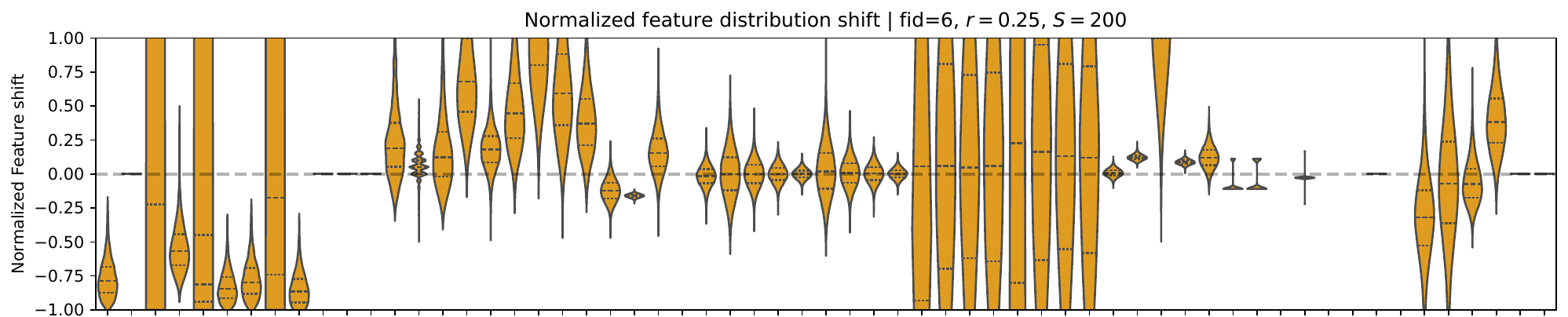}
    \includegraphics[width=.8\linewidth,trim=0cm .7cm 0cm 0cm,clip]{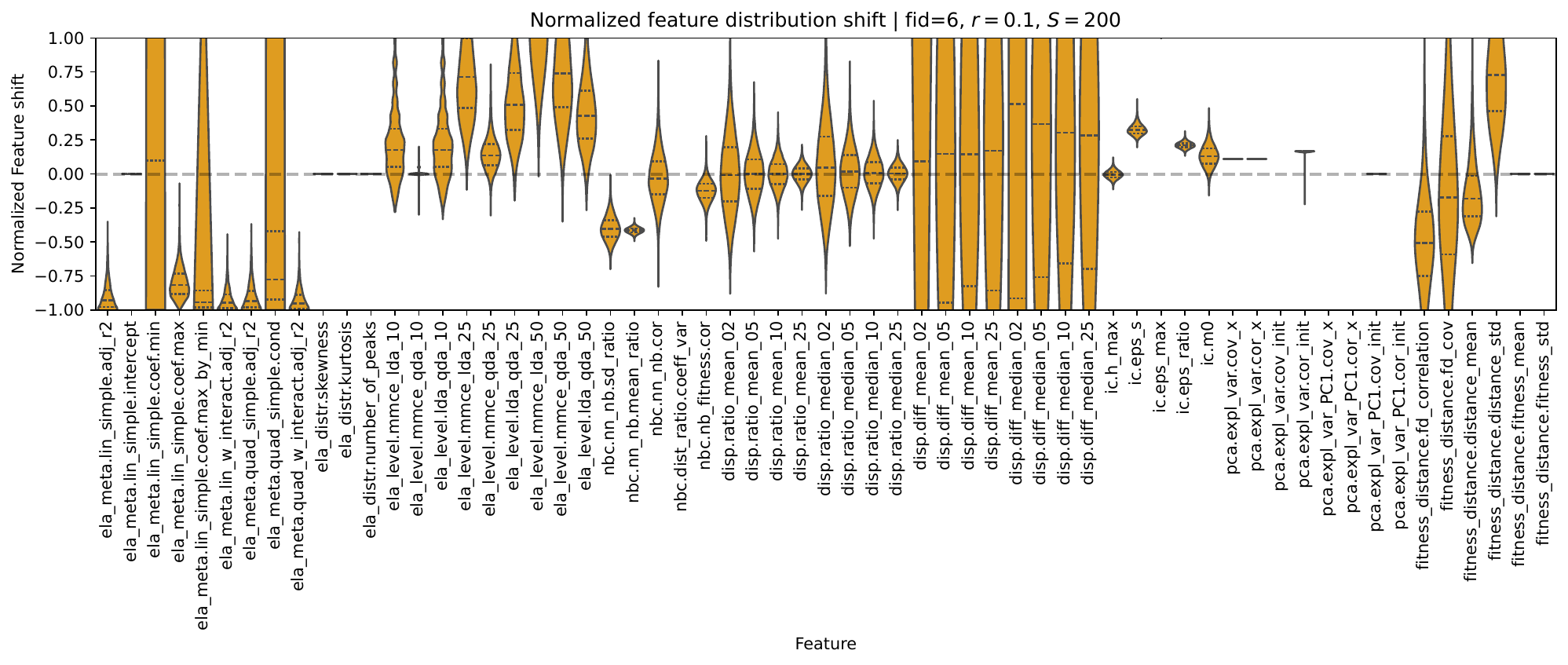}
  \caption{
  Same as above, for $\boldsymbol{S=200}$.
  }
    \label{fig:violin_f6_n200}
\end{figure}

\begin{figure}[hbtp]
    \centering
    \includegraphics[width=.8\linewidth,trim=0cm 7.5cm 0cm 0cm,clip]{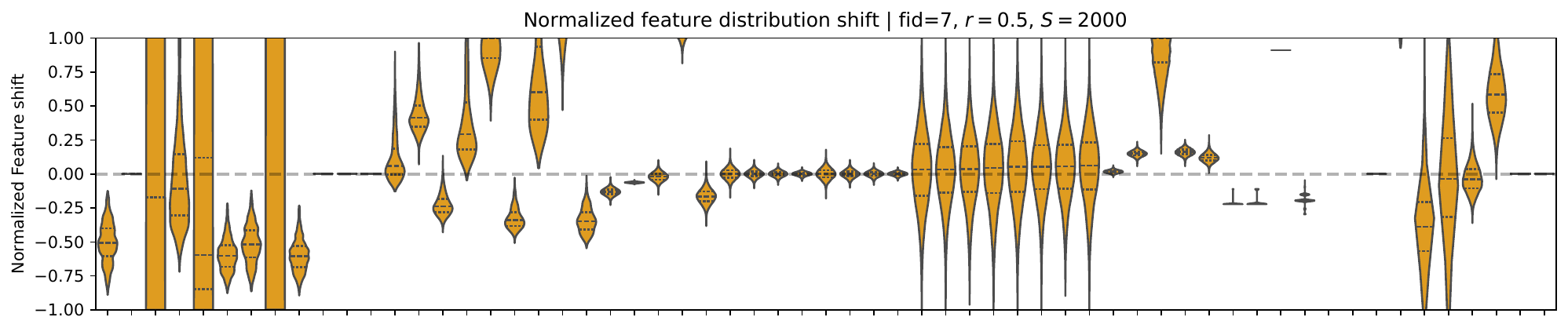}
    \includegraphics[width=.8\linewidth,trim=0cm 7.5cm 0cm 0cm,clip]{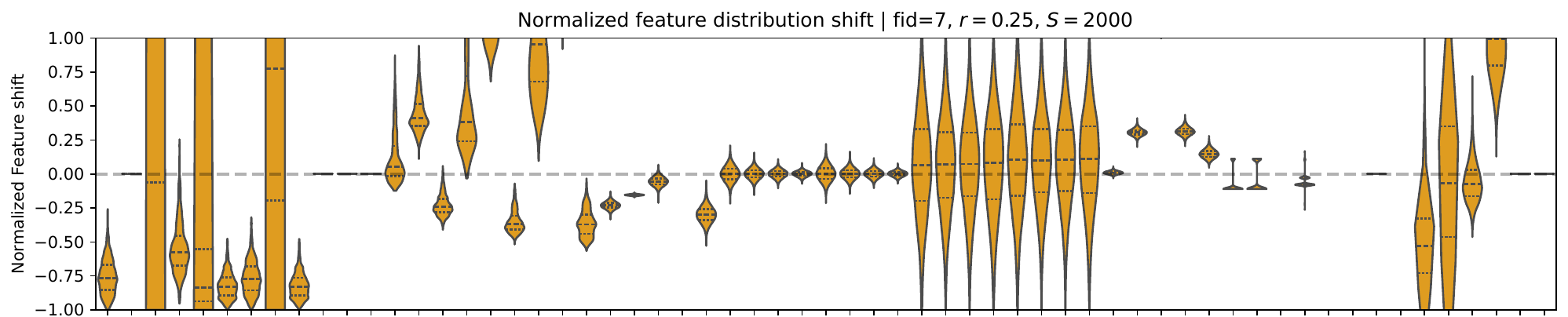}
    \includegraphics[width=.8\linewidth,trim=0cm .7cm 0cm 0cm,clip]{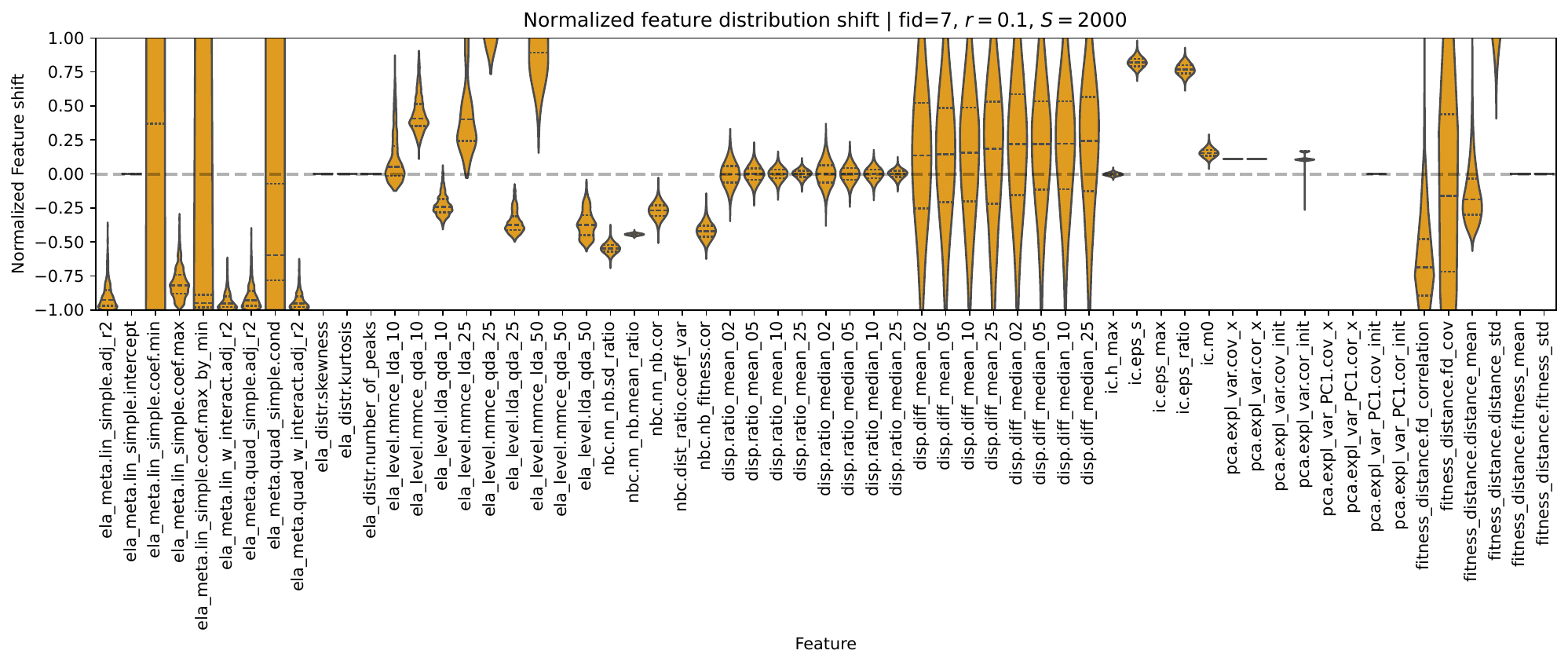}
    \caption{
    Normalized aggregated feature distribution shift of \textbf{Step-ellipsoid (f7) function} with $\boldsymbol{S=2000}$ for compression ratios $r=\{0.5,0.25,0.1\}$. The dashed line denotes a normalized reference corresponding to the median of each feature distribution in the original search space. To enhance visualization, the limits of the Normalized Feature shift has been set to $[-1,1 ]$.
    }
    \label{fig:violin_f7_n2000}
\end{figure}

\begin{figure}[hbtp]\ContinuedFloat
    \centering
    \includegraphics[width=.8\linewidth,trim=0cm 7.5cm 0cm 0cm,clip]{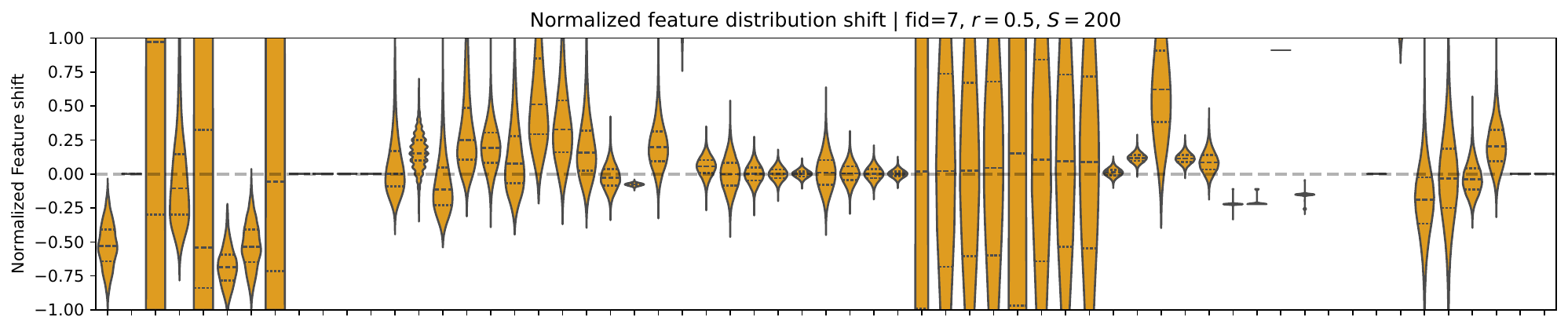}
    \includegraphics[width=.8\linewidth,trim=0cm 7.5cm 0cm 0cm,clip]{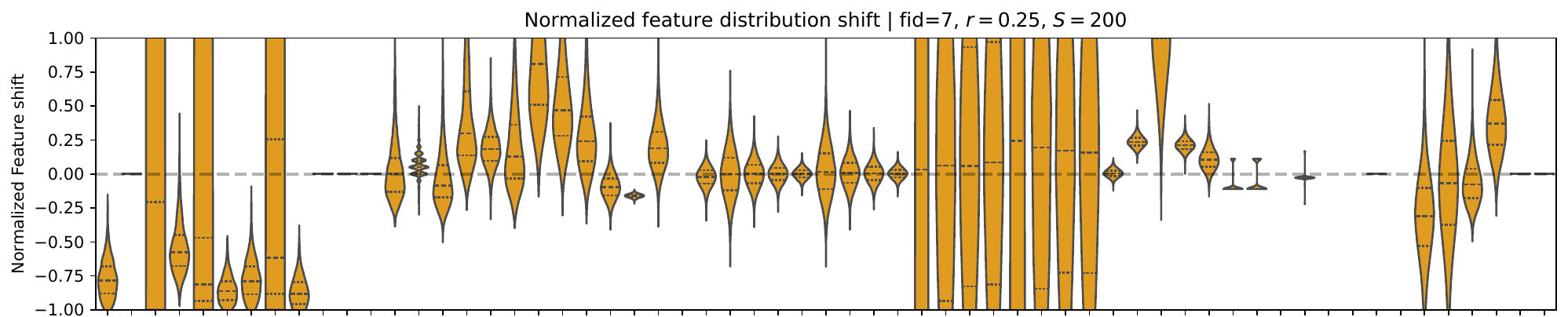}
    \includegraphics[width=.8\linewidth,trim=0cm .7cm 0cm 0cm,clip]{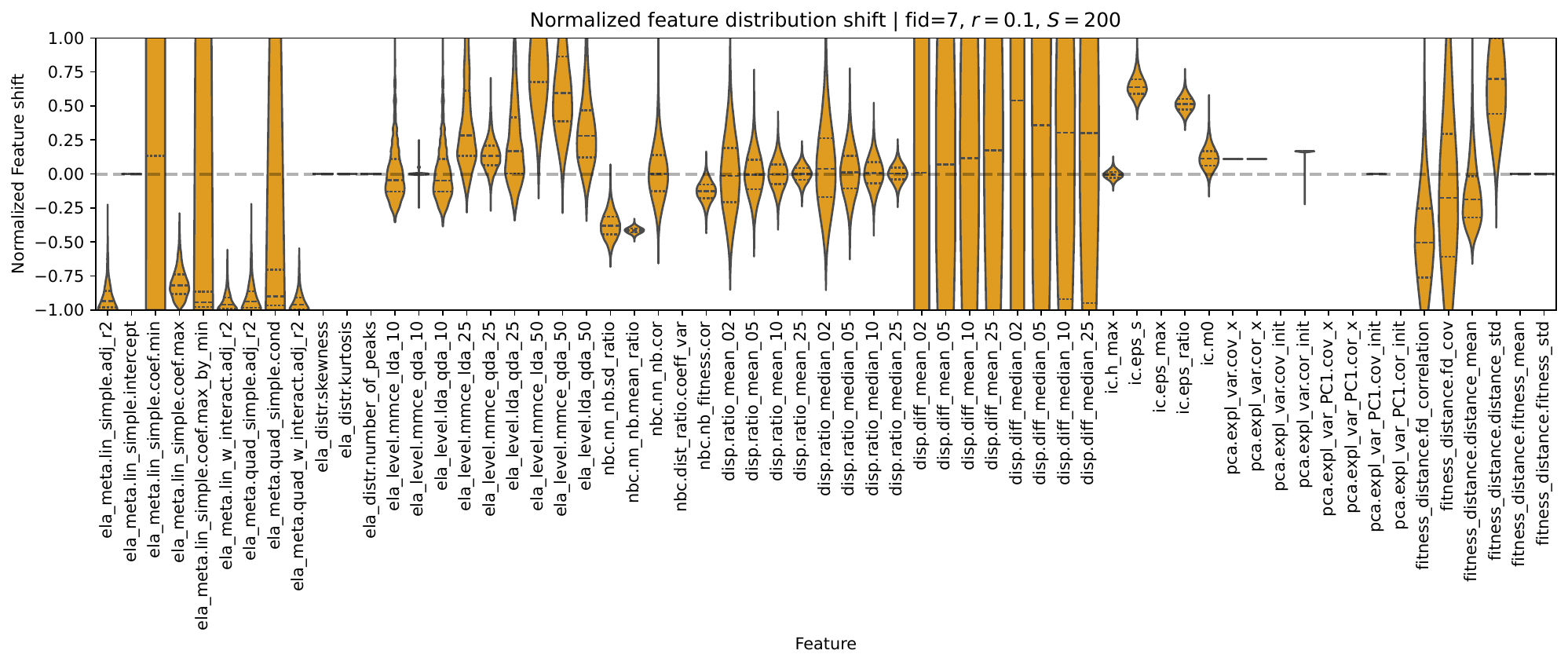}
  \caption{
  Same as above, for $\boldsymbol{S=200}$.
  }
    \label{fig:violin_f7_n200}
\end{figure}


\begin{figure}[hbtp]
    \centering
    \includegraphics[width=.8\linewidth,trim=0cm 7.5cm 0cm 0cm,clip]{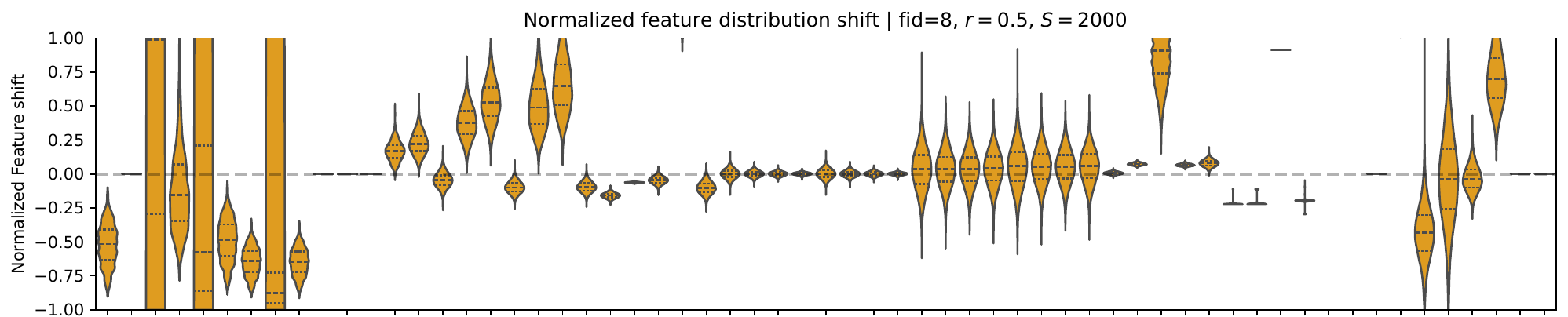}
    \includegraphics[width=.8\linewidth,trim=0cm 7.5cm 0cm 0cm,clip]{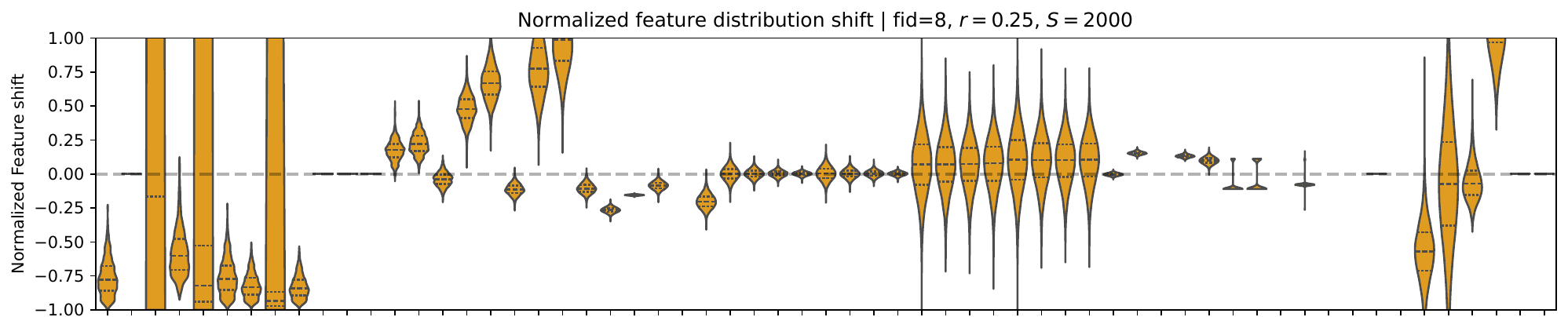}
    \includegraphics[width=.8\linewidth,trim=0cm .7cm 0cm 0cm,clip]{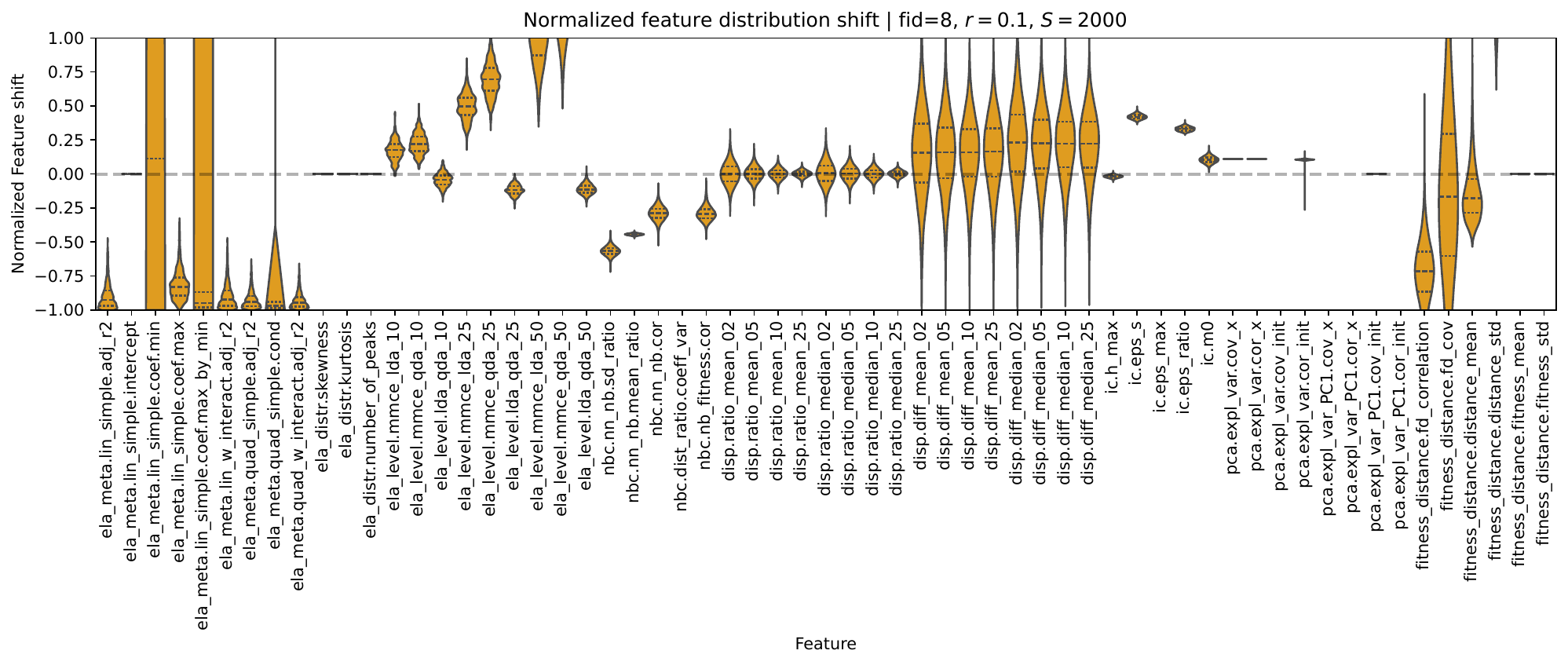}
       \caption{Normalized aggregated feature distribution shift of \textbf{Rosenbrock (f8) function} with $\boldsymbol{S=2000}$ for compression ratios $r=\{0.5,0.25,0.1\}$. The horizontal dashed line denotes a normalized reference corresponding to the median of each feature distribution in the original search space. To enhance visualization, the limits of the Normalized Feature shift has been set to $[-1,1 ]$.}
    \label{fig:violin_f8_n2000}
\end{figure}

\begin{figure}[hbtp]\ContinuedFloat
    \centering
    \includegraphics[width=.8\linewidth,trim=0cm 7.5cm 0cm 0cm,clip]{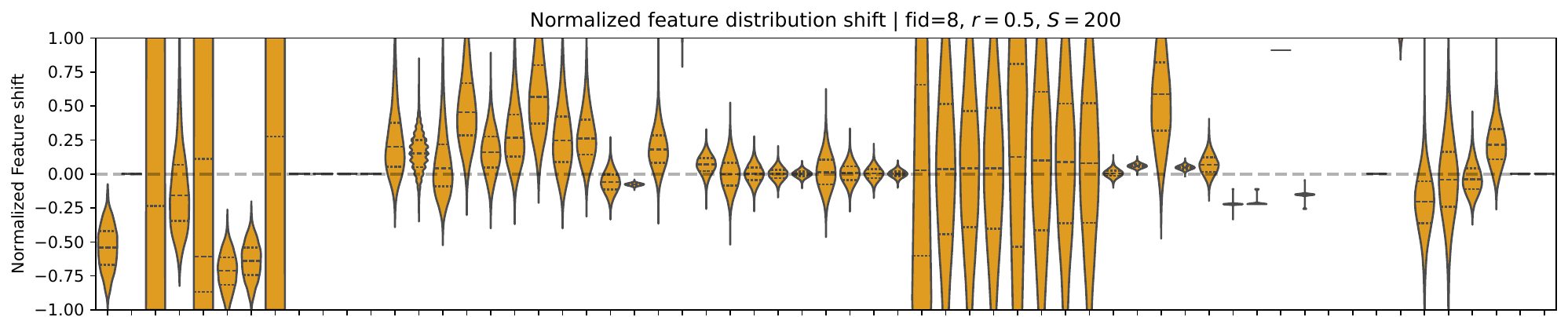}
    \includegraphics[width=.8\linewidth,trim=0cm 7.5cm 0cm 0cm,clip]{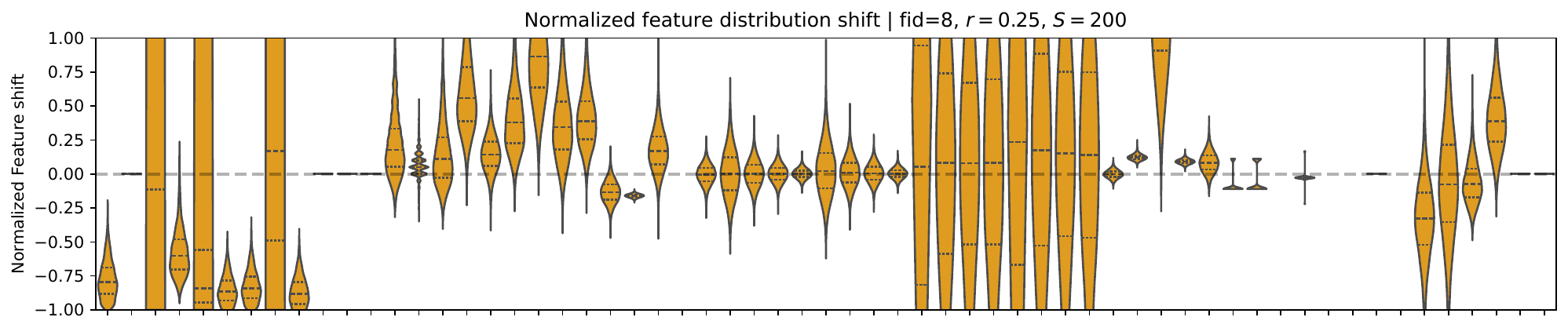}
    \includegraphics[width=.8\linewidth,trim=0cm .7cm 0cm 0cm,clip]{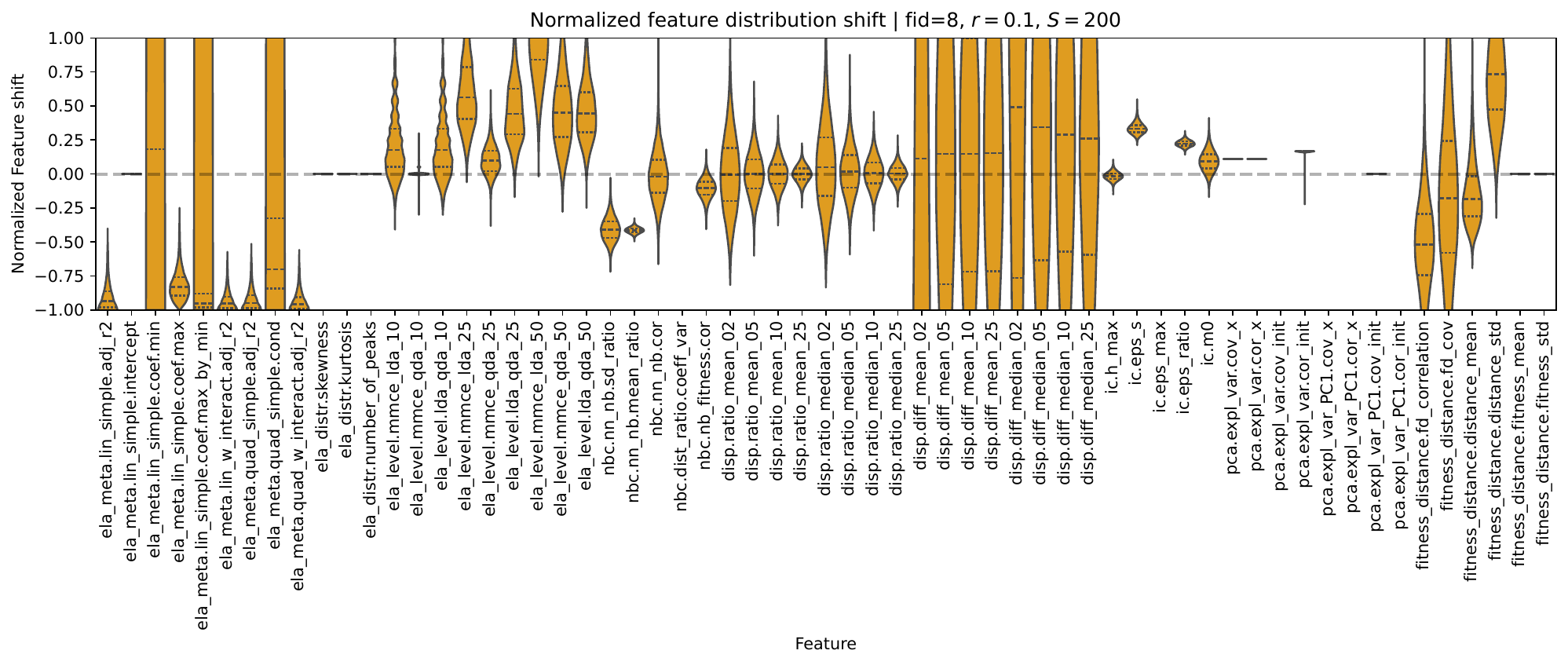}
   \caption{Same as above, for $\boldsymbol{S=200}$.}
    \label{fig:violin_f8_n200}
\end{figure}


\begin{figure}[hbtp]
    \centering
    \includegraphics[width=.8\linewidth,trim=0cm 7.5cm 0cm 0cm,clip]{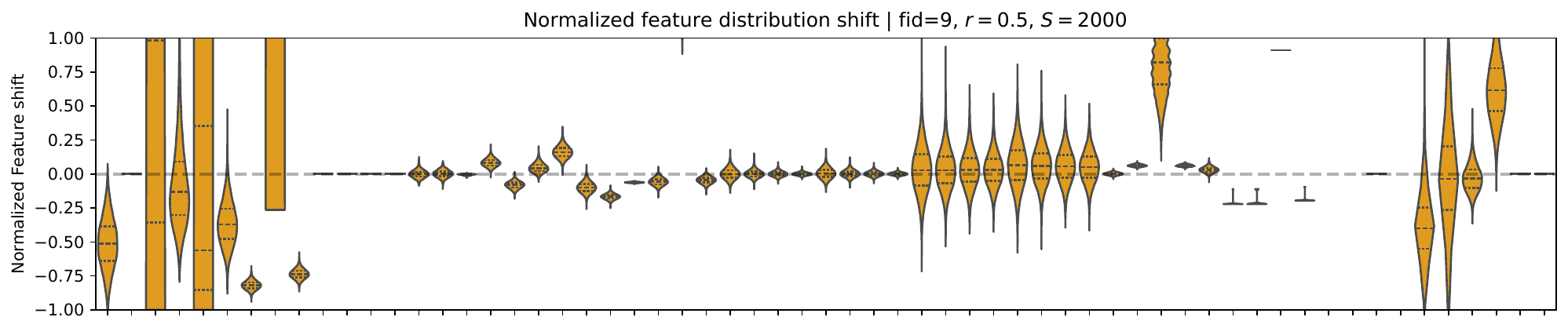}
    \includegraphics[width=.8\linewidth,trim=0cm 7.5cm 0cm 0cm,clip]{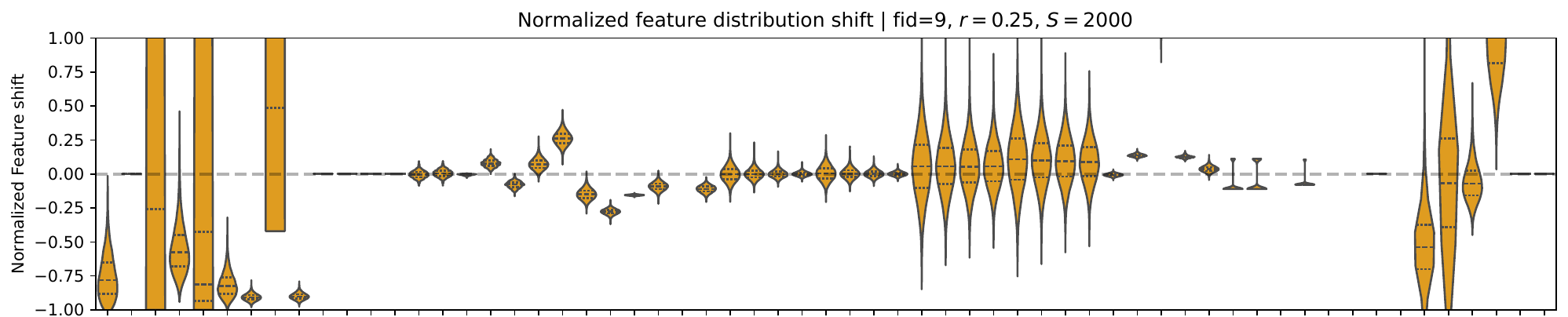}
    \includegraphics[width=.8\linewidth,trim=0cm .7cm 0cm 0cm,clip]{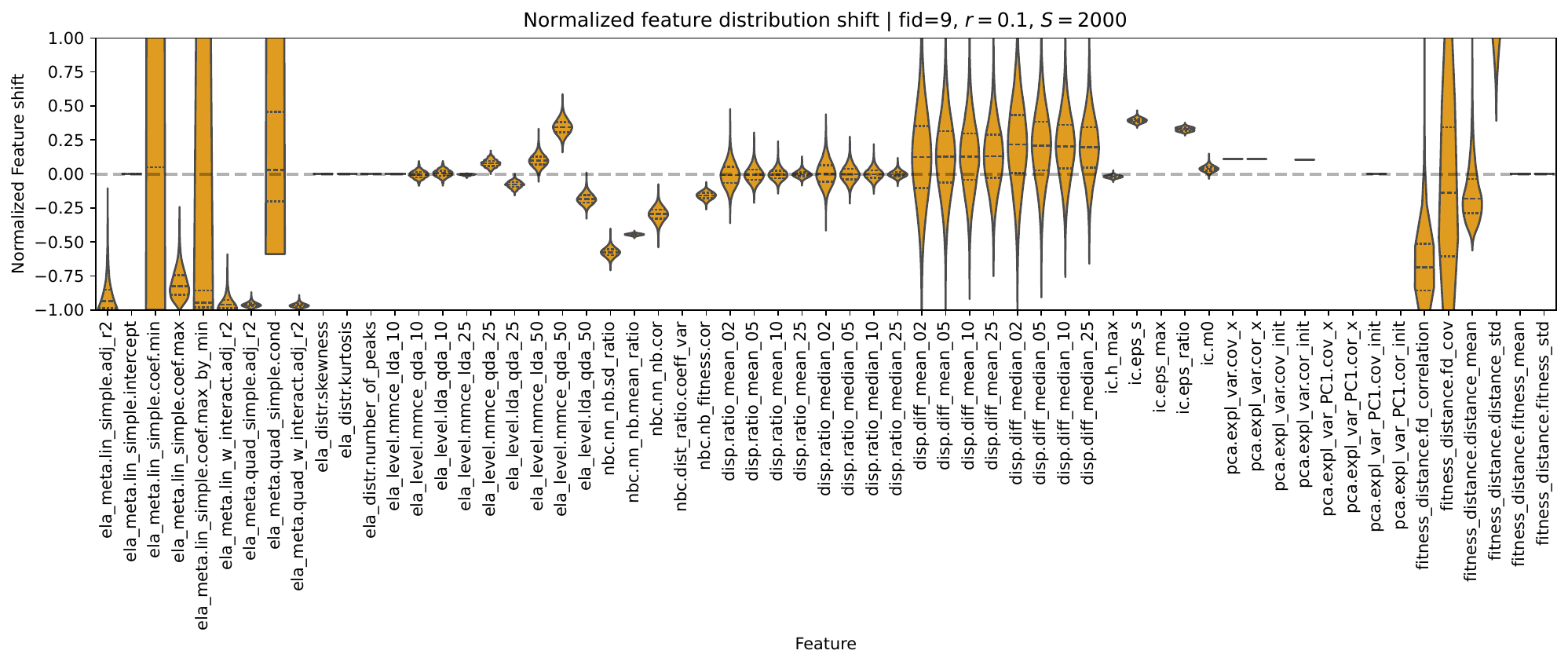}
       \caption{Normalized aggregated feature distribution shift of \textbf{Rosenbrock rotated (f9) function} with $\boldsymbol{S=2000}$ for compression ratios $r=\{0.5,0.25,0.1\}$. The horizontal dashed line denotes a normalized reference corresponding to the median of each feature distribution in the original search space. To enhance visualization, the limits of the Normalized Feature shift has been set to $[-1,1 ]$.}
    \label{fig:violin_f9_n2000}
\end{figure}

\begin{figure}[hbtp]\ContinuedFloat
    \centering
    \includegraphics[width=.8\linewidth,trim=0cm 7.5cm 0cm 0cm,clip]{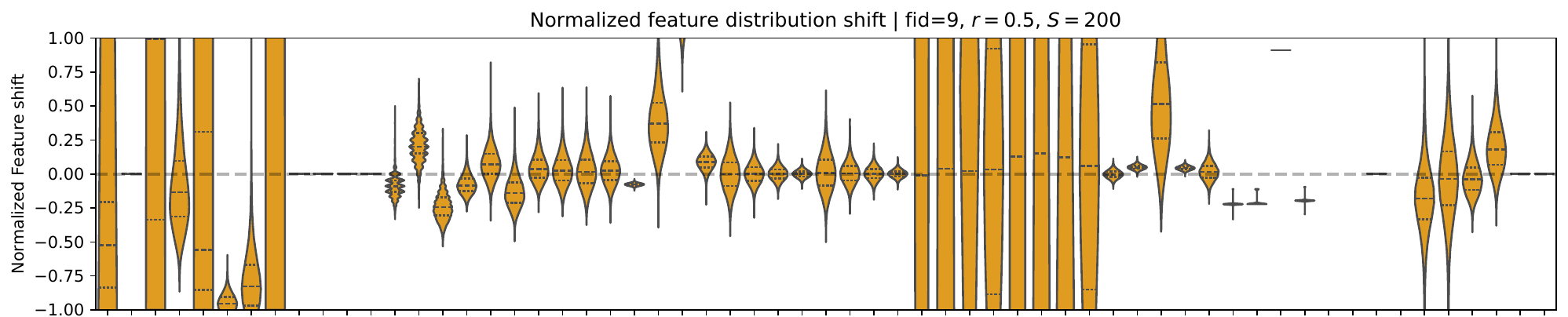}
    \includegraphics[width=.8\linewidth,trim=0cm 7.5cm 0cm 0cm,clip]{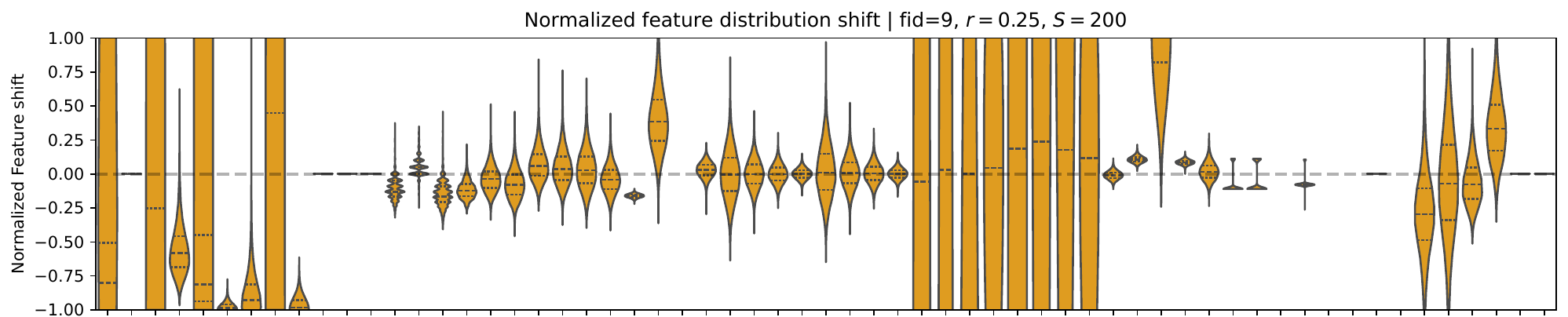}
    \includegraphics[width=.8\linewidth,trim=0cm .7cm 0cm 0cm,clip]{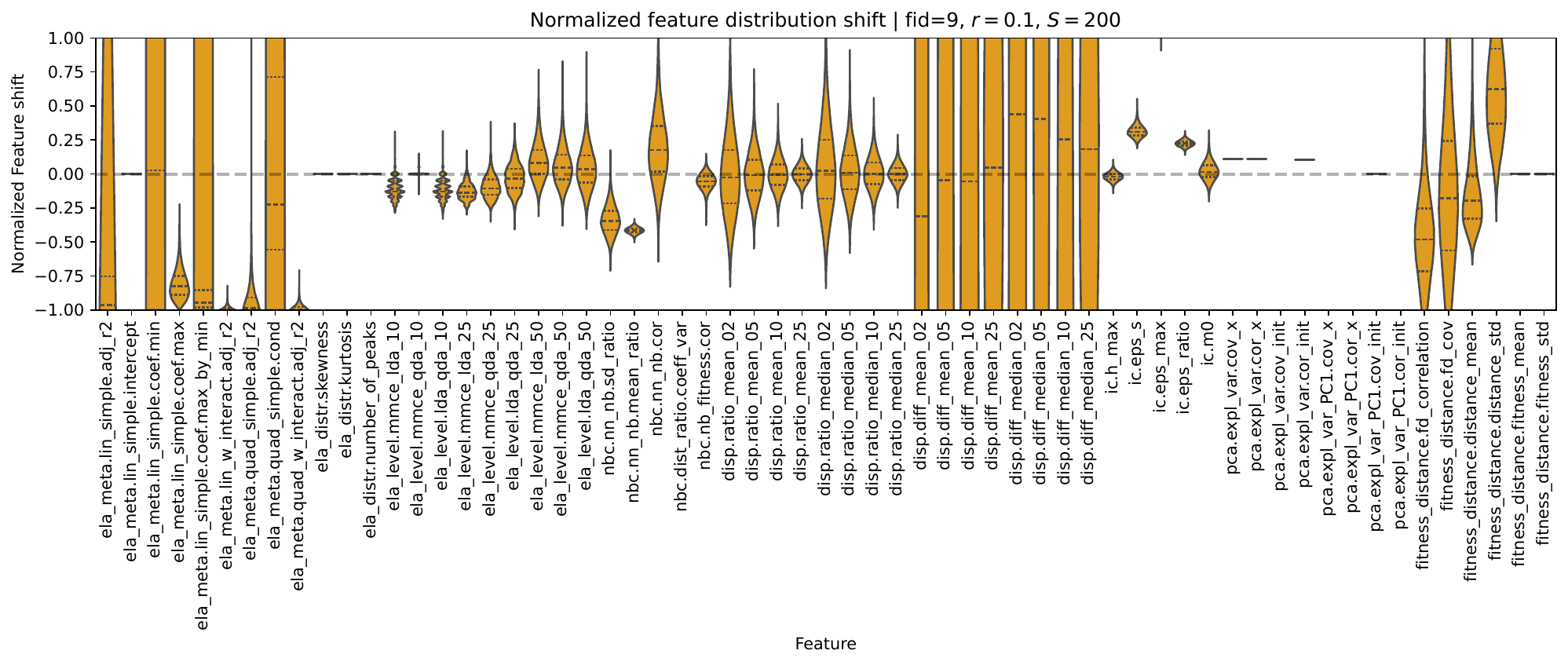}
   \caption{Same as above, for $\boldsymbol{S=200}$.}
    \label{fig:violin_f9_n200}
\end{figure}

 \begin{figure}[hbtp]
    \centering
    \includegraphics[width=.8\linewidth,trim=0cm 7.5cm 0cm 0cm,clip]{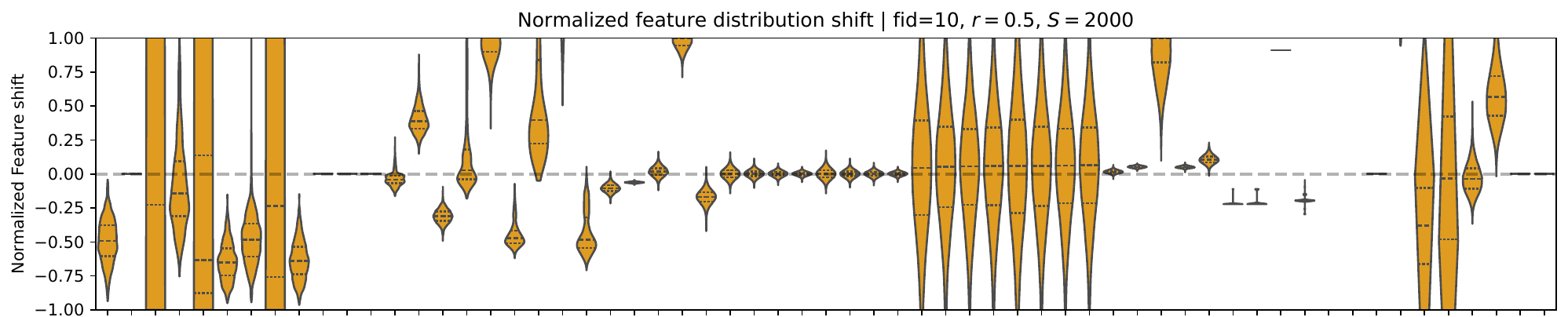}
    \includegraphics[width=.8\linewidth,trim=0cm 7.5cm 0cm 0cm,clip]{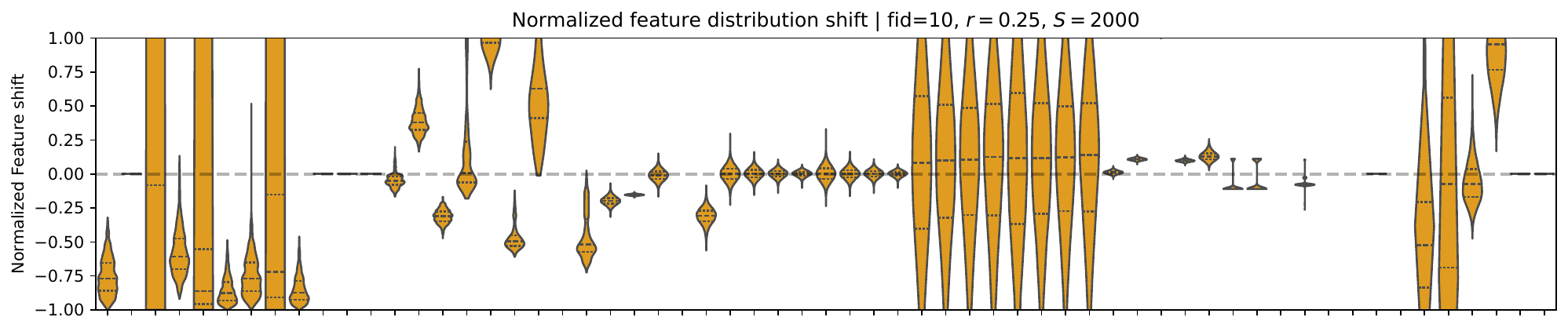}
    \includegraphics[width=.8\linewidth,trim=0cm .7cm 0cm 0cm,clip]{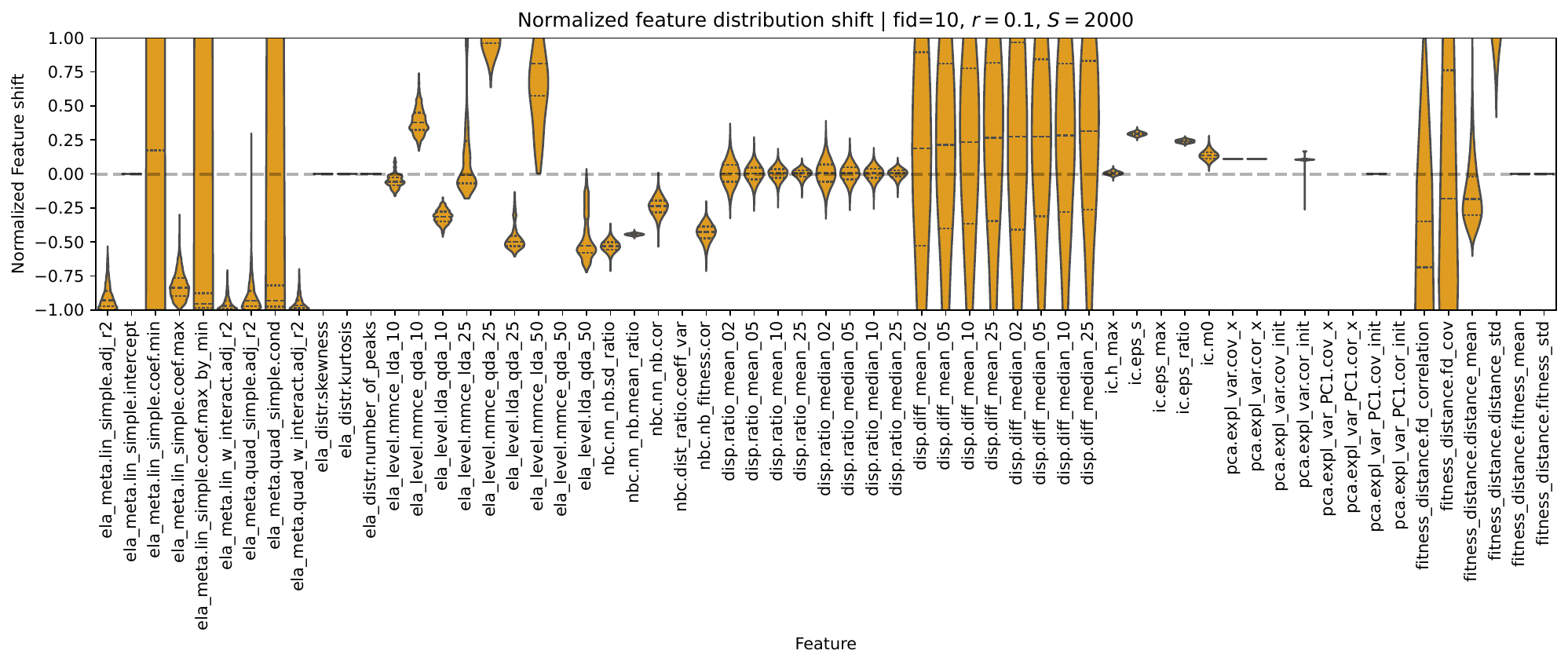}
       \caption{Normalized aggregated feature distribution shift of \textbf{Ellipsoid (f10) function} with $\boldsymbol{S=2000}$ for compression ratios $r=\{0.5,0.25,0.1\}$. The horizontal dashed line denotes a normalized reference corresponding to the median of each feature distribution in the original search space. To enhance visualization, the limits of the Normalized Feature shift has been set to $[-1,1 ]$.}
    \label{fig:violin_f10_n2000}
\end{figure}

\begin{figure}[hbtp]\ContinuedFloat
    \centering
    \includegraphics[width=.8\linewidth,trim=0cm 7.5cm 0cm 0cm,clip]{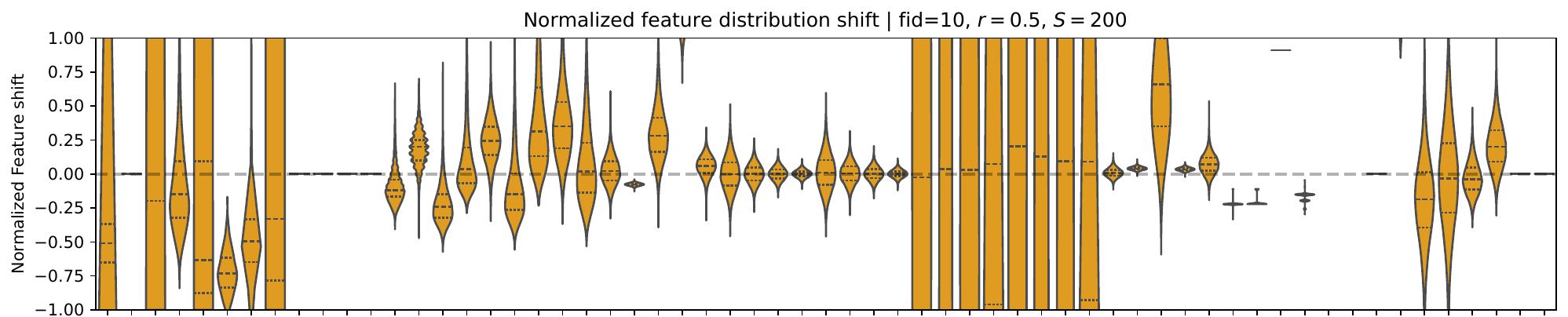}
    \includegraphics[width=.8\linewidth,trim=0cm 7.5cm 0cm 0cm,clip]{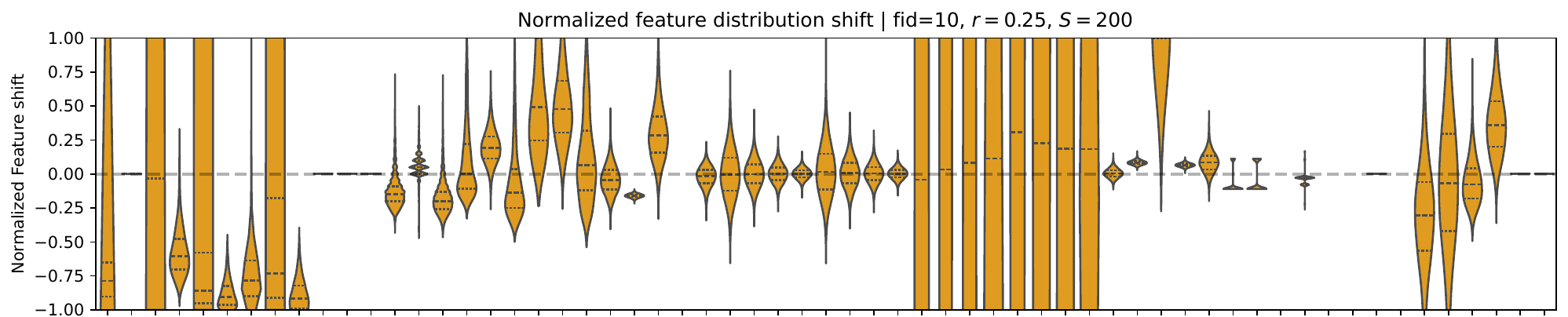}
    \includegraphics[width=.8\linewidth,trim=0cm .7cm 0cm 0cm,clip]{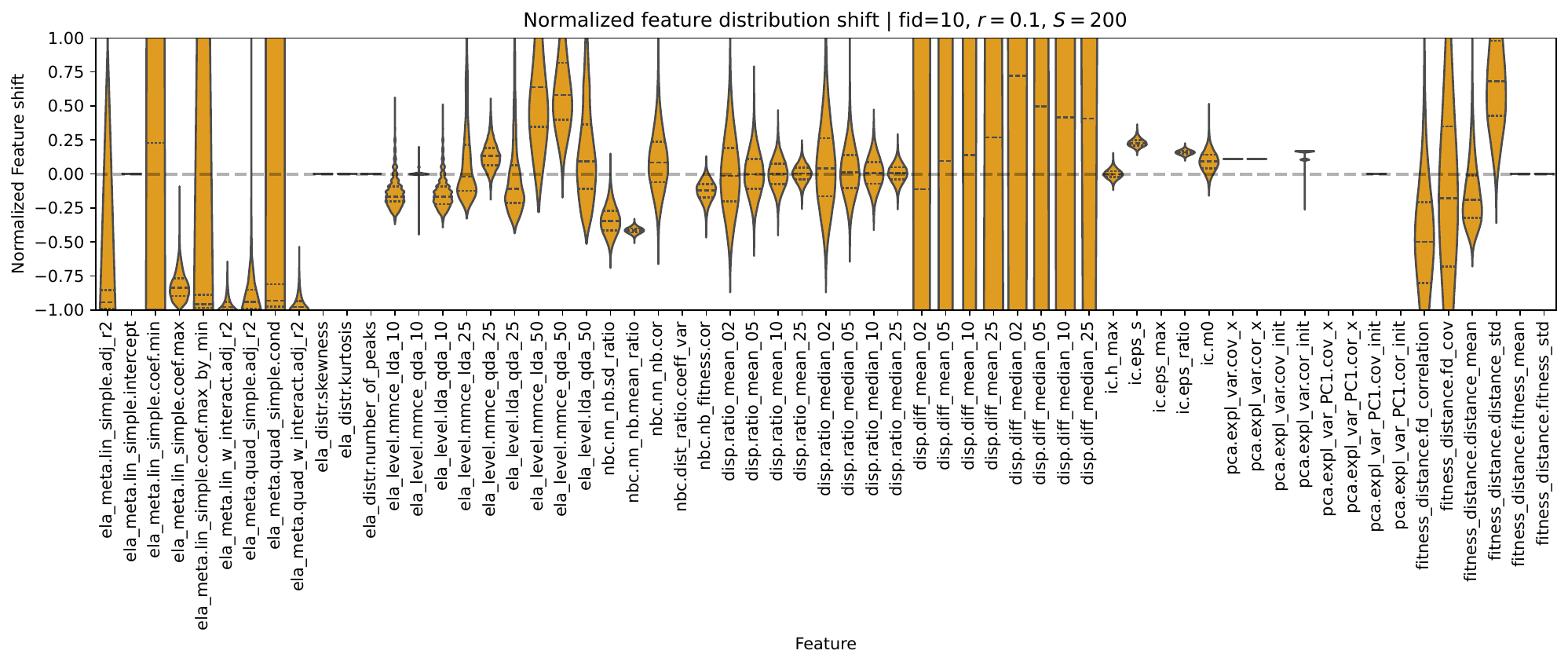}
   \caption{Same as above, for $\boldsymbol{S=200}$.}
    \label{fig:violin_f10_n200}
\end{figure}


\begin{figure}[hbtp]
    \centering
    \includegraphics[width=.8\linewidth,trim=0cm 7.5cm 0cm 0cm,clip]{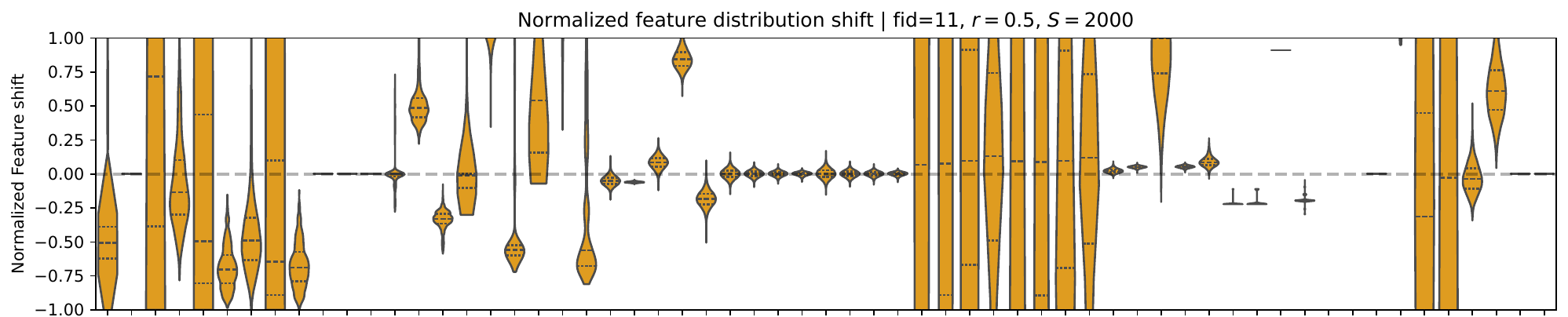}
    \includegraphics[width=.8\linewidth,trim=0cm 7.5cm 0cm 0cm,clip]{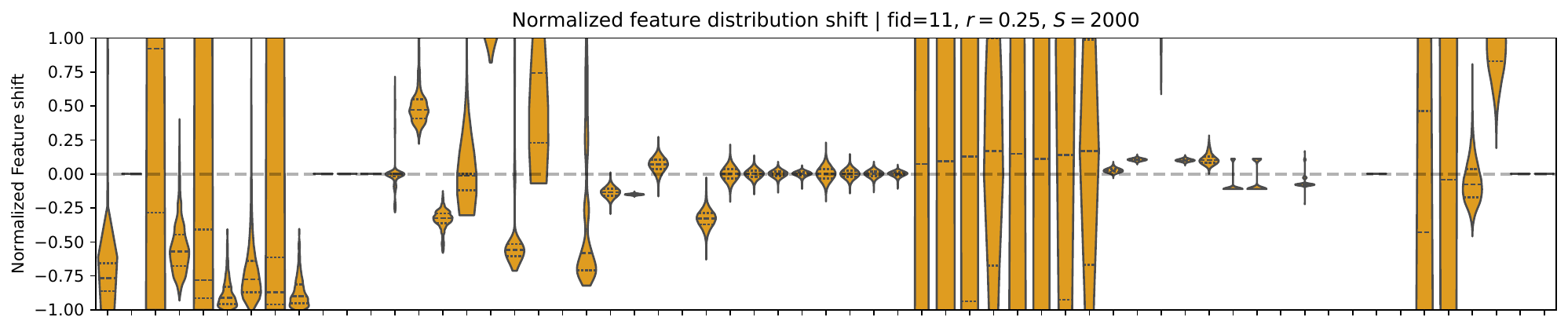}
    \includegraphics[width=.8\linewidth,trim=0cm .7cm 0cm 0cm,clip]{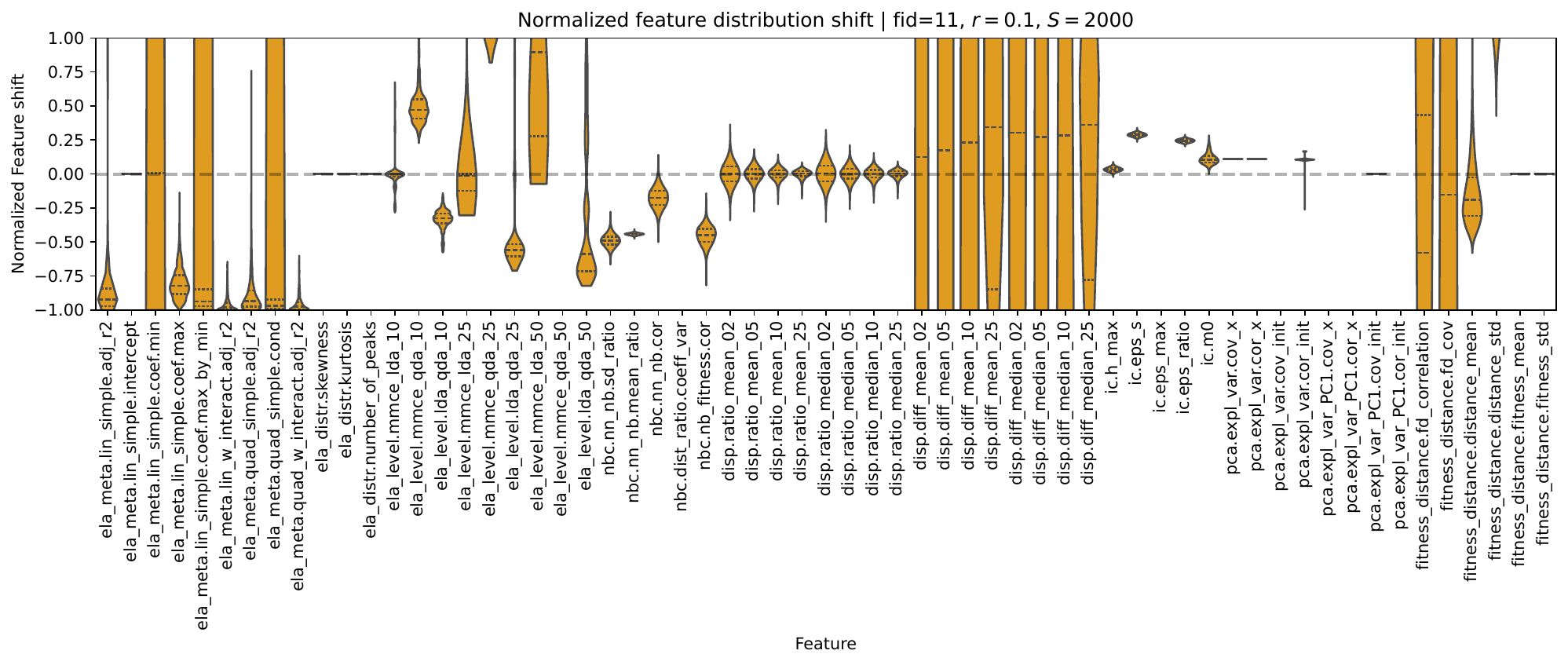}
       \caption{Normalized aggregated feature distribution shift of \textbf{Discus (f11) function} with $\boldsymbol{S=2000}$ for compression ratios $r=\{0.5,0.25,0.1\}$. The horizontal dashed line denotes a normalized reference corresponding to the median of each feature distribution in the original search space. To enhance visualization, the limits of the Normalized Feature shift has been set to $[-1,1 ]$.}
    \label{fig:violin_f11_n2000}
\end{figure}

\begin{figure}[hbtp]\ContinuedFloat
    \centering
    \includegraphics[width=.8\linewidth,trim=0cm 7.5cm 0cm 0cm,clip]{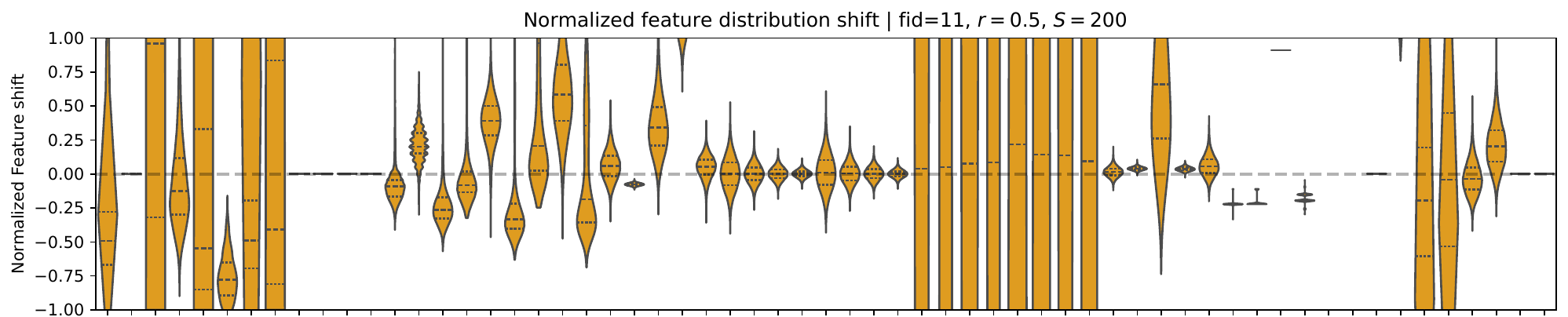}
    \includegraphics[width=.8\linewidth,trim=0cm 7.5cm 0cm 0cm,clip]{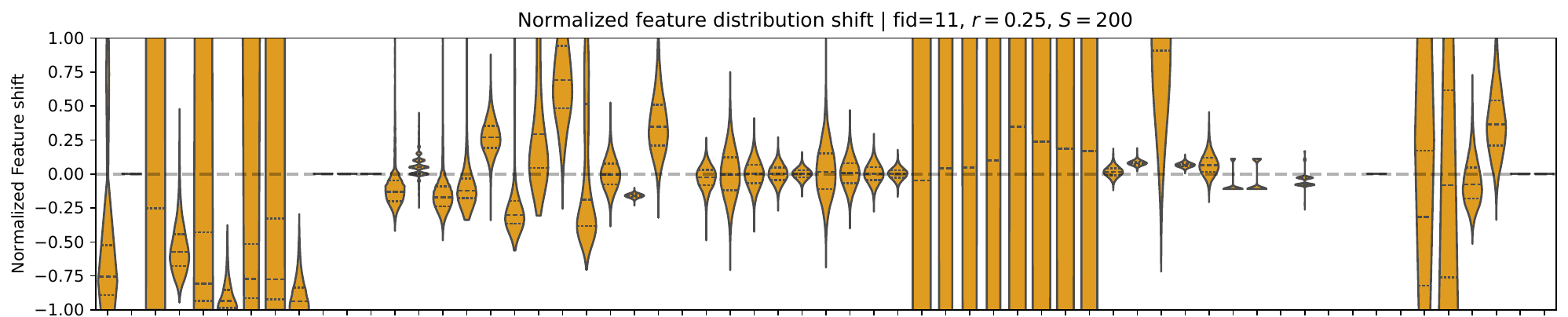}
    \includegraphics[width=.8\linewidth,trim=0cm .7cm 0cm 0cm,clip]{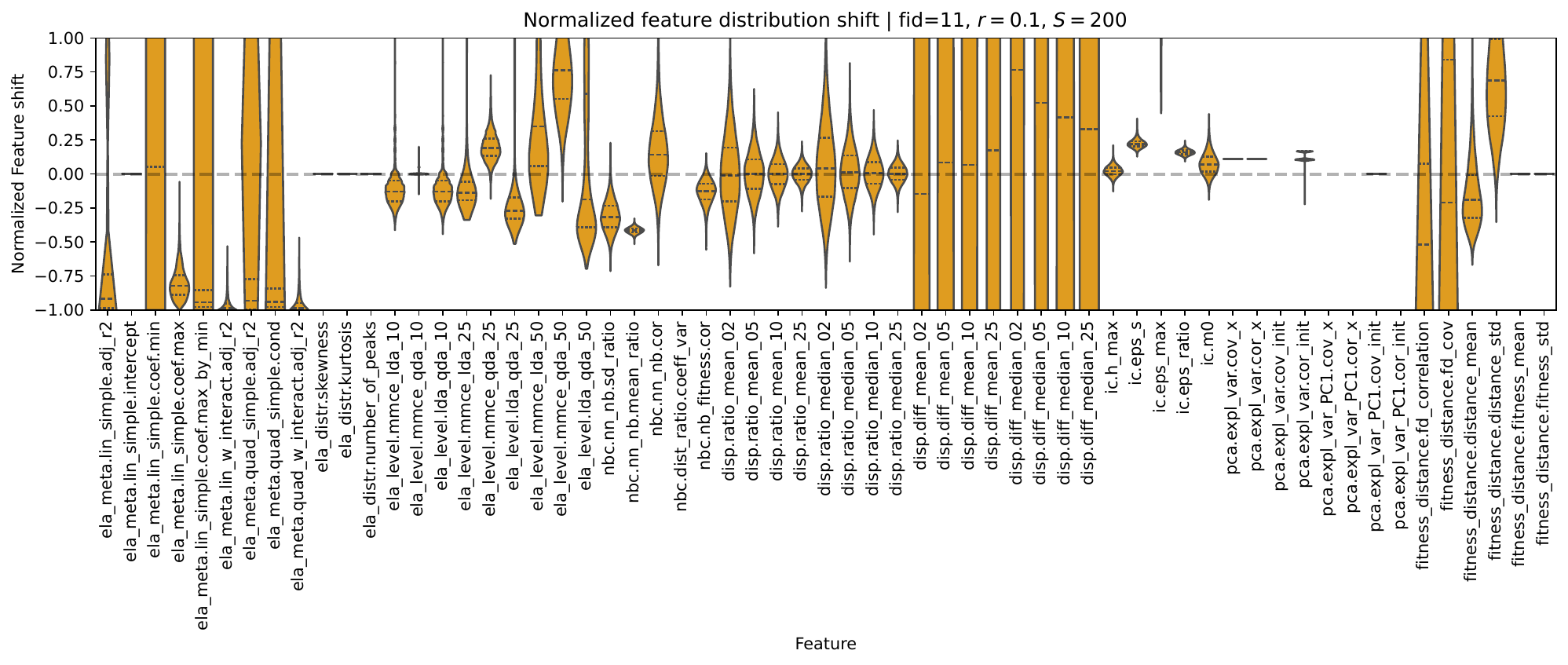}
    \caption{Same as above, for $\boldsymbol{S=200}$.}
    \label{fig:violin_f11_n200}
\end{figure}


\begin{figure}[hbtp]
    \centering
    \includegraphics[width=.8\linewidth,trim=0cm 7.5cm 0cm 0cm,clip]{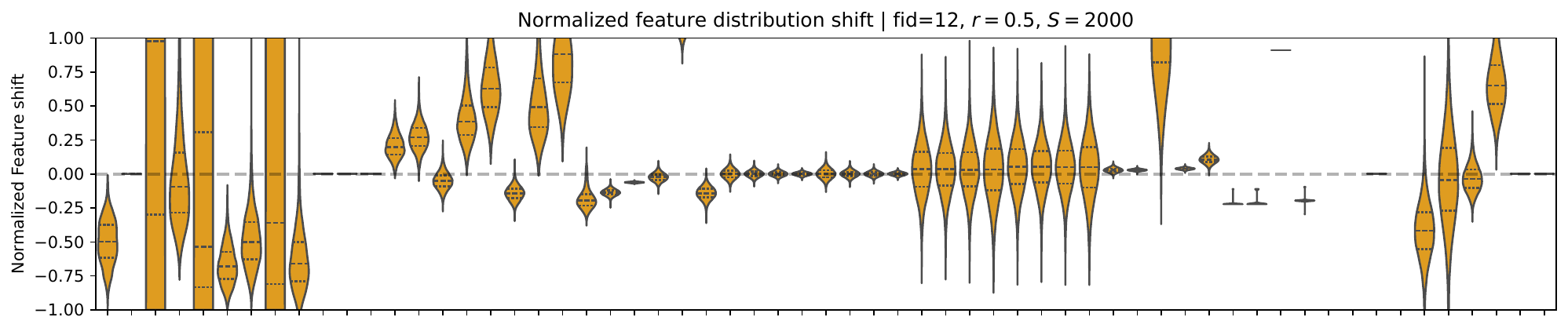}
    \includegraphics[width=.8\linewidth,trim=0cm 7.5cm 0cm 0cm,clip]{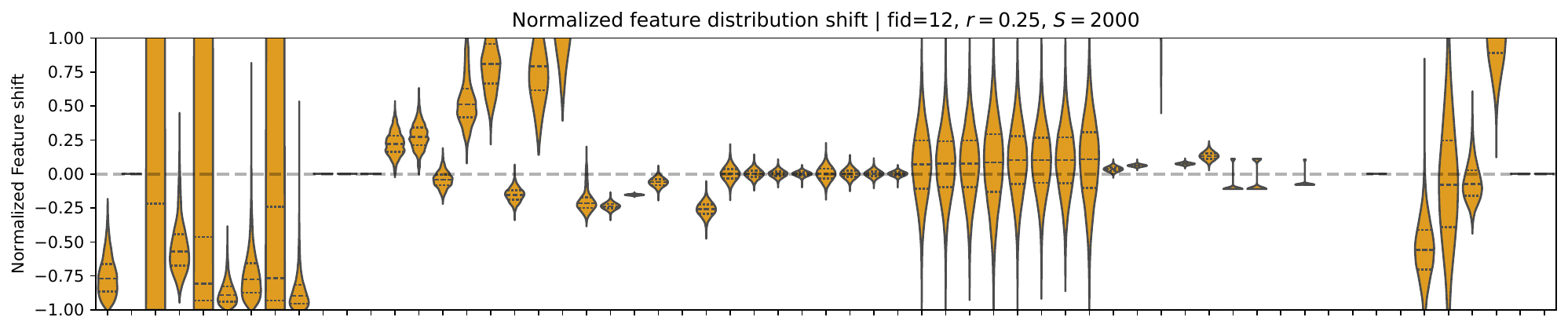}
    \includegraphics[width=.8\linewidth,trim=0cm .7cm 0cm 0cm,clip]{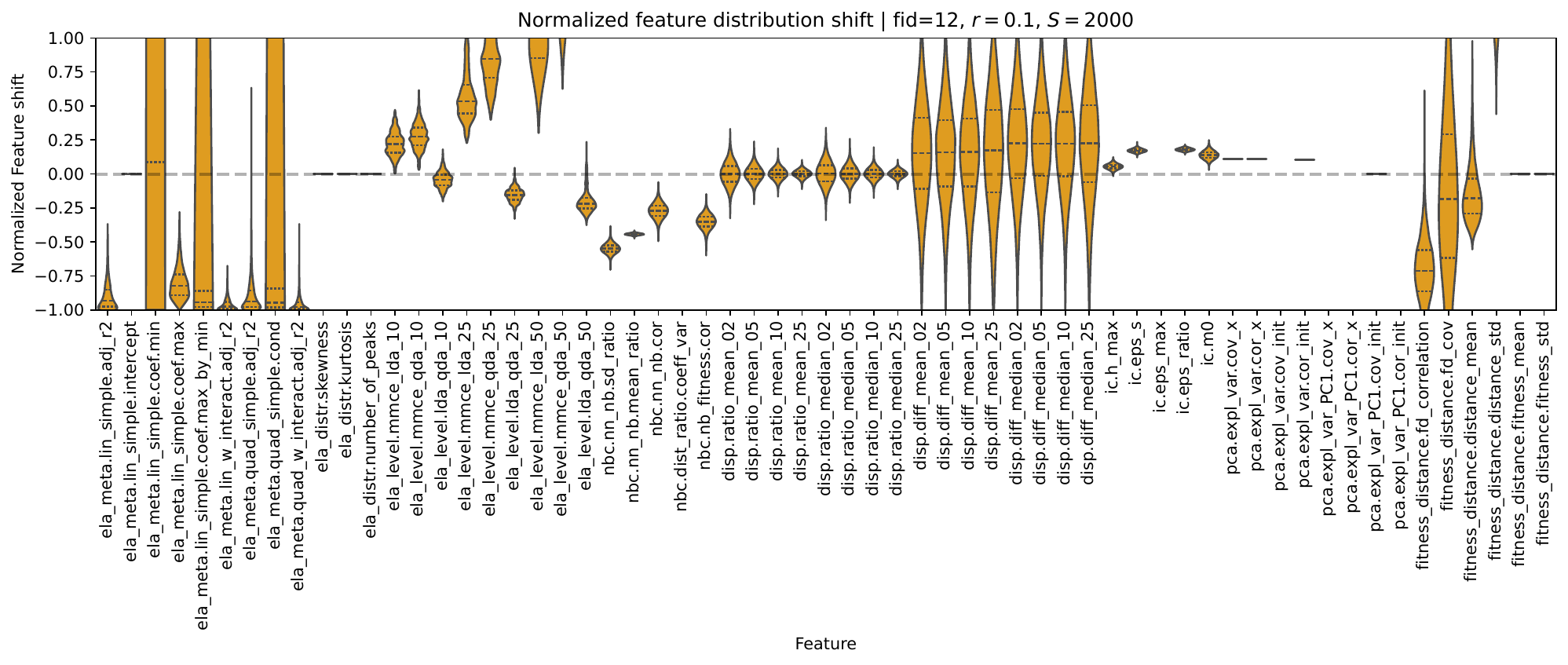}
       \caption{Normalized aggregated feature distribution shift of \textbf{Bent cigar (f12) function} with $\boldsymbol{S=2000}$ for compression ratios $r=\{0.5,0.25,0.1\}$. The horizontal dashed line denotes a normalized reference corresponding to the median of each feature distribution in the original search space. To enhance visualization, the limits of the Normalized Feature shift has been set to $[-1,1 ]$.}
    \label{fig:violin_f12_n2000}
\end{figure}

\begin{figure}[hbtp]\ContinuedFloat
    \centering
    \includegraphics[width=.8\linewidth,trim=0cm 7.5cm 0cm 0cm,clip]{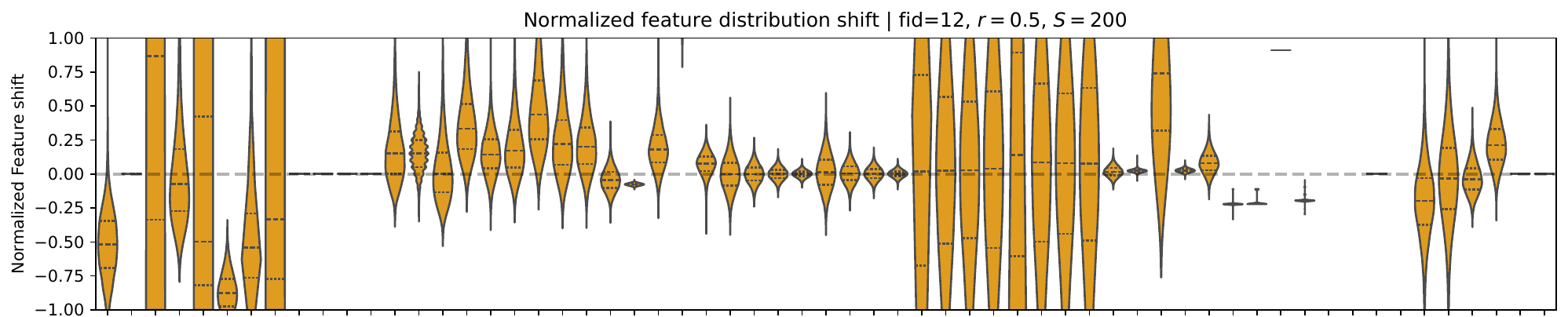}
    \includegraphics[width=.8\linewidth,trim=0cm 7.5cm 0cm 0cm,clip]{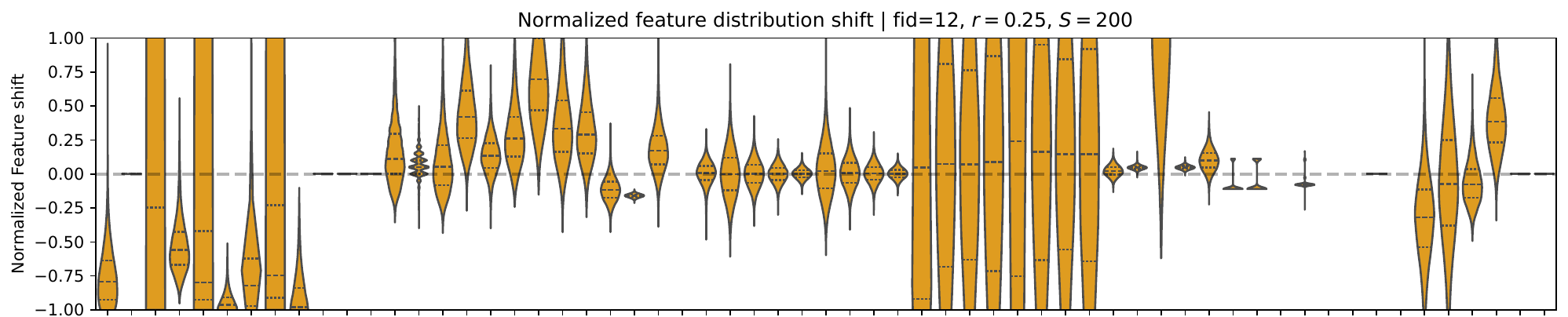}
    \includegraphics[width=.8\linewidth,trim=0cm .7cm 0cm 0cm,clip]{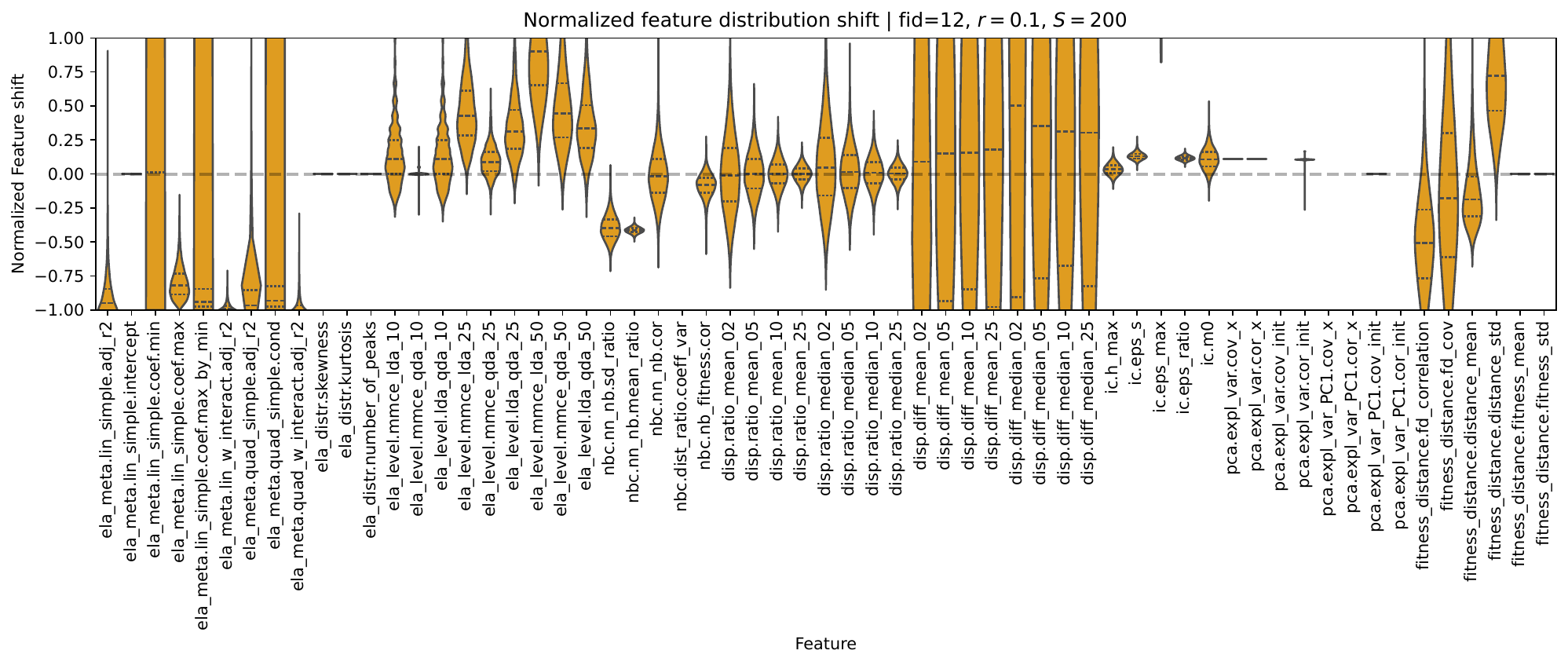}
    \caption{Same as above, for $\boldsymbol{S=200}$.}
    \label{fig:violin_f12_n200}
\end{figure}


\begin{figure}[hbtp]
    \centering
    \includegraphics[width=.8\linewidth,trim=0cm 7.5cm 0cm 0cm,clip]{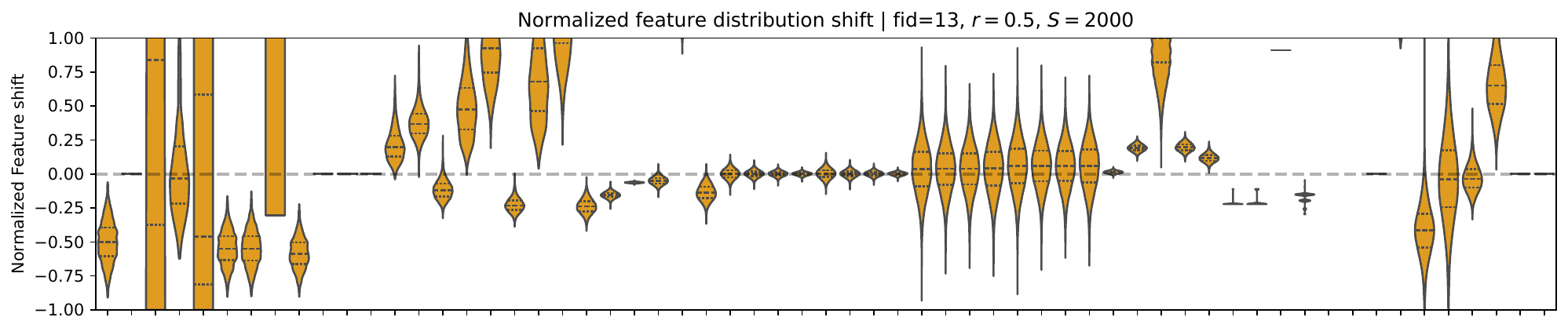}
    \includegraphics[width=.8\linewidth,trim=0cm 7.5cm 0cm 0cm,clip]{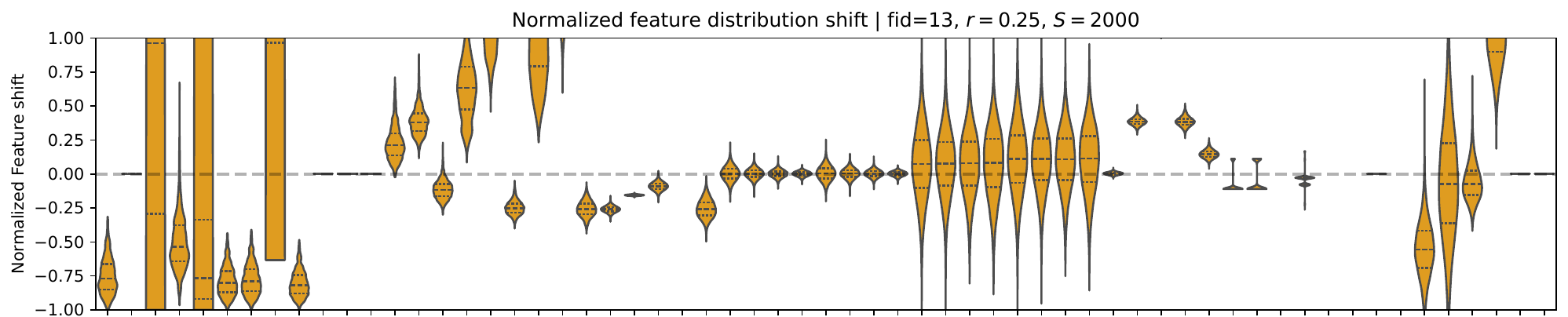}
    \includegraphics[width=.8\linewidth,trim=0cm .7cm 0cm 0cm,clip]{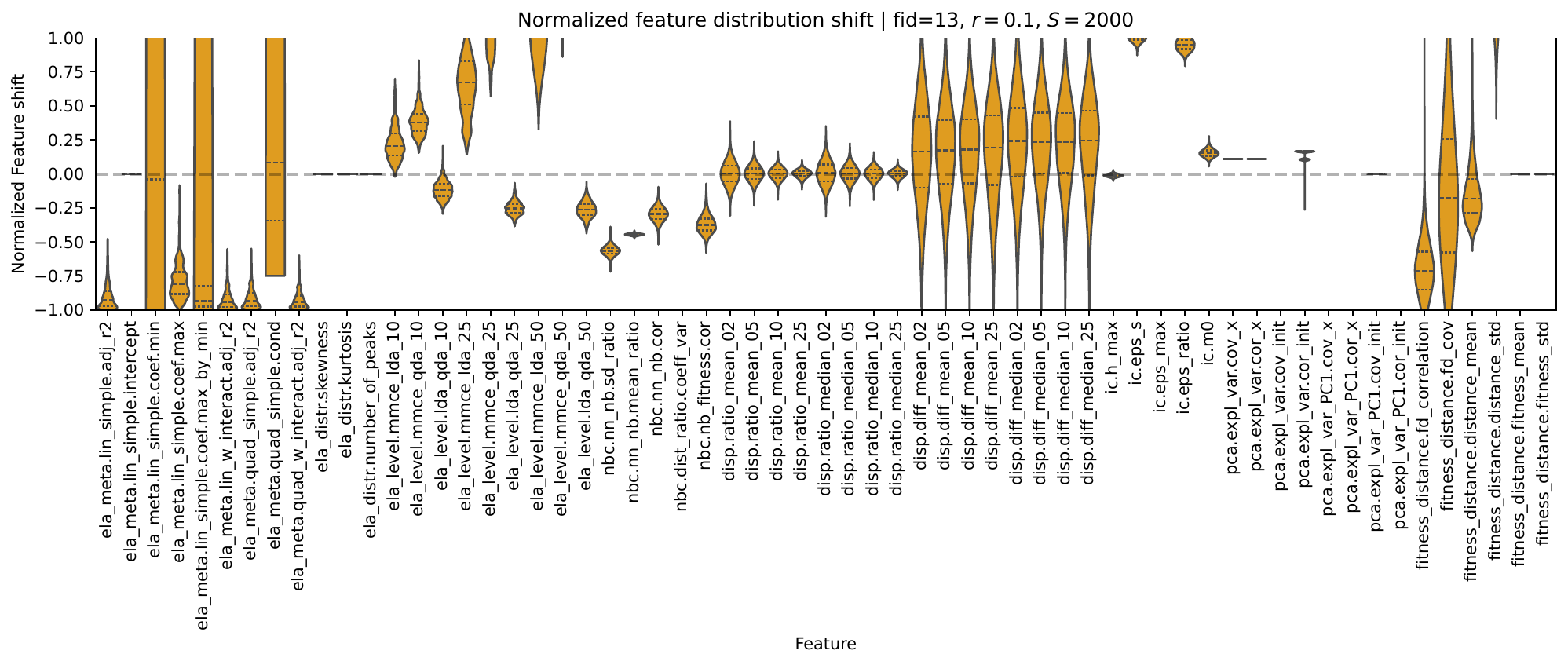}
       \caption{Normalized aggregated feature distribution shift of \textbf{Sharp ridge (f13) function} with $\boldsymbol{S=2000}$ for compression ratios $r=\{0.5,0.25,0.1\}$. The horizontal dashed line denotes a normalized reference corresponding to the median of each feature distribution in the original search space. To enhance visualization, the limits of the Normalized Feature shift has been set to $[-1,1 ]$.}
    \label{fig:violin_f13_n2000}
\end{figure}

\begin{figure}[hbtp]\ContinuedFloat
    \centering
    \includegraphics[width=.8\linewidth,trim=0cm 7.5cm 0cm 0cm,clip]{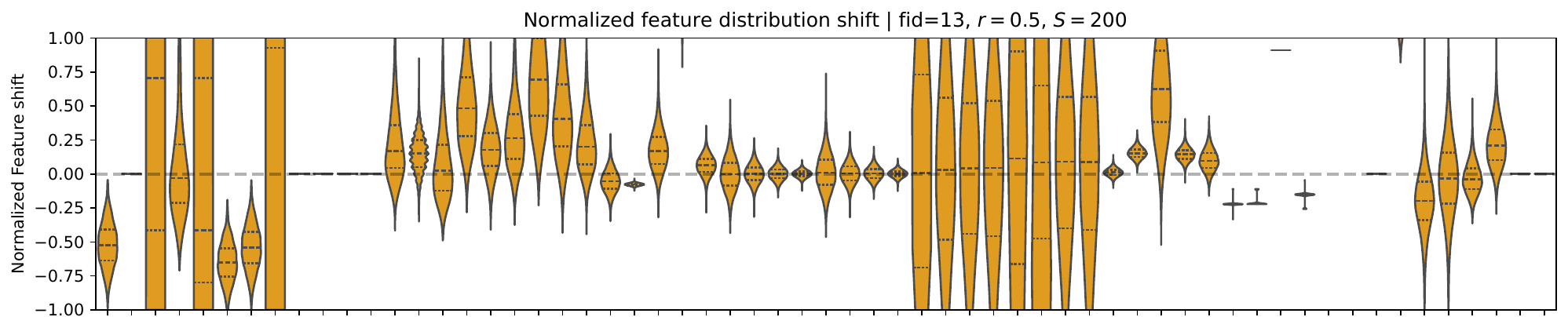}
    \includegraphics[width=.8\linewidth,trim=0cm 7.5cm 0cm 0cm,clip]{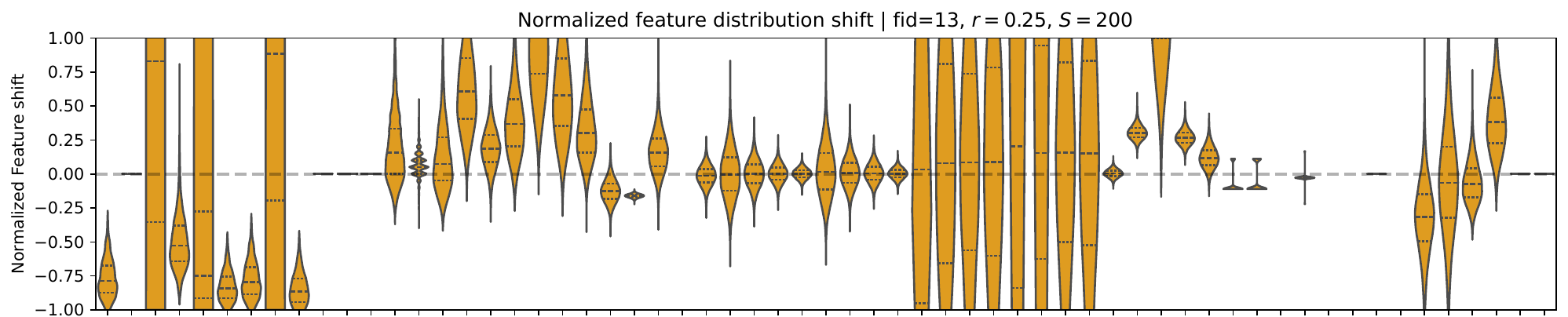}
    \includegraphics[width=.8\linewidth,trim=0cm .7cm 0cm 0cm,clip]{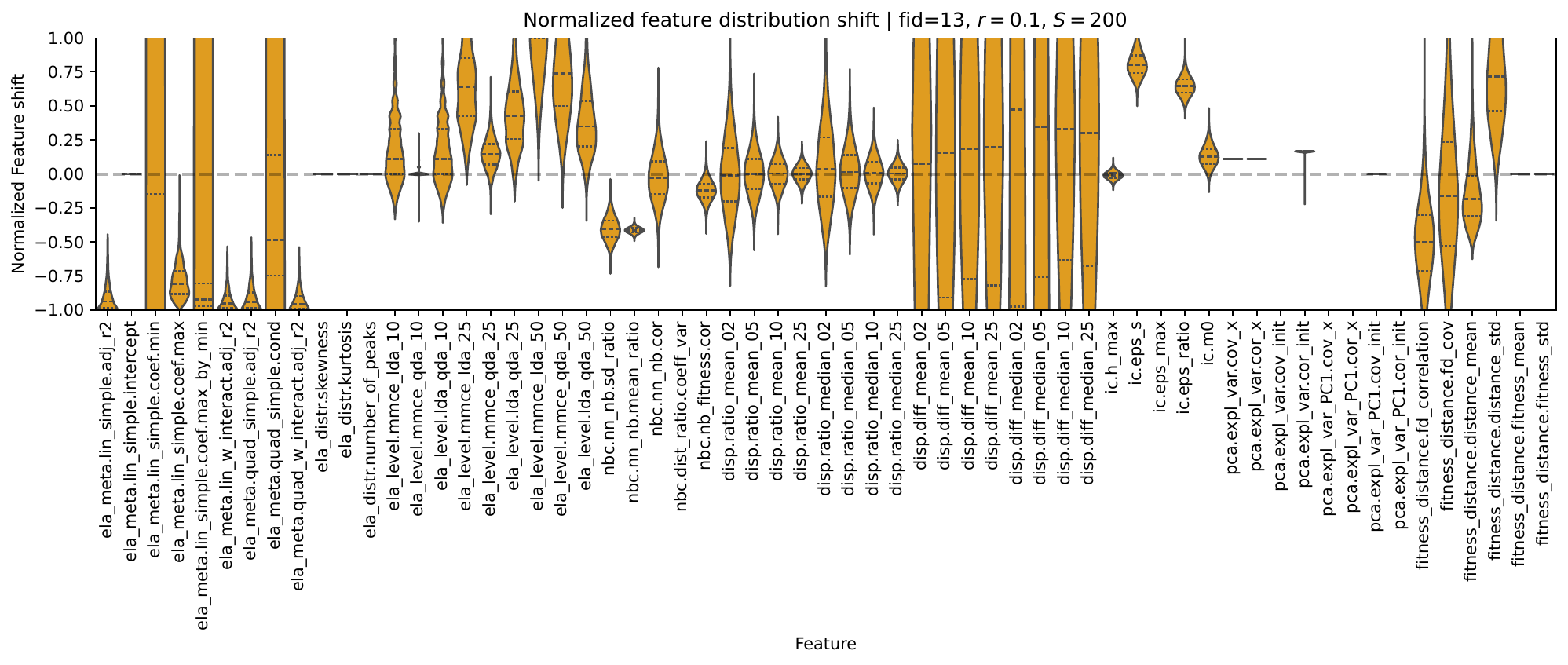}
    \caption{Same as above, for $\boldsymbol{S=200}$.}
    \label{fig:violin_f13_n200}
\end{figure}


\begin{figure}[hbtp]
    \centering
    \includegraphics[width=.8\linewidth,trim=0cm 7.5cm 0cm 0cm,clip]{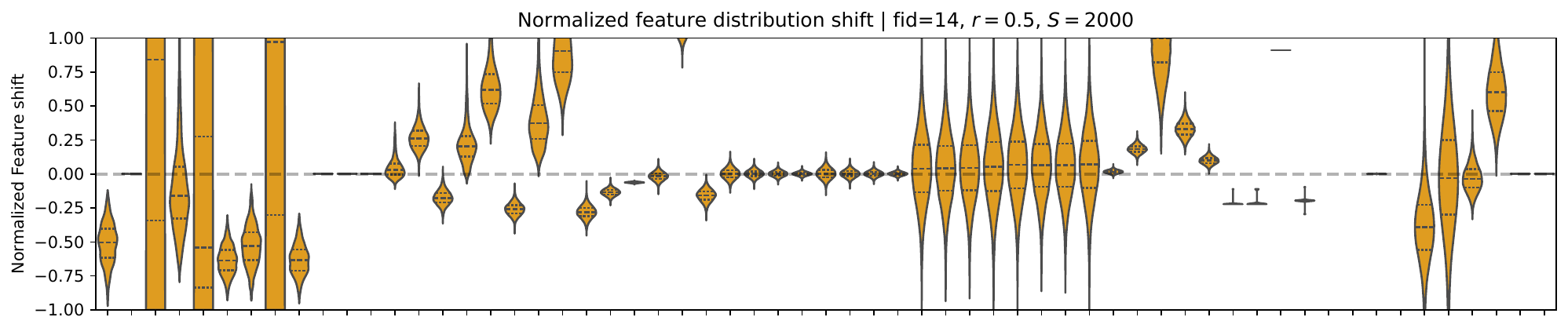}
    \includegraphics[width=.8\linewidth,trim=0cm 7.5cm 0cm 0cm,clip]{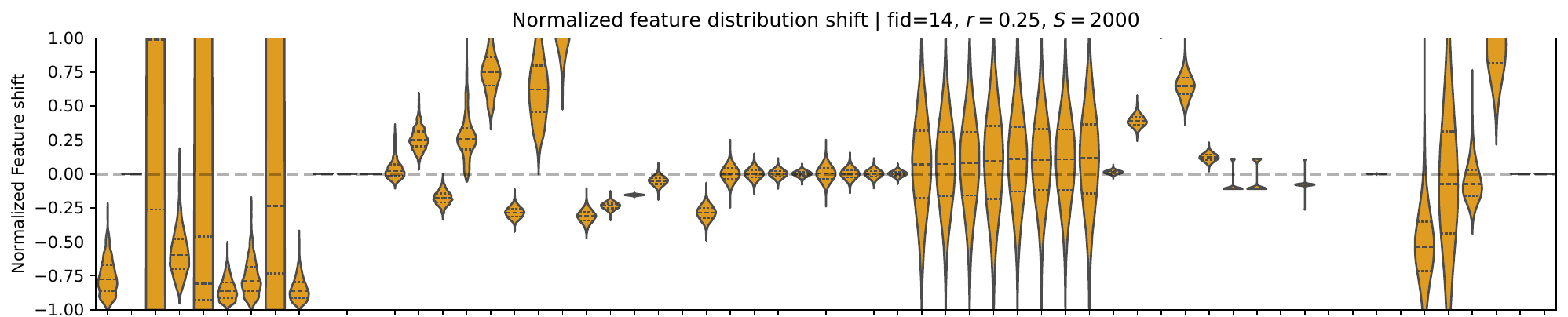}
    \includegraphics[width=.8\linewidth,trim=0cm .7cm 0cm 0cm,clip]{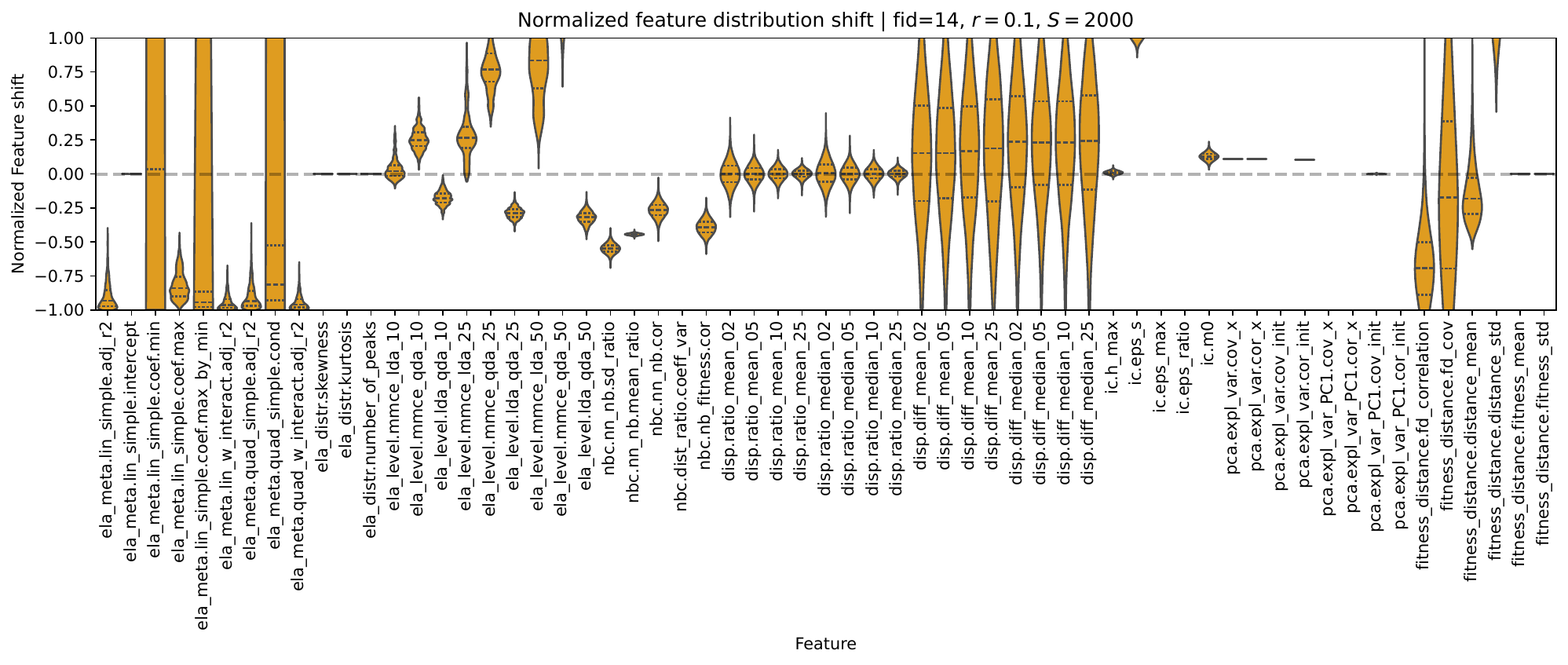}
       \caption{Normalized aggregated feature distribution shift of \textbf{Sum of different powers (f14) function} with $\boldsymbol{S=2000}$ for compression ratios $r=\{0.5,0.25,0.1\}$. The horizontal dashed line denotes a normalized reference corresponding to the median of each feature distribution in the original search space. To enhance visualization, the limits of the Normalized Feature shift has been set to $[-1,1 ]$.}
    \label{fig:violin_f14_n2000}
\end{figure}

\begin{figure}[hbtp]\ContinuedFloat
    \centering
    \includegraphics[width=.8\linewidth,trim=0cm 7.5cm 0cm 0cm,clip]{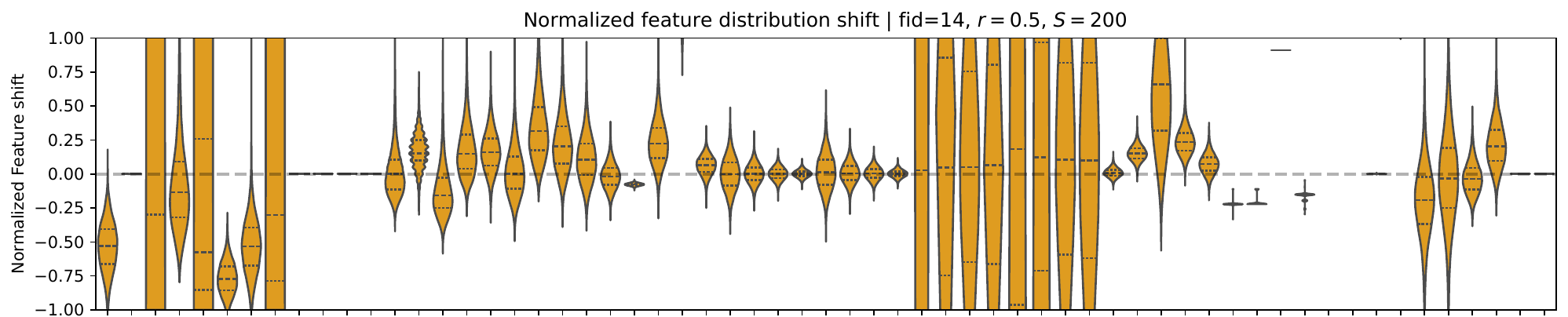}
    \includegraphics[width=.8\linewidth,trim=0cm 7.5cm 0cm 0cm,clip]{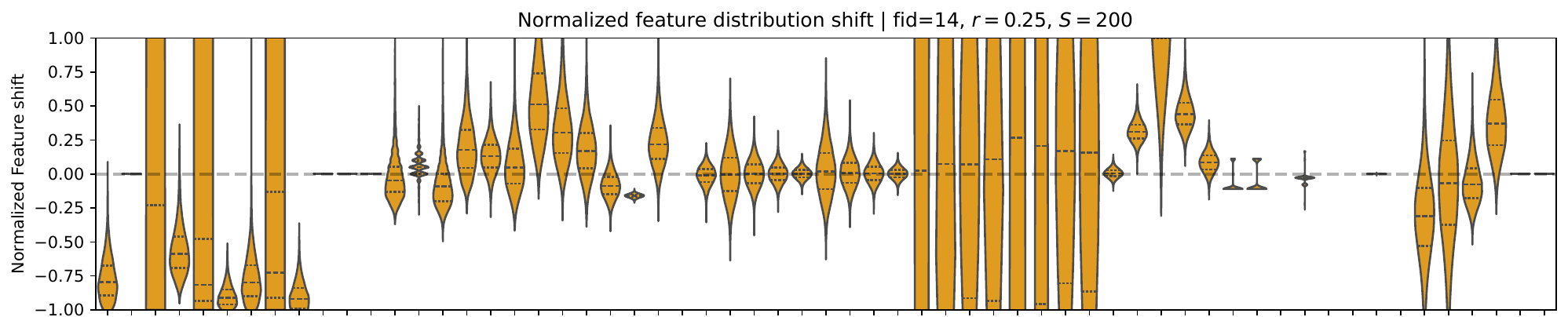}
    \includegraphics[width=.8\linewidth,trim=0cm .7cm 0cm 0cm,clip]{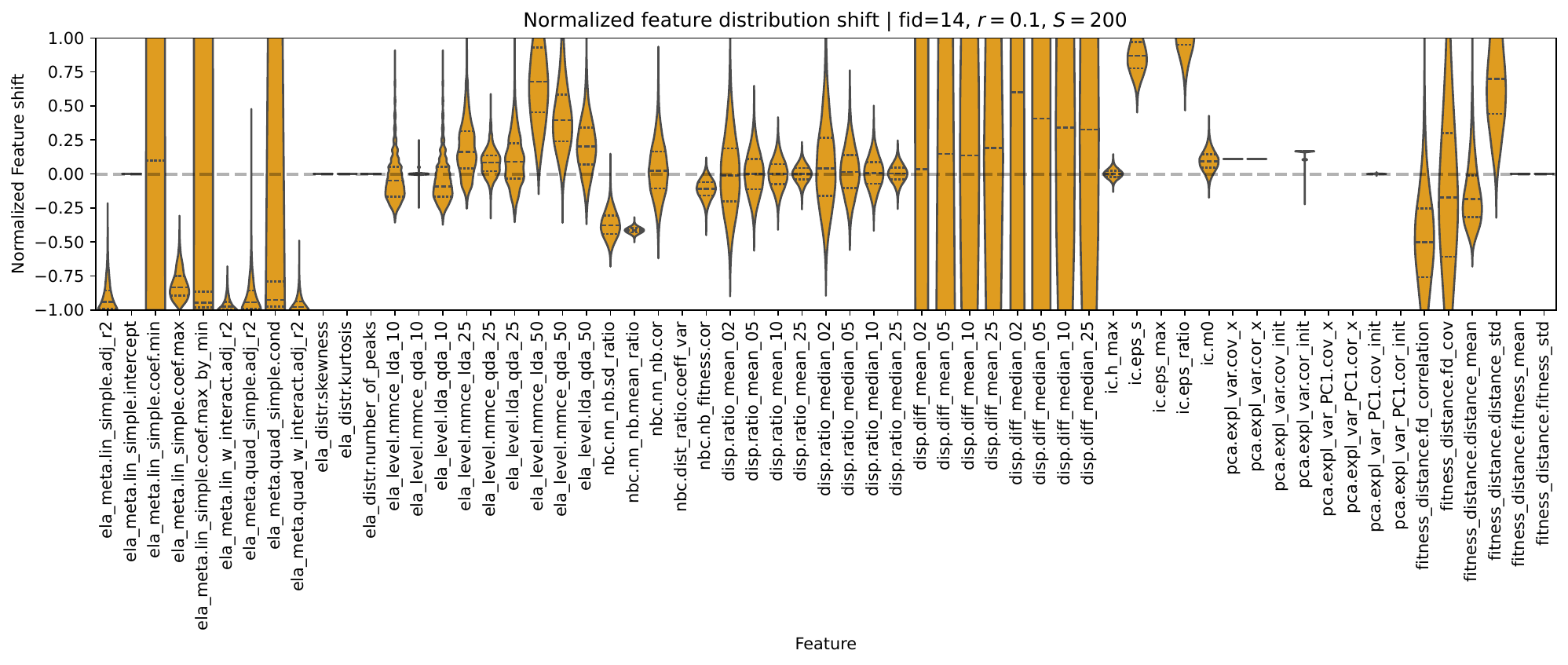}
    \caption{Same as above, for $\boldsymbol{S=200}$.}
    \label{fig:violin_f14_n200}
\end{figure}


\begin{figure}[hbtp]
    \centering
    \includegraphics[width=.8\linewidth,trim=0cm 7.5cm 0cm 0cm,clip]{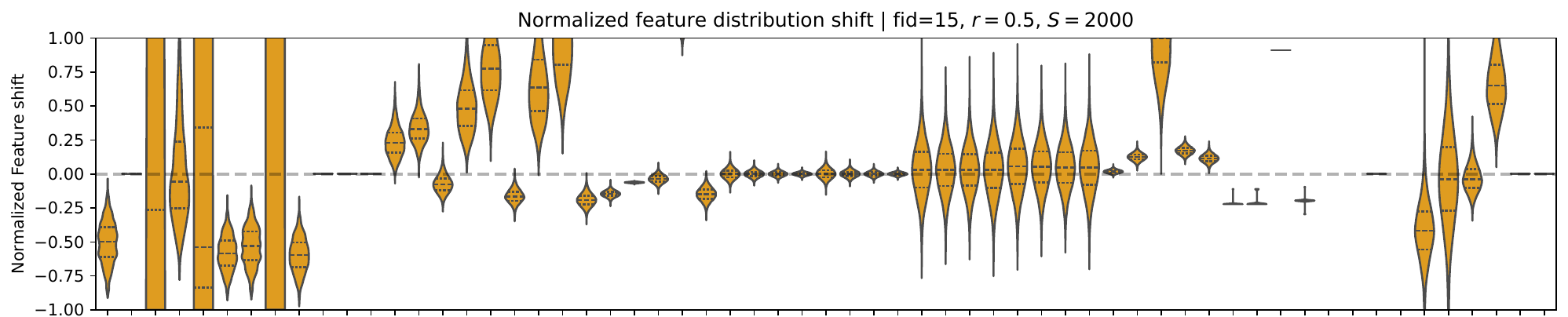}
    \includegraphics[width=.8\linewidth,trim=0cm 7.5cm 0cm 0cm,clip]{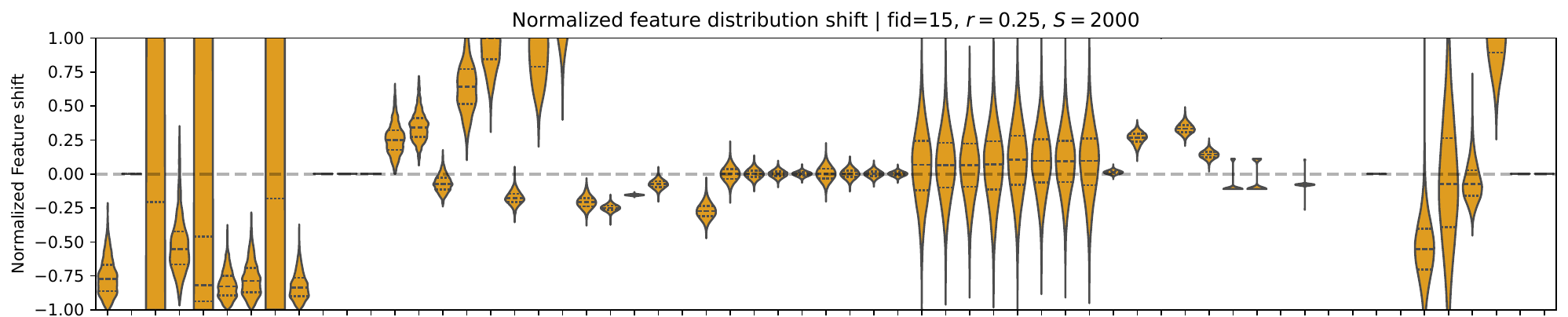}
    \includegraphics[width=.8\linewidth,trim=0cm .7cm 0cm 0cm,clip]{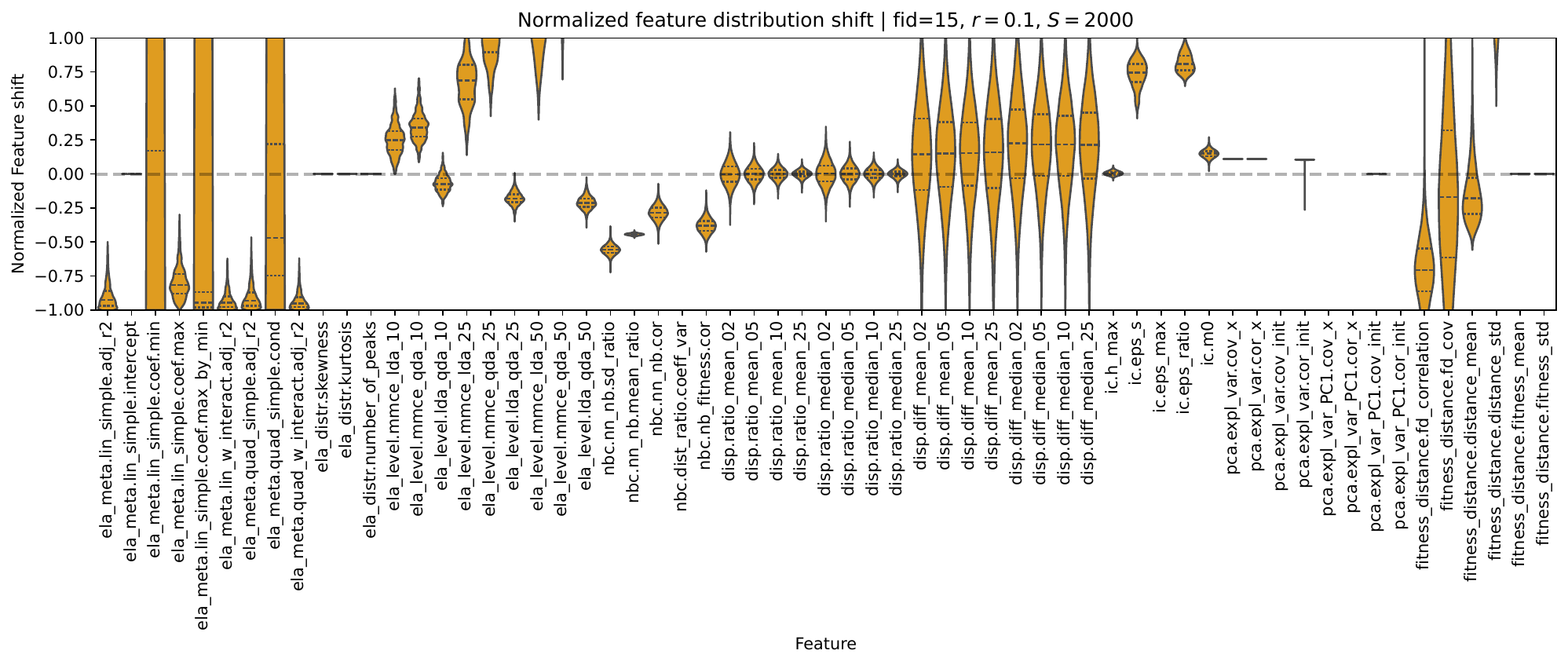}
       \caption{Normalized aggregated feature distribution shift of \textbf{Rastrigin (f15) function} with $\boldsymbol{S=2000}$ for compression ratios $r=\{0.5,0.25,0.1\}$. The horizontal dashed line denotes a normalized reference corresponding to the median of each feature distribution in the original search space. To enhance visualization, the limits of the Normalized Feature shift has been set to $[-1,1 ]$.}
    \label{fig:violin_f15_n2000}
\end{figure}

\begin{figure}[hbtp]\ContinuedFloat
    \centering
    \includegraphics[width=.8\linewidth,trim=0cm 7.5cm 0cm 0cm,clip]{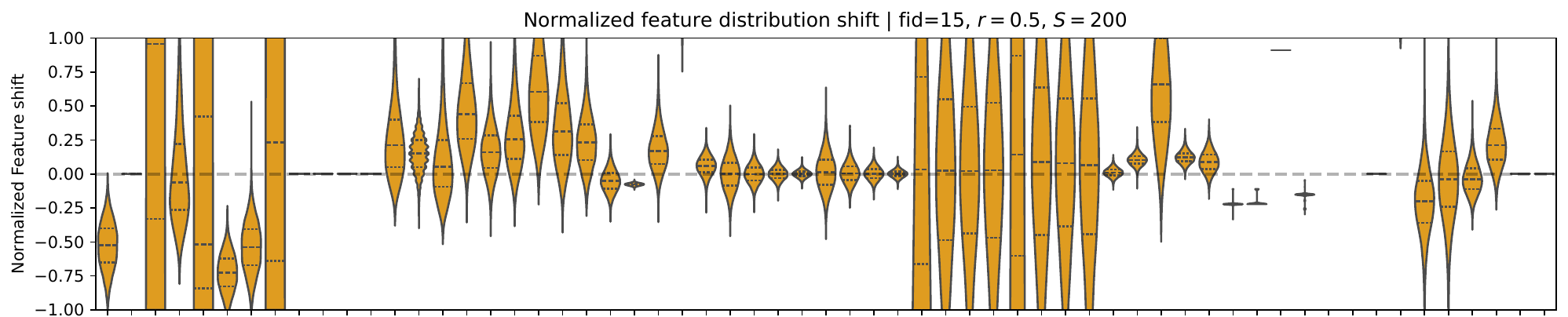}
    \includegraphics[width=.8\linewidth,trim=0cm 7.5cm 0cm 0cm,clip]{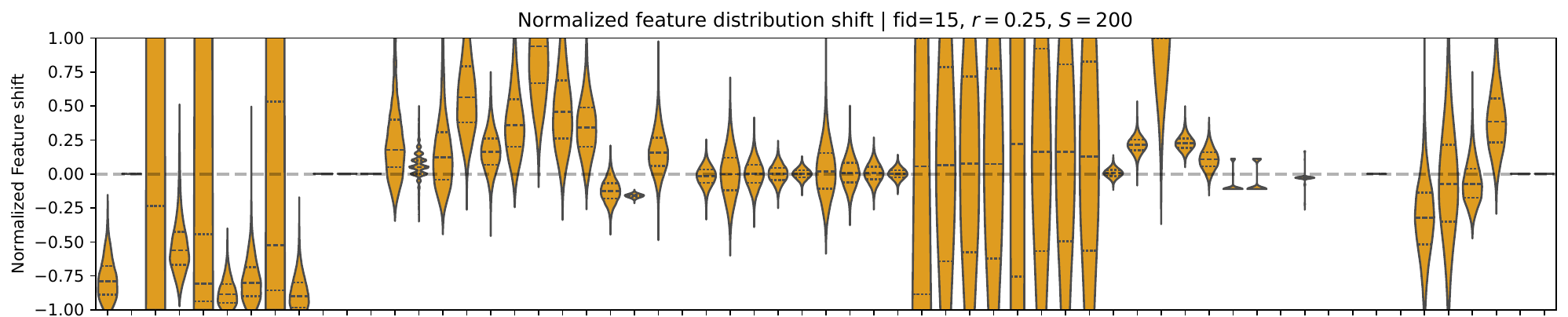}
    \includegraphics[width=.8\linewidth,trim=0cm .7cm 0cm 0cm,clip]{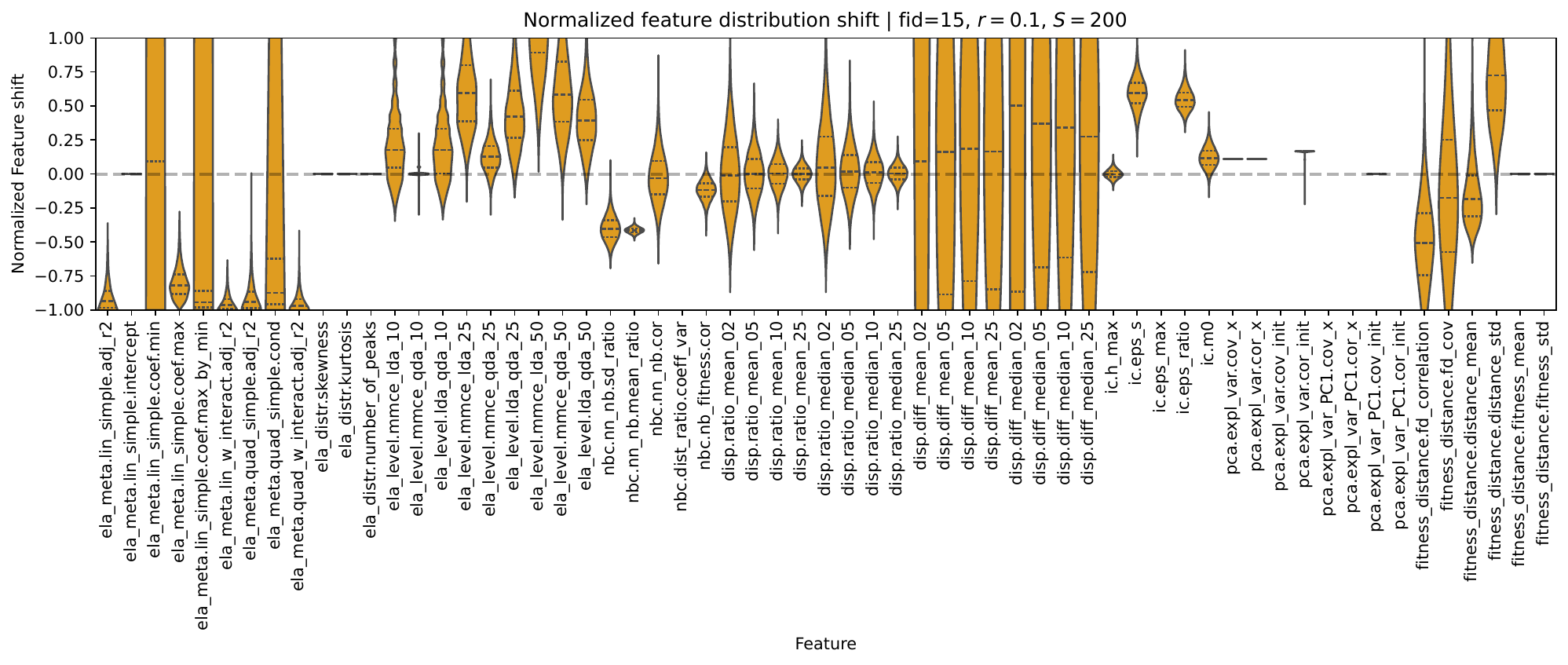}
    \caption{Same as above, for $\boldsymbol{S=200}$.}
    \label{fig:violin_f15_n200}
\end{figure}


\begin{figure}[hbtp]
    \centering
    \includegraphics[width=.8\linewidth,trim=0cm 7.5cm 0cm 0cm,clip]{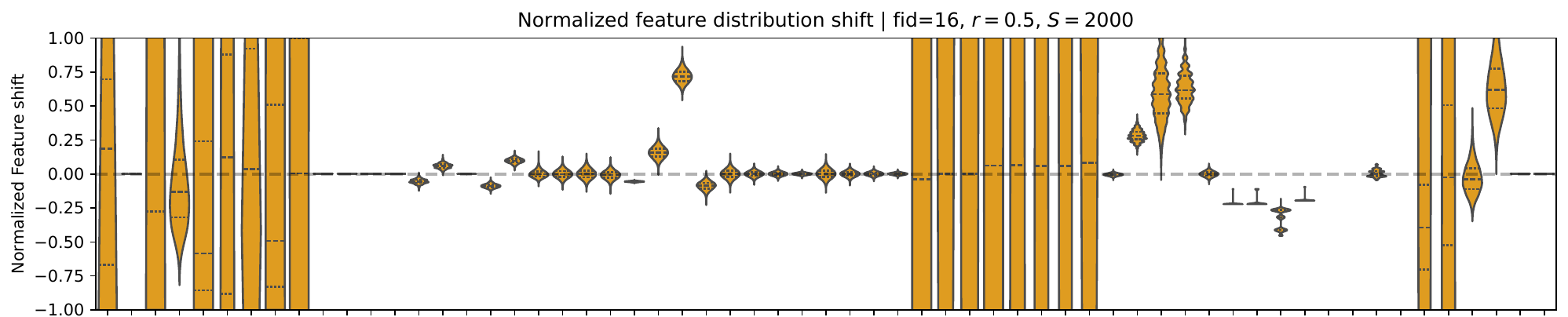}
    \includegraphics[width=.8\linewidth,trim=0cm 7.5cm 0cm 0cm,clip]{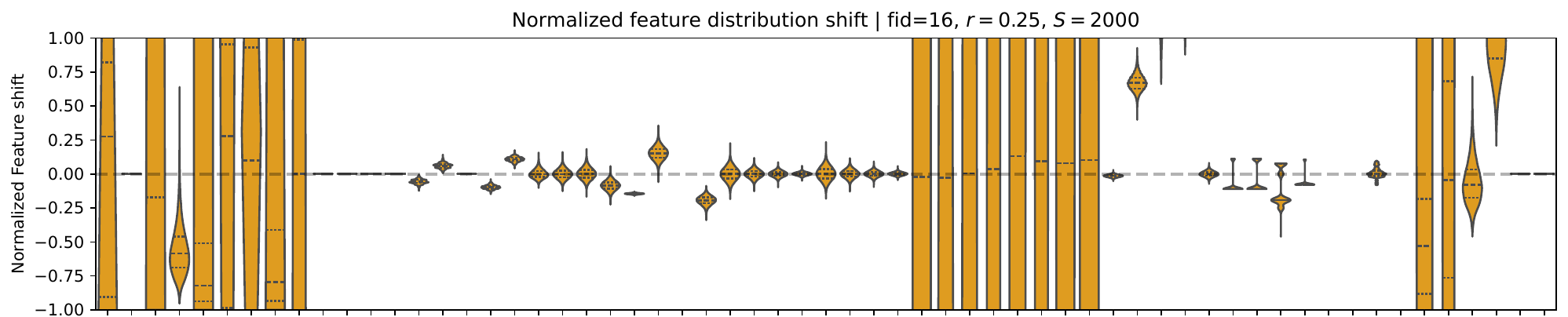}
    \includegraphics[width=.8\linewidth,trim=0cm .7cm 0cm 0cm,clip]{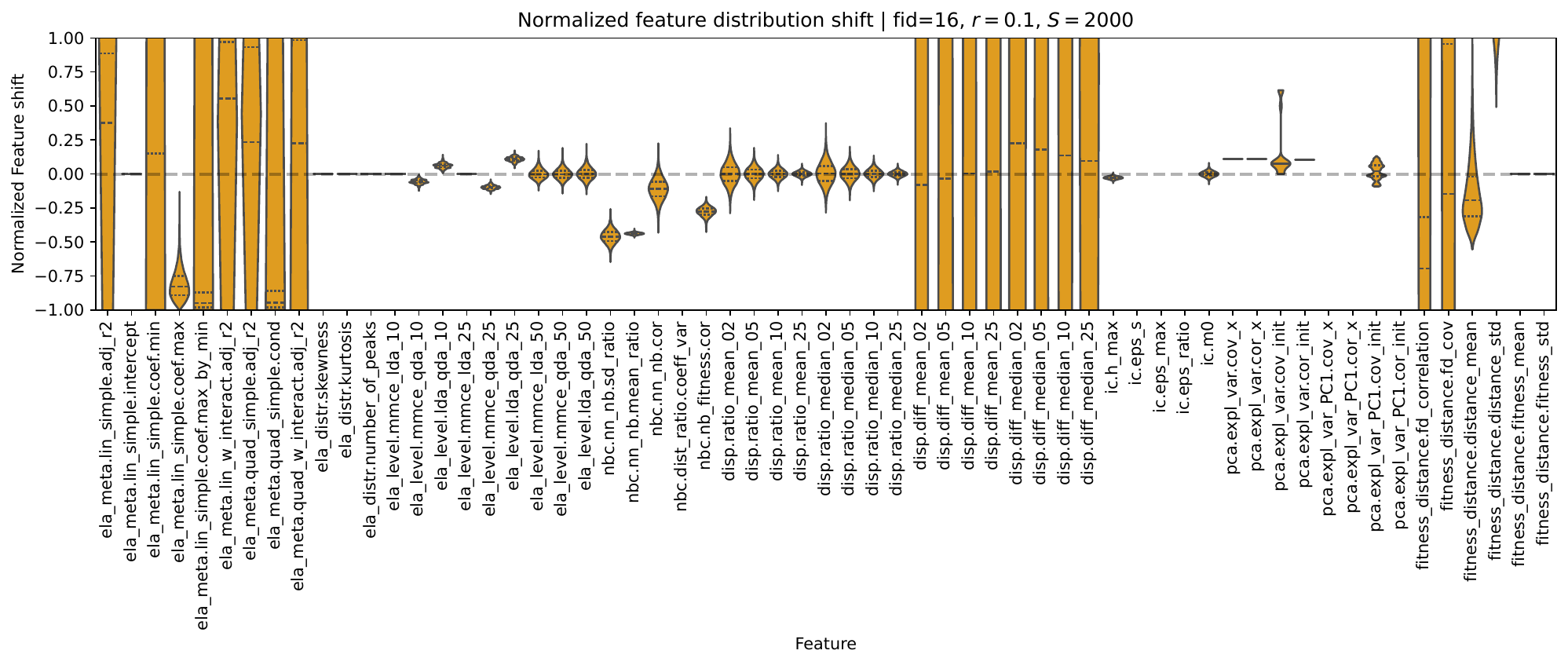}
    \caption{Normalized aggregated feature distribution shift of \textbf{Weierstrass (f16) function} with $\boldsymbol{S=2000}$ for compression ratios $r=\{0.5,0.25,0.1\}$. The horizontal dashed line denotes a normalized reference corresponding to the median of each feature distribution in the original search space. To enhance visualization, the limits of the Normalized Feature shift has been set to $[-1,1 ]$.}
    \label{fig:violin_f16_n2000}
\end{figure}

\begin{figure}[hbtp]
    \centering
    \includegraphics[width=.8\linewidth,trim=0cm 7.5cm 0cm 0cm,clip]{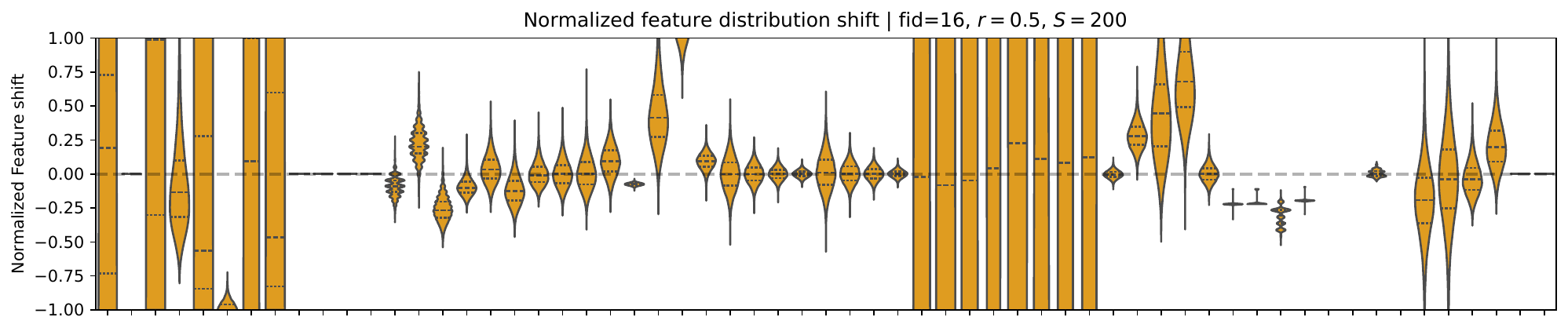}
    \includegraphics[width=.8\linewidth,trim=0cm 7.5cm 0cm 0cm,clip]{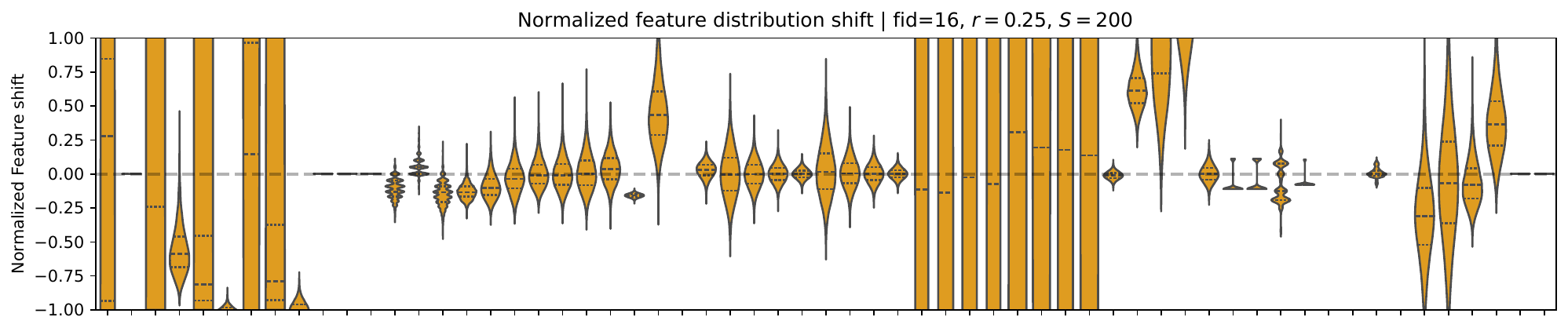}
    \includegraphics[width=.8\linewidth,trim=0cm .7cm 0cm 0cm,clip]{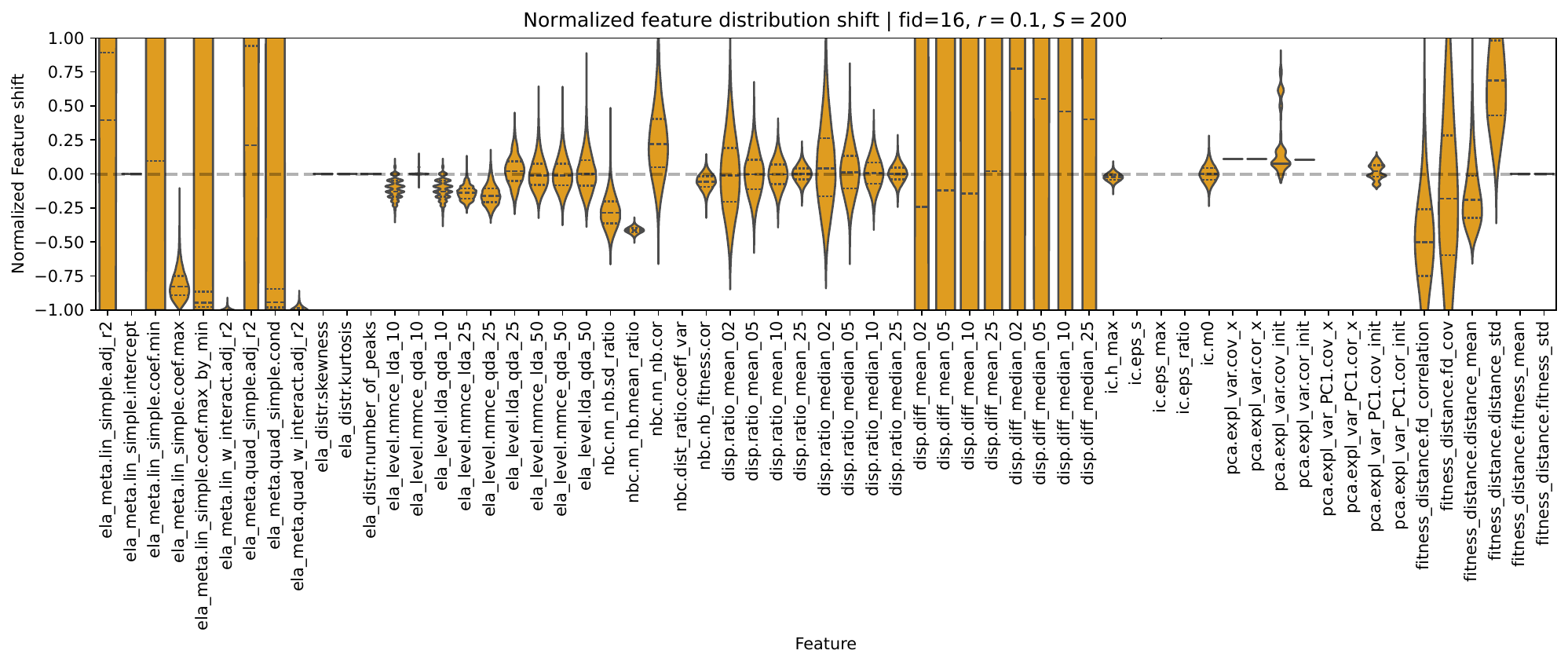}
    \caption{Same as above, for $\boldsymbol{S=200}$.}
    \label{fig:violin_f16_n200}
\end{figure}


\begin{figure}[hbtp]
    \centering
    \includegraphics[width=.8\linewidth,trim=0cm 7.5cm 0cm 0cm,clip]{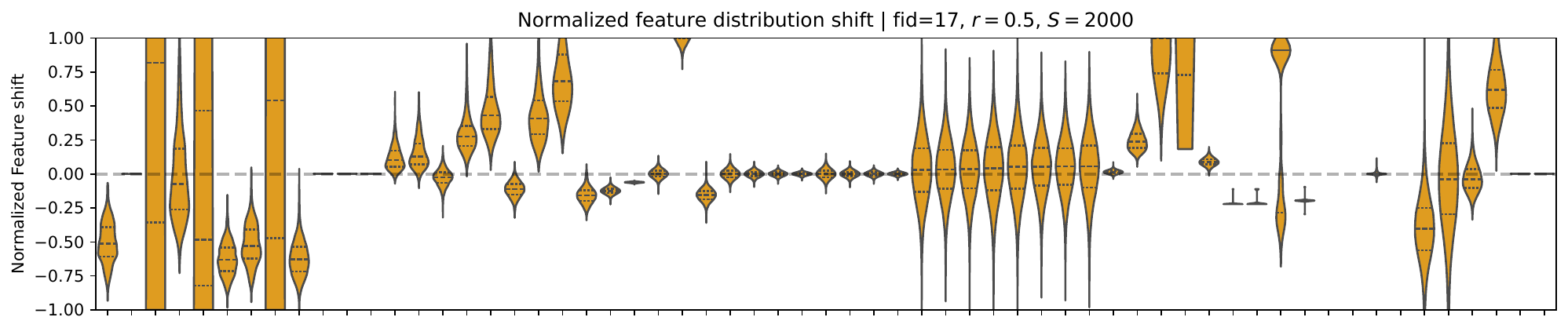}
    \includegraphics[width=.8\linewidth,trim=0cm 7.5cm 0cm 0cm,clip]{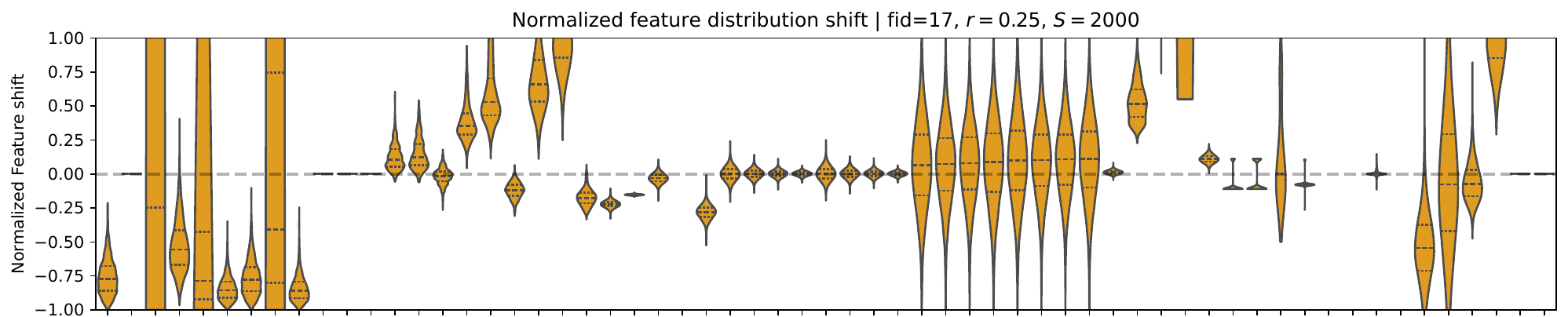}
    \includegraphics[width=.8\linewidth,trim=0cm .7cm 0cm 0cm,clip]{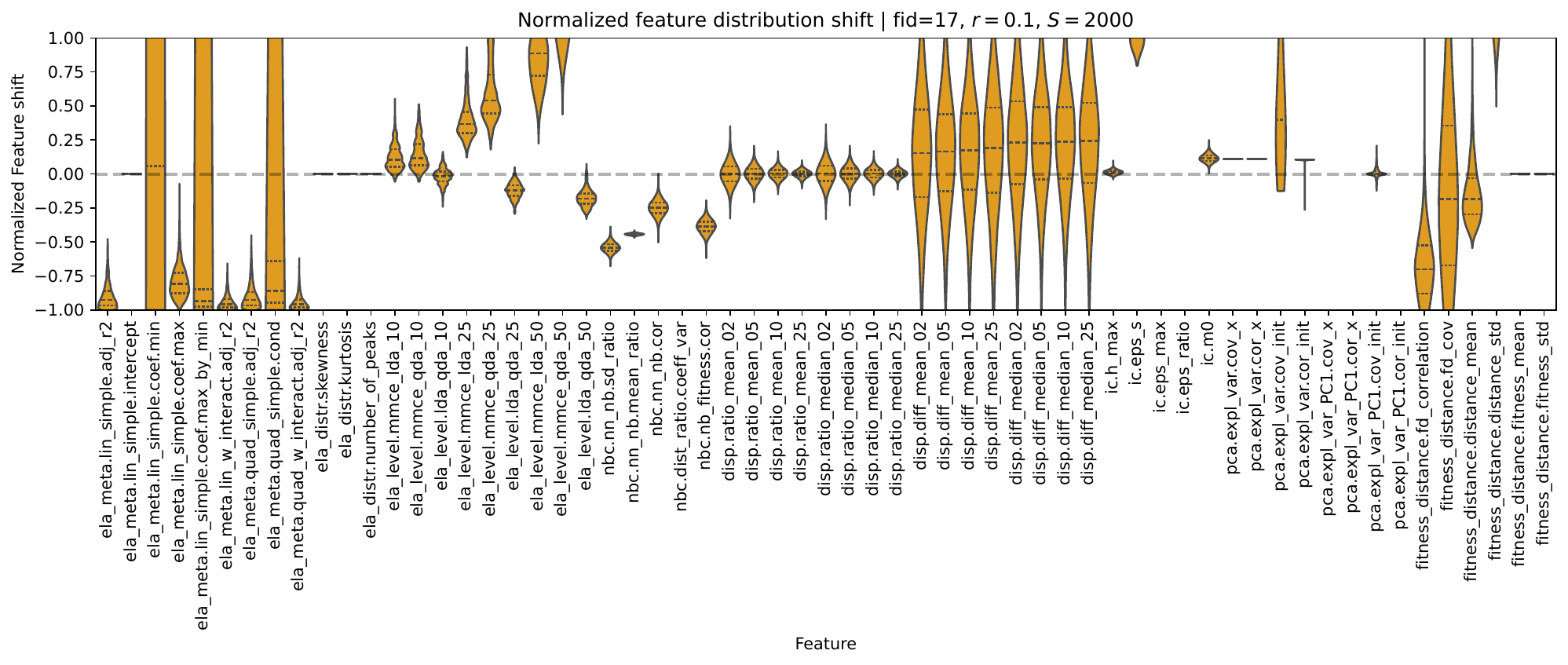}
    \caption{Normalized aggregated feature distribution shift of \textbf{Schaffer F7, condition 10 (f17) function} with $\boldsymbol{S=2000}$ for compression ratios $r=\{0.5,0.25,0.1\}$. The horizontal dashed line denotes a normalized reference corresponding to the median of each feature distribution in the original search space. To enhance visualization, the limits of the Normalized Feature shift has been set to $[-1,1 ]$.}
    \label{fig:violin_f17_n2000}
\end{figure}

\begin{figure}[hbtp]
    \centering
    \includegraphics[width=.8\linewidth,trim=0cm 7.5cm 0cm 0cm,clip]{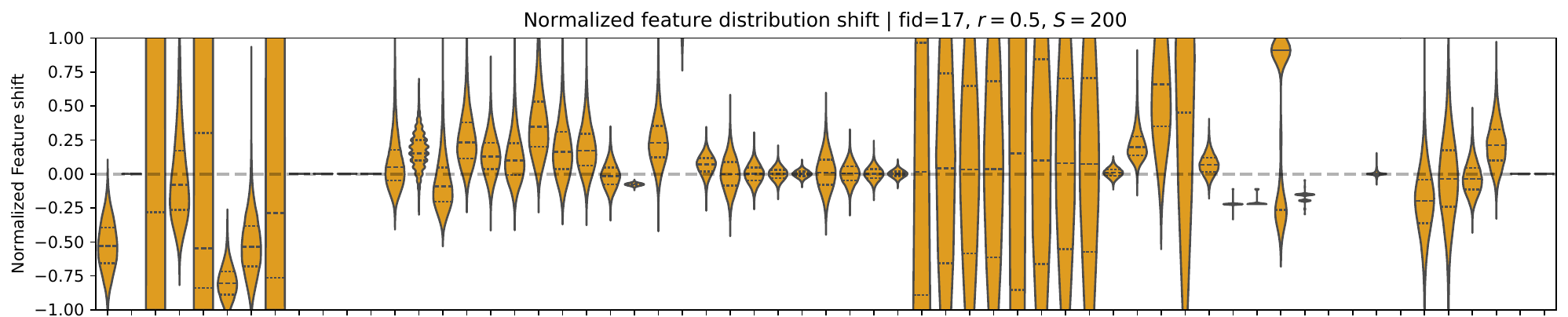}
    \includegraphics[width=.8\linewidth,trim=0cm 7.5cm 0cm 0cm,clip]{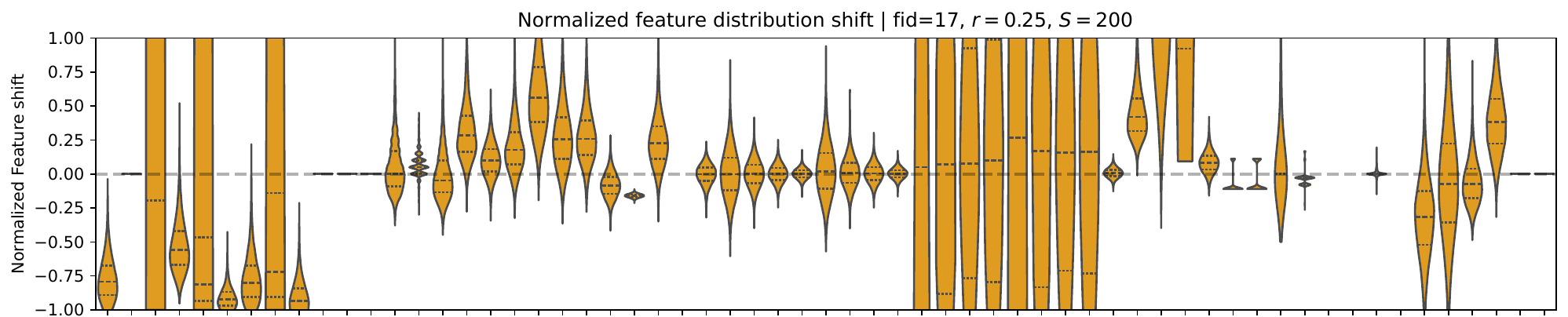}
    \includegraphics[width=.8\linewidth,trim=0cm .7cm 0cm 0cm,clip]{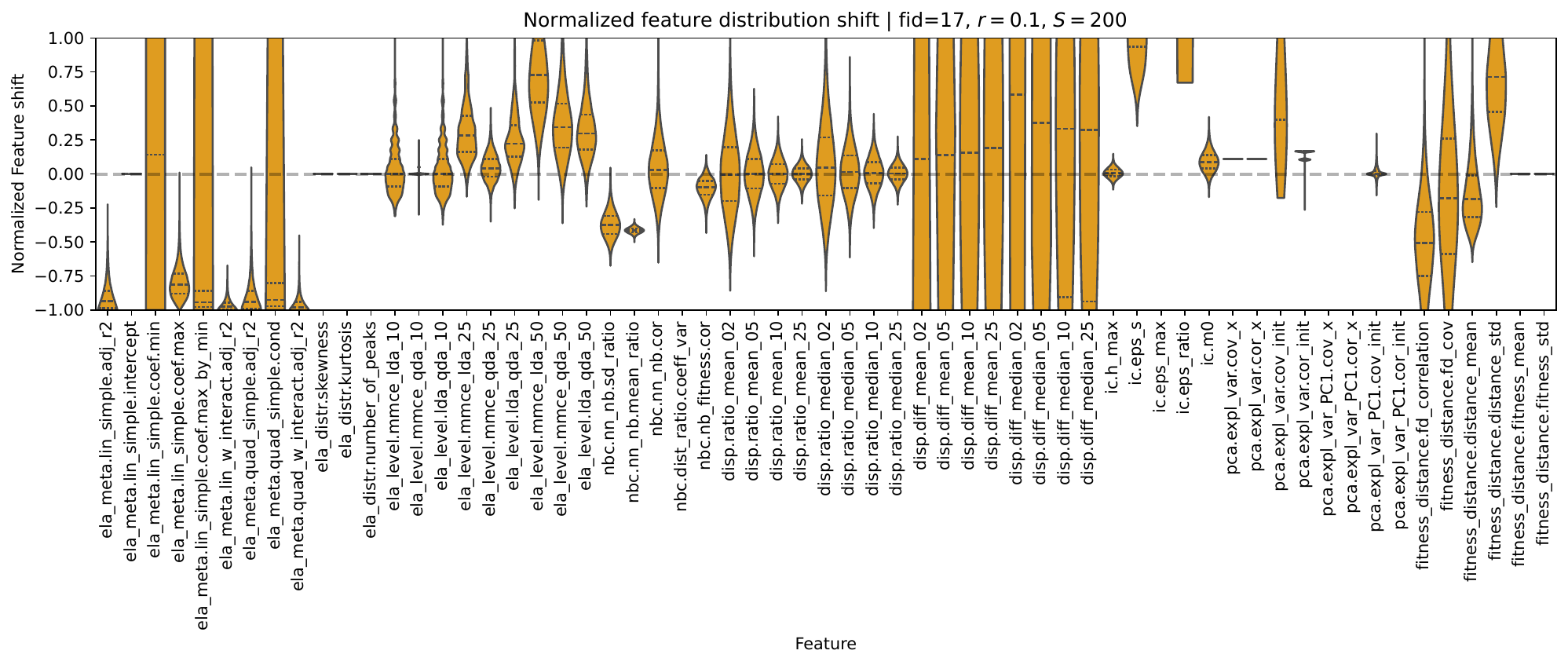}
    \caption{Same as above, for $\boldsymbol{S=200}$.}
    \label{fig:violin_f17_n200}
\end{figure}


\begin{figure}[hbtp]
    \centering
    \includegraphics[width=.8\linewidth,trim=0cm 7.5cm 0cm 0cm,clip]{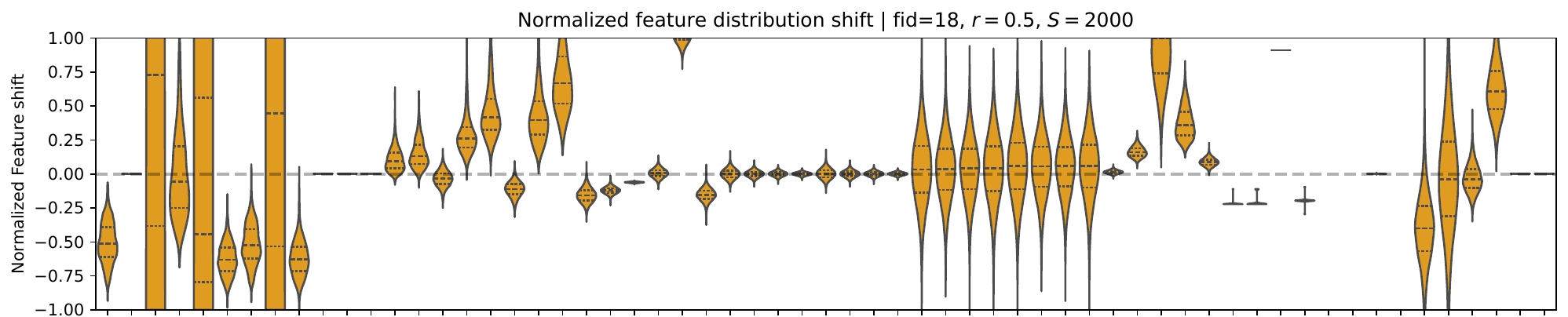}
    \includegraphics[width=.8\linewidth,trim=0cm 7.5cm 0cm 0cm,clip]{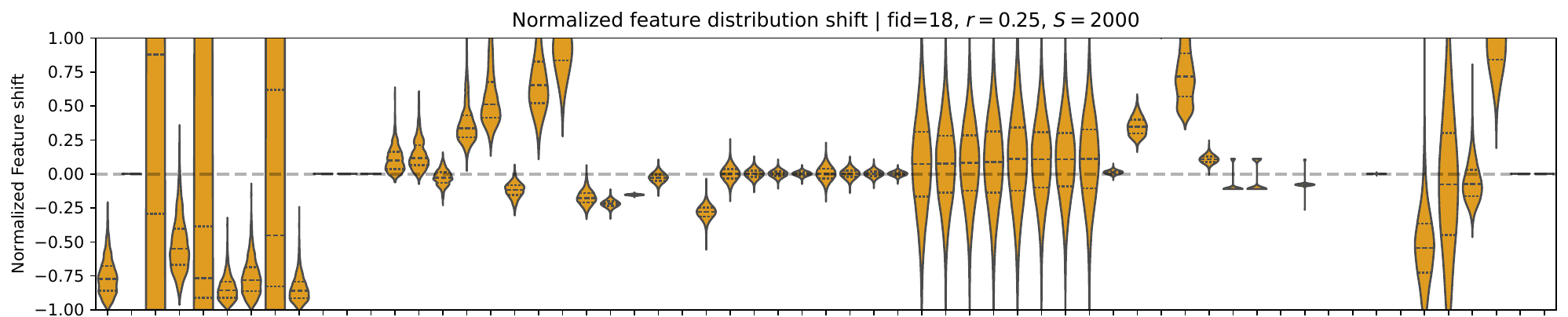}
    \includegraphics[width=.8\linewidth,trim=0cm .7cm 0cm 0cm,clip]{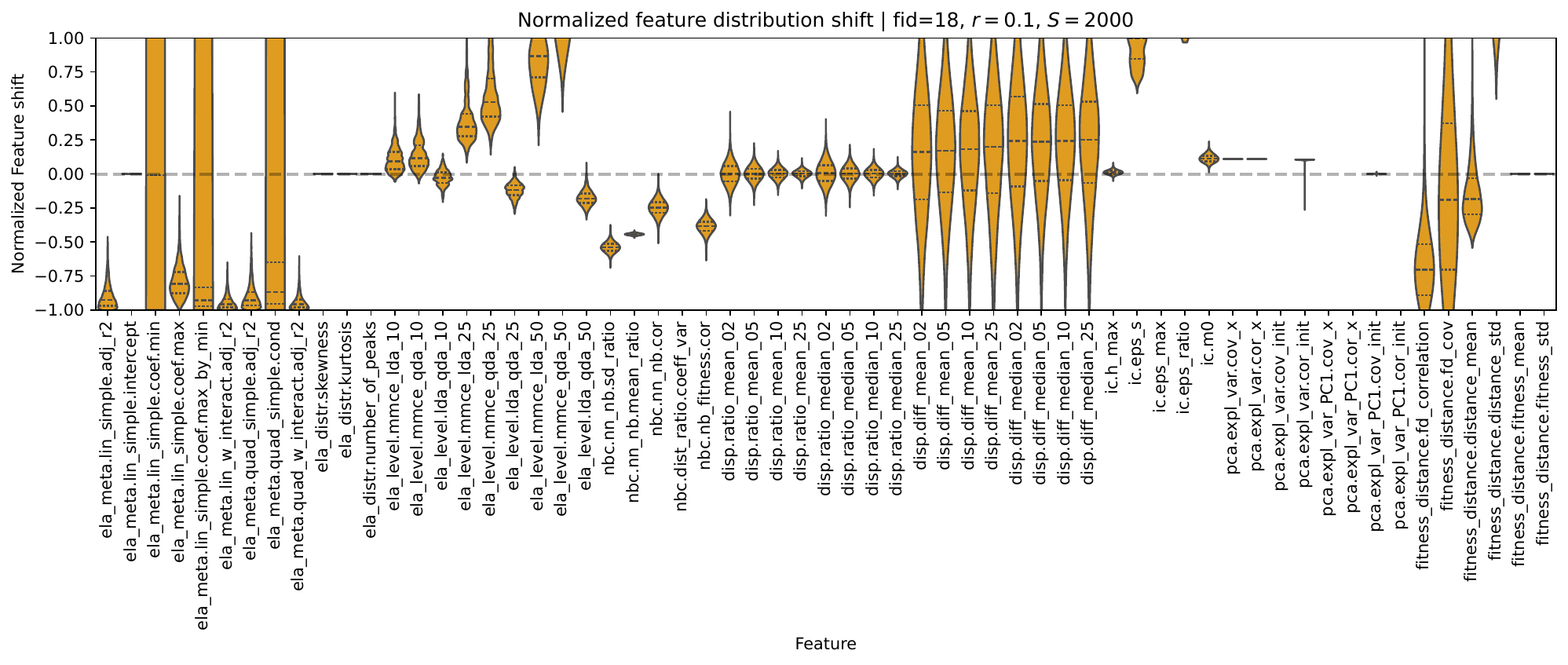}
    \caption{Normalized aggregated feature distribution shift of \textbf{Schaffer F7, condition 1000 (f18) function} with $\boldsymbol{S=2000}$ for compression ratios $r=\{0.5,0.25,0.1\}$. The horizontal dashed line denotes a normalized reference corresponding to the median of each feature distribution in the original search space. To enhance visualization, the limits of the Normalized Feature shift has been set to $[-1,1 ]$.}
    \label{fig:violin_f18_n2000}
\end{figure}

\begin{figure}[hbtp]
    \centering
    \includegraphics[width=.8\linewidth,trim=0cm 7.5cm 0cm 0cm,clip]{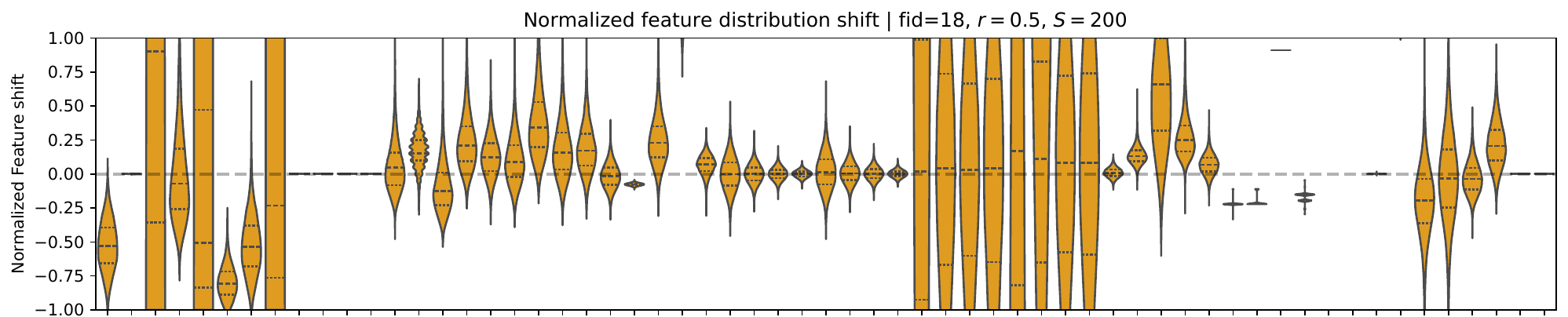}
    \includegraphics[width=.8\linewidth,trim=0cm 7.5cm 0cm 0cm,clip]{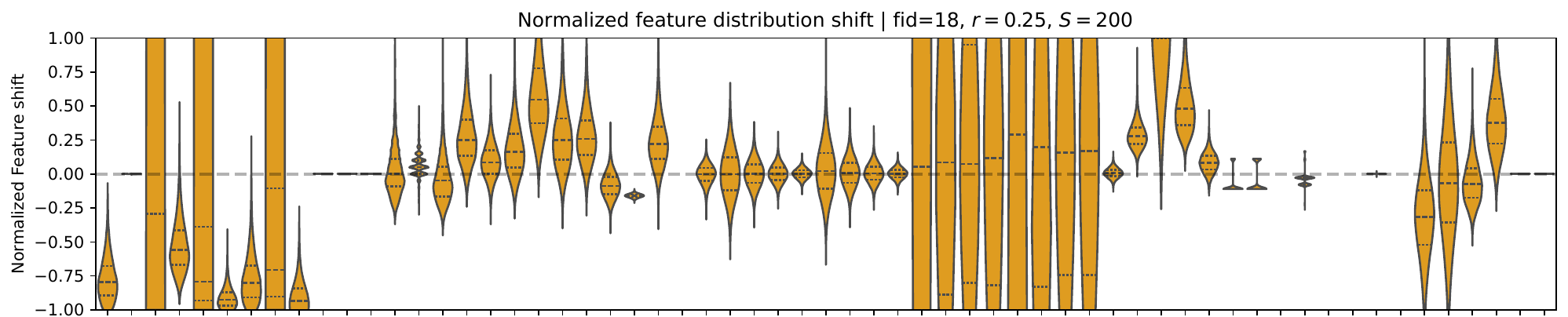}
    \includegraphics[width=.8\linewidth,trim=0cm .7cm 0cm 0cm,clip]{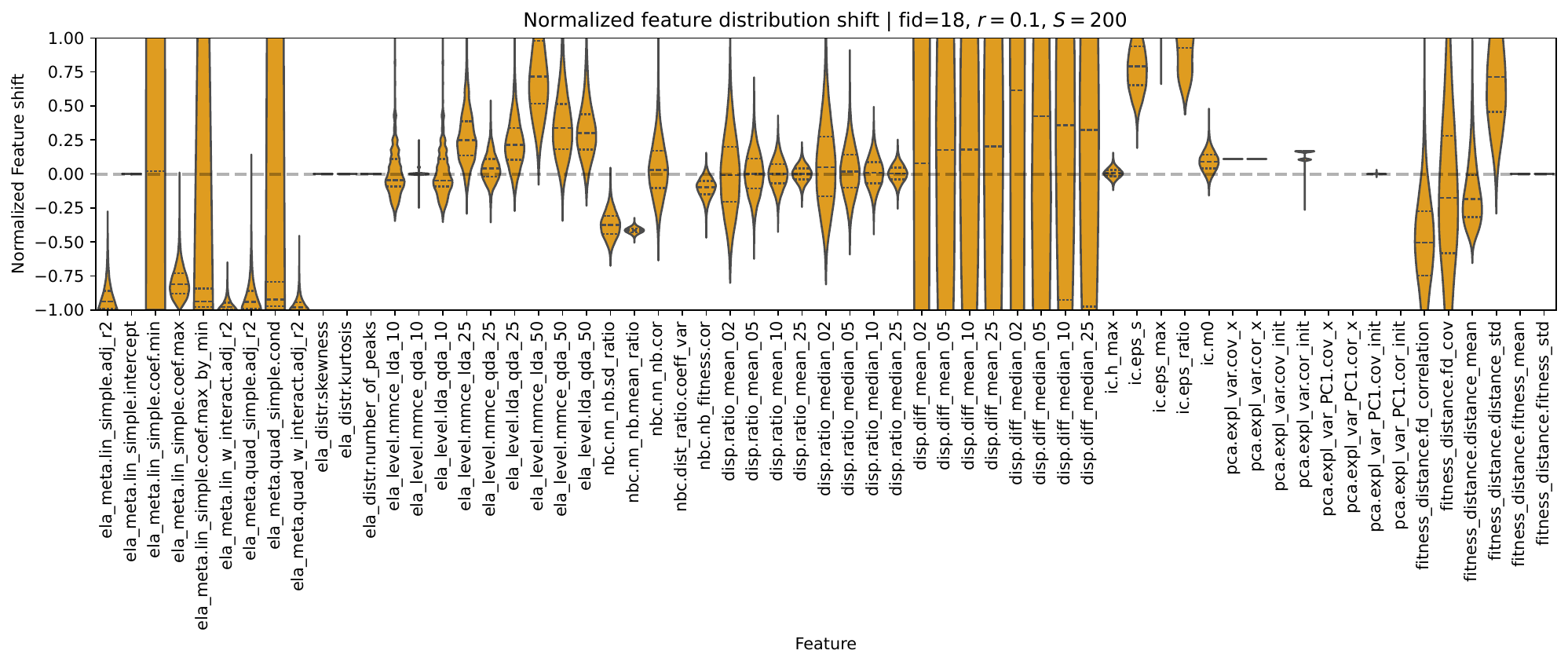}
    \caption{Same as above, for $\boldsymbol{S=200}$.}
    \label{fig:violin_f18_n200}
\end{figure}


\begin{figure}[hbtp]
    \centering
    \includegraphics[width=.8\linewidth,trim=0cm 7.5cm 0cm 0cm,clip]{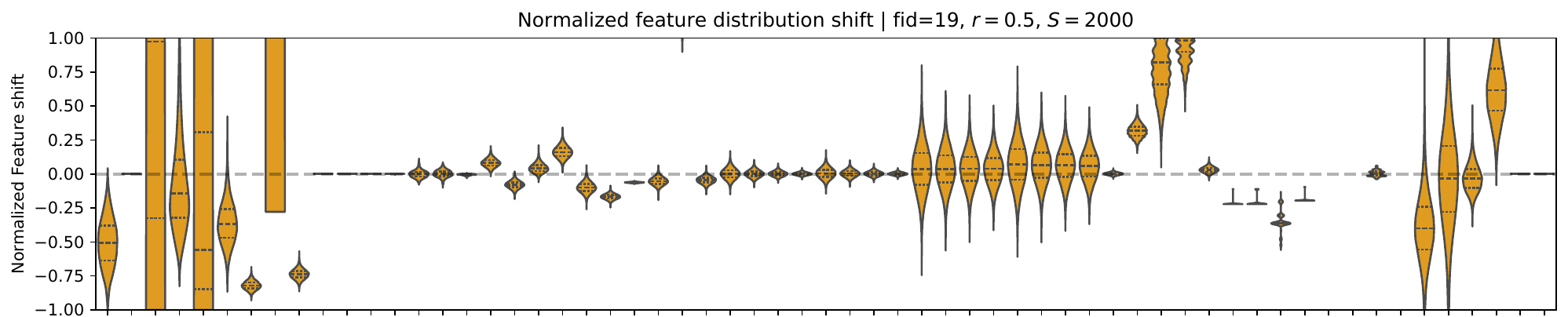}
    \includegraphics[width=.8\linewidth,trim=0cm 7.5cm 0cm 0cm,clip]{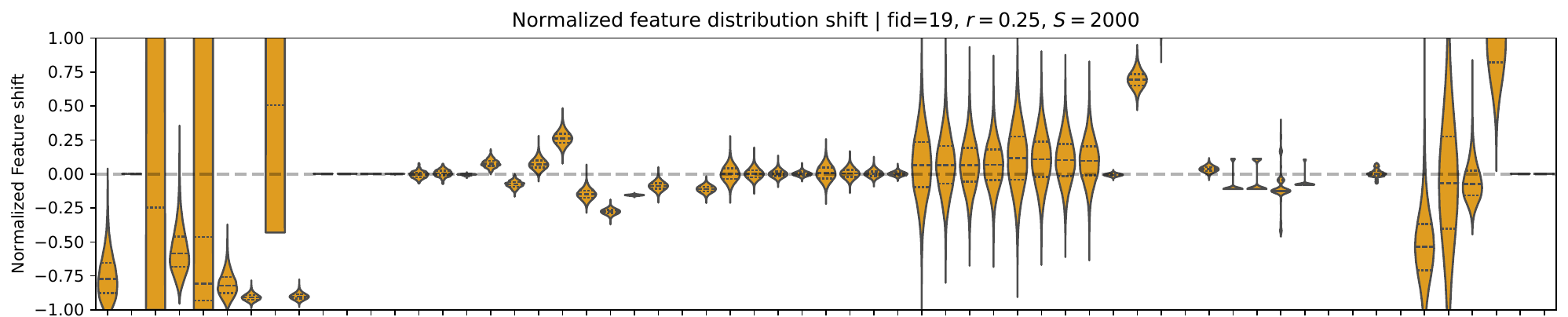}
    \includegraphics[width=.8\linewidth,trim=0cm .7cm 0cm 0cm,clip]{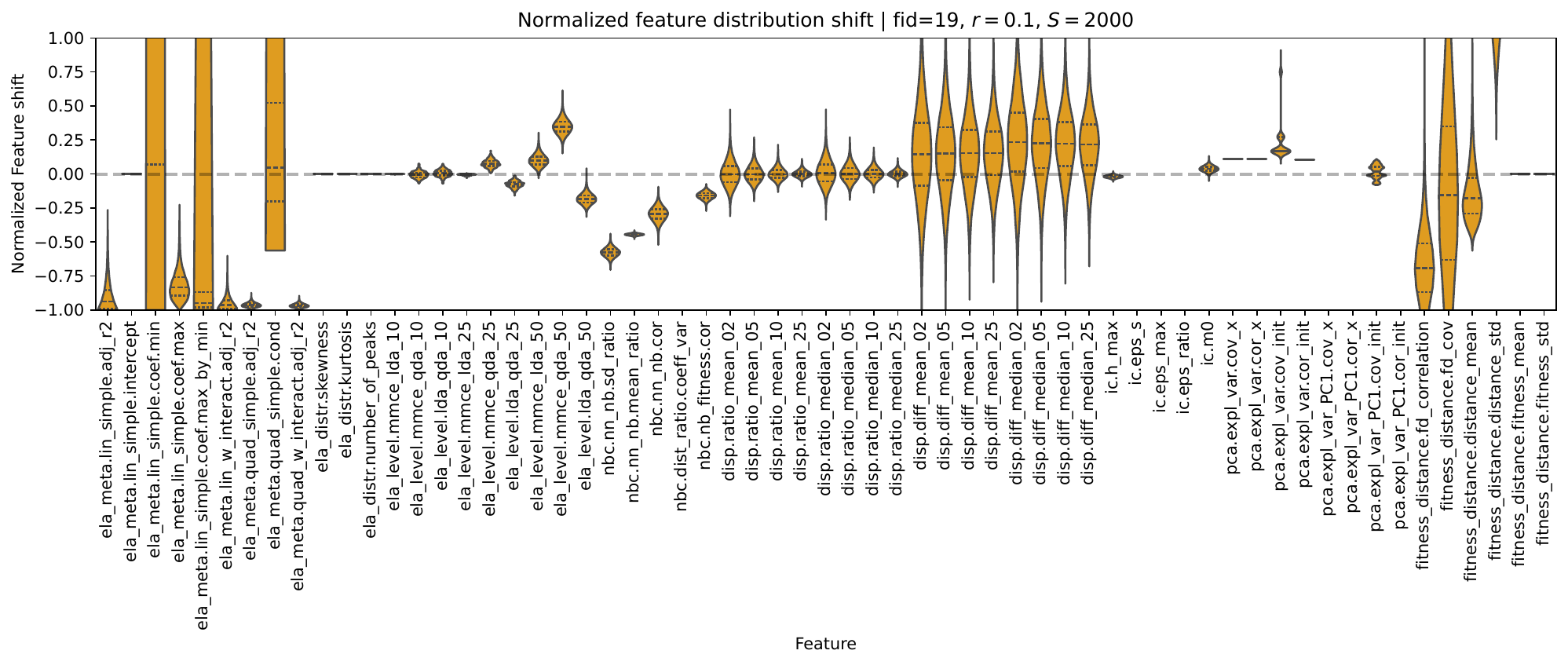}
    \caption{Normalized aggregated feature distribution shift of \textbf{Griewank-Rosenbrock F8F2 (f19) function} with $\boldsymbol{S=2000}$ for compression ratios $r=\{0.5,0.25,0.1\}$. The horizontal dashed line denotes a normalized reference corresponding to the median of each feature distribution in the original search space. To enhance visualization, the limits of the Normalized Feature shift has been set to $[-1,1 ]$.}
    \label{fig:violin_f19_n2000}
\end{figure}

\begin{figure}[hbtp]
    \centering
    \includegraphics[width=.8\linewidth,trim=0cm 7.5cm 0cm 0cm,clip]{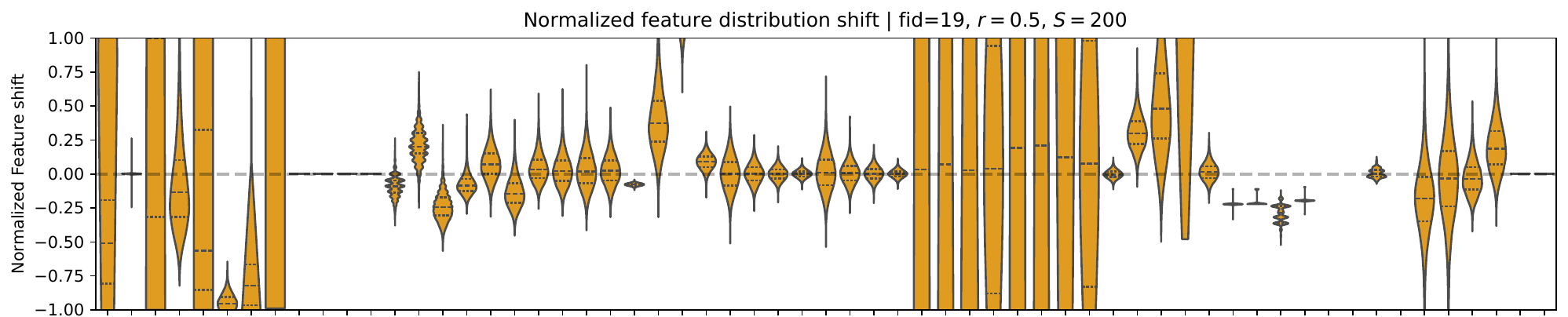}
    \includegraphics[width=.8\linewidth,trim=0cm 7.5cm 0cm 0cm,clip]{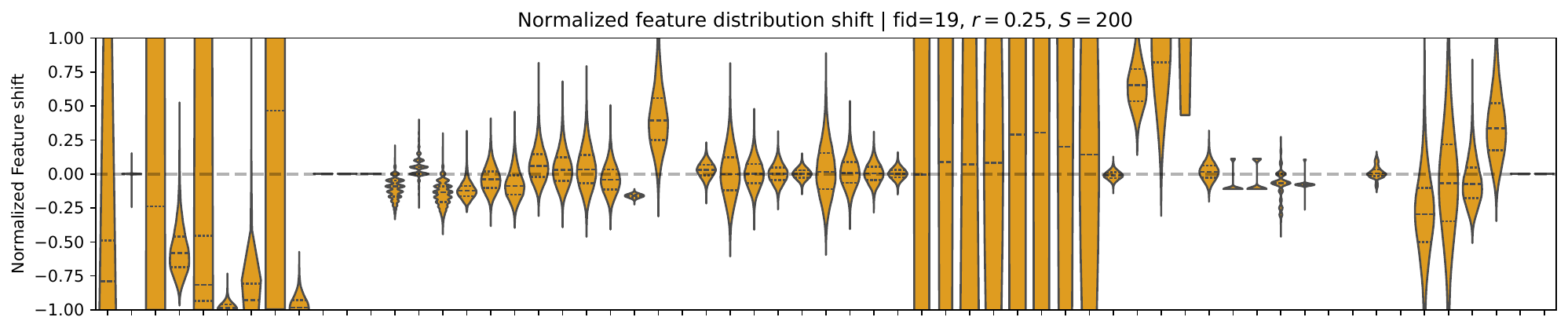}
    \includegraphics[width=.8\linewidth,trim=0cm .7cm 0cm 0cm,clip]{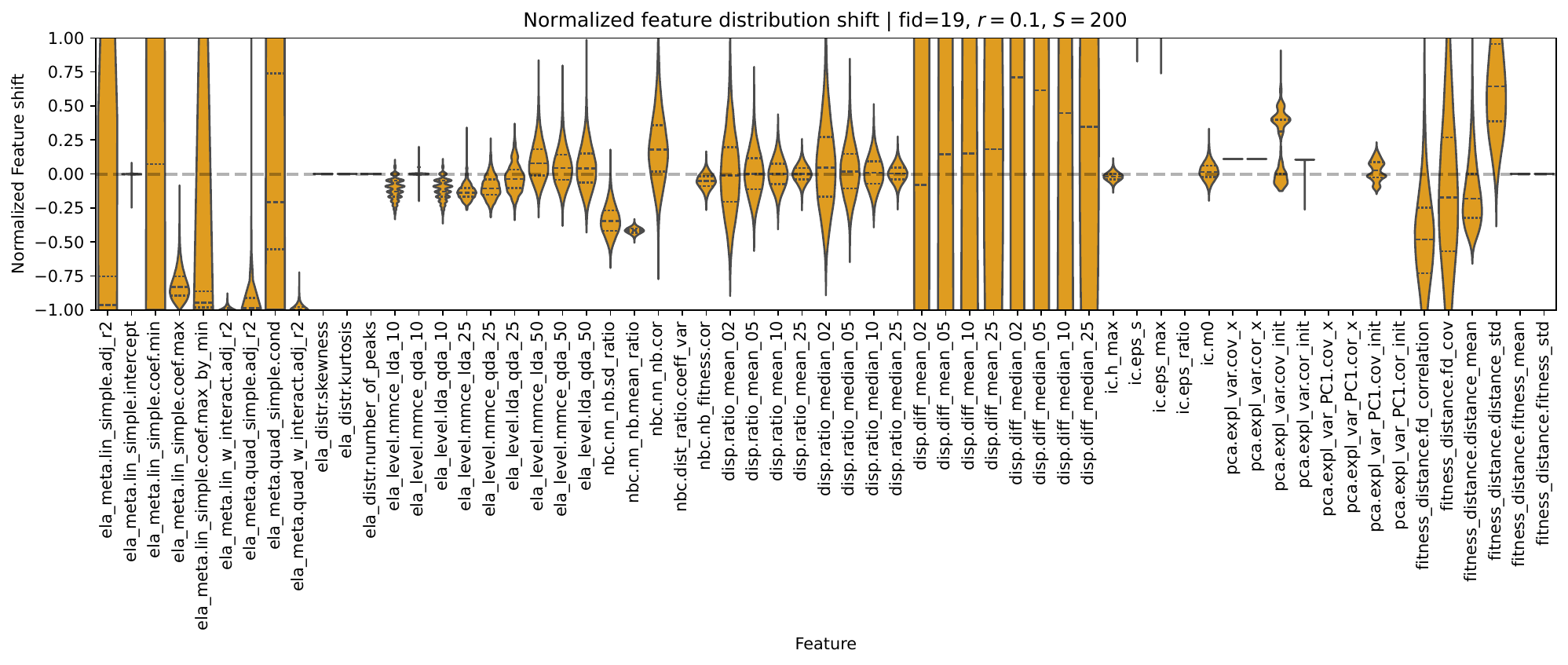}
    \caption{Same as above, for $\boldsymbol{S=200}$.}
    \label{fig:violin_f19_n200}
\end{figure}


\begin{figure}[hbtp]
    \centering
    \includegraphics[width=.8\linewidth,trim=0cm 7.5cm 0cm 0cm,clip]{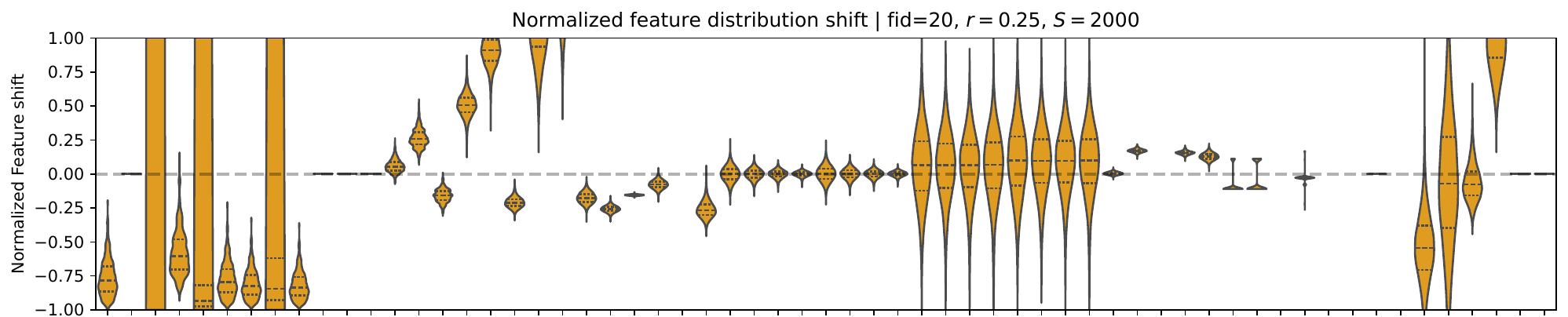}
    \includegraphics[width=.8\linewidth,trim=0cm .7cm 0cm 0cm,clip]{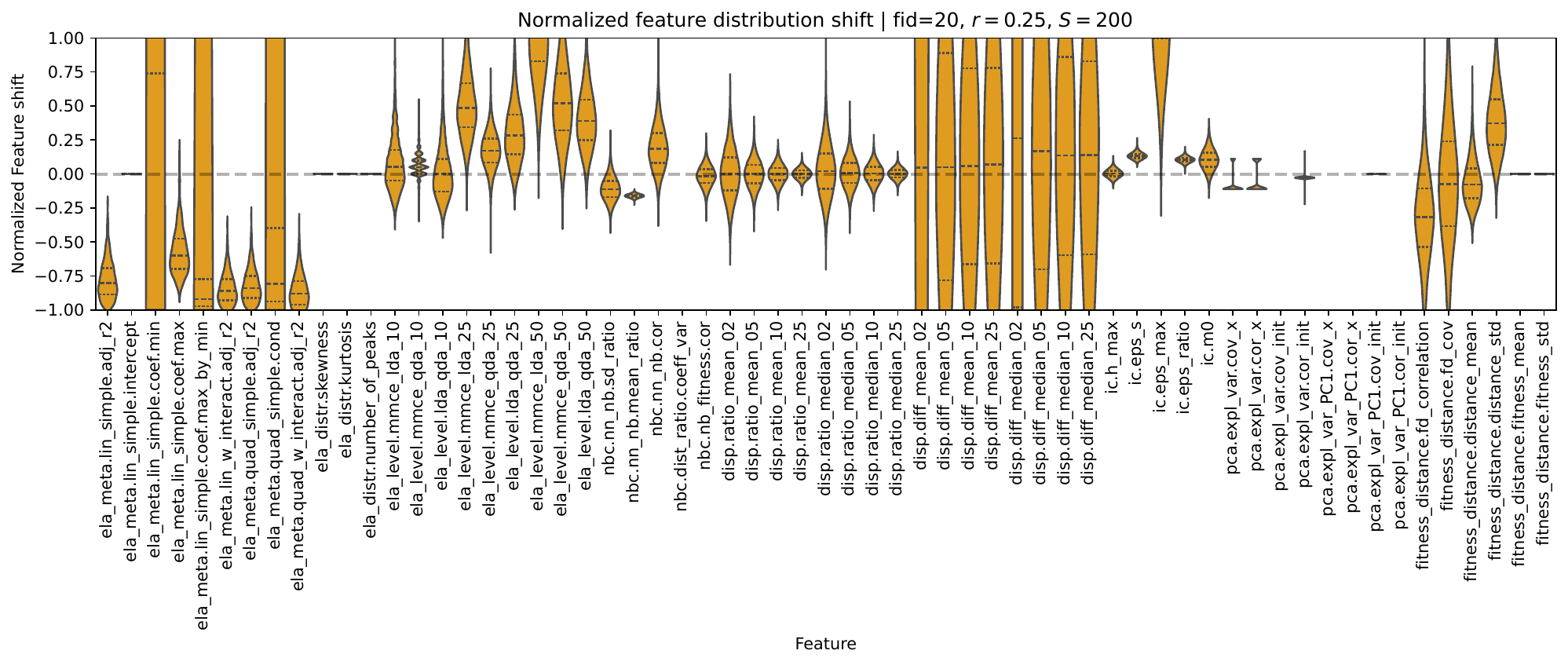}
    \caption{Normalized aggregated feature distribution shift of \textbf{Schwefel (f20) function} with $\boldsymbol{S=2000}$ and $\boldsymbol{S=200}$ for compression ratio $r=0.25$. The horizontal dashed line denotes a normalized reference corresponding to the median of each feature distribution in the original search space. To enhance visualization, the limits of the Normalized Feature shift has been set to $[-1,1 ]$. Figure for $\boldsymbol{S=200}$ is in the main paper.}
    \label{fig:violin_f20_r0.25}
\end{figure}

\begin{figure}[hbtp]
    \centering
    \includegraphics[width=.8\linewidth,trim=0cm 7.5cm 0cm 0cm,clip]{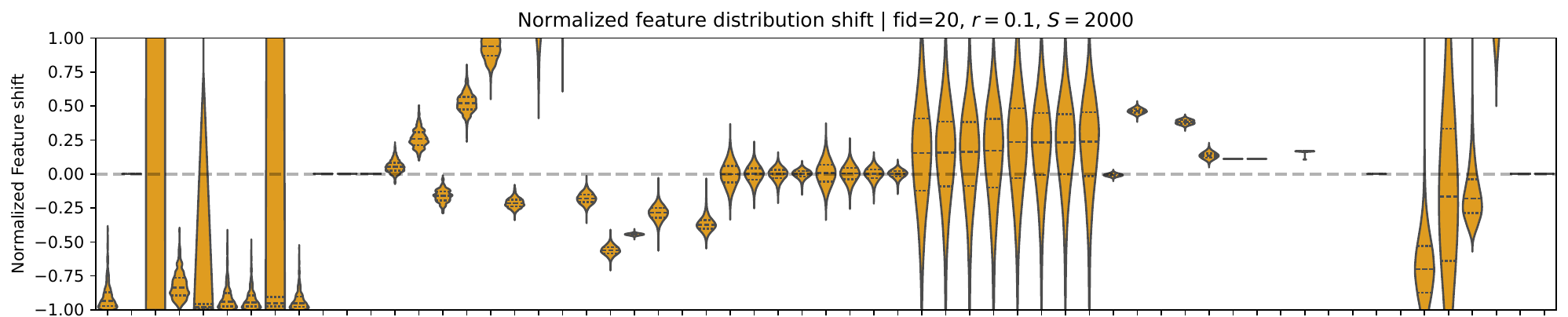}
    \includegraphics[width=.8\linewidth,trim=0cm .7cm 0cm 0cm,clip]{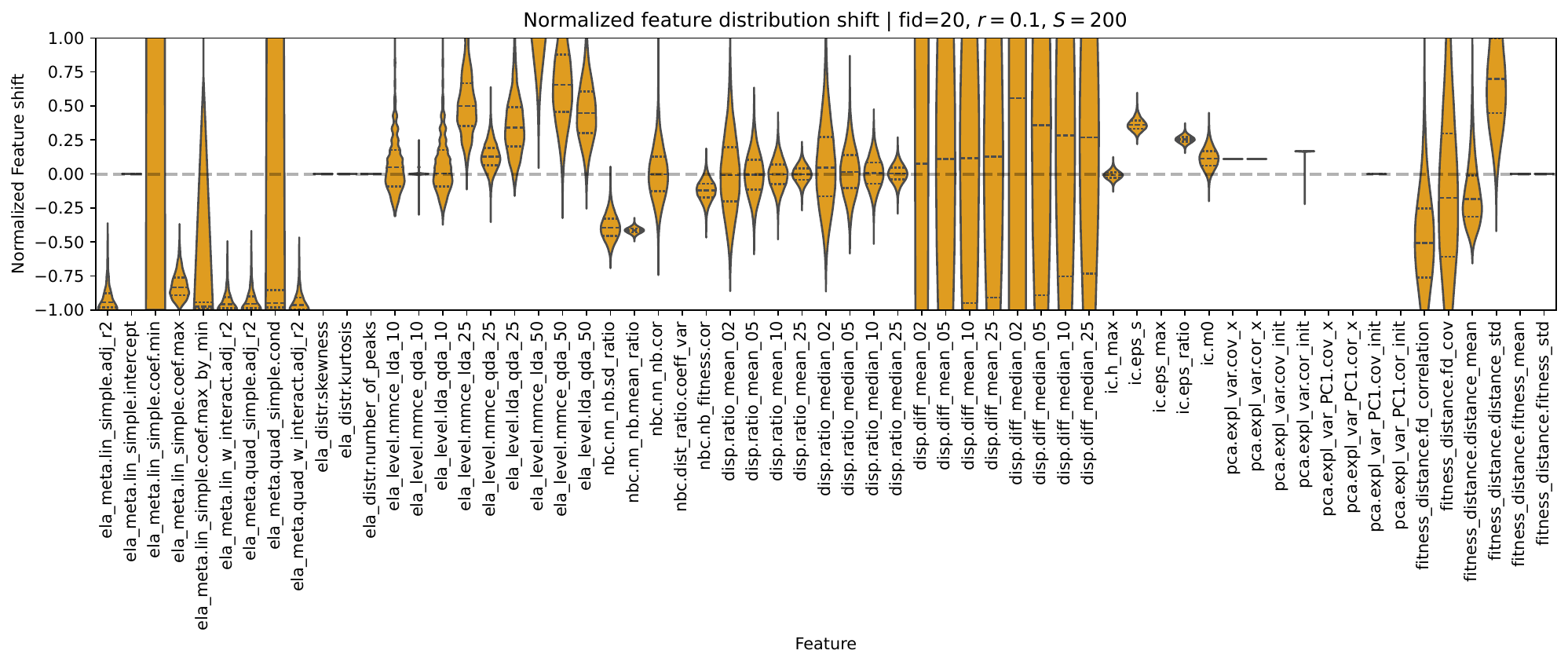}
    \caption{Normalized aggregated feature distribution shift of \textbf{Schwefel (f20) function} with $\boldsymbol{S=2000}$ and $\boldsymbol{S=200}$ for compression ratio $r=0.1$. The horizontal dashed line denotes a normalized reference corresponding to the median of each feature distribution in the original search space. To enhance visualization, the limits of the Normalized Feature shift has been set to $[-1,1 ]$. Figure for $\boldsymbol{S=200}$ is in the main paper.}
    \label{fig:violin_f20_r0.25}
\end{figure}


\begin{figure}[hbtp]
    \centering
    \includegraphics[width=.8\linewidth,trim=0cm 7.5cm 0cm 0cm,clip]{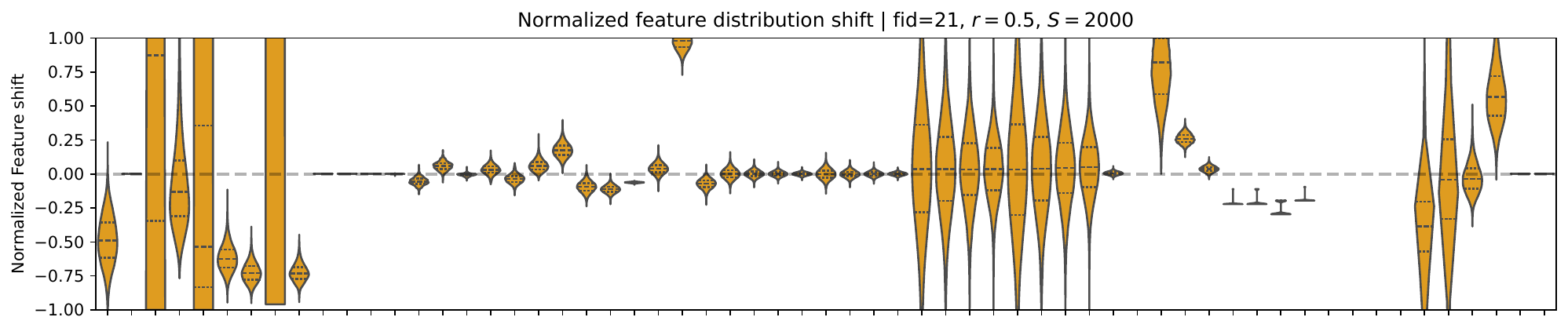}
    \includegraphics[width=.8\linewidth,trim=0cm 7.5cm 0cm 0cm,clip]{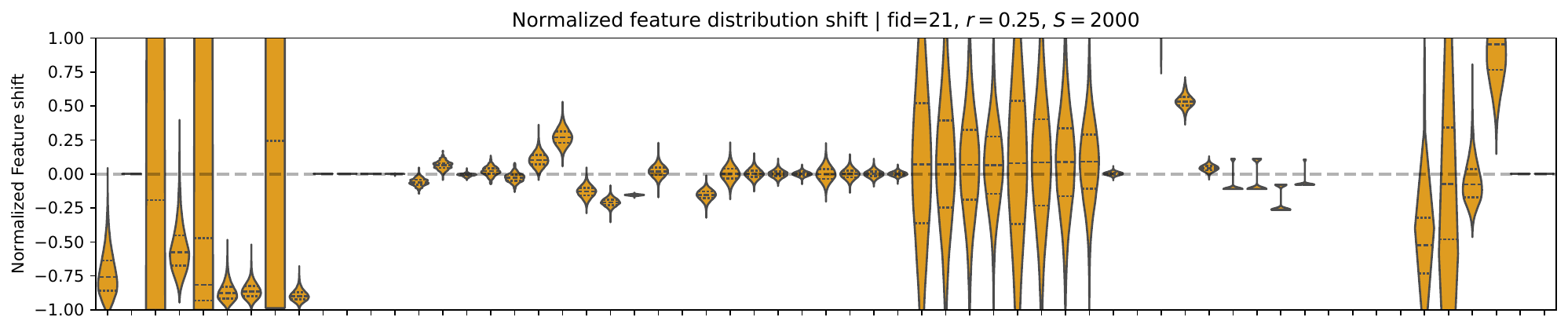}
    \includegraphics[width=.8\linewidth,trim=0cm .7cm 0cm 0cm,clip]{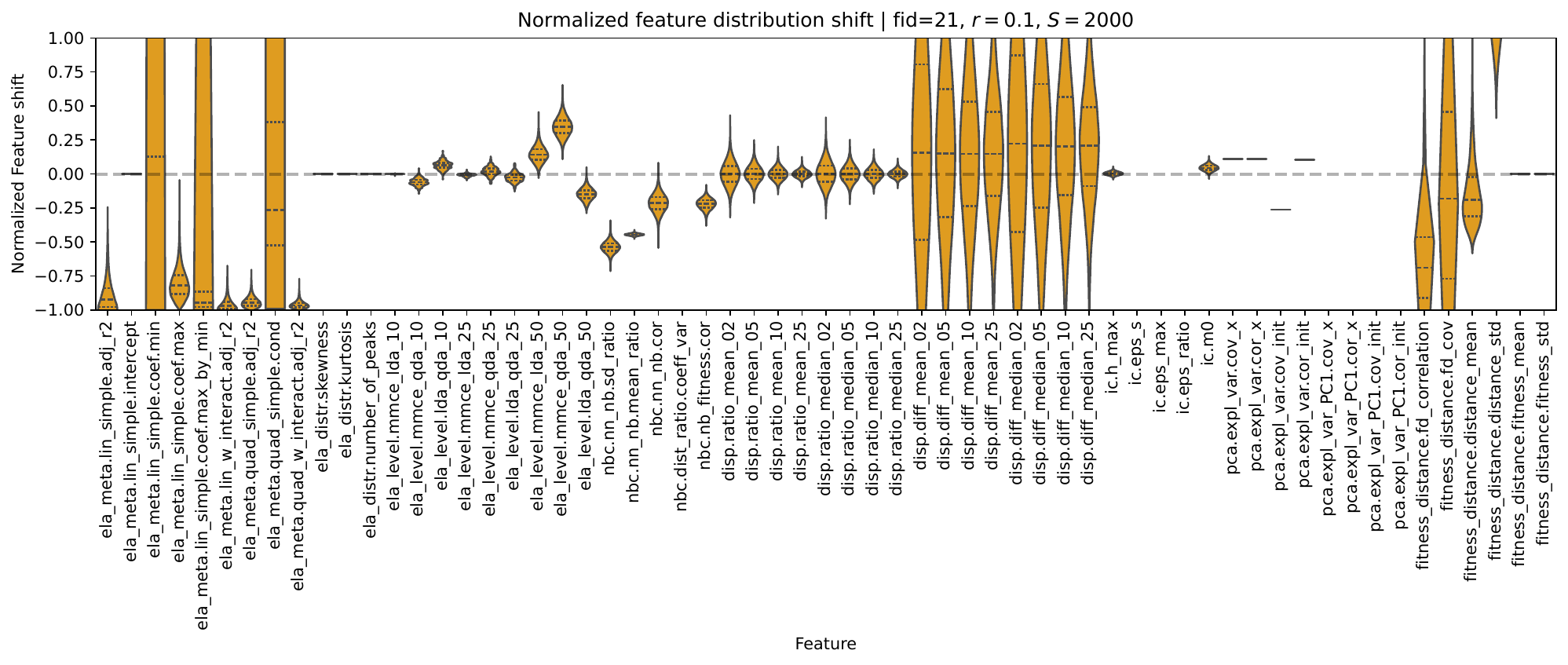}
    \caption{Normalized aggregated feature distribution shift of \textbf{Gallagher 101 peaks (f21) function} with $\boldsymbol{S=2000}$ for compression ratios $r=\{0.5,0.25,0.1\}$. The horizontal dashed line denotes a normalized reference corresponding to the median of each feature distribution in the original search space. To enhance visualization, the limits of the Normalized Feature shift has been set to $[-1,1 ]$.}
    \label{fig:violin_f21_n2000}
\end{figure}

\begin{figure}[hbtp]
    \centering
    \includegraphics[width=.8\linewidth,trim=0cm 7.5cm 0cm 0cm,clip]{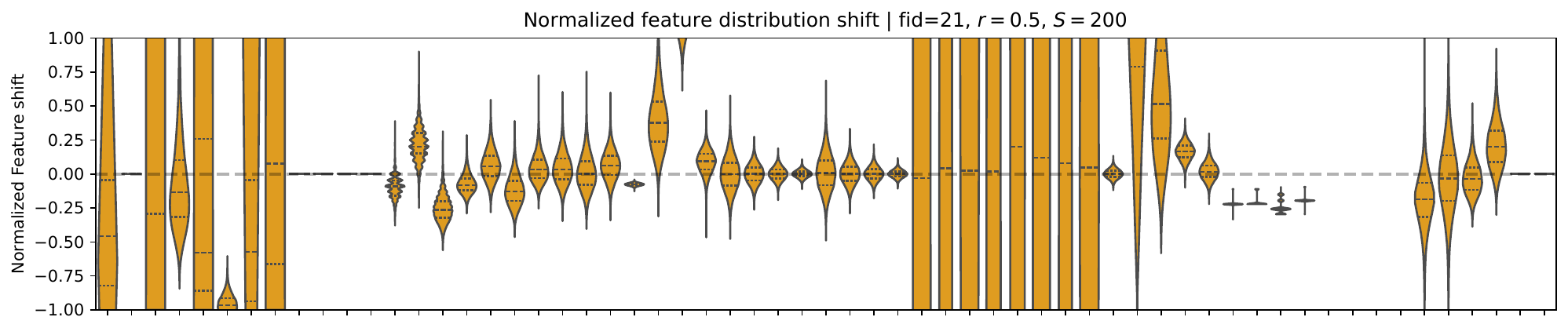}
    \includegraphics[width=.8\linewidth,trim=0cm 7.5cm 0cm 0cm,clip]{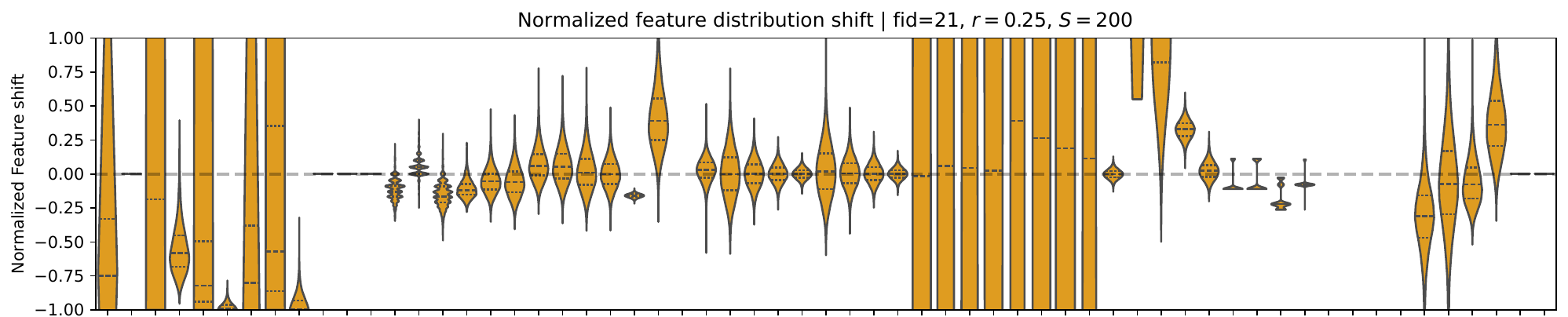}
    \includegraphics[width=.8\linewidth,trim=0cm .7cm 0cm 0cm,clip]{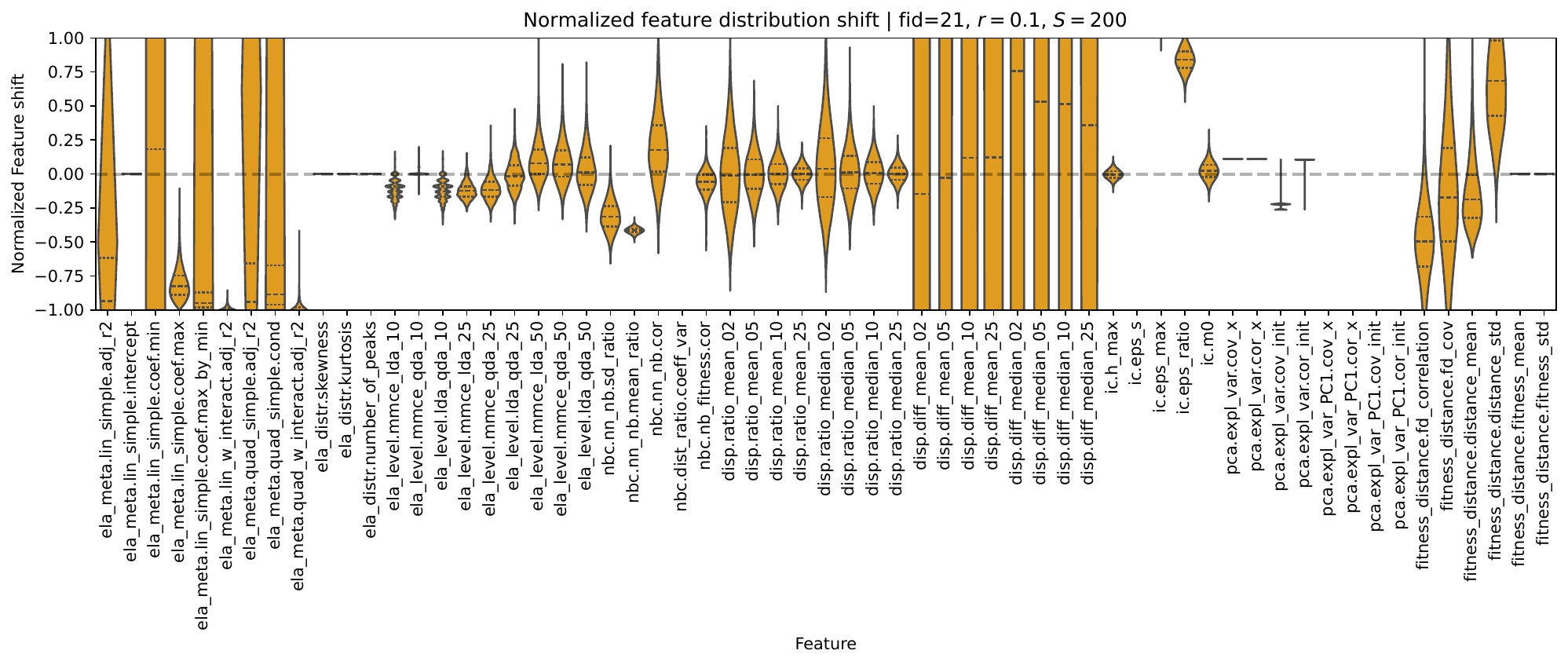}
    \caption{Same as above, for $\boldsymbol{S=200}$.}
    \label{fig:violin_f21_n200}
\end{figure}


\begin{figure}[hbtp]
    \centering
    \includegraphics[width=.8\linewidth,trim=0cm 7.5cm 0cm 0cm,clip]{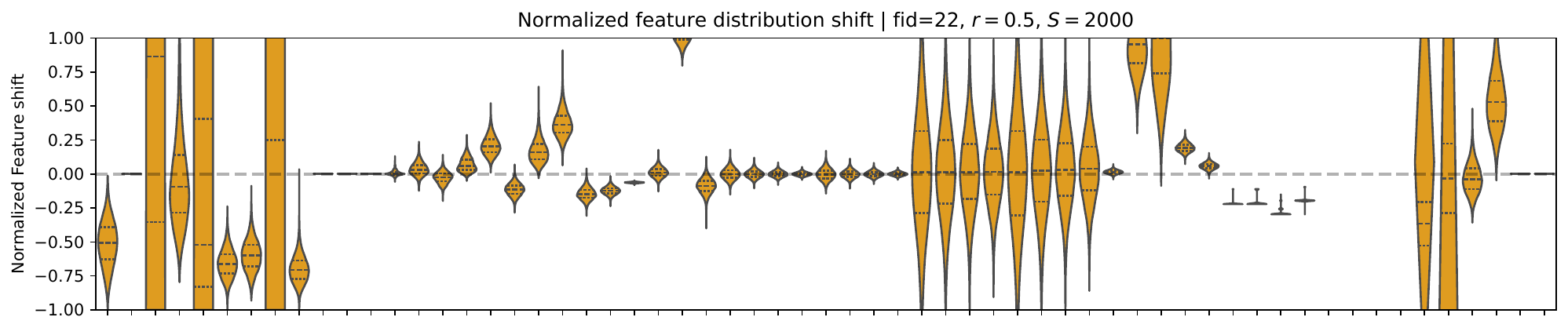}
    \includegraphics[width=.8\linewidth,trim=0cm 7.5cm 0cm 0cm,clip]{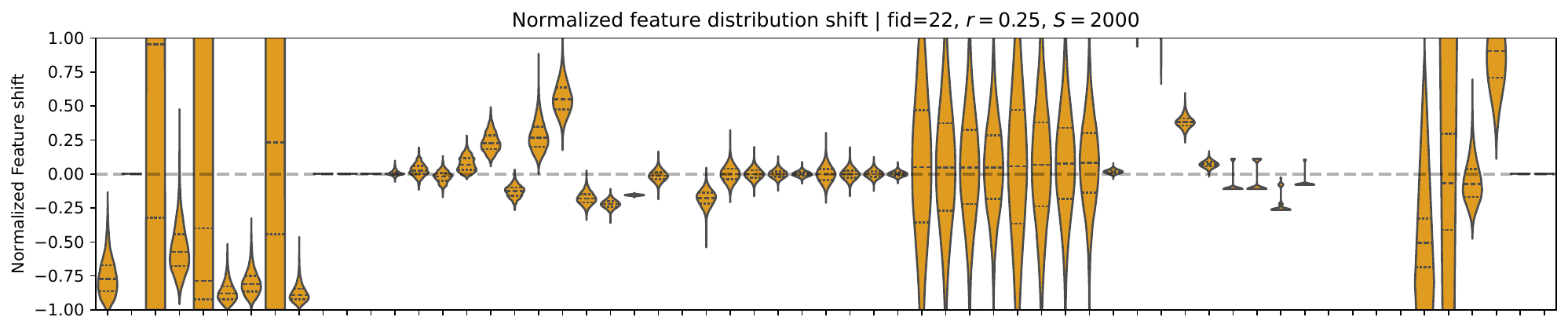}
    \includegraphics[width=.8\linewidth,trim=0cm .7cm 0cm 0cm,clip]{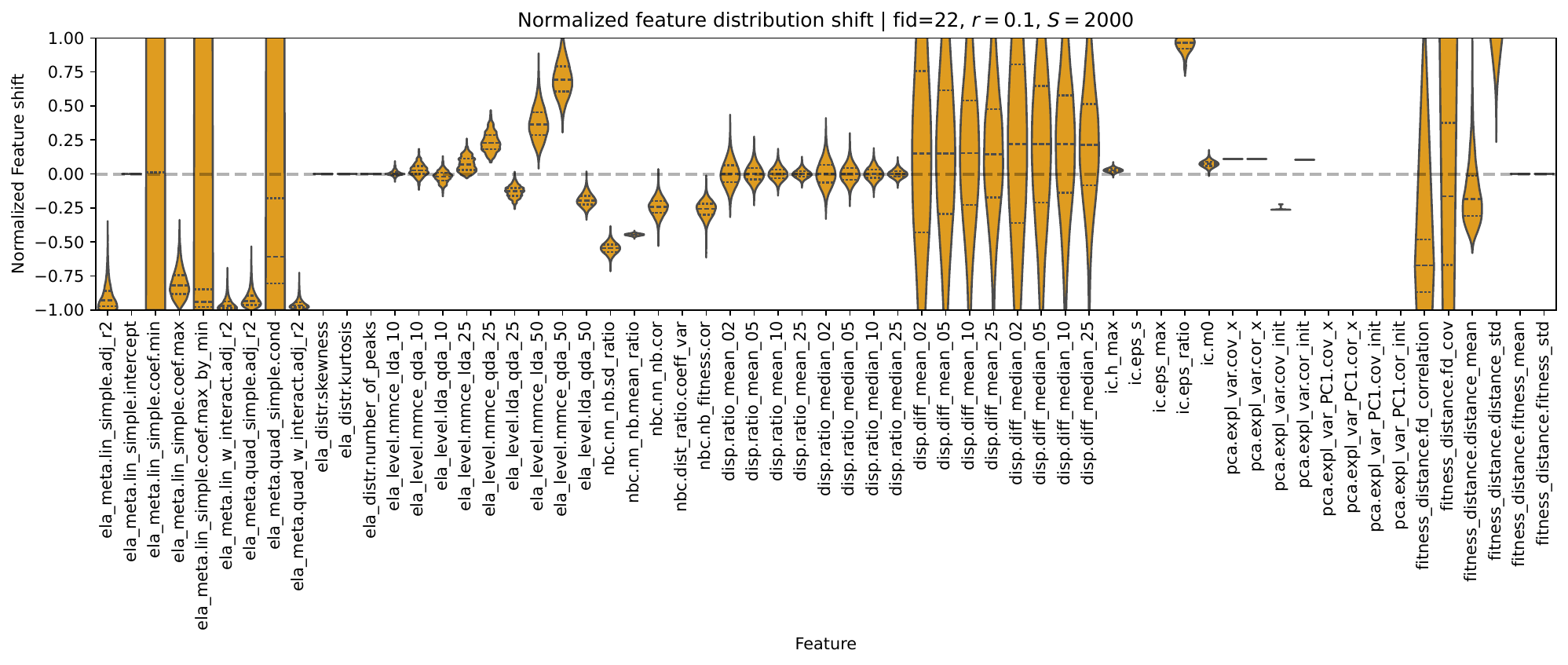}
    \caption{Normalized aggregated feature distribution shift of \textbf{Gallagher 21 peaks (f22) function} with $\boldsymbol{S=2000}$ for compression ratios $r=\{0.5,0.25,0.1\}$. The horizontal dashed line denotes a normalized reference corresponding to the median of each feature distribution in the original search space. To enhance visualization, the limits of the Normalized Feature shift has been set to $[-1,1 ]$.}
    \label{fig:violin_f22_n2000}
\end{figure}

\begin{figure}[hbtp]
    \centering
    \includegraphics[width=.8\linewidth,trim=0cm 7.5cm 0cm 0cm,clip]{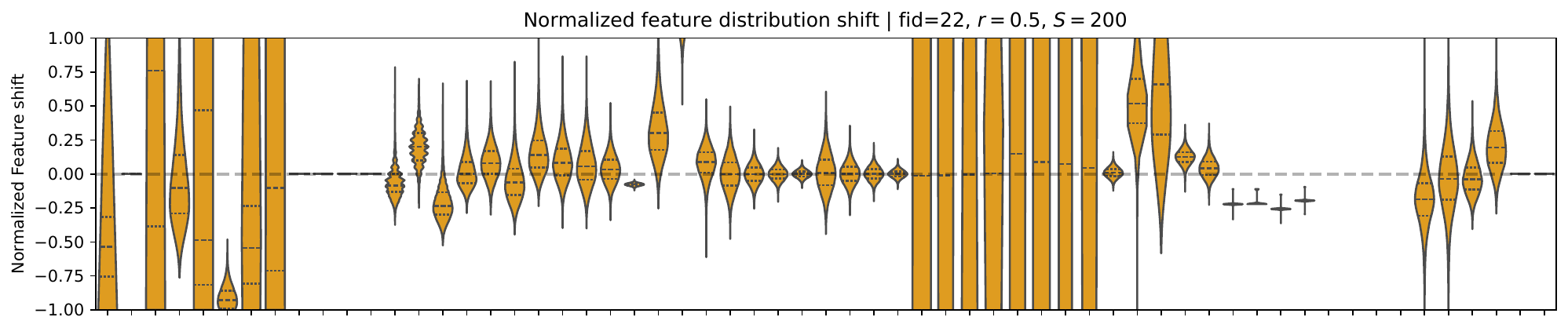}
    \includegraphics[width=.8\linewidth,trim=0cm 7.5cm 0cm 0cm,clip]{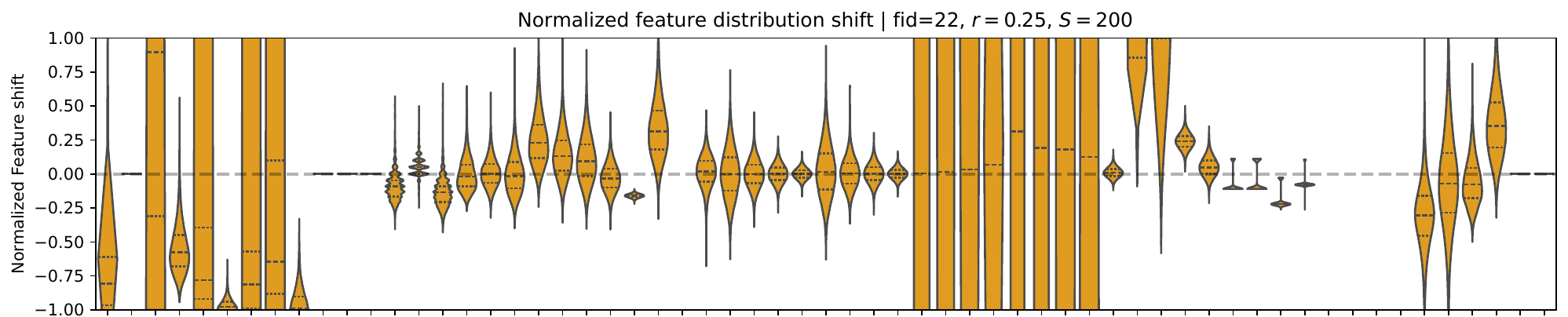}
    \includegraphics[width=.8\linewidth,trim=0cm .7cm 0cm 0cm,clip]{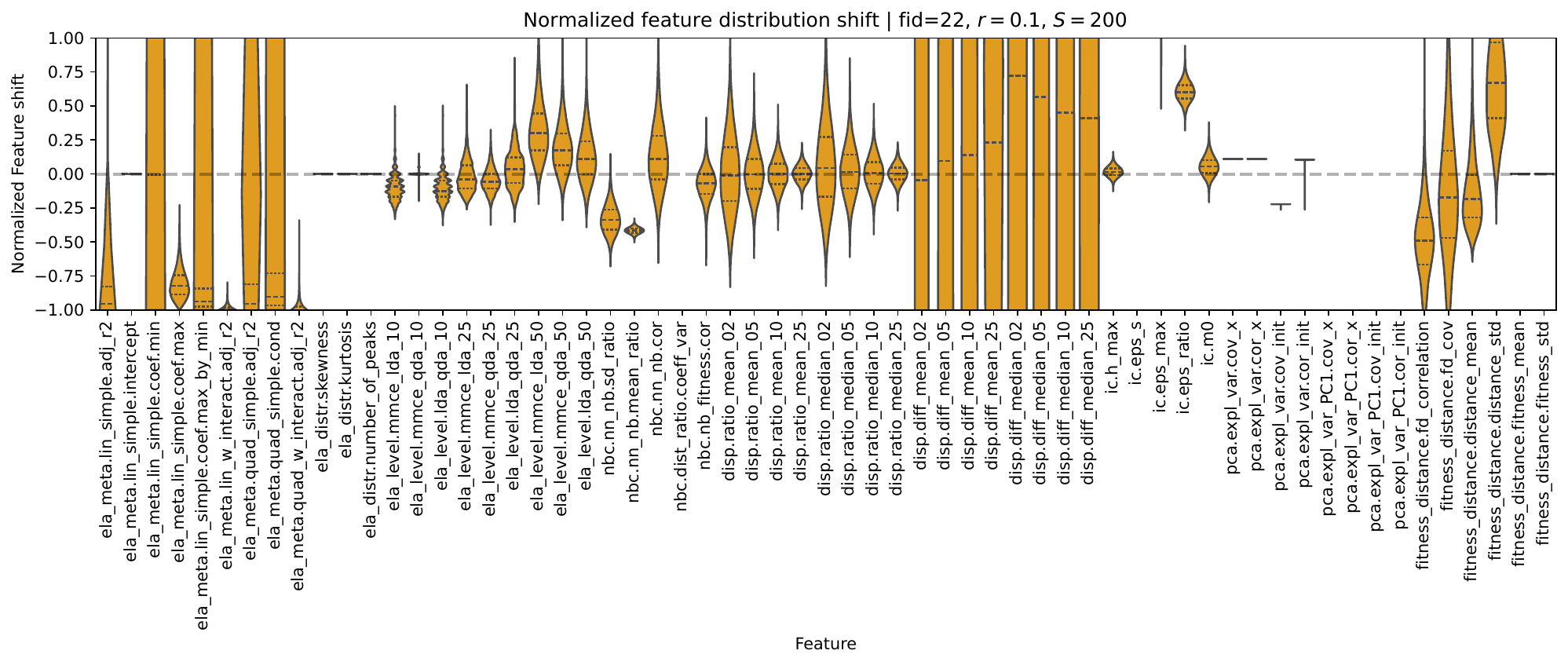}
    \caption{Same as above, for $\boldsymbol{S=200}$.}
    \label{fig:violin_f22_n200}
\end{figure}


\begin{figure}[hbtp]
    \centering
    \includegraphics[width=.8\linewidth,trim=0cm 7.5cm 0cm 0cm,clip]{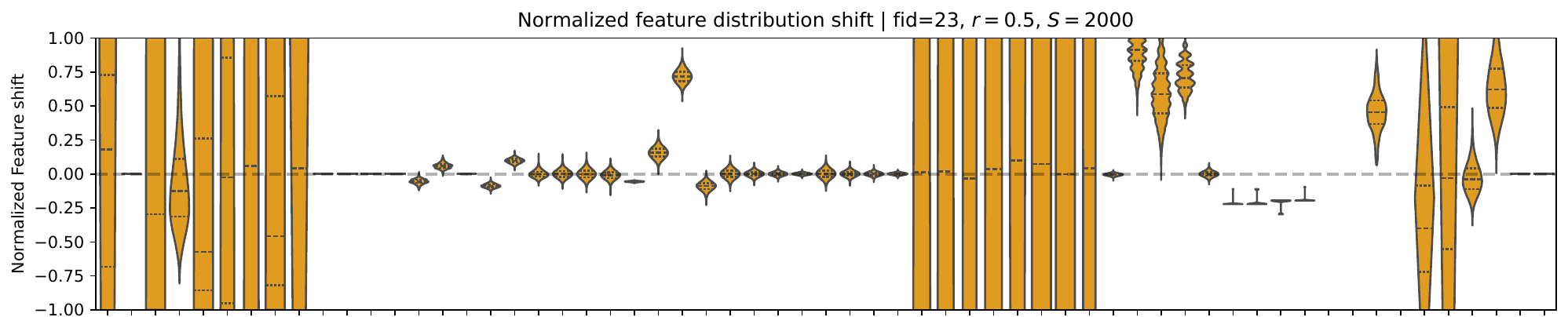}
    \includegraphics[width=.8\linewidth,trim=0cm 7.5cm 0cm 0cm,clip]{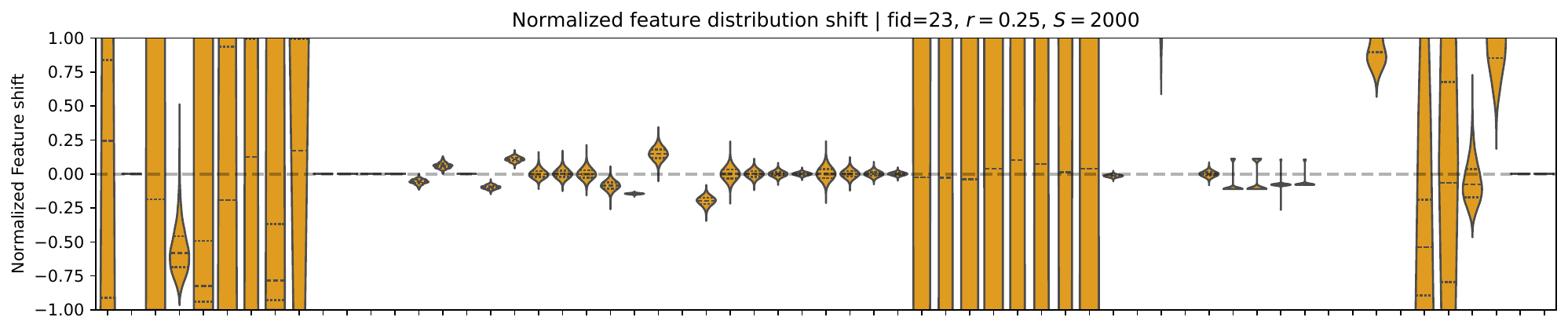}
    \includegraphics[width=.8\linewidth,trim=0cm .7cm 0cm 0cm,clip]{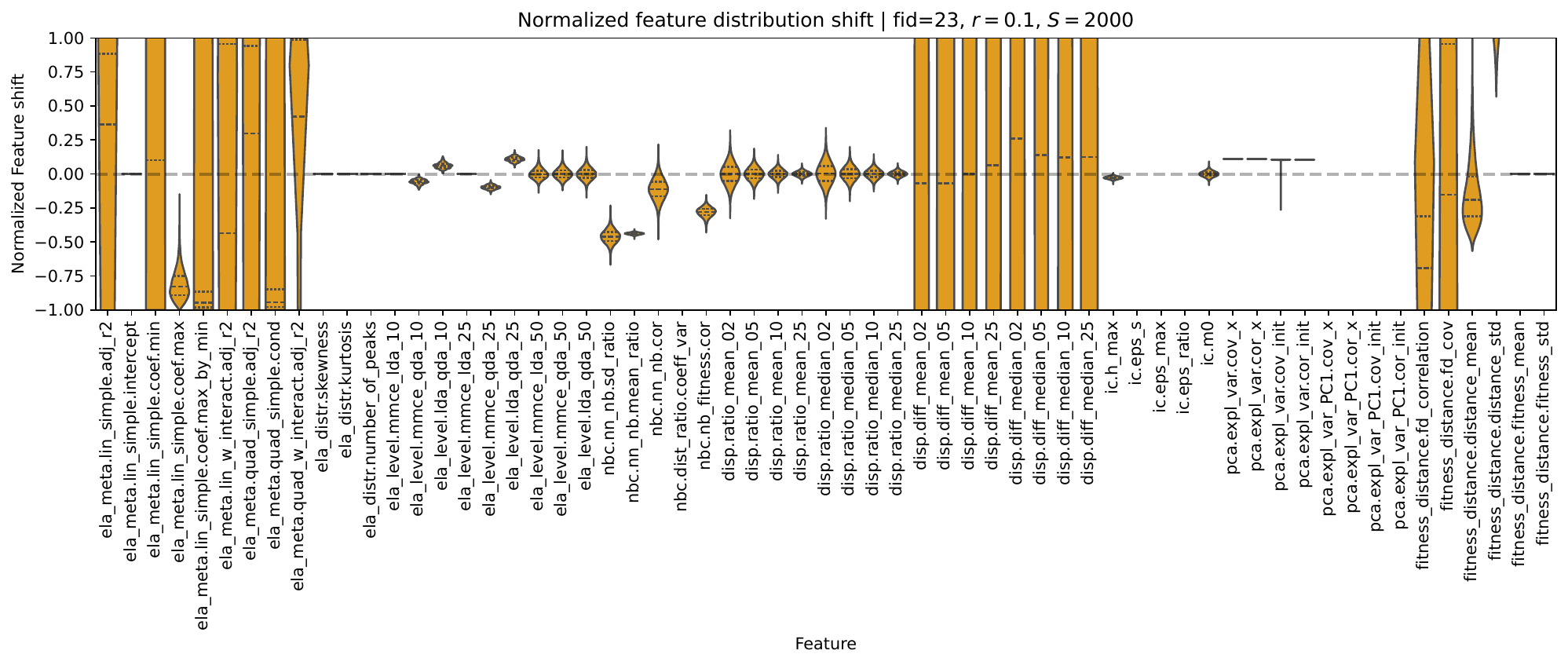}
    \caption{Normalized aggregated feature distribution shift of \textbf{Katsuura (f23) function} with $\boldsymbol{S=2000}$ for compression ratios $r=\{0.5,0.25,0.1\}$. The horizontal dashed line denotes a normalized reference corresponding to the median of each feature distribution in the original search space. To enhance visualization, the limits of the Normalized Feature shift has been set to $[-1,1 ]$.}
    \label{fig:violin_f23_n2000}
\end{figure}

\begin{figure}[hbtp]
    \centering
    \includegraphics[width=.8\linewidth,trim=0cm 7.5cm 0cm 0cm,clip]{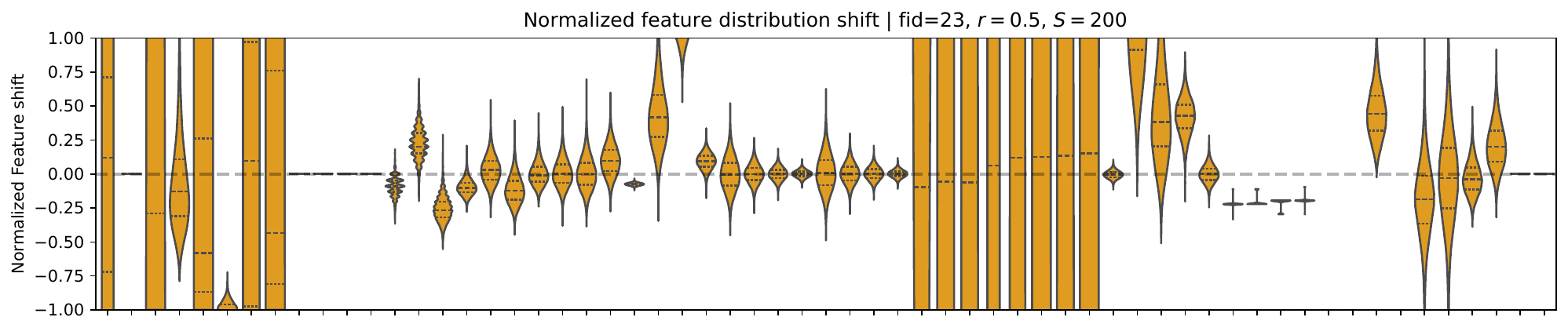}
    \includegraphics[width=.8\linewidth,trim=0cm 7.5cm 0cm 0cm,clip]{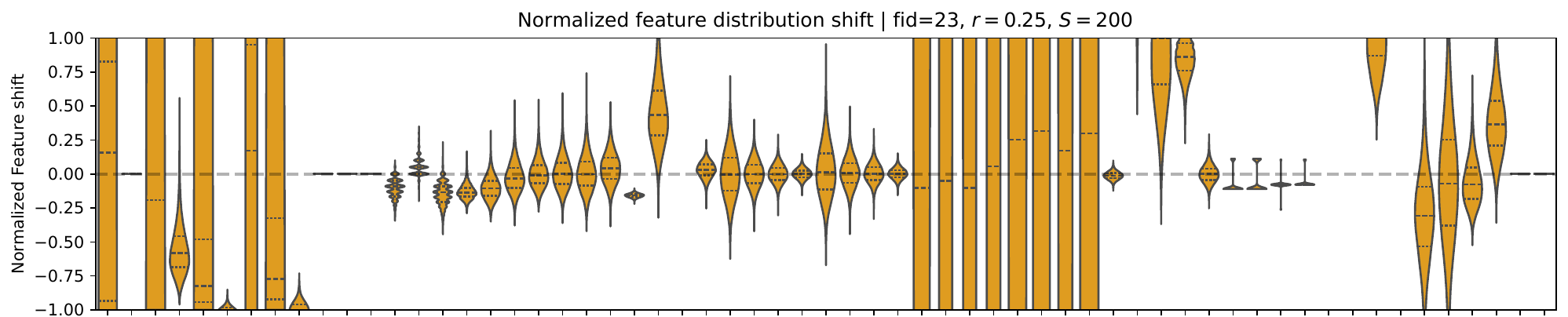}
    \includegraphics[width=.8\linewidth,trim=0cm .7cm 0cm 0cm,clip]{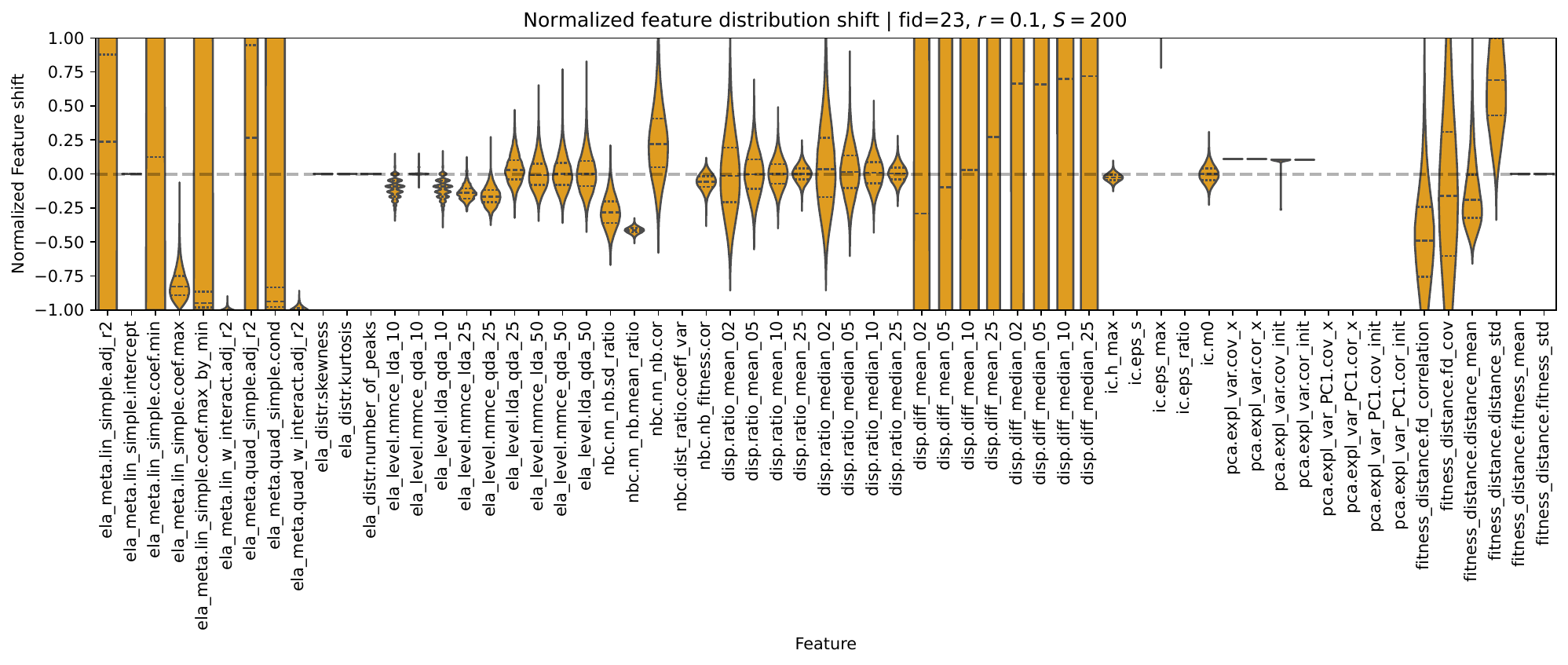}
    \caption{Same as above, for $\boldsymbol{S=200}$.}
    \label{fig:violin_f23_n200}
\end{figure}


\begin{figure}[hbtp]
    \centering
    \includegraphics[width=.8\linewidth,trim=0cm 7.5cm 0cm 0cm,clip]{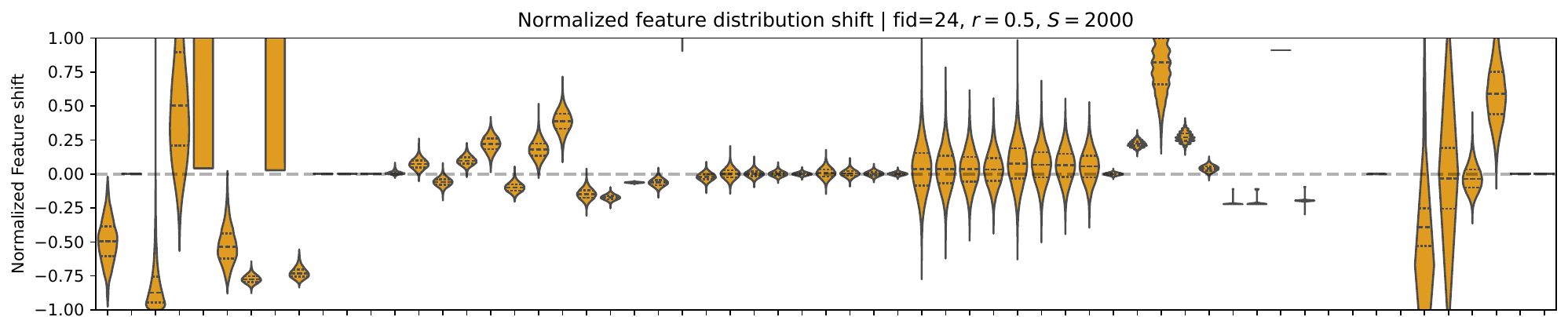}
    \includegraphics[width=.8\linewidth,trim=0cm 7.5cm 0cm 0cm,clip]{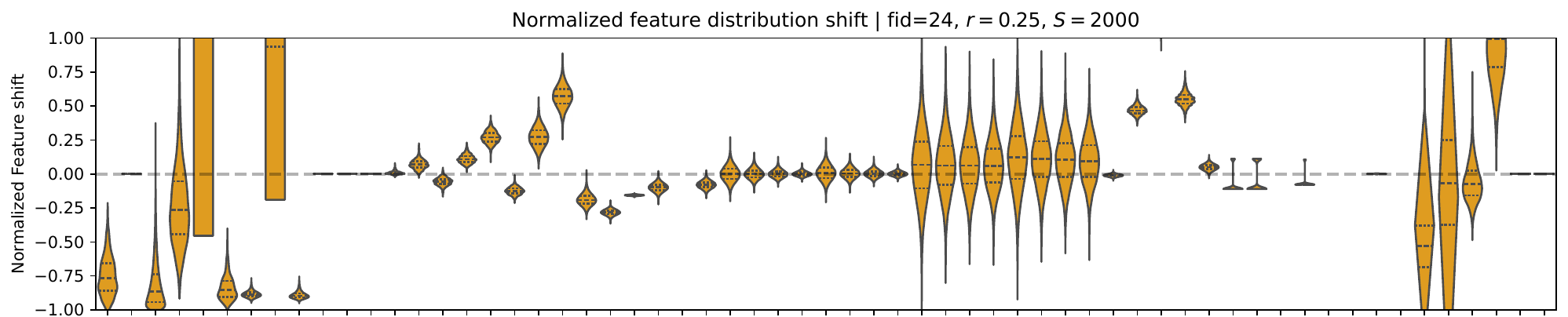}
    \includegraphics[width=.8\linewidth,trim=0cm .7cm 0cm 0cm,clip]{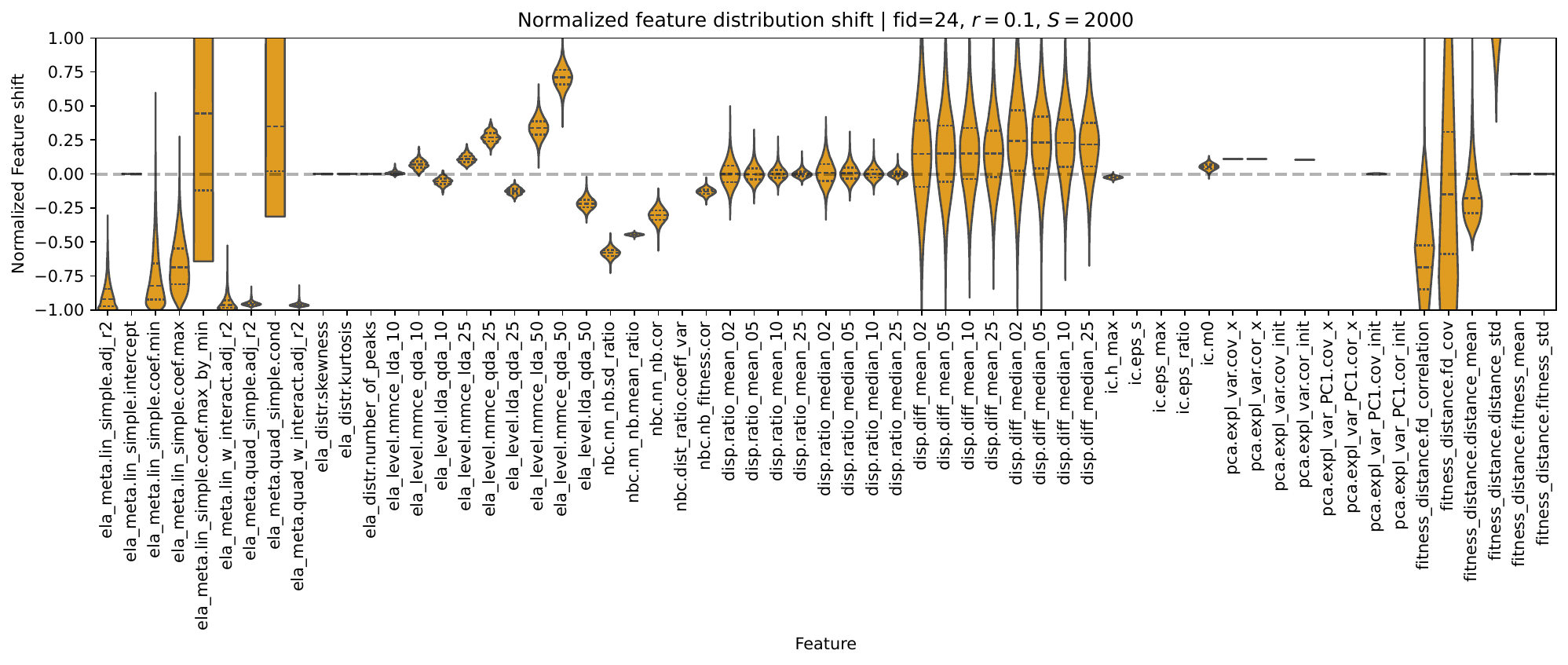}
    \caption{Normalized aggregated feature distribution shift of \textbf{Lunacek bi-Rastrigin (f24) function} with $\boldsymbol{S=2000}$ for compression ratios $r=\{0.5,0.25,0.1\}$. The horizontal dashed line denotes a normalized reference corresponding to the median of each feature distribution in the original search space. To enhance visualization, the limits of the Normalized Feature shift has been set to $[-1,1 ]$.}
    \label{fig:violin_f24_n2000}
\end{figure}

\begin{figure}[hbtp]
    \centering
    \includegraphics[width=.8\linewidth,trim=0cm 7.5cm 0cm 0cm,clip]{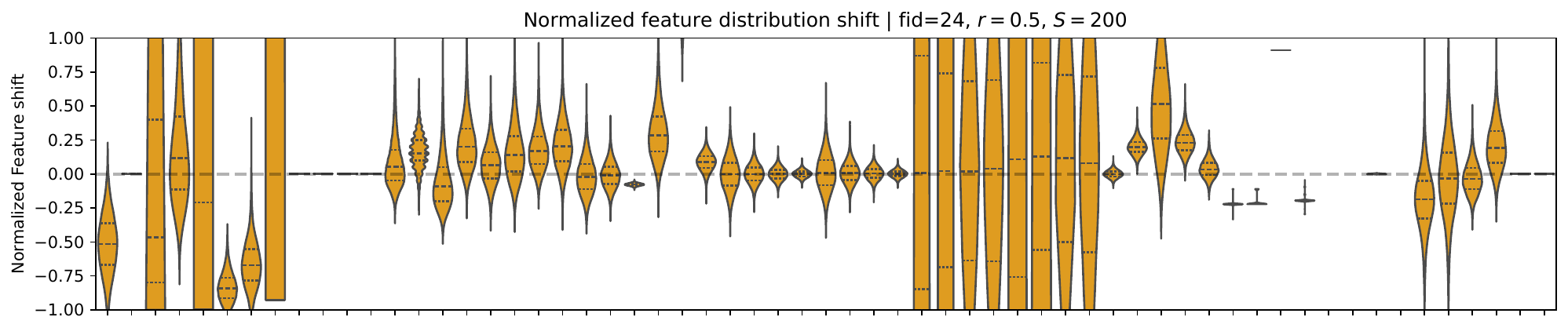}
    \includegraphics[width=.8\linewidth,trim=0cm 7.5cm 0cm 0cm,clip]{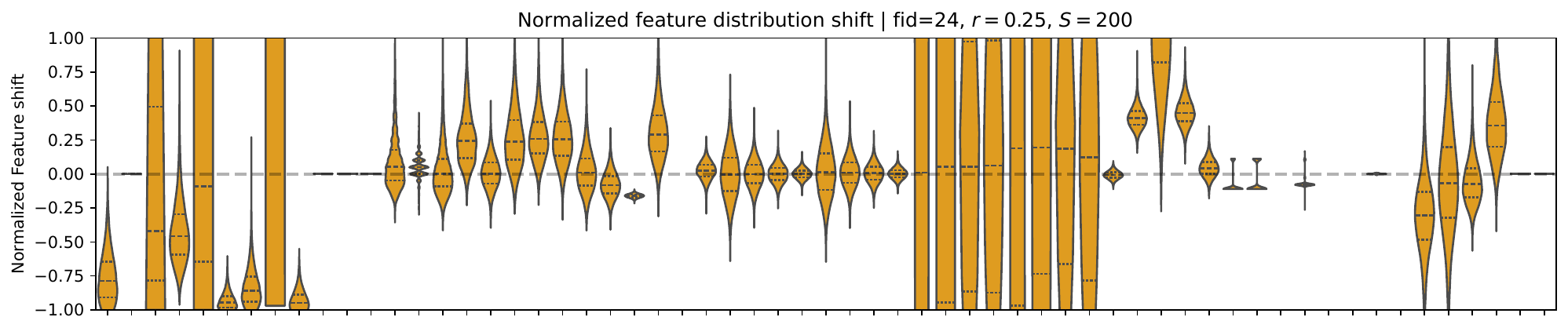}
    \includegraphics[width=.8\linewidth,trim=0cm .7cm 0cm 0cm,clip]{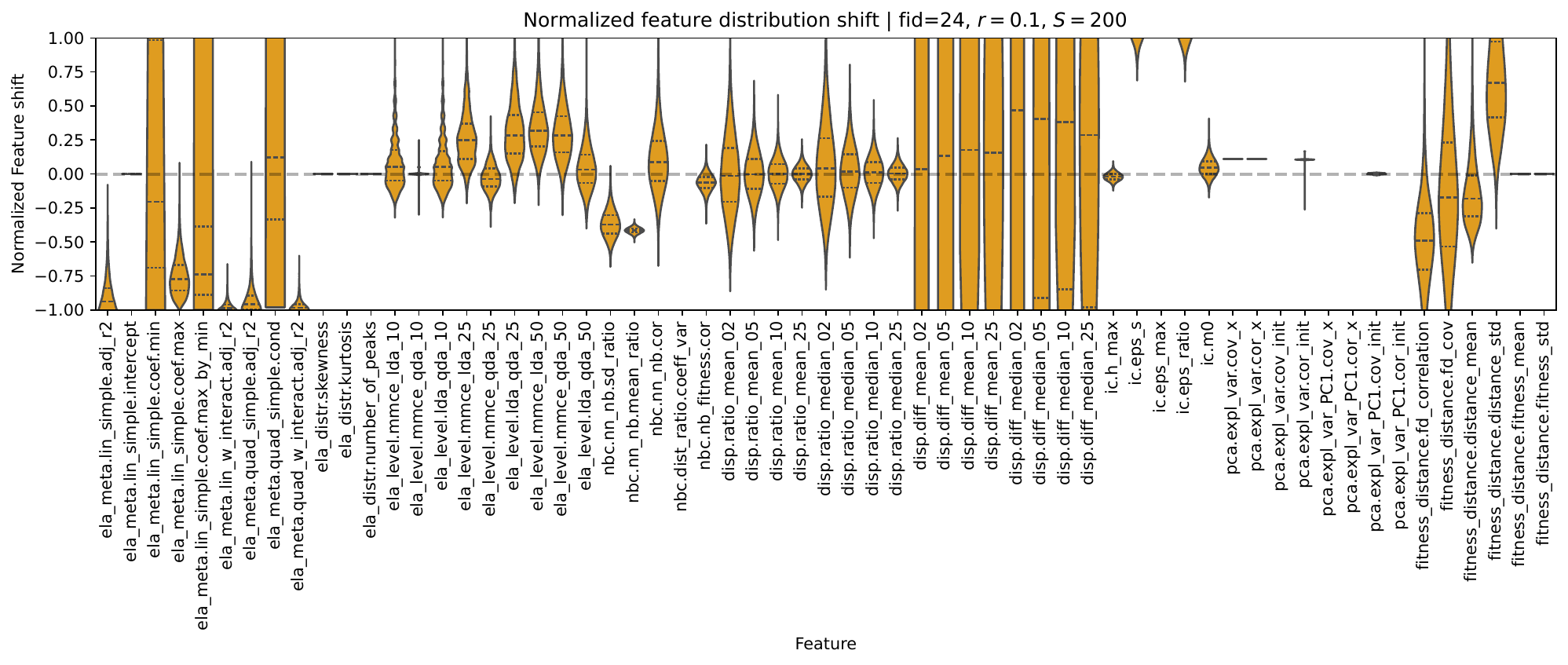}
    \caption{Same as above, for $\boldsymbol{S=200}$.}
    \label{fig:violin_f24_n200}
\end{figure}

\clearpage
\subsection{Distributions with resampling}

\begin{figure}[hbtp]
    \centering
    \includegraphics[width=.7\linewidth]{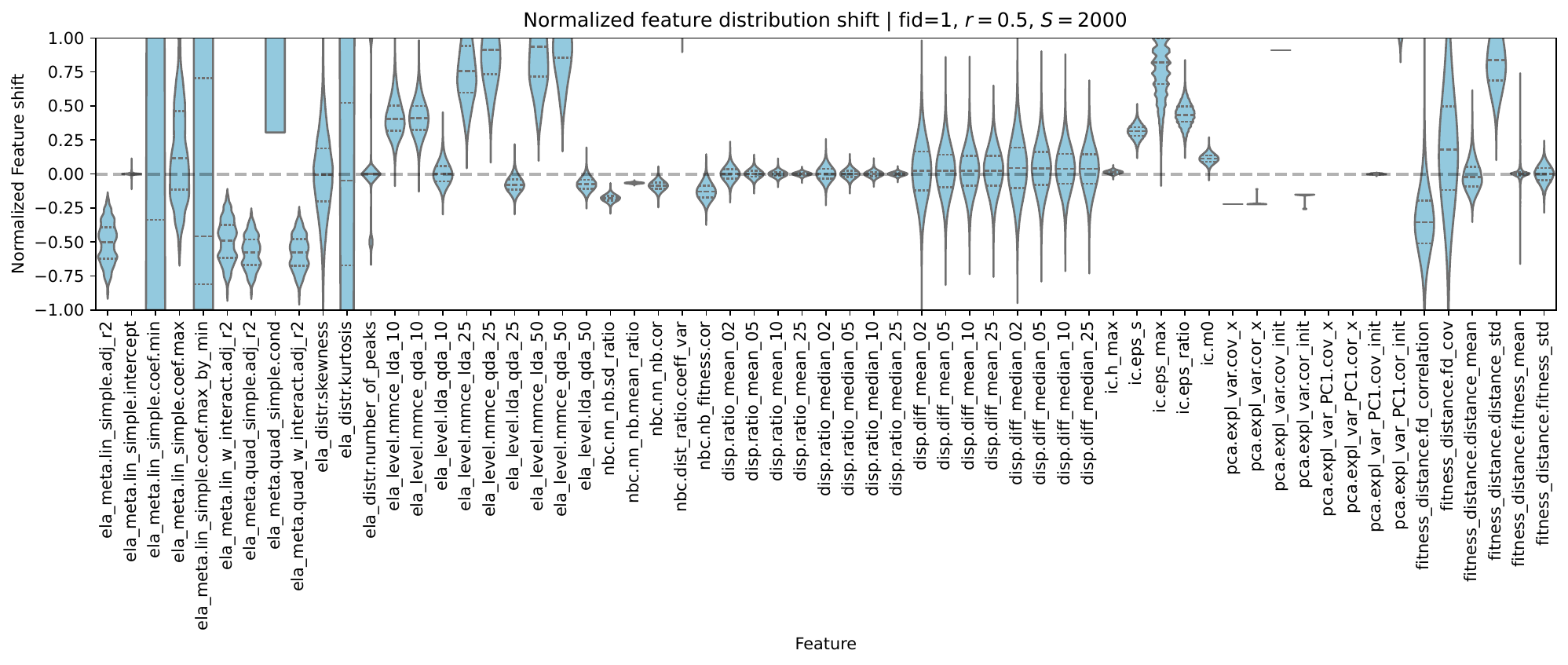}
    \caption{Normalized aggregated feature distribution shift of \textbf{Sphere (f1) function} with $\boldsymbol{S=2000}$ and for compression ratio $\boldsymbol{r=0.5}$, with \textbf{resampling}. The horizontal dashed line denotes a normalized reference corresponding to the median of each feature distribution in the original search space. To enhance visualization, the limits of the Normalized Feature shift has been set to $[-1,1 ]$.}
    \label{fig:violin_f1_resampled_r0.5_n2000}
\end{figure}

\begin{figure}[hbtp]
    \centering
    \includegraphics[width=.7\linewidth]{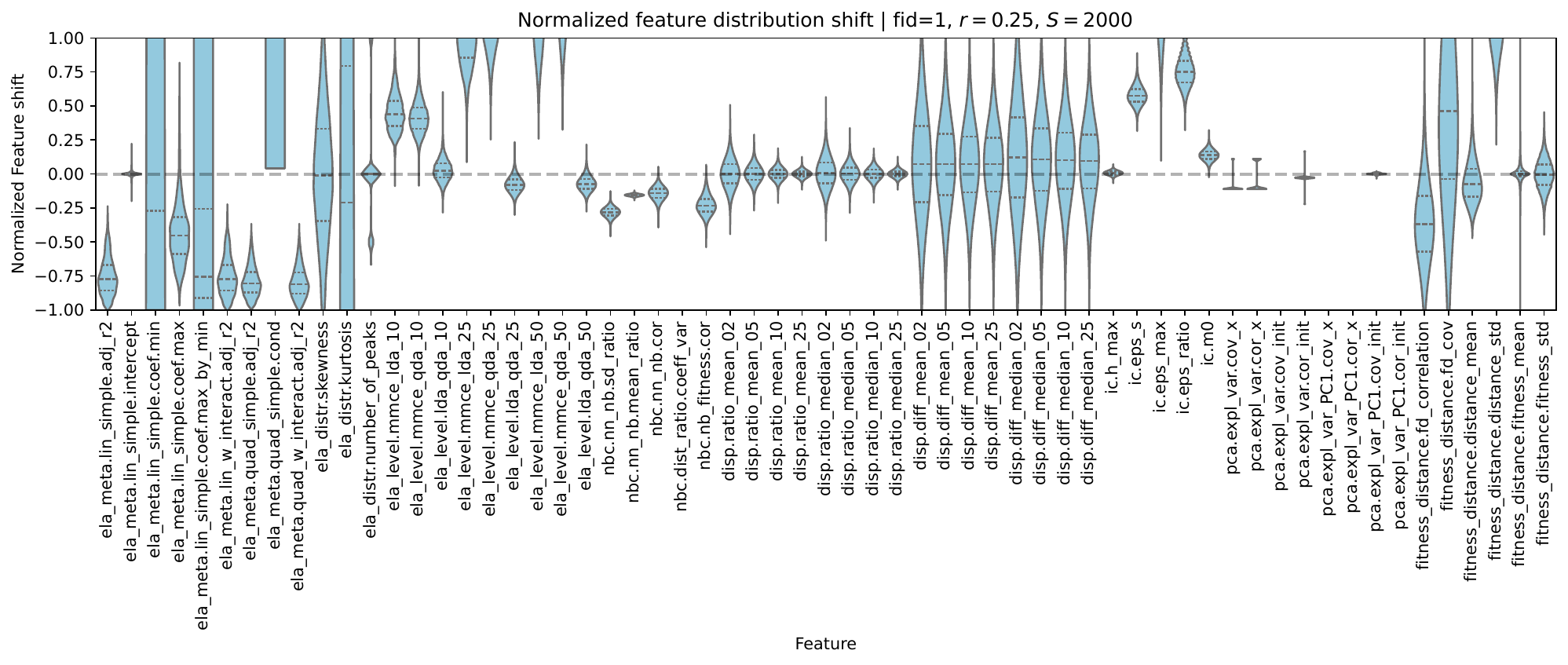}
    \caption{Same as above for $\boldsymbol{S=2000}$ and $\boldsymbol{r=0.25}$.}
    \label{fig:violin_f1_resampled_r0.25_n2000}
\end{figure}

\begin{figure}[hbtp]
    \centering
    \includegraphics[width=.7\linewidth]{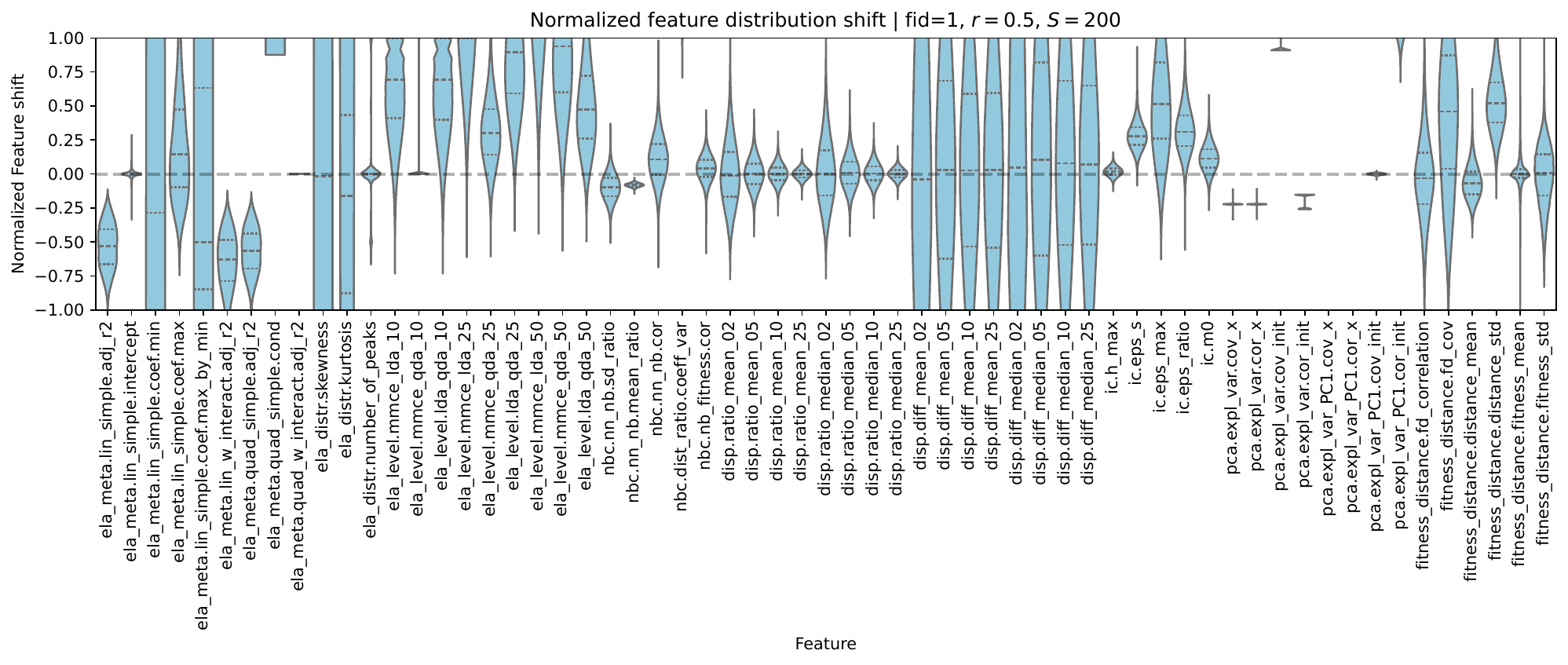}
    \caption{Same as above for $\boldsymbol{S=200}$ and $\boldsymbol{r=0.5}$.}
    \label{fig:violin_f1_resampled_r0.5_n200}
\end{figure}


\begin{figure}[hbtp]
    \centering
    \includegraphics[width=.7\linewidth]{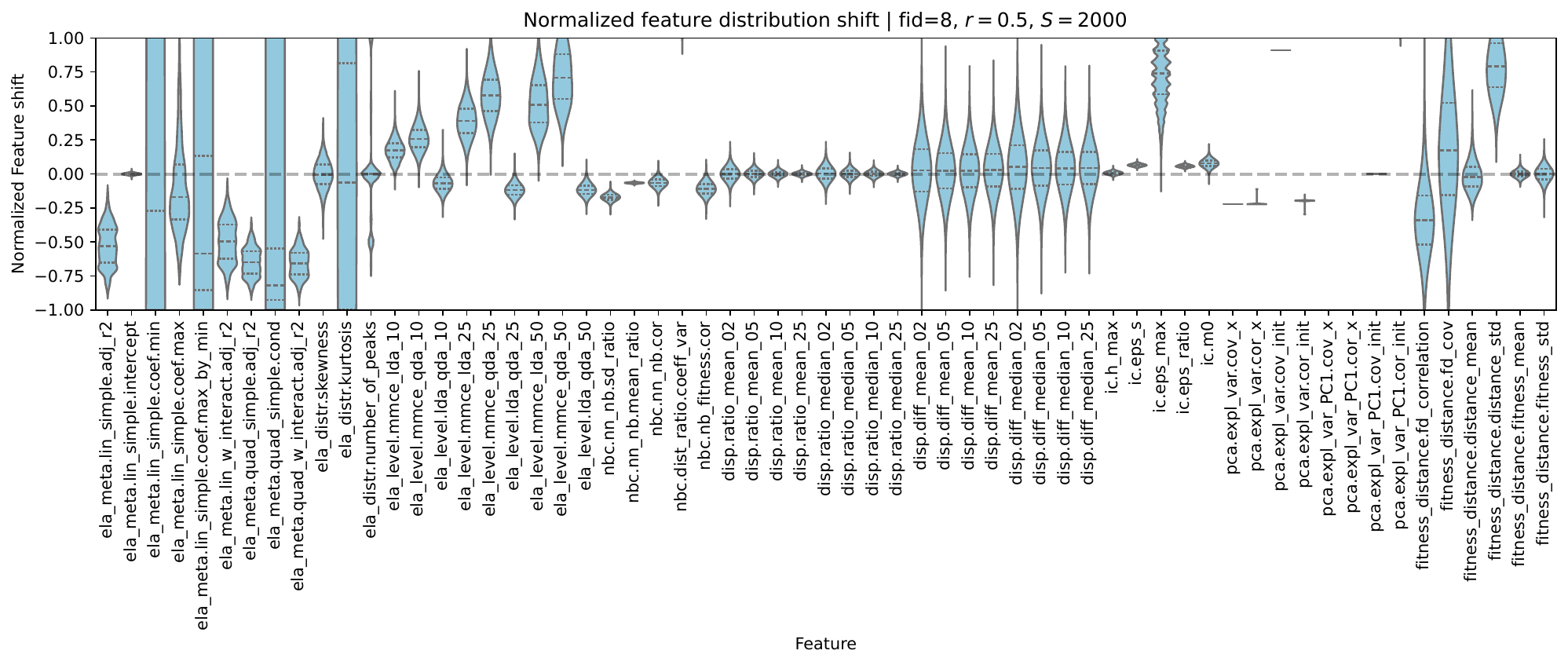}
    \caption{Normalized aggregated feature distribution shift of \textbf{Rosenbrock (f8) function} with $\boldsymbol{S=2000}$ and for compression ratio $\boldsymbol{r=0.5}$, with \textbf{resampling}. The horizontal dashed line denotes a normalized reference corresponding to the median of each feature distribution in the original search space. To enhance visualization, the limits of the Normalized Feature shift has been set to $[-1,1 ]$.}
    \label{fig:violin_f8_resampled_r0.5_n2000}
\end{figure}

\begin{figure}[hbtp]
    \centering
    \includegraphics[width=.7\linewidth]{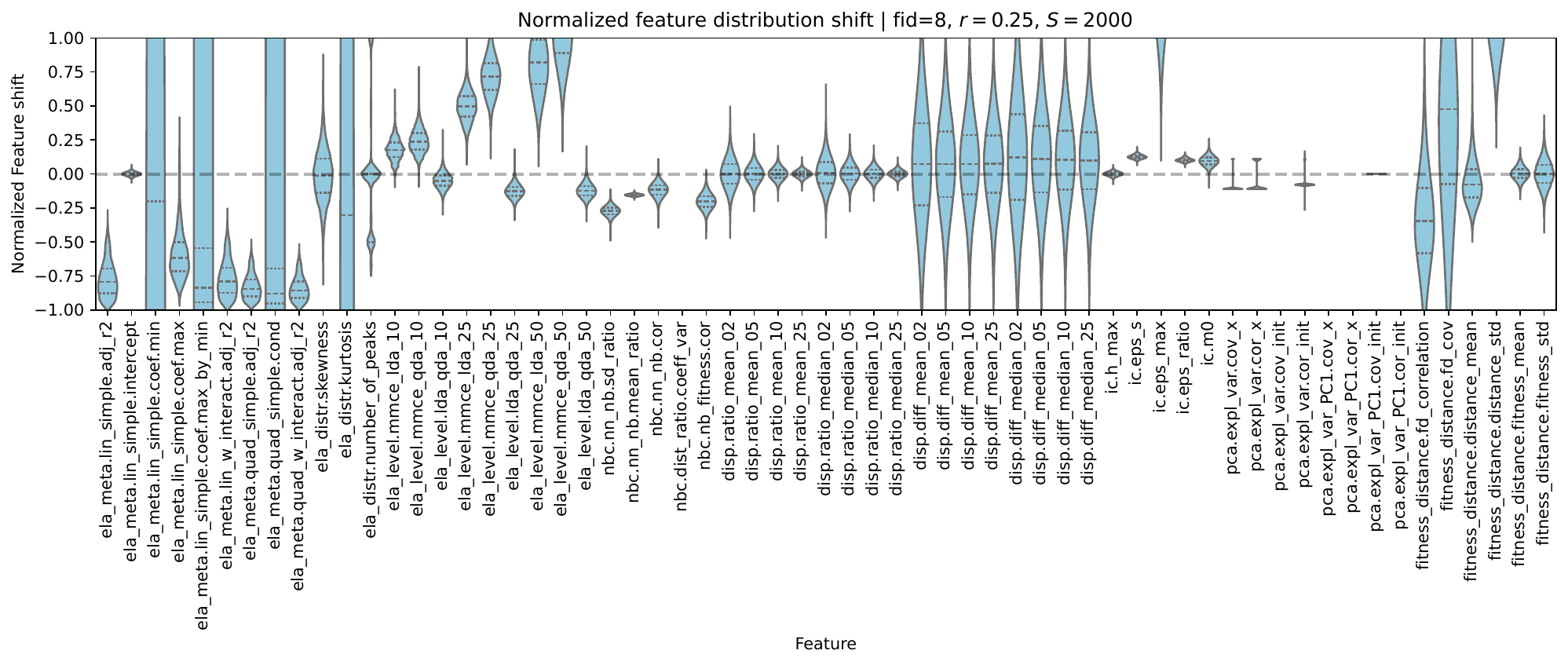}
    \caption{Same as above for $\boldsymbol{S=2000}$ and $\boldsymbol{r=0.25}$.}
    \label{fig:violin_f8_resampled_r0.25_n2000}
\end{figure}

\begin{figure}[hbtp]
    \centering
    \includegraphics[width=.7\linewidth]{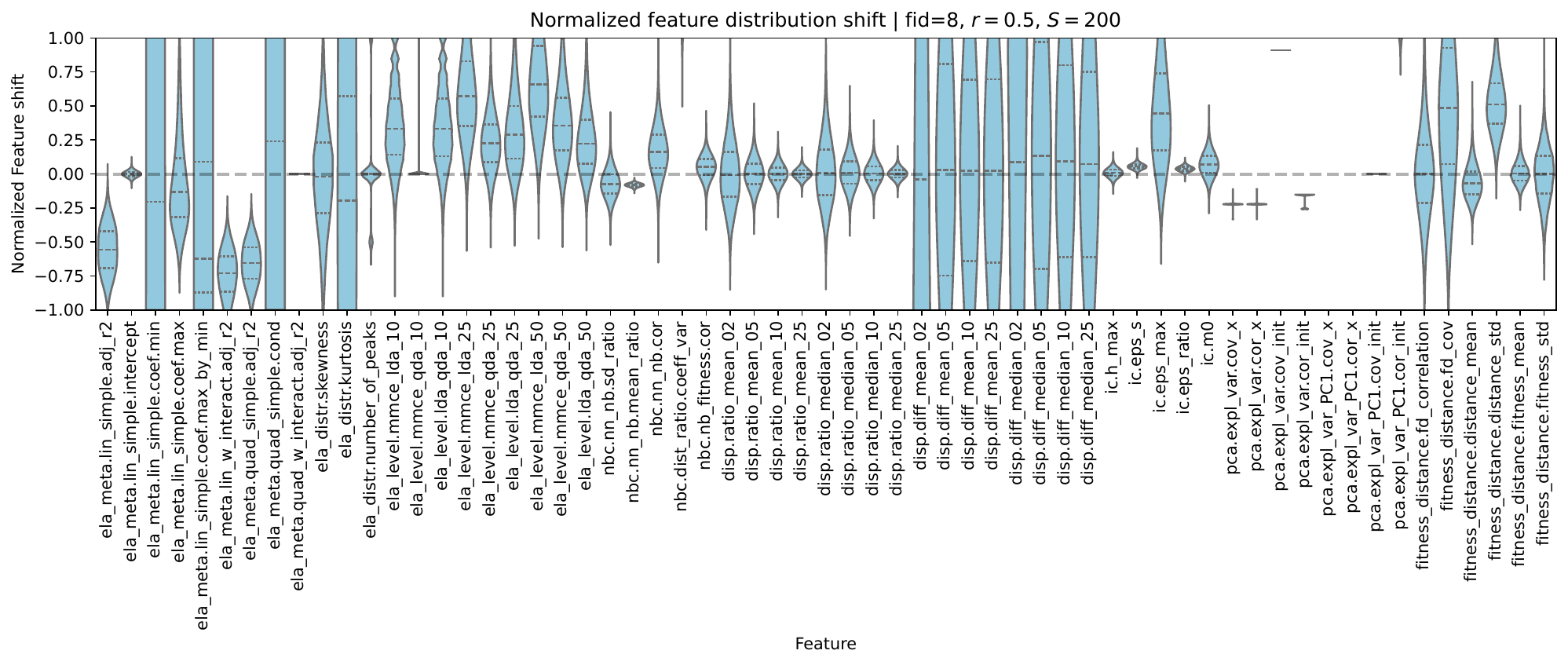}
    \caption{Same as above for $\boldsymbol{S=200}$ and $\boldsymbol{r=0.5}$.}
    \label{fig:violin_f8_resampled_r0.5_n200}
\end{figure}


\begin{figure}[hbtp]
    \centering
    \includegraphics[width=.7\linewidth]{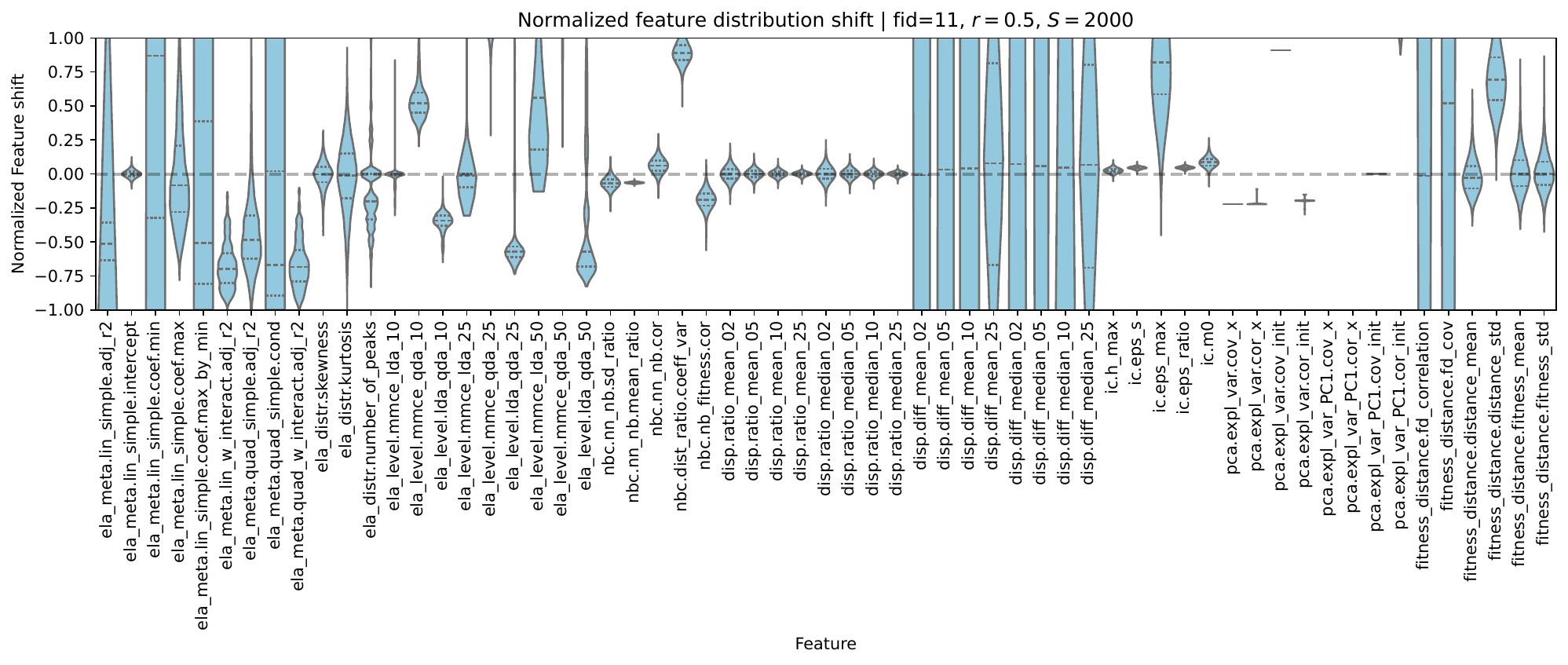}
    \caption{Normalized aggregated feature distribution shift of \textbf{Discus (f11) function} with $\boldsymbol{S=2000}$ and for compression ratio $\boldsymbol{r=0.5}$, with \textbf{resampling}. The horizontal dashed line denotes a normalized reference corresponding to the median of each feature distribution in the original search space. To enhance visualization, the limits of the Normalized Feature shift has been set to $[-1,1 ]$.}
    \label{fig:violin_f11_resampled_r0.5_n2000}
\end{figure}

\begin{figure}[hbtp]
    \centering
    \includegraphics[width=.7\linewidth]{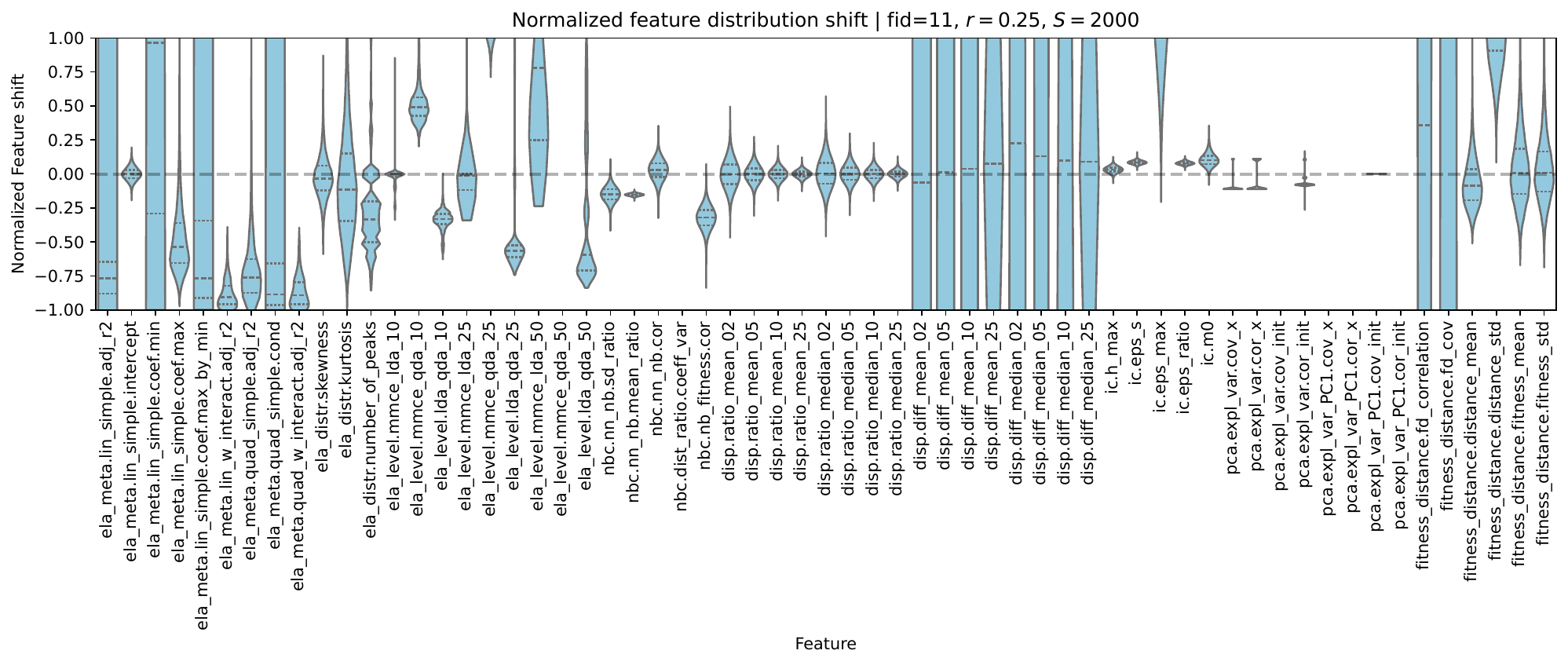}
    \caption{Same as above for $\boldsymbol{S=2000}$ and $\boldsymbol{r=0.25}$.}
    \label{fig:violin_f11_resampled_r0.25_n2000}
\end{figure}

\begin{figure}[hbtp]
    \centering
    \includegraphics[width=.7\linewidth]{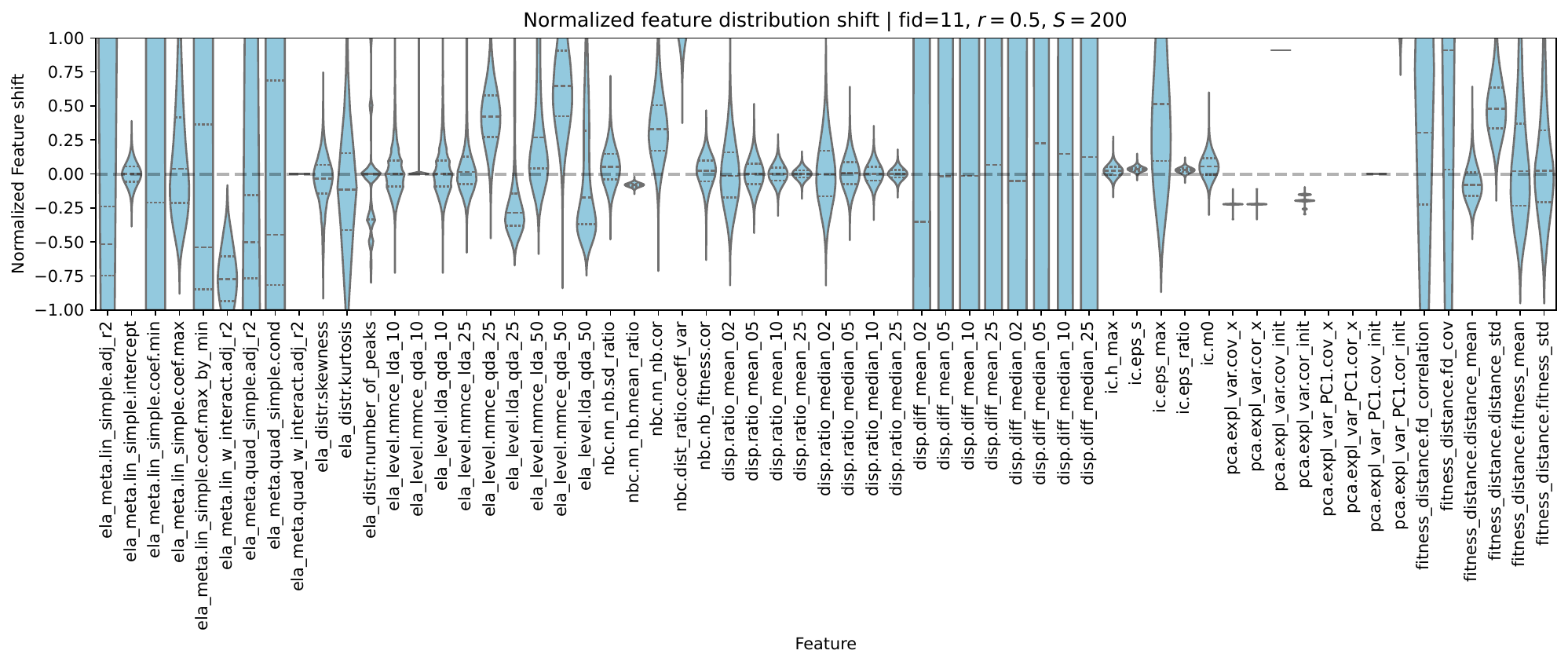}
    \caption{Same as above for $\boldsymbol{S=200}$ and $\boldsymbol{r=0.5}$.}
    \label{fig:violin_f11_resampled_r0.5_n200}
\end{figure}


\begin{figure}[hbtp]
    \centering
    \includegraphics[width=.7\linewidth]{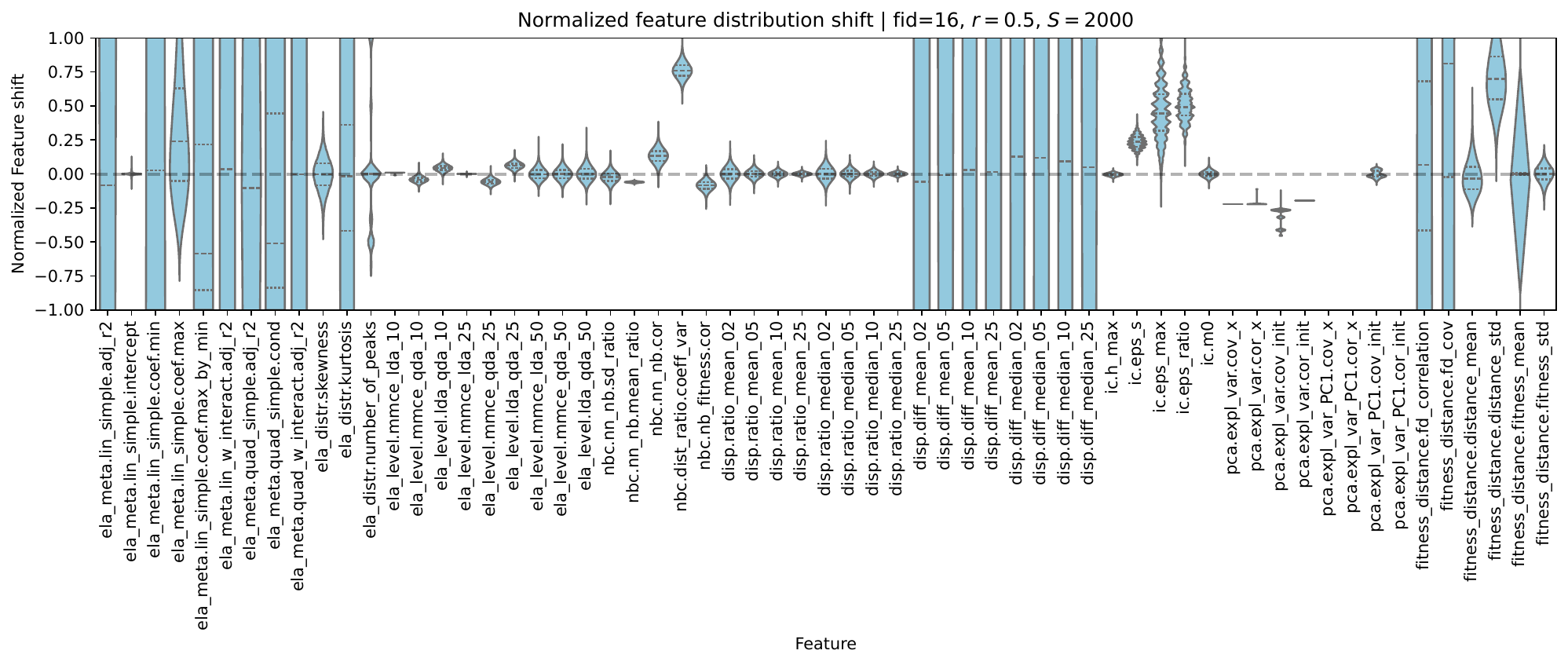}
    \caption{Normalized aggregated feature distribution shift of \textbf{Weierstrass (f16) function} with $\boldsymbol{S=2000}$ and for compression ratio $\boldsymbol{r=0.5}$, with \textbf{resampling}. The horizontal dashed line denotes a normalized reference corresponding to the median of each feature distribution in the original search space. To enhance visualization, the limits of the Normalized Feature shift has been set to $[-1,1 ]$.}
    \label{fig:violin_f16_resampled_r0.5_n2000}
\end{figure}

\begin{figure}[hbtp]
    \centering
    \includegraphics[width=.7\linewidth]{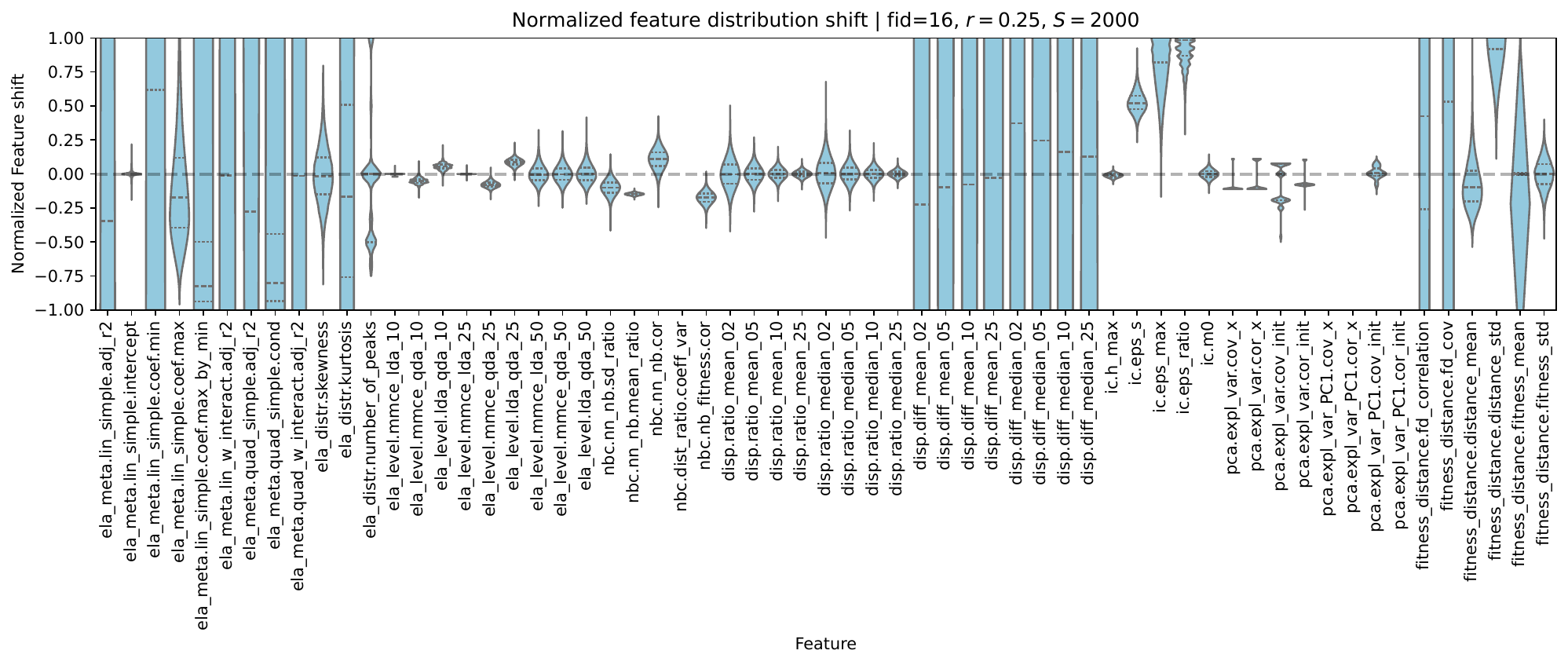}
    \caption{Same as above for $\boldsymbol{S=2000}$ and $\boldsymbol{r=0.25}$.}
    \label{fig:violin_f16_resampled_r0.25_n2000}
\end{figure}

\begin{figure}[hbtp]
    \centering
    \includegraphics[width=.7\linewidth]{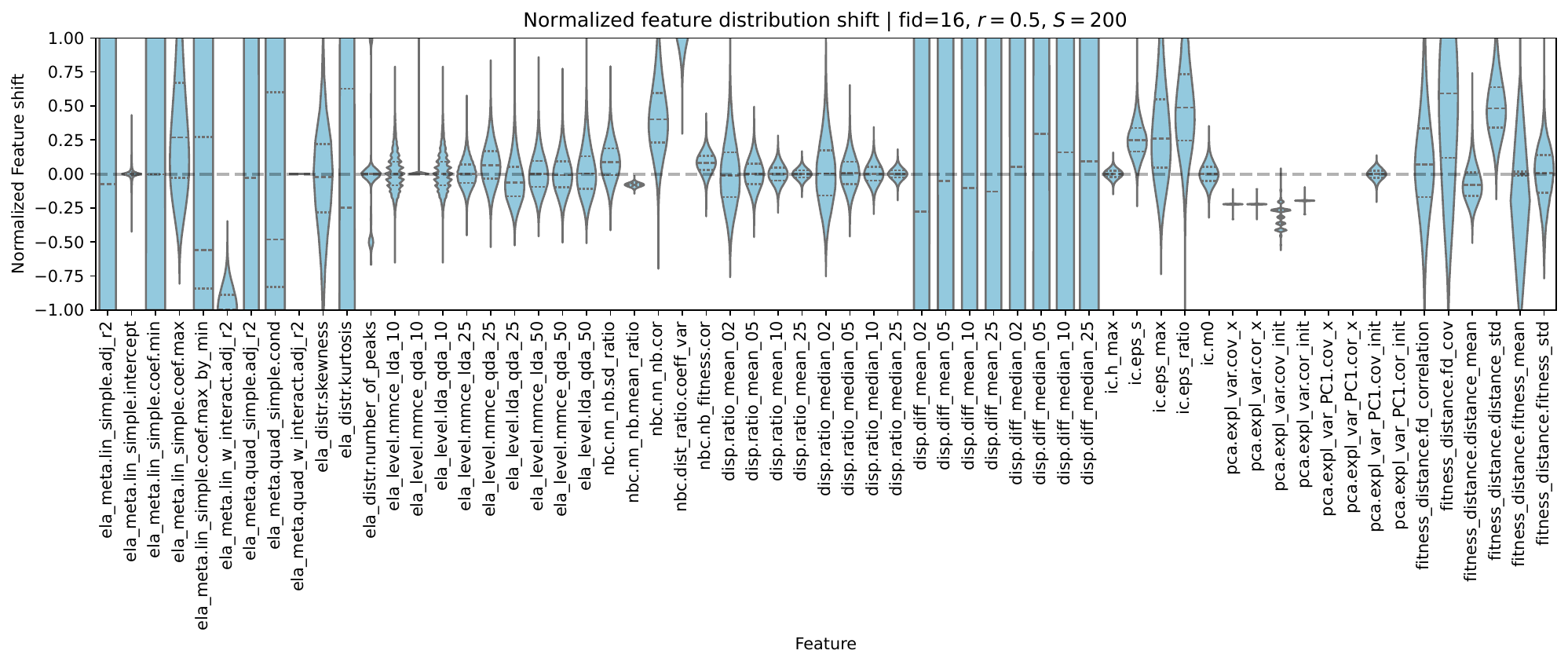}
    \caption{Same as above for $\boldsymbol{S=200}$ and $\boldsymbol{r=0.5}$.}
    \label{fig:violin_f16_resampled_r0.5_n200}
\end{figure}


\begin{figure}[hbtp]
    \centering
    \includegraphics[width=.7\linewidth]{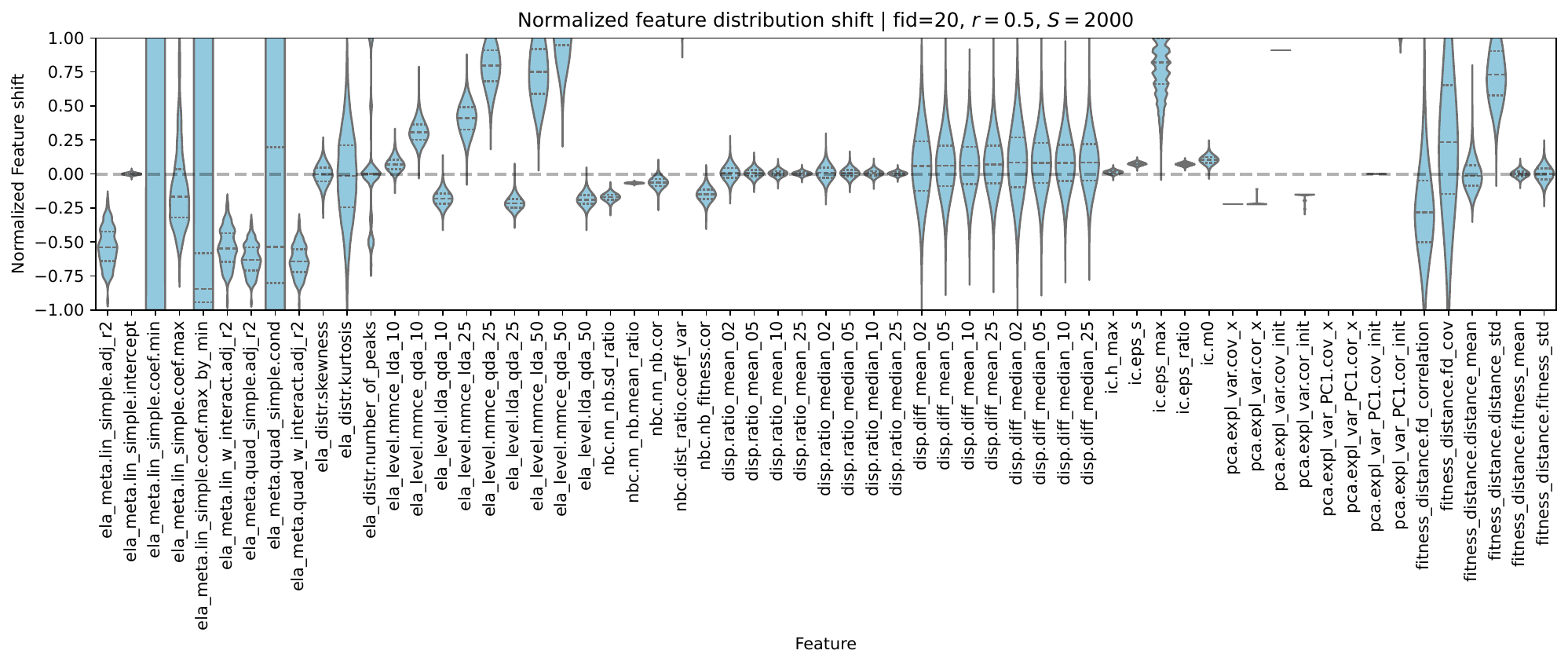}
    \caption{Normalized aggregated feature distribution shift of \textbf{Schwefel (f20) function} with $\boldsymbol{S=2000}$ and for compression ratio $\boldsymbol{r=0.5}$, with \textbf{resampling}. The horizontal dashed line denotes a normalized reference corresponding to the median of each feature distribution in the original search space. To enhance visualization, the limits of the Normalized Feature shift has been set to $[-1,1 ]$.}
    \label{fig:violin_f20_resampled_r0.5_n2000}
\end{figure}

\begin{figure}[hbtp]
    \centering
    \includegraphics[width=.7\linewidth]{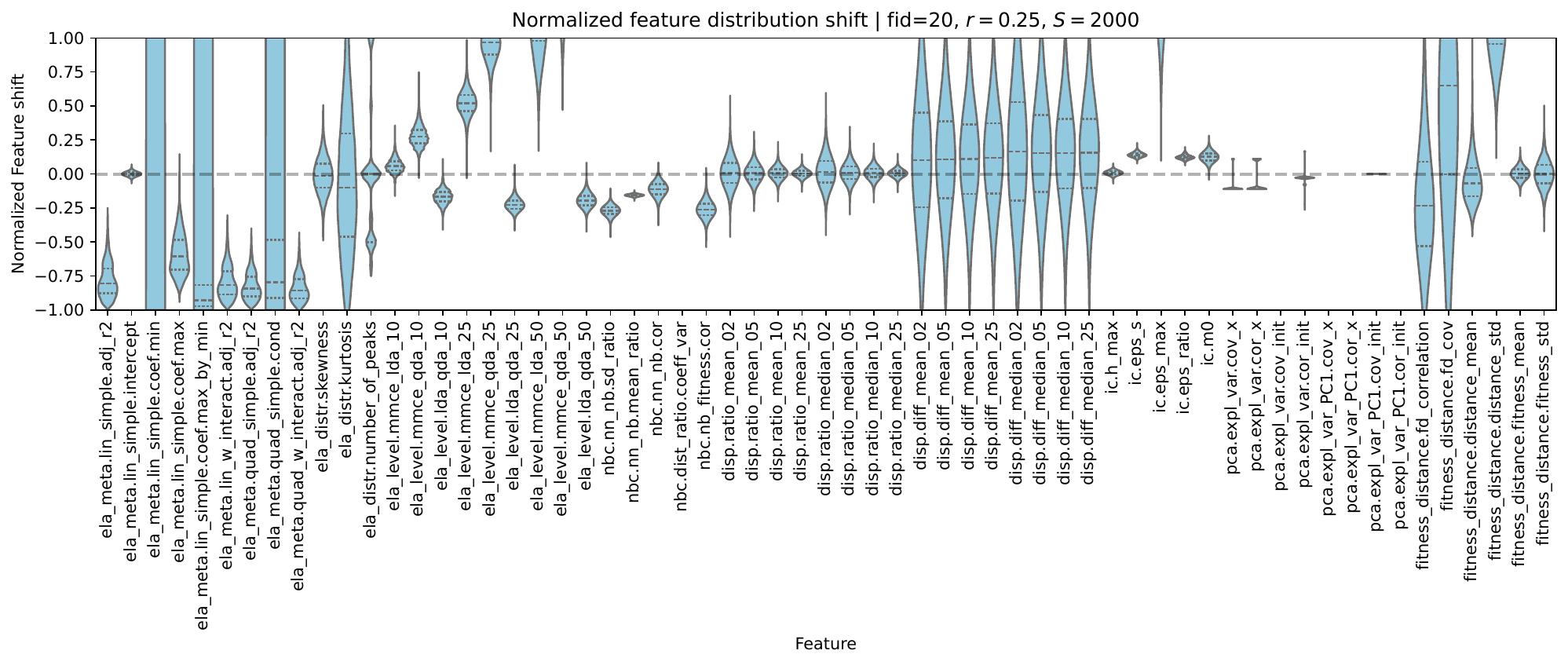}
    \caption{Same as above for $\boldsymbol{S=2000}$ and $\boldsymbol{r=0.25}$.}
    \label{fig:violin_f20_resampled_r0.25_n2000}
\end{figure}

\begin{figure}[hbtp]
    \centering
    \includegraphics[width=.7\linewidth]{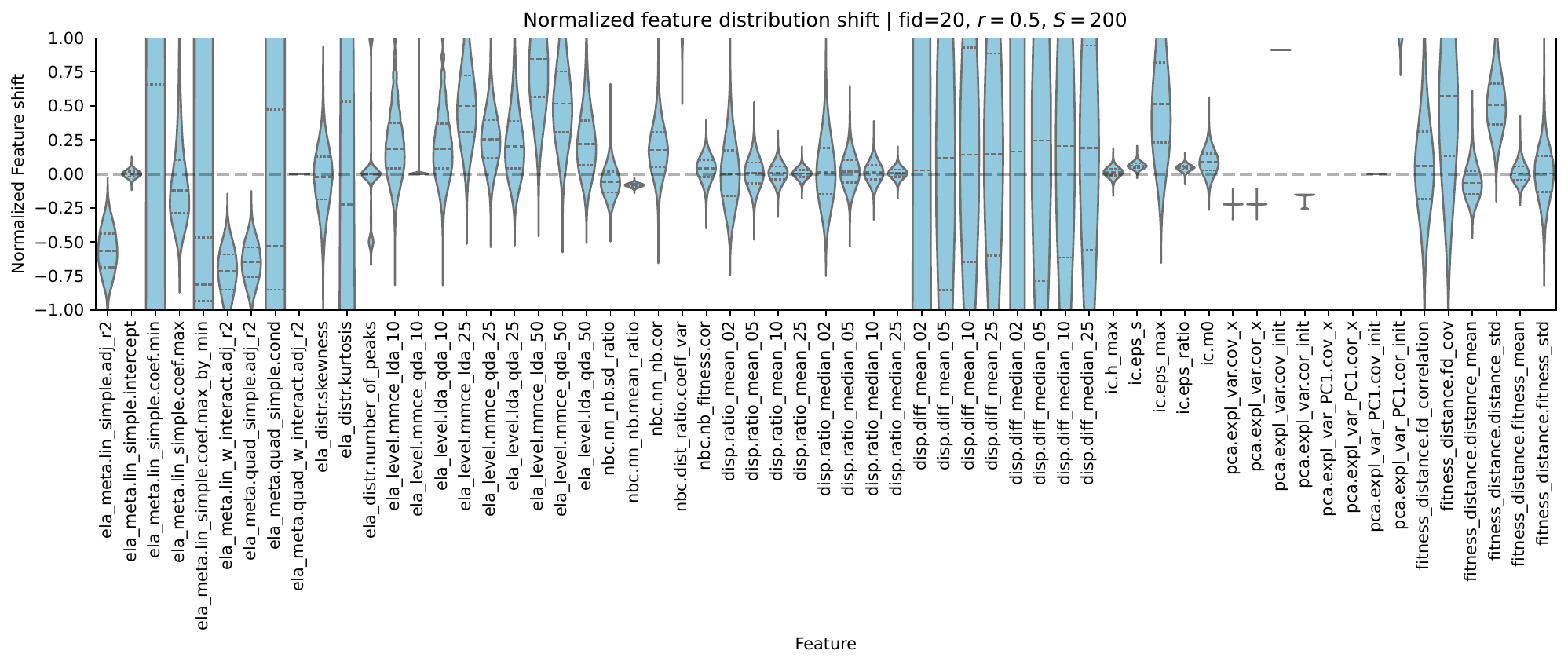}
    \caption{Same as above for $\boldsymbol{S=200}$ and $\boldsymbol{r=0.5}$.}
    \label{fig:violin_f20_resampled_r0.5_n200}
\end{figure}
